%% file: main.tex
\documentclass{article}
\usepackage{iclr2026_conference,times}
\usepackage{amsmath,amssymb,amsfonts}
\usepackage{graphicx}
\usepackage{booktabs}
\usepackage{subcaption}
\usepackage{wrapfig}
\usepackage{xcolor}
\usepackage{microtype}
\usepackage{hyperref}
\usepackage{url}
\usepackage{float}   % 프리앰블
\usepackage{array}
\usepackage{placeins} 
\input{math_commands.tex}  % keep if your project uses it

\title{FROST: Training-Free Few-Shot Segmentation with Frozen Features and Nonparametric Statistics}
\author{\makebox[\textwidth][c]{%
\begin{tabular}{c}
\rule{0pt}{3em}Junghwan Park \\   % 이름 줄 위에 3em 높이의 보이지 않는 받침
\mdseries TelePIX \\
\mdseries \texttt{junghwan@telepix.net}
\end{tabular}}}

\iclrfinalcopy

\begin{document}
\maketitle

\lhead{}      % ICLR 헤더 문구 제거
\renewcommand{\headrulewidth}{0pt}   % 머리글 밑줄 제거

\begin{abstract}
Few-shot segmentation asks a model to delineate a target class in a query image from only a handful of annotated examples, a setting most acute in remote sensing, where labels are scarce and the imagery departs sharply from the natural images on which vision backbones are pretrained. Prevailing approaches either train a segmenter on labelled episodes, which raises accuracy within the training distribution but binds the model to it, or reduce each class to a lossy summary of frozen features, a single prototype, a few cluster prototypes, or a discrete clustering, none of which preserves the internal structure of a multimodal class. We argue that a class is better described by a distribution than by a point, and that frozen self-supervised features already carry enough structure to estimate that distribution directly. We introduce FROST, a training-free few-shot segmenter that treats the reference foreground and background as two point clouds on the unit sphere of frozen DINOv3 features and labels each query token by a nonparametric density ratio, with a threshold the Bayes rule fixes at zero under equal priors. Because the variance of a density estimate shrinks as its sample grows, the decision sharpens as references accumulate, and every remaining quantity from the kernel bandwidth to the spatial gate is read from the support set rather than tuned. We develop FROST for overhead imagery, where a class is typically a scatter of many small and dissimilar instances that a density tracks but a lossy summary blurs. Across seventeen remote-sensing benchmarks FROST surpasses both training-free and learning-based methods, leading by 5.6 mIoU from a single annotated example and widening its lead as the support set grows, all while remaining among the smallest models compared. Code is available at \url{https://github.com/jhpark-ai/FROST}.
\end{abstract}

\section{Introduction}

Producing a segmentation map for a new remote-sensing task usually begins with drawing dense pixel labels by hand, an effort that grows with the scenes and the expertise the target classes demand, and that recurs whenever the sensor, the region, or the set of classes changes. Label efficiency is therefore not a convenience but a precondition for the field, and the obstacle is sharper than in natural imagery, because a model pretrained on everyday photographs meets an unfamiliar viewpoint, scale, and spectral makeup over the Earth, so competence on one collection seldom survives the move to another without new labels and renewed training.

\begin{figure}[t]
\centering
% ===================== Figure 1: concept (left half) =====================
\begin{minipage}[t]{0.5\textwidth}
\centering
\setcounter{subfigure}{0}%
\includegraphics[width=0.95\linewidth]{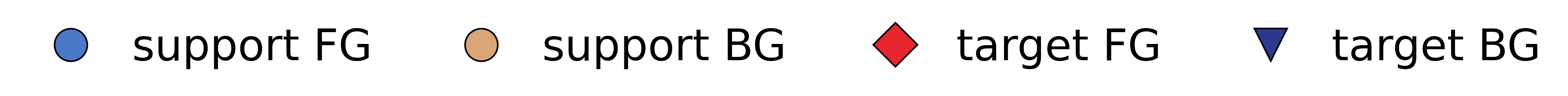}\\[3pt]
\begin{subfigure}[b]{0.32\linewidth}\centering
  \includegraphics[height=2.1cm]{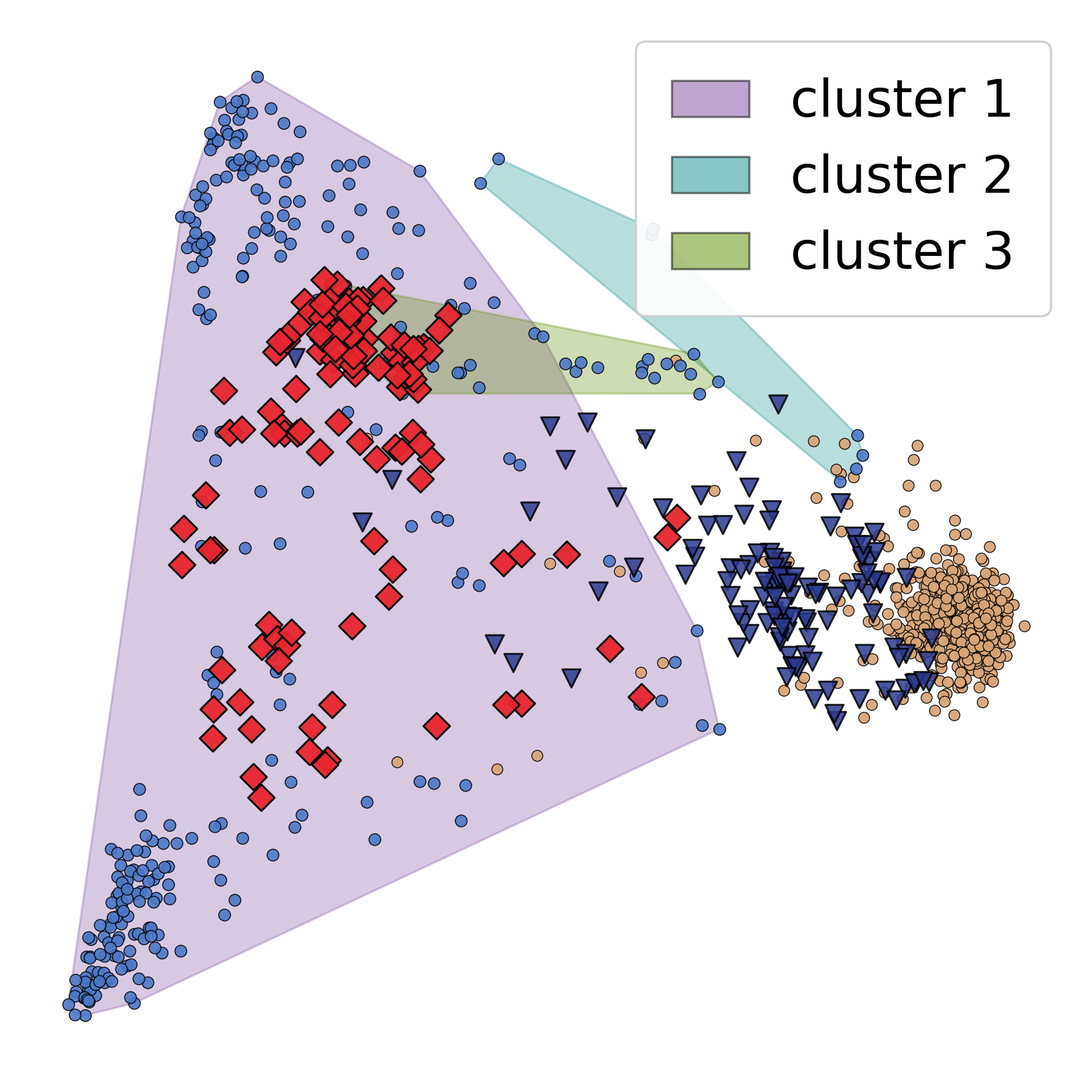}
  \caption{INSID3}\label{fig:c_insid3}
\end{subfigure}\hfill
\begin{subfigure}[b]{0.32\linewidth}\centering
  \includegraphics[height=2.1cm]{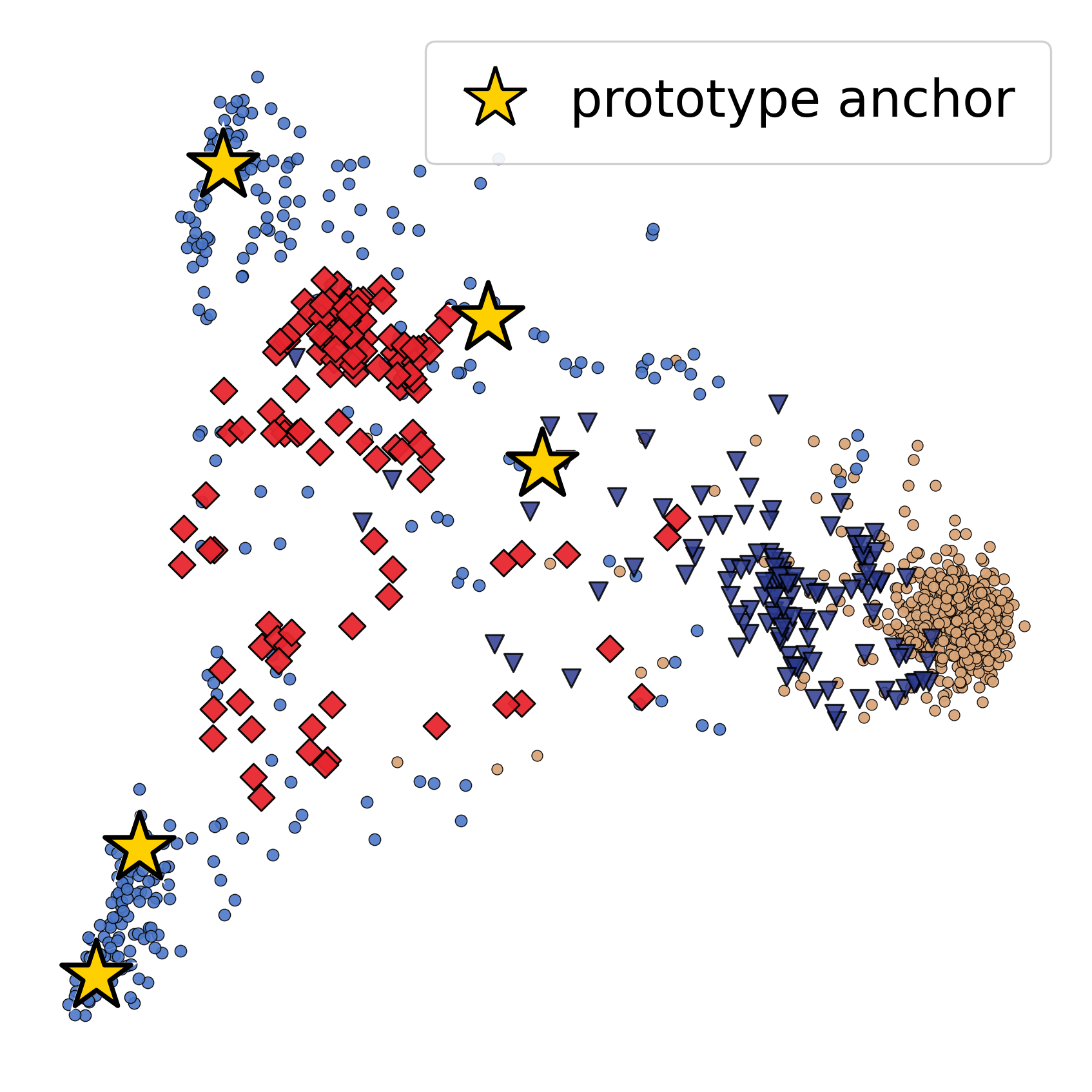}
  \caption{FSSDINO}\label{fig:c_fssdino}
\end{subfigure}\hfill
\begin{subfigure}[b]{0.32\linewidth}\centering
  \includegraphics[height=2.1cm]{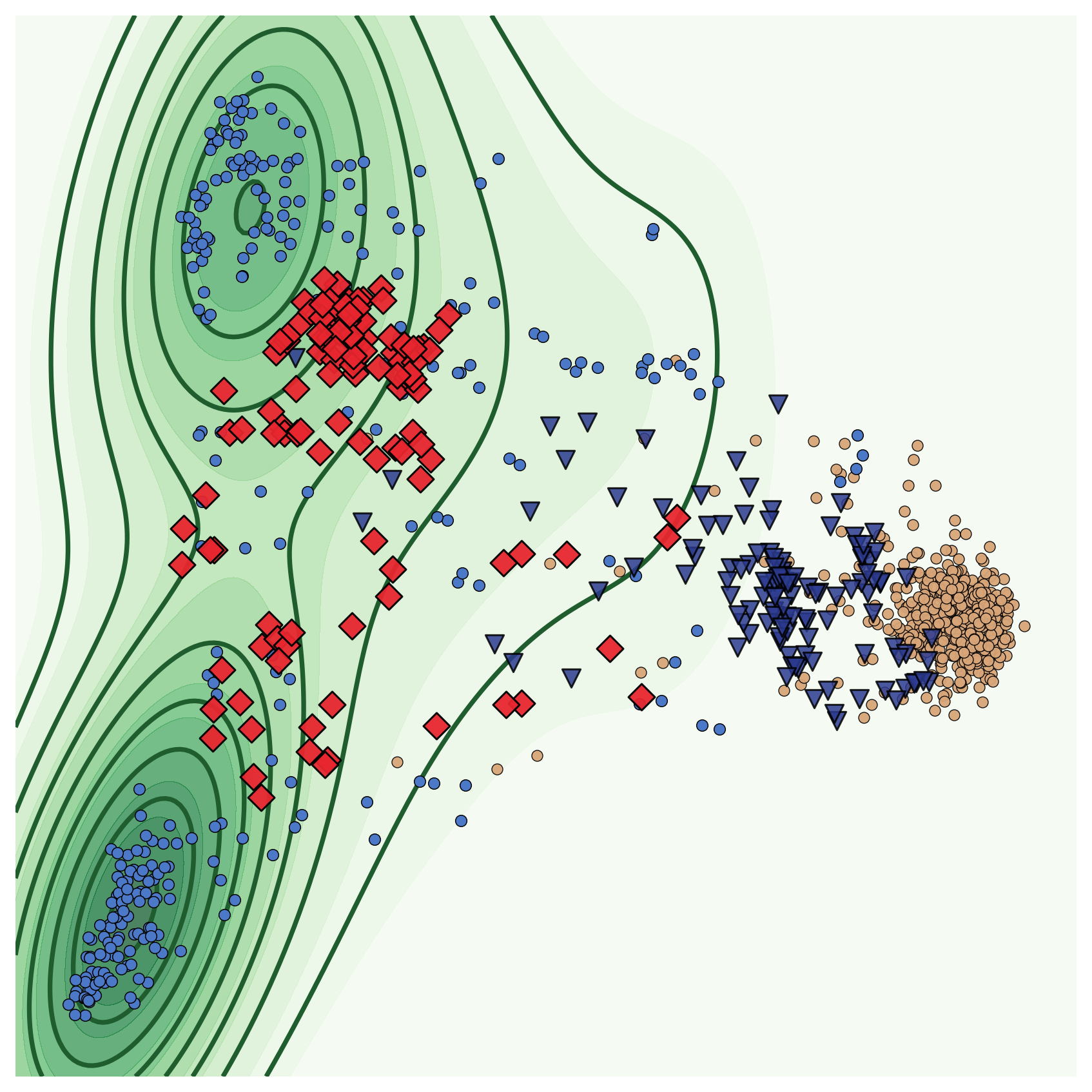}
  \caption{FROST}\label{fig:c_frost}
\end{subfigure}
\caption{On a shared PCA embedding of one remote-sensing support and query episode, INSID3 forms clusters, FSSDINO keeps prototypes, and FROST keeps full densities.}
\label{fig:concept}
\end{minipage}\hfill
% ===================== Figure 2: overview (right half) =====================
\begin{minipage}[t]{0.48\textwidth}
\centering
\setcounter{subfigure}{0}%
\includegraphics[width=0.95\linewidth]{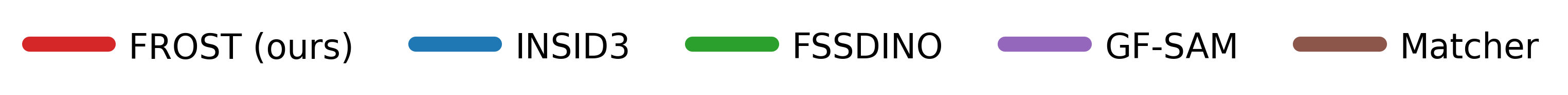}\\[3pt]
\begin{subfigure}[b]{0.56\linewidth}\centering
  \includegraphics[height=2.1cm]{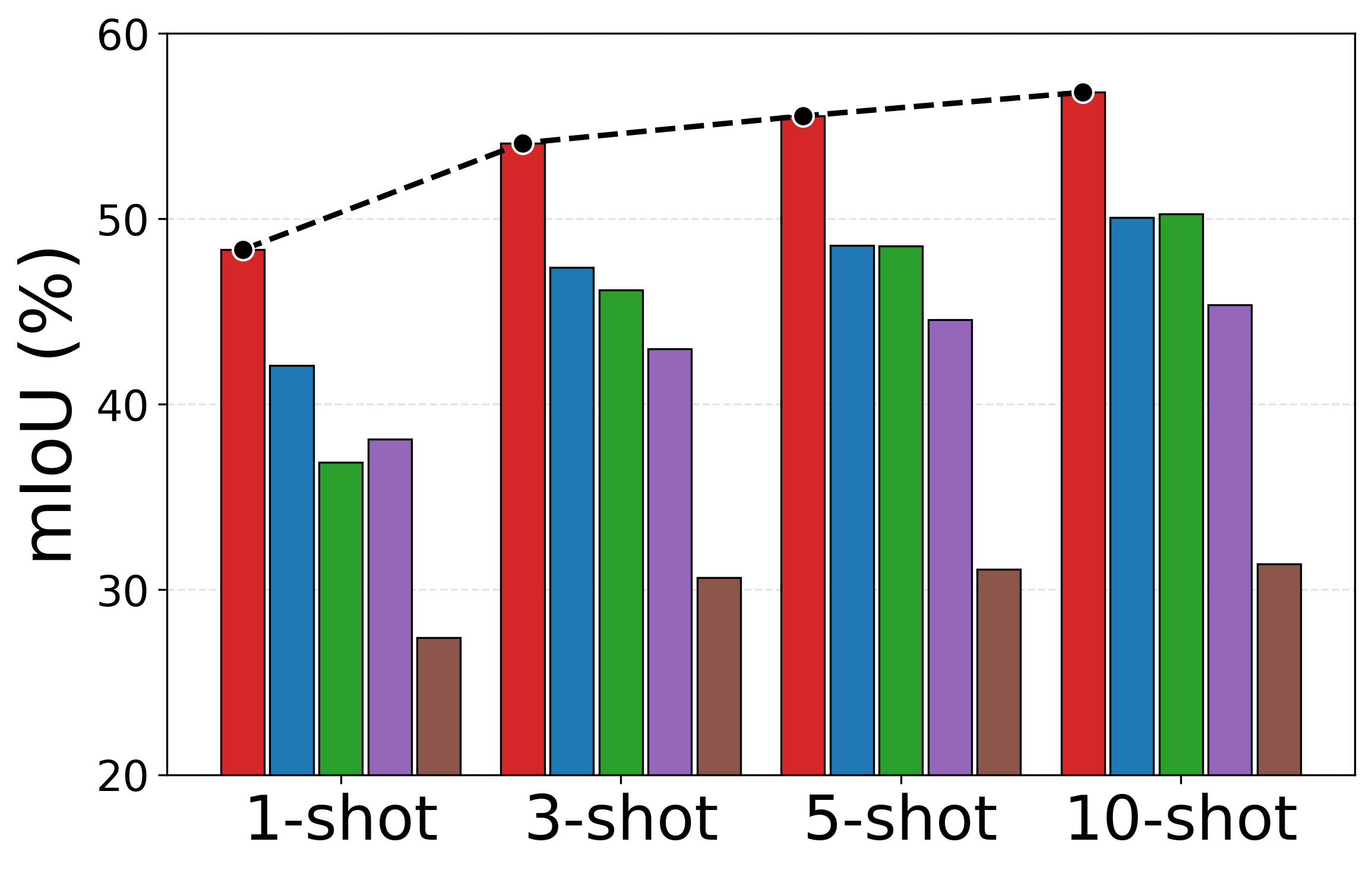}
  \caption{Per support size}\label{fig:o_bars}
\end{subfigure}\hfill
\begin{subfigure}[b]{0.40\linewidth}\centering
  \includegraphics[height=2.1cm]{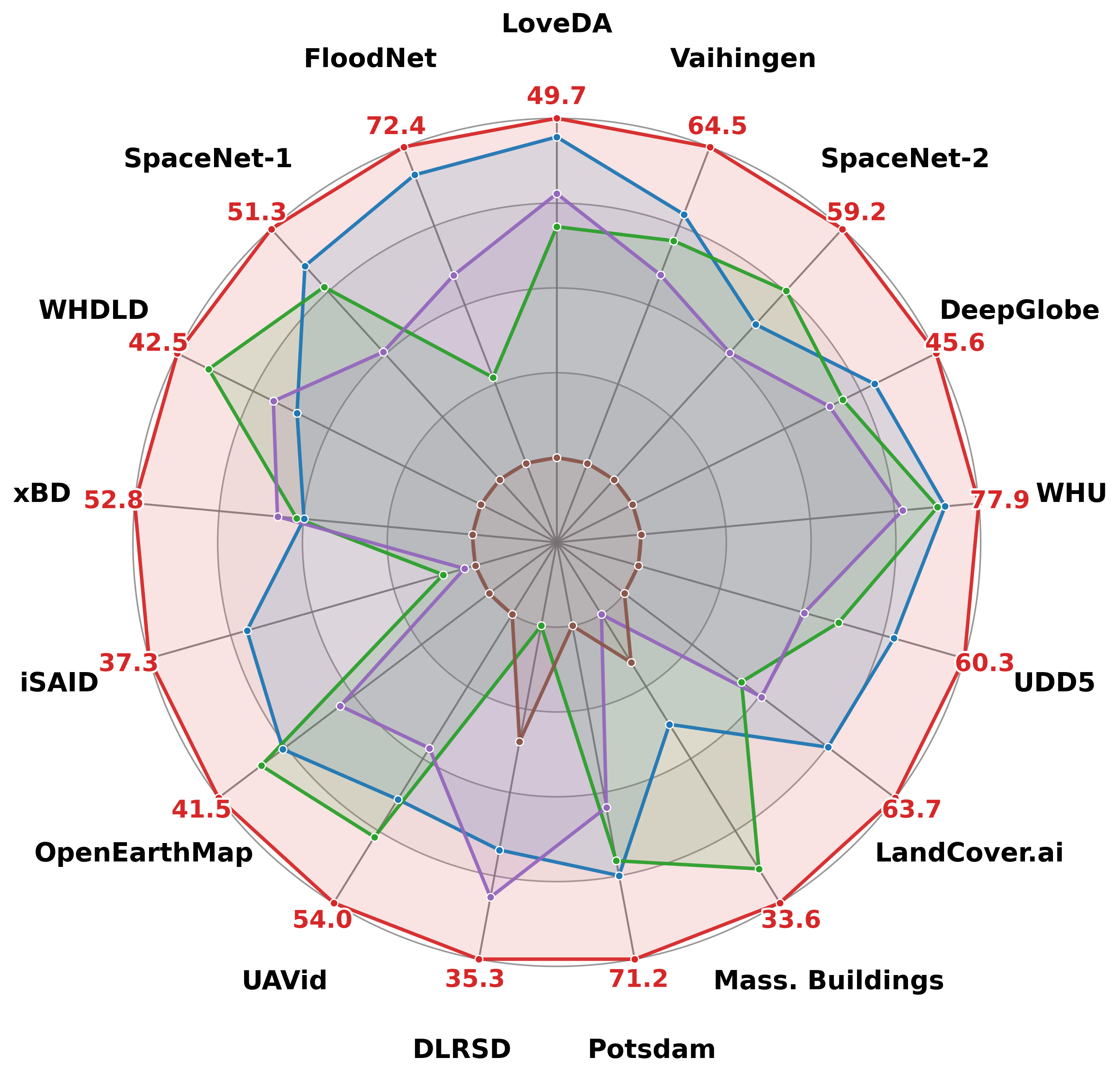}
  \caption{Per benchmark}\label{fig:o_radar}
\end{subfigure}
\caption{On the seventeen remote-sensing benchmarks, FROST leads the strongest training-free few-shot segmenters at every support size $k \in \{1, 3, 5, 10\}$ and on benchmarks.}
\label{fig:overview}
\end{minipage}
\end{figure}

Few-shot segmentation is the response built for this constraint, asking a model to reproduce a class from only a few reference masks. The established route teaches this skill by optimizing over labelled episodes, but the resulting parameters carry the imprint of the classes and scenes they are fit to, so a new domain reopens the need for annotation and retraining \citep{min2021hypercorrelation, hong2022cost}. A newer route drops the optimization and reads the answer off a frozen self-supervised encoder's features, whose neighbourhoods already place semantically related regions together with no segmentation label ever seen \citep{simeoni2025dinov3}. These frozen-feature methods still reduce a class to a lossy summary, however, whether by collapsing the support into one or a few prototypes \citep{zakir2026revealing} or by committing the scene to a handful of discrete clusters \citep{cuttano2026insid3}, discarding the internal structure of a class that appears as many small and dissimilar pieces, the rule rather than the exception in overhead imagery. Figure~\ref{fig:concept} makes this loss concrete on one episode embedded by principal component analysis, where the FSSDINO prototypes average across the appearance modes and miss the smaller ones, the INSID3 clusters impose hard boundaries that drop a mode or absorb background, and the FROST density tracks the foreground across its modes.

Our position is that the ingredient still missing from frozen-feature segmentation is not more training but a sharper decision, one that represents a class by its full distribution rather than a lossy summary. We introduce FROST, a training-free few-shot segmenter assembled from quantities estimated on the support set. A frozen DINOv3 encoder embeds the support and query images, the reference masks split the support tokens into foreground and background anchors, and each anchor set defines a kernel density on the feature sphere, so a query token is labelled foreground whenever its foreground density exceeds its background density. This is a likelihood-ratio test whose threshold sits at zero by the Bayes rule under equal priors rather than at a tuned value, and because a density estimate tightens as its sample grows, the decision sharpens as reference masks accumulate, a scaling property the lossy summaries lack. The kernel width and a spatial gate are likewise read from the support alone, leaving no quantity to calibrate per dataset. We develop these choices for overhead imagery, where a class is a scatter of many small and dissimilar instances rather than one dominant object, so a single prototype averages across appearance modes that the density retains, while the foreground fills a roughly balanced fraction of a tile that matches the equal-prior threshold. Both properties hold far less often in object-centric natural images, which is why we expect, and later observe, the advantage of FROST to be largest in the overhead setting.

Our contributions are threefold. First, we present FROST, a fully training-free few-shot segmentation pipeline that returns high-resolution masks from a frozen backbone and lightweight statistics computed on the support set. Second, we cast labelling as a kernel density ratio test with a Bayes threshold and a bandwidth chosen by a leave-one-out class margin, and adapt its feature stage to the scarce-reference regime with a shrinkage Mahalanobis whitening of the within-class scatter. Third, we evaluate FROST on seventeen remote-sensing benchmarks, where it overtakes both training-free and learning-based methods from a single mask and widens its lead as the support set grows, and on natural-image and cross-domain benchmarks, where it overtakes the strongest training-free methods as references accumulate, the smaller advantage there tracing to its prior-free and equal-prior design. Figure~\ref{fig:overview} summarizes this outcome, where FROST leads on sixteen of the seventeen remote-sensing benchmarks while remaining among the smallest models compared.

\section{Related Work}

\paragraph{Few-shot segmentation.} The task hands a model only a handful of labelled references and asks it to segment the same class elsewhere, and most solutions learn how. Prototype methods average the support foreground into a few vectors and match query pixels against them \citep{wang2019panet, tian2020prior}, correlation methods learn dense affinity or hypercorrelation between support and query \citep{min2021hypercorrelation, hong2022cost}, and generalist in-context models predict a mask from a visual example \citep{wang2023seggpt}. A distinct group leaves a promptable foundation model untouched and feeds it prompts from the references, so its mask decoder produces the output \citep{kirillov2023segment, zhang2024personalize, liu2023matcher}. Each of these designs pays for its accuracy with a learning stage on segmentation data or with a decoder already trained on masks, and that dependence is what erodes performance once the deployment domain drifts. Because the classes and scenes of that learning stage are natural images, the imprint they leave transfers poorly to overhead imagery, where the viewpoint, scale, and spectral makeup all shift at once. FROST keeps neither and reads its prediction from frozen features and reference statistics.

\paragraph{Segmentation from frozen self-supervised features.} Self-supervised vision transformers expose dense features whose neighbourhoods respect object extent and cross-image correspondence, which has supported unsupervised discovery and matching with no fine-tuning \citep{caron2021emerging, simeoni2021localizing, amir2021deep}, and the localized features of DINOv3 are accurate enough to carry segmentation from a single frozen model \citep{simeoni2025dinov3}. INSID3 segments in context from these features by grouping a scene into discrete regions and transferring labels through correspondence, after cancelling a positional bias it finds in the similarities so that matching proceeds on content \citep{cuttano2026insid3}. FSSDINO probes the few-shot capacity of the same features with cluster prototypes refined by Gram matrices, and finds the last layer is not the most informative one \citep{zakir2026revealing}. FROST is training-free in the same sense and includes a positional debiasing step we attribute to INSID3 in Section~\ref{sec:refine}, but its decision rule sets it apart. In place of a prototype or a discrete clustering, FROST forms a density for the foreground and the background and compares them as a ratio, which supplies a threshold of zero without calibration and an accuracy that rises as references accumulate, which neither a fixed prototype nor a hard clustering offers.

% \paragraph{Open-vocabulary and text-prompted segmentation.} Another family names the target in words and segments by aligning dense visual features with a text embedding \citep{luddecke2022image, ghiasi2022scaling, liang2023open}, with remote-sensing adaptations that recover lost spatial detail or marshal additional foundation models to firm up boundaries \citep{li2025segearth}. These systems request no annotation when run and range over an open set of names, yet the accuracy they can reach on a given class is settled during pretraining and cannot be raised by supplying a mask. That is a different supply of supervision from the few labelled masks we assume, so we compare FROST only with mask-based few-shot methods.

\paragraph{Nonparametric and regularized statistics.} FROST is built from established tools. Deciding between classes by competing kernel density estimates, and tuning the kernel width from the data, are standard in nonparametric statistics \citep{silverman2018density}, while pulling a noisy covariance estimate toward a structured target is the classical remedy when the dimension is large relative to the sample \citep{ledoit2004well}. FROST carries this machinery onto the dense features of a foundation model, and the gains we measure from heavy shrinkage and from each added anchor are the effects this theory leads one to expect.

\section{Method}
\label{sec:method}

FROST assembles a few-shot segmenter from a frozen encoder and statistics computed on the support set alone, with no training and no quantity tuned per dataset. We first fix the setting and notation and sketch the pipeline, then describe its three stages in turn, the feature refinement that conditions the geometry of the frozen tokens, the density-ratio rule that labels each query token, and the spatial refinement that turns the per-token scores into a full-resolution mask.

\begin{figure}[t]
\centering
\includegraphics[width=\linewidth]{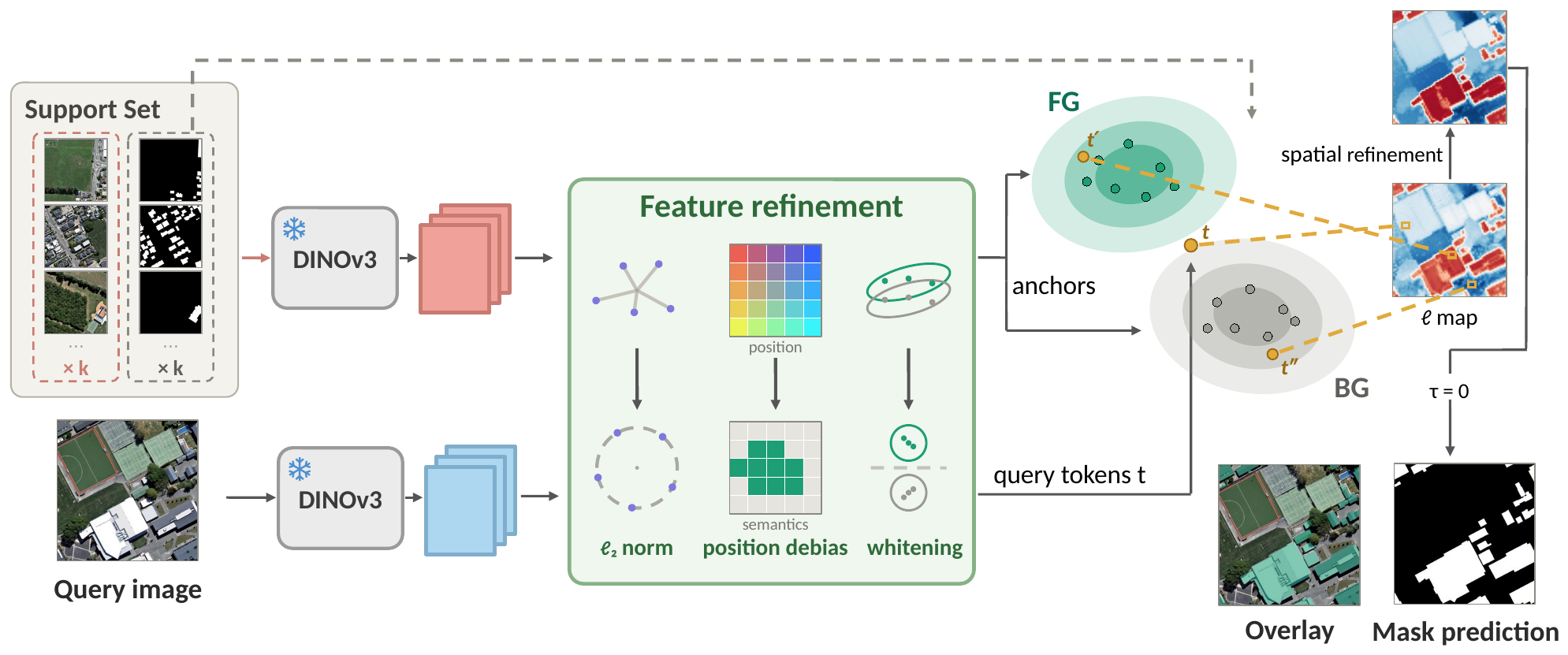}
\caption{The FROST pipeline, from frozen DINOv3 tokens to a mask through feature refinement, a density ratio over foreground and background anchors, spatial smoothing and gating, and upsampling.}
\label{fig:arch}
\end{figure}

\subsection{Setting and overview}
We are given a query image and a small support set of $k$ images with binary masks marking a target class, and we predict a mask for that class in the query image. A frozen DINOv3 ViT-L/16 encoder maps every image to a grid of $h \times w$ patch tokens of dimension $C$, with $h = w = 64$ and $C = 1024$ at the $1024 \times 1024$ input we use, and we double the support by horizontal flipping. FROST refines these tokens, splits them into foreground and background anchors with the reference masks, scores each query token by a ratio of two anchor densities, smooths and gates the resulting scalar field over space, and upsamples it into a full-resolution mask, as Figure~\ref{fig:arch} shows. These choices suit the structure of overhead imagery, where a class is often many small and dissimilar instances. No quantity in FROST is tuned per dataset, since each is either read from the support set, such as the whitening transform, the kernel bandwidth, and the gate radius, or held to one value across all benchmarks, such as the projection rank and the shrinkage intensity, and Section~\ref{sec:ablation} ablates each. Throughout, the index $j$ ranges over the spatial positions of the token grid and $m$ over feature directions.

\subsection{Feature refinement}
\label{sec:refine}
A query token reaches a decision only through its similarities with the anchors, so the feature-space geometry governs every later step, and we condition it in three stages before estimating any density.
\paragraph{Normalization.} We first place every token on the unit sphere by $L^2$ normalization, which removes the variation in token magnitude across images so that the comparisons below depend on direction, the cosine similarity that the von Mises--Fisher kernel of Section~\ref{sec:kde} assumes.

\paragraph{Positional debiasing.} We then apply the positional debiasing of INSID3 \citep{cuttano2026insid3}, which finds that the cross-image similarity of frozen tokens is driven partly by absolute position rather than content. Being shared across images, this component raises the similarity between unrelated tokens at the same location and would pull the density ratio toward position rather than appearance, so following INSID3 we estimate it by passing a content-free Gaussian image $X_{\mathrm{noise}} \sim \mathcal{N}(0, I)$ through the encoder, taking the $r$ leading directions $u_1, \dots, u_r$ of its token covariance, and projecting them out of every token by $v \mapsto v - \sum_{m=1}^{r} (v^\top u_m)\, u_m$ before renormalizing, with $r = 250$.

\paragraph{Shrinkage whitening.} The reference tokens vary along a few high-variance directions that track global scene appearance rather than the target class, and we suppress them so that the kernel of Section~\ref{sec:kde} responds to class-discriminative variation. We measure this nuisance variation by the within-class scatter rather than the total covariance, because centring each class on its own mean keeps the foreground-to-background mean difference intact, whereas a total covariance would absorb it into its leading directions and the whitening would then erase the very signal the kernel must read. Splitting the support tokens by the reference masks into a foreground group $\mathcal{S}_{\mathrm{FG}}$ and a background group $\mathcal{S}_{\mathrm{BG}}$ of sizes $N_{\mathrm{FG}}$ and $N_{\mathrm{BG}}$ with means $\bar{a}_{\mathrm{FG}}$ and $\bar{a}_{\mathrm{BG}}$, we form
\begin{equation}
\Sigma_W = \frac{1}{N_{\mathrm{FG}} + N_{\mathrm{BG}} - 2}\Big( \sum_{a \in \mathcal{S}_{\mathrm{FG}}} (a - \bar{a}_{\mathrm{FG}})(a - \bar{a}_{\mathrm{FG}})^\top + \sum_{a \in \mathcal{S}_{\mathrm{BG}}} (a - \bar{a}_{\mathrm{BG}})(a - \bar{a}_{\mathrm{BG}})^\top \Big),
\end{equation}
and because the anchors are far fewer than the feature dimension $C$ we regularize it by heavy shrinkage toward a scaled identity, forming $\Sigma_{W,\lambda} = (1 - \lambda)\, \Sigma_W + \lambda\, \mu_W\, I_C$ with the mean eigenvalue $\mu_W = \tfrac{1}{C}\operatorname{tr}(\Sigma_W)$ and intensity $\lambda = 0.95$, so the empirical scatter retains only the weight $1 - \lambda = 0.05$ \citep{ledoit2004well}. We whiten every token by $\Sigma_{W,\lambda}^{-1/2}$ and keep the whitened scale rather than renormalizing, since the bilateral term of Section~\ref{sec:spatial} reads the whitened token directly and benefits from that scale, while the density ratio and the prototype compare directions through cosine similarity and need no unit norm. We write $t_j$ for the whitened token at position $j$.

\paragraph{Anchors.} The reference masks, downsampled to the token grid, label each support token, so after whitening the foreground tokens form the anchor set $\mathcal{A}_{\mathrm{FG}}$ of size $N_{\mathrm{FG}}$ and the background tokens form $\mathcal{A}_{\mathrm{BG}}$ of size $N_{\mathrm{BG}}$, the same positions used for the scatter above, and the horizontal flips of the support enlarge both sets.

\subsection{Density ratio classification}
\label{sec:kde}
\paragraph{Density ratio.} We treat each anchor set as a sample from a class-conditional density on the sphere and estimate it with the exponential of the cosine similarity $\cos(u,v) = \tfrac{u^\top v}{\lVert u \rVert\, \lVert v \rVert}$, the von Mises--Fisher kernel with concentration $1/\sigma$, so that for a query token $t_j$ the two estimates are
\begin{equation}
\scalebox{0.85}{$\displaystyle
\hat{p}(t_j \mid \mathrm{FG}) = \frac{1}{N_{\mathrm{FG}}} \sum_{a \in \mathcal{A}_{\mathrm{FG}}} \frac{1}{Z_\sigma} \exp\!\Big(\frac{\cos(t_j, a)}{\sigma}\Big),
\qquad
\hat{p}(t_j \mid \mathrm{BG}) = \frac{1}{N_{\mathrm{BG}}} \sum_{a \in \mathcal{A}_{\mathrm{BG}}} \frac{1}{Z_\sigma} \exp\!\Big(\frac{\cos(t_j, a)}{\sigma}\Big)
$}
\end{equation}
where $\sigma$ is the kernel bandwidth and $Z_\sigma$ its normalizing constant, and unlike a prototype, which collapses a class to a single point, this sum retains every anchor and represents a multimodal class by all of its modes. This matters in overhead imagery, where a class is commonly a scatter of small and dissimilar instances whose modes a single prototype averages over, while the density offers less over a prototype when a class is one coherent object, as more often in natural images. We label the query token by the log ratio of the two densities,
\begin{equation}
\ell_j = \log \hat{p}(t_j \mid \mathrm{FG}) - \log \hat{p}(t_j \mid \mathrm{BG}),
\label{eq:ratio}
\end{equation}
in which $Z_\sigma$ cancels, and we assign the token to the foreground when $\ell_j$ exceeds a threshold $\tau$, which the Bayes rule places at $\tau = 0$ under equal class priors rather than at a tuned value. Because the variance of a kernel density estimate falls as its sample grows and the anchor counts $N_{\mathrm{FG}}$ and $N_{\mathrm{BG}}$ grow with the number of reference masks, each added mask lowers the variance of $\ell_j$ and sharpens the decision without retraining.

\paragraph{Bandwidth selection.} The densities depend on $\sigma$, which we fix from the support set before any query is scored by a leave-one-out class margin over a small grid,
\begin{equation}
M(\sigma) = \frac{1}{N_{\mathrm{FG}} + N_{\mathrm{BG}}} \sum_{a} s_a\, \ell^{\setminus a}_\sigma(a),
\end{equation}
where $s_a = +1$ for a foreground anchor and $-1$ for a background one and $\ell^{\setminus a}_\sigma(a)$ is the log ratio of Equation~\ref{eq:ratio} at $a$ with $a$ removed from its own class set, and we keep the width that maximizes this signed margin. We choose $\sigma$ by this margin rather than by held-out likelihood because the margin is what the threshold acts on, and it reads only from the references and tunes nothing per dataset.

\subsection{Spatial refinement}
\label{sec:spatial}
The classifier so far scores each token on its own, and three inexpensive spatial operations turn the resulting field into a clean mask.

\paragraph{Bilateral propagation.} A token mislabelled in isolation is usually surrounded by similar-looking neighbours that are labelled correctly, so we let the score field exchange evidence among them through a bilateral affinity that couples feature and colour similarity within a window,
\begin{equation}
B_{jj'} = \exp\!\Big(\frac{t_j^\top t_{j'}}{\gamma_f}\Big)\, \exp\!\Big(-\frac{\lVert c_j - c_{j'} \rVert^2}{\gamma_c}\Big),
\qquad \lVert g_j - g_{j'} \rVert_\infty \le d_{\max},
\end{equation}
where $t_j$ is the whitened token at position $j$, $c_j$ the mean colour of its patch, $g_j$ its grid coordinate, $\gamma_f$ and $\gamma_c$ feature and colour temperatures, and the affinity vanishes outside the window of radius $d_{\max}$. We row-normalize $B$ into a row-stochastic matrix $P$ and iterate $\ell^{(\nu+1)}_j = (1 - \beta)\, \ell_j + \beta \sum_{j'} P_{jj'}\, \ell^{(\nu)}_{j'}$ from $\ell^{(0)} = \ell$ for a fixed number of steps with mixing weight $\beta = 0.70$, writing $\tilde{\ell}$ for the smoothed field, where the colour term keeps the propagation from bleeding across object boundaries.

\paragraph{Forward and backward candidate gating.} A dense overhead scene contains structures that resemble the target locally and raise isolated false positives, which we suppress with a forward and a backward test that demand support-query agreement and together mark which positions are admissible. The forward test admits a position only when its token has positive cosine similarity to the foreground prototype, the mean of the foreground anchors, discarding positions that already favour the background, and this prototype serves as a coarse pre-filter, not the classifier, which remains the density ratio of Equation~\ref{eq:ratio}. The backward test, adapted from the reciprocal matching of INSID3 \citep{cuttano2026insid3}, keeps a position only when it lies among the $k_b$ nearest query positions of the anchor it matches, with $k_b = 3$, and a spatial gate of radius $\rho = \lfloor \rho_{\max}\, \pi_{\mathrm{FG}} \rceil$, with $\pi_{\mathrm{FG}} = N_{\mathrm{FG}} / (N_{\mathrm{FG}} + N_{\mathrm{BG}})$ and $\rho_{\max} = 4$, restricts this agreement to a neighbourhood that widens for a class occupying more of the scene. The foreground fraction $\pi_{\mathrm{FG}}$ therefore enters the prediction only through the spatial extent of the admissible region and not through the per-token threshold, which the equal-prior assumption keeps at $\tau = 0$, an assumption well matched to overhead tiles in which the foreground fills a substantial and roughly balanced fraction of the scene. The two tests are computed on the $64 \times 64$ token grid and intersected into a candidate mask, applied at the final step below.

\paragraph{Upsampling and thresholding.} Thresholding $\tilde{\ell}$ on the $64 \times 64$ grid would quantize the boundary to patch borders, so we upsample the scalar field rather than the prediction, bilinearly interpolating $\tilde{\ell}$ to the $1024 \times 1024$ resolution and thresholding afterwards, so the boundary falls on the zero crossing at sub-patch precision. We then intersect it with the candidate mask carried to the same resolution to yield the final mask. Only the scalar field and the binary candidate mask are upsampled, never the high-dimensional tokens, which keeps the step inexpensive.

\section{Experiments}
\label{sec:exp}

We test whether labelling query tokens by a density ratio over frozen features matches or surpasses methods that train, across the breadth of remote-sensing imagery. We first describe the benchmarks, the evaluation protocol, the baselines, and the implementation, then report the main comparison at the four support sizes, study how the accuracy of each method scales as reference masks accumulate, and ablate the components of FROST.

\providecommand{\panw}{0.085\textwidth}
\begin{figure}[t]
\centering
\setlength{\tabcolsep}{0.4pt}
\renewcommand{\arraystretch}{0.4}
\resizebox{\textwidth}{!}{% 두 그룹을 한 줄로 묶어 본문 폭에 맞춤 (밀림/넘침 방지)
\begin{tabular}{@{}c@{\hspace{5pt}}|@{\hspace{5pt}}c@{}} % 가운데 | 로 좌/우 구분
\begin{tabular}{@{}m{1em} *{5}{>{\centering\arraybackslash}m{\panw}}@{}}
 & \scriptsize Reference & \scriptsize FSSDINO & \scriptsize INSID3 & \scriptsize FROST & \scriptsize GT \\
\rotatebox{90}{\scriptsize Building}
  & \includegraphics[width=\panw]{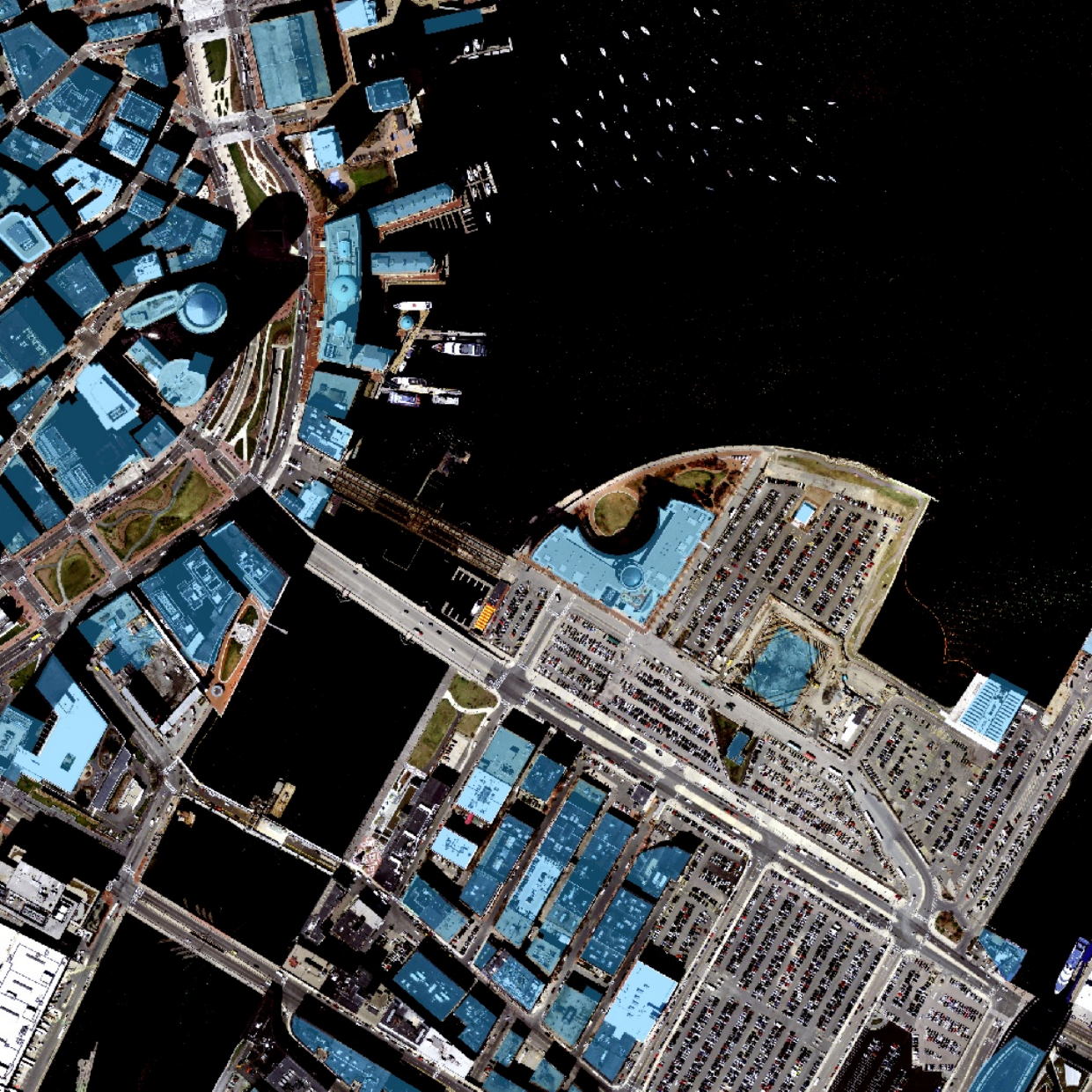}
  & \includegraphics[width=\panw]{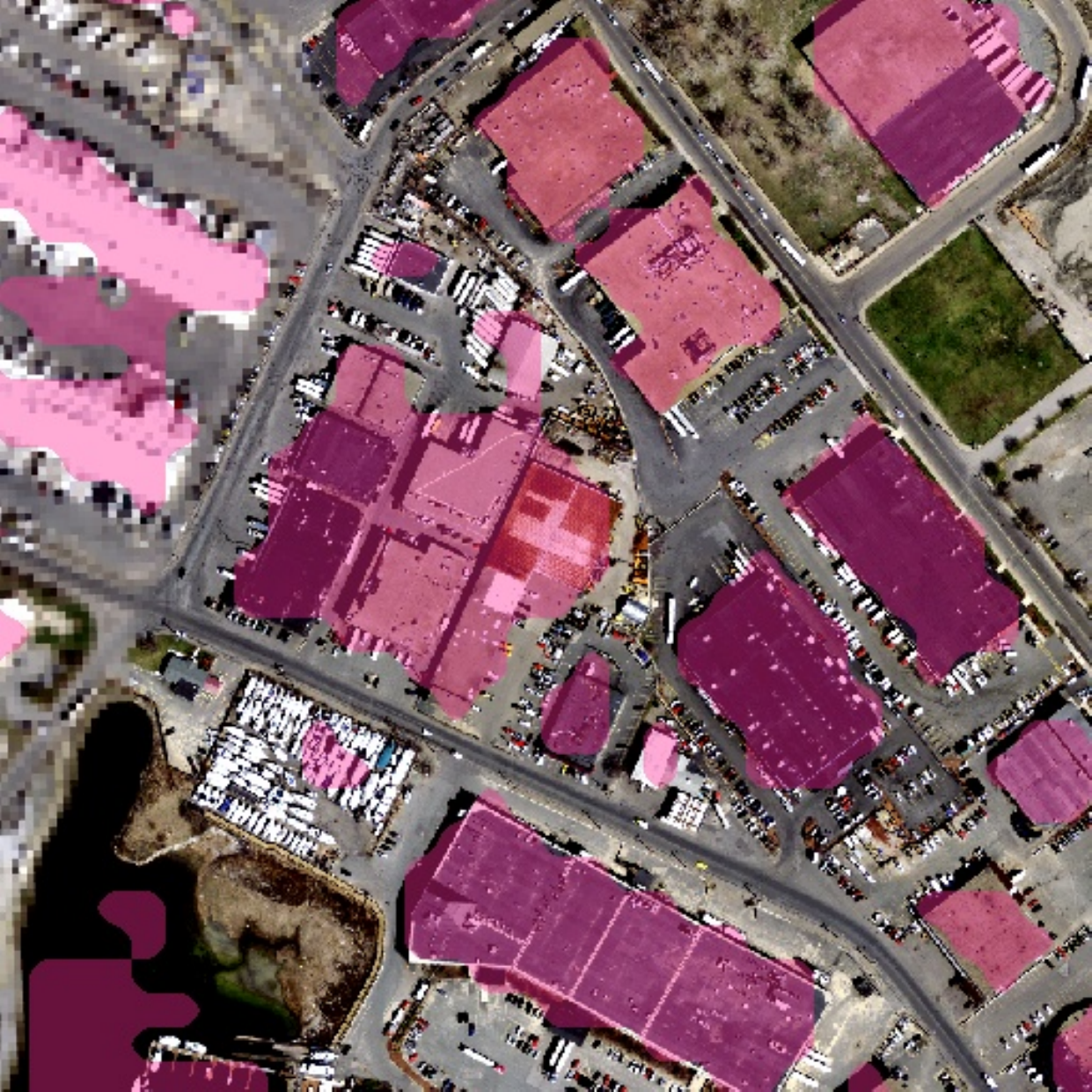}
  & \includegraphics[width=\panw]{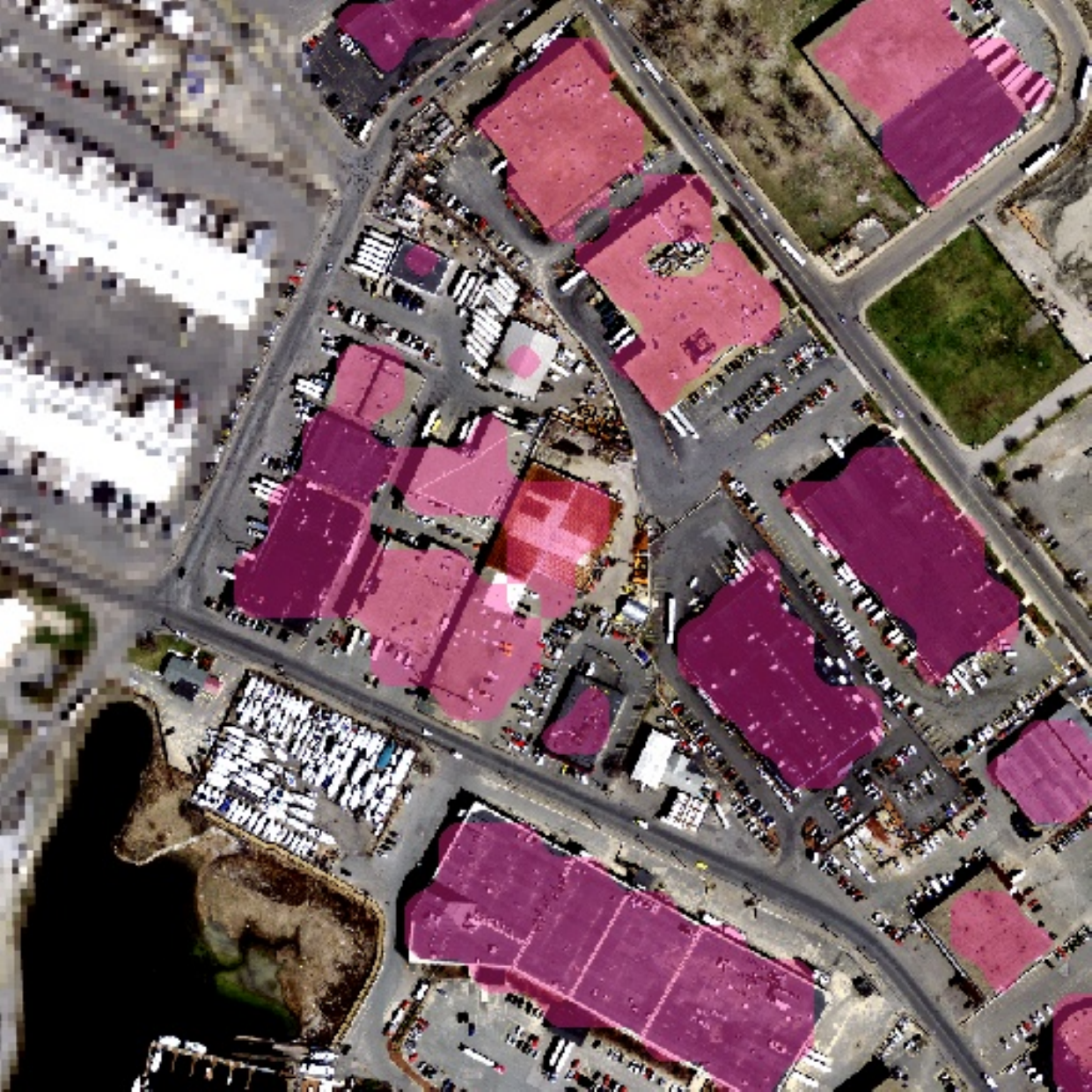}
  & \includegraphics[width=\panw]{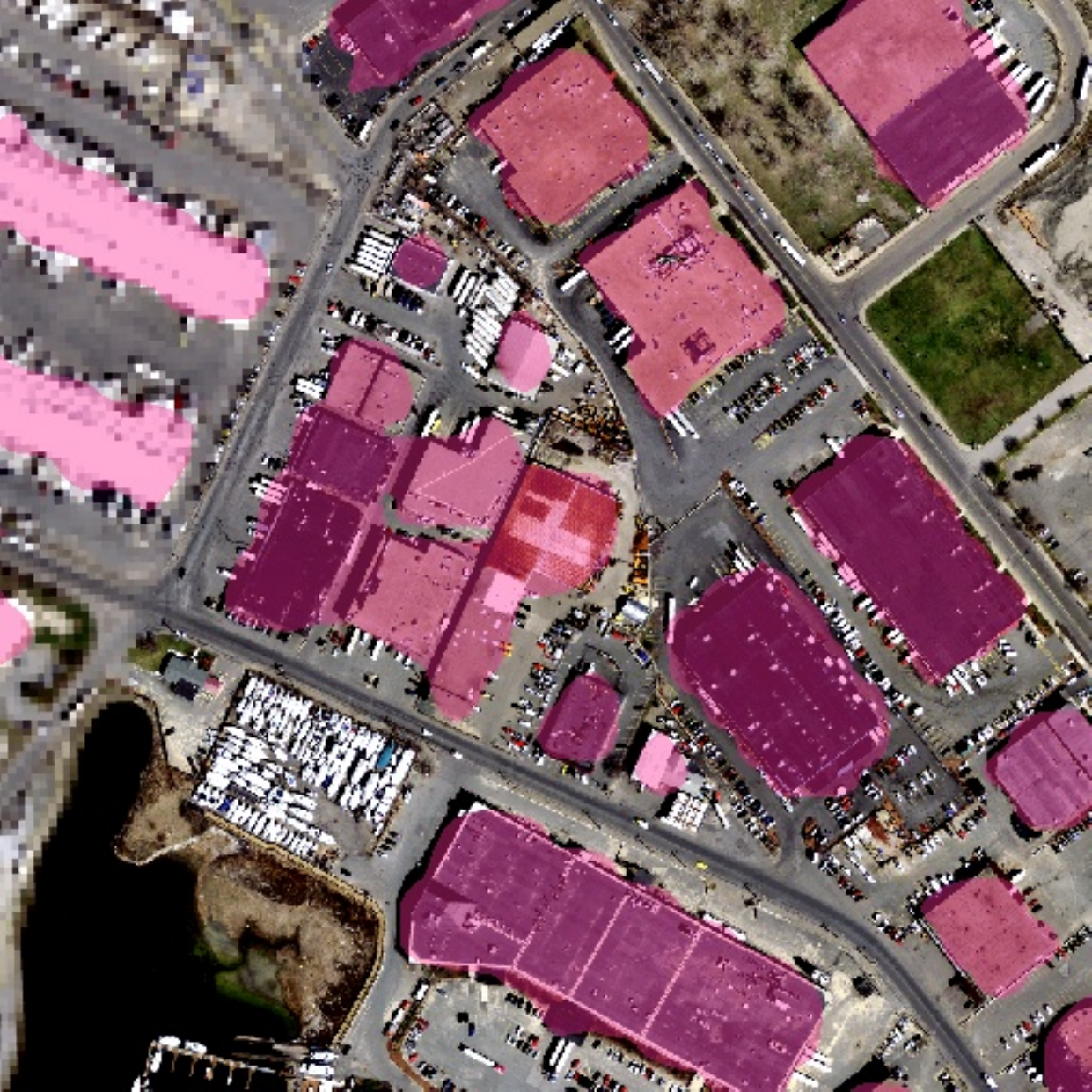}
  & \includegraphics[width=\panw]{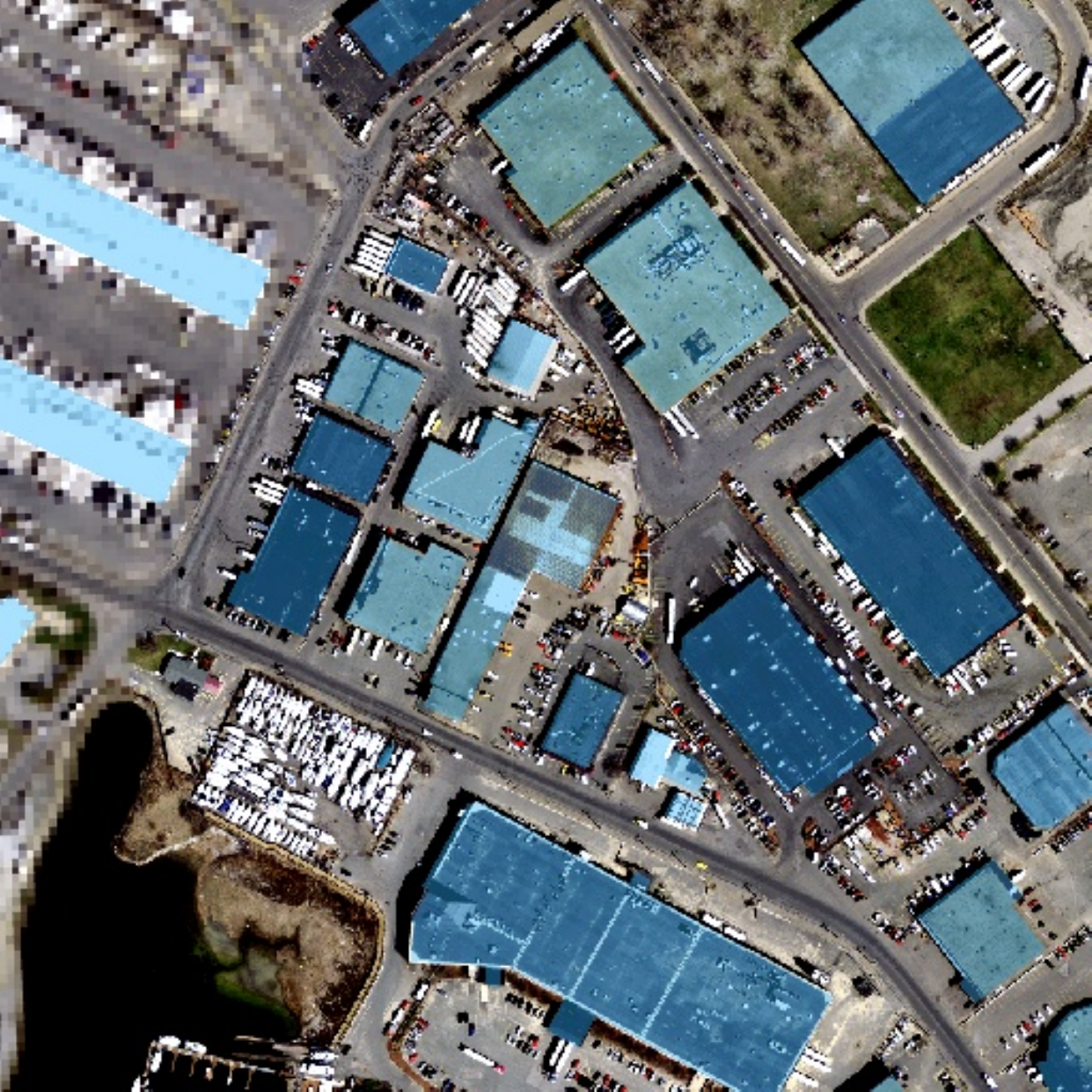} \\
\rotatebox{90}{\scriptsize Building}
  & \includegraphics[width=\panw]{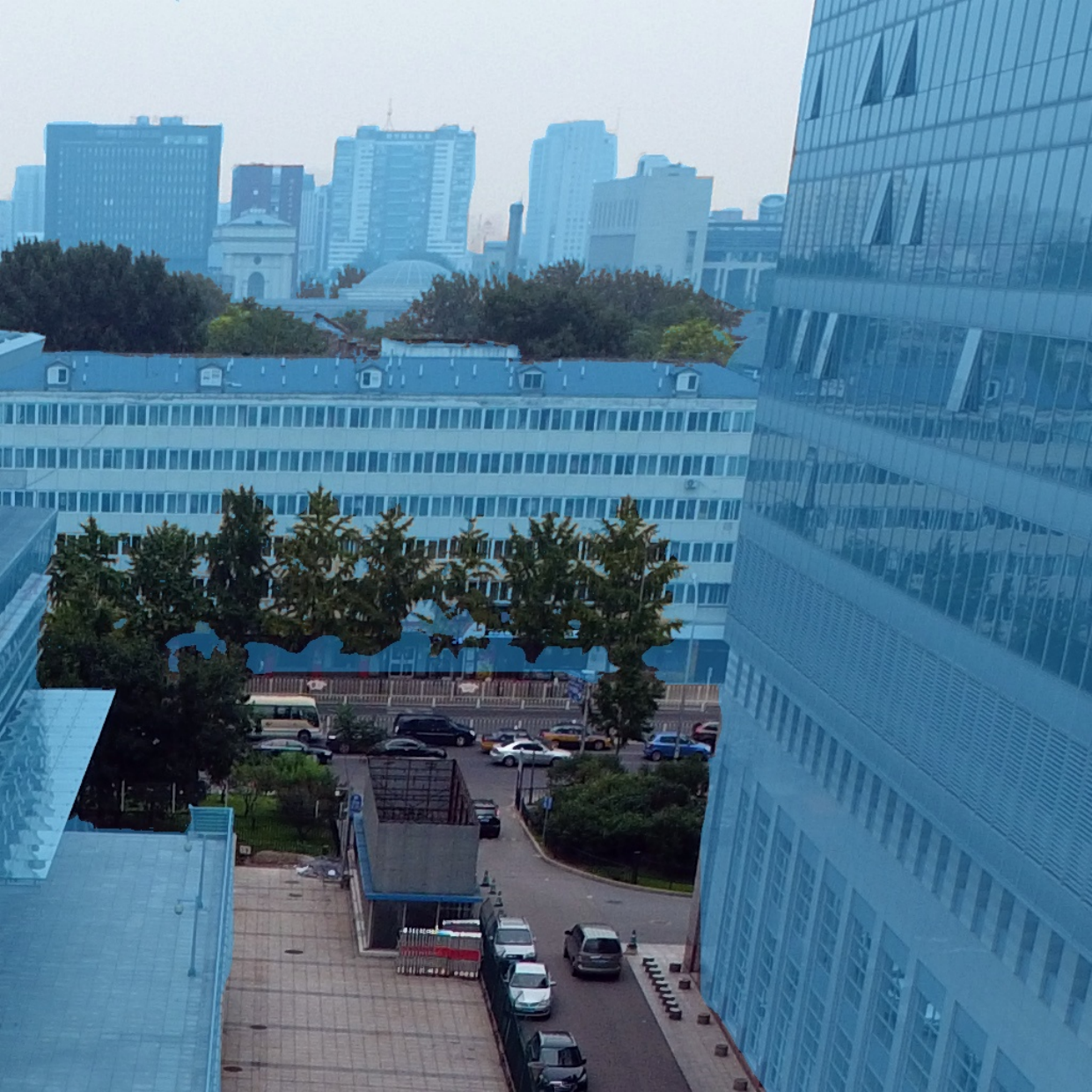}
  & \includegraphics[width=\panw]{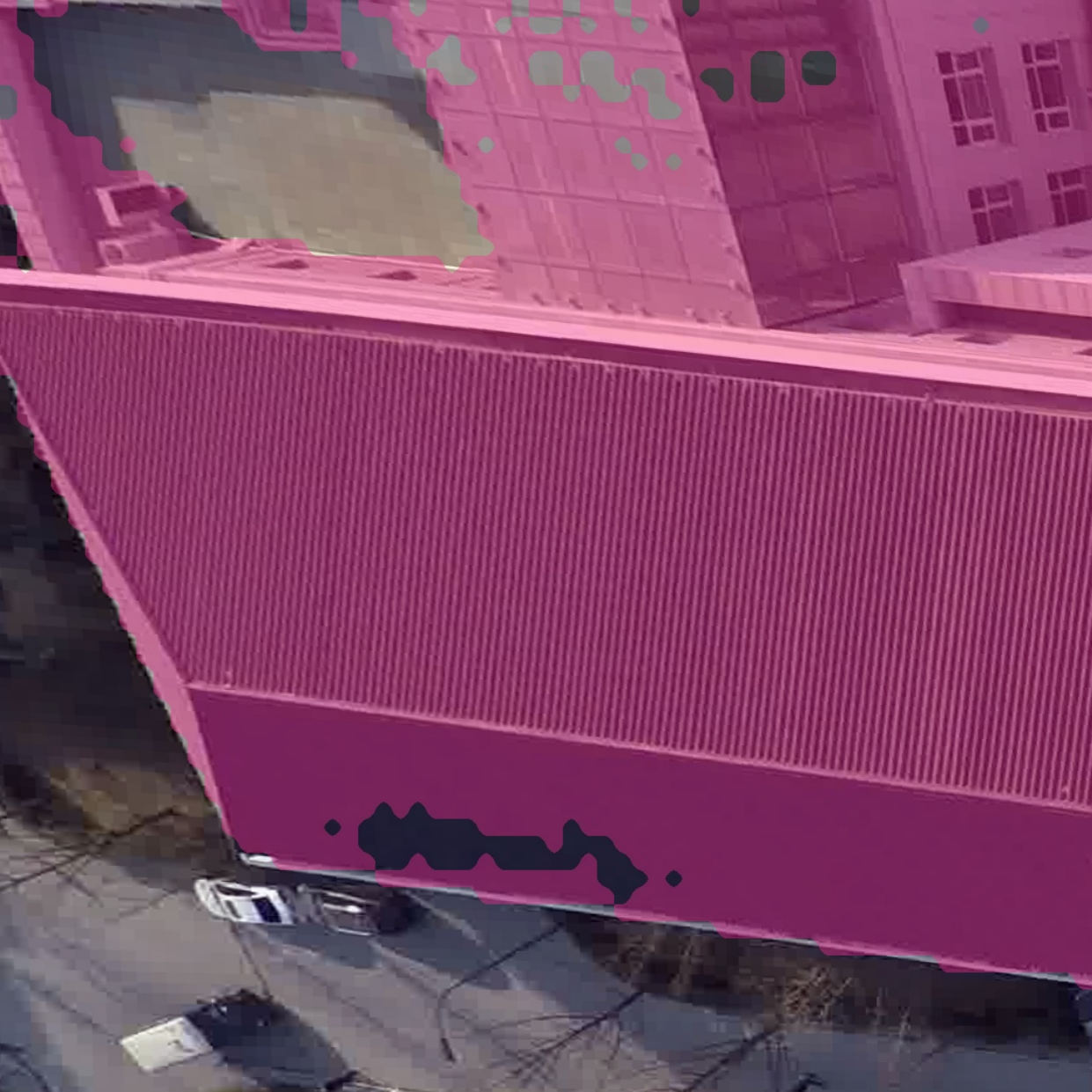}
  & \includegraphics[width=\panw]{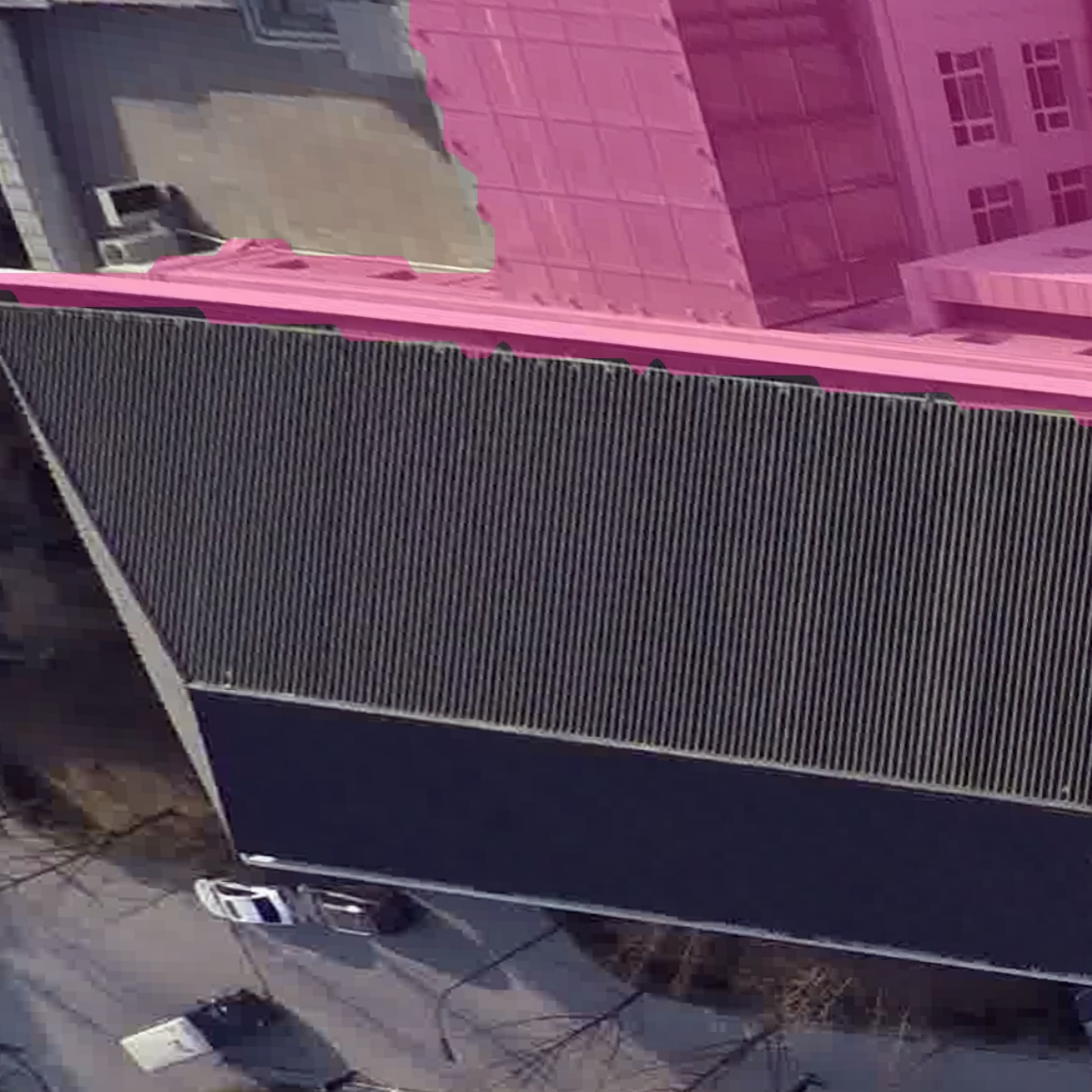}
  & \includegraphics[width=\panw]{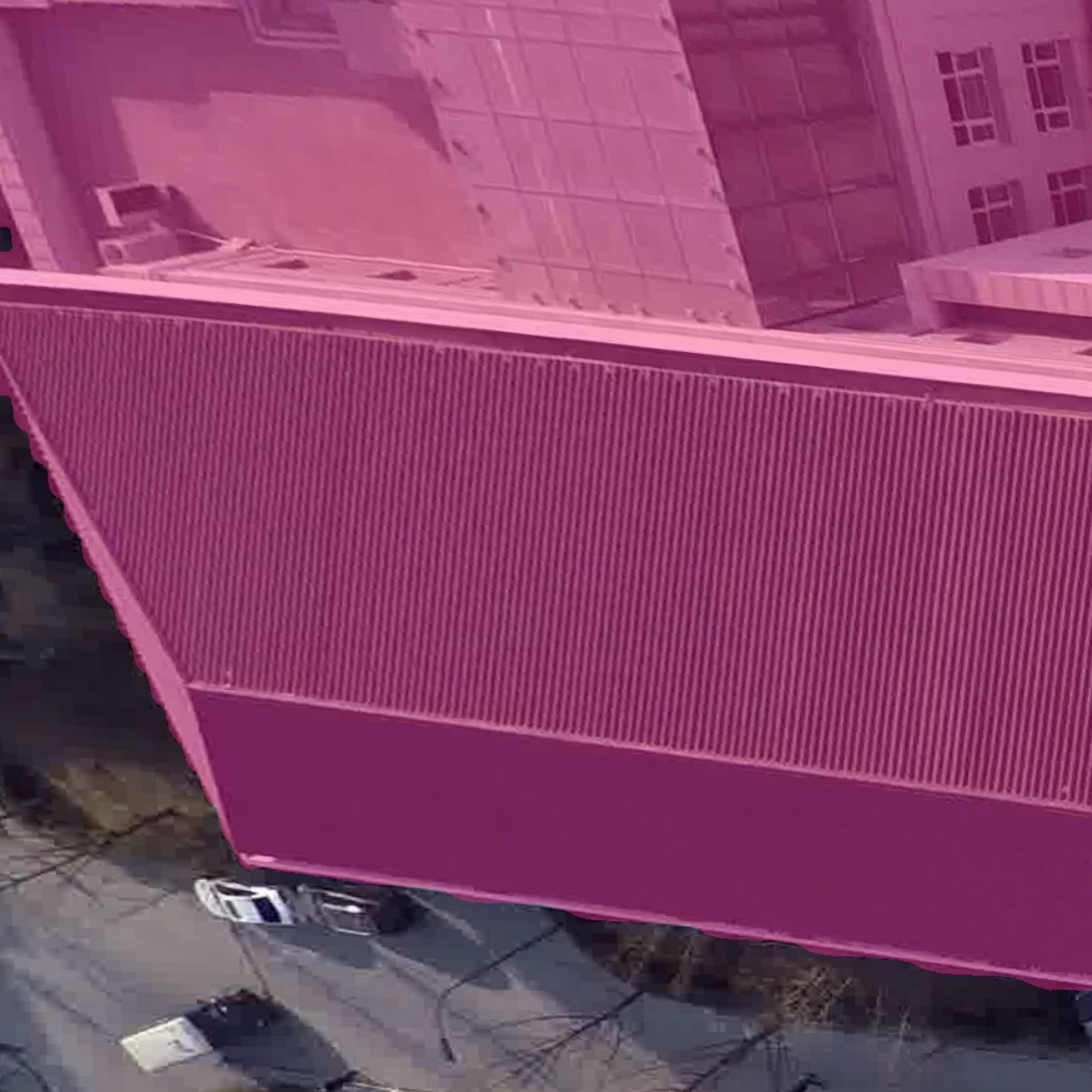}
  & \includegraphics[width=\panw]{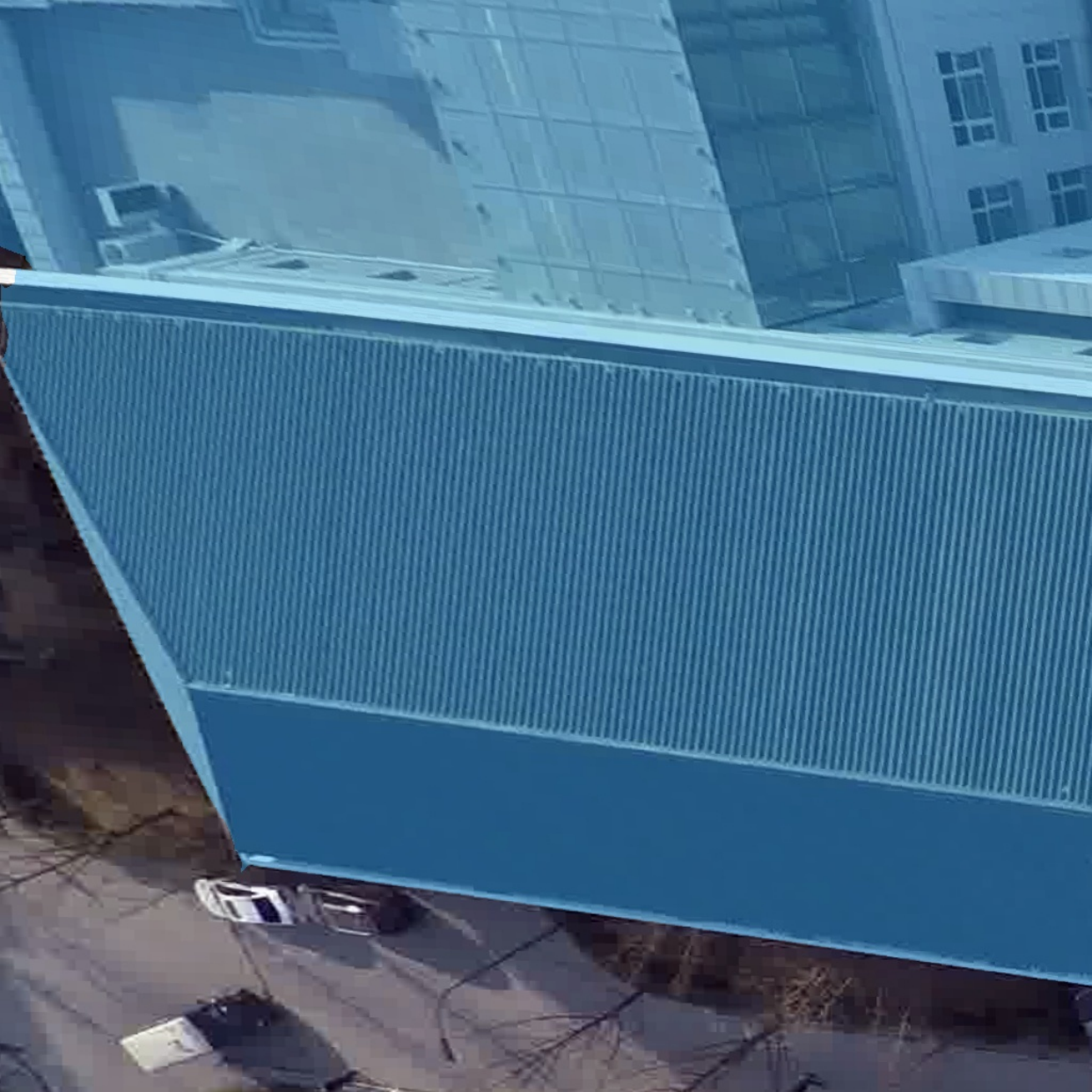} \\
\rotatebox{90}{\scriptsize Agriculture}
  & \includegraphics[width=\panw]{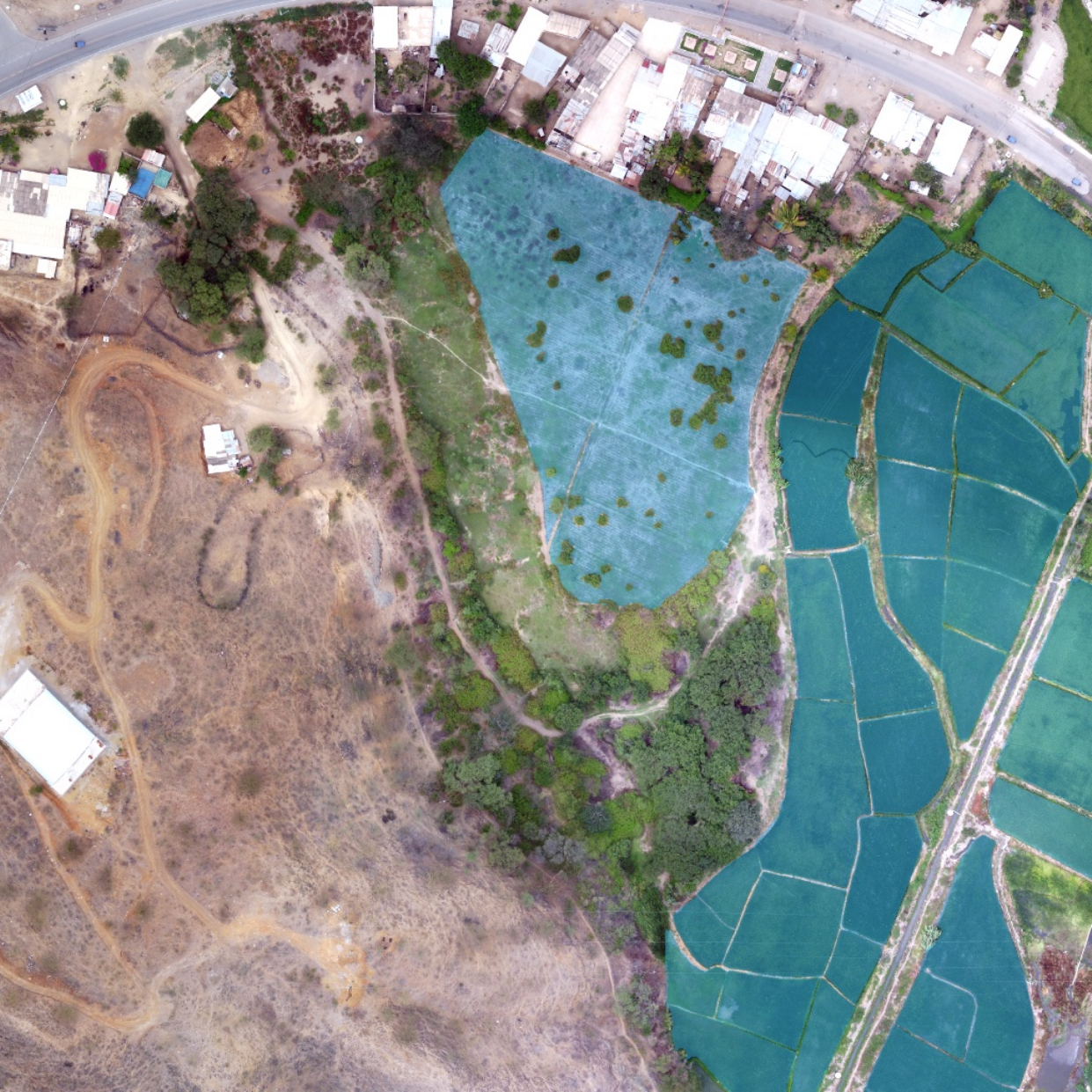}
  & \includegraphics[width=\panw]{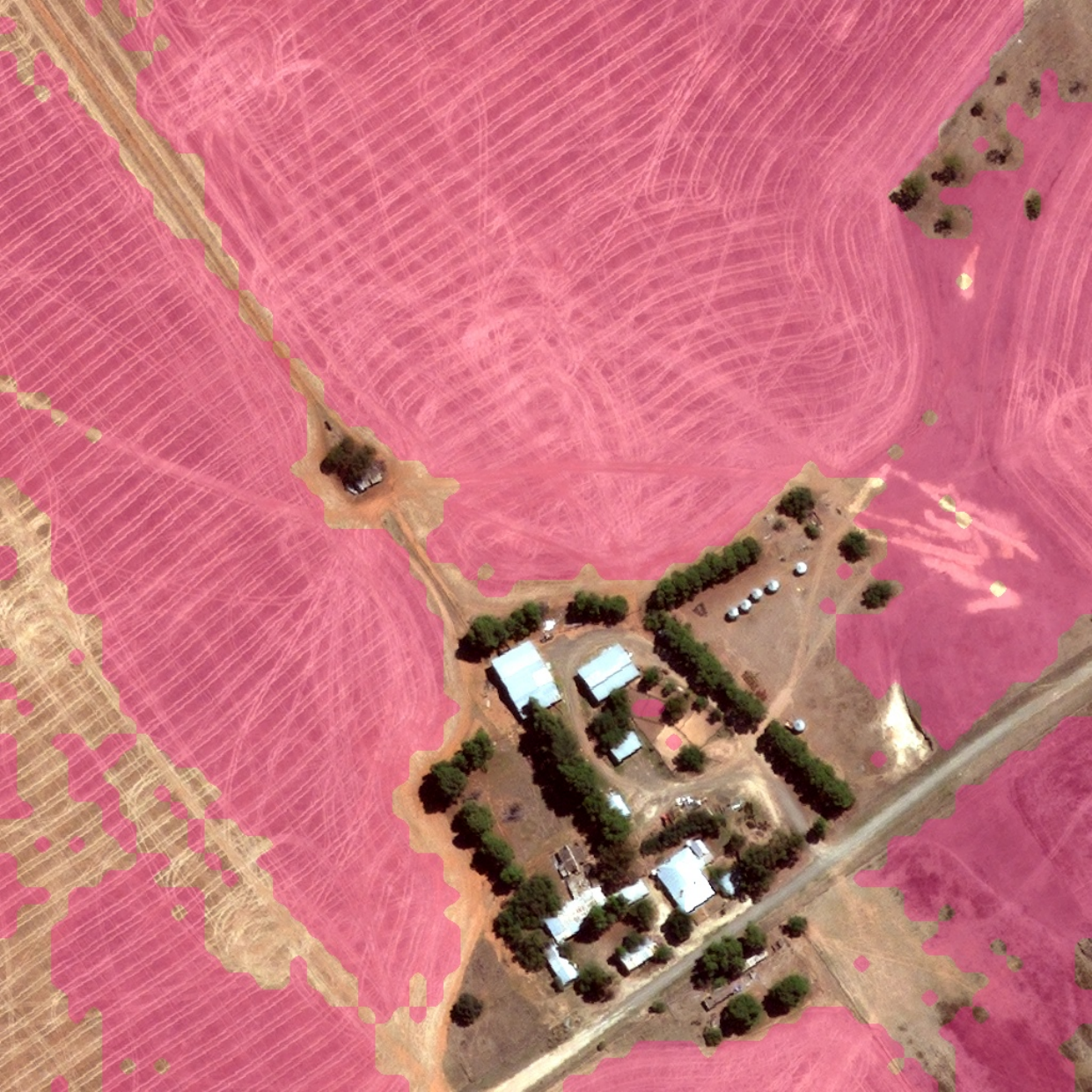}
  & \includegraphics[width=\panw]{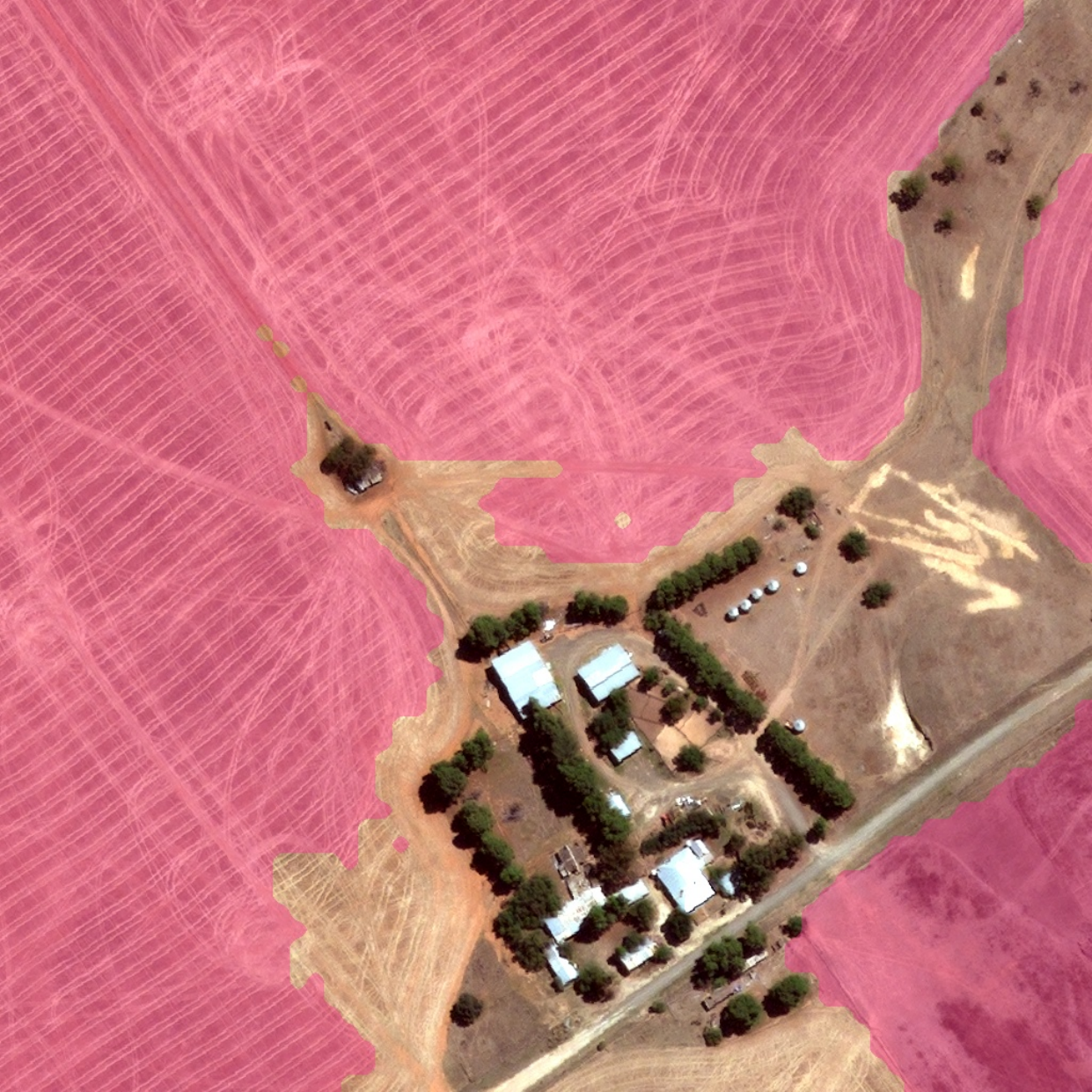}
  & \includegraphics[width=\panw]{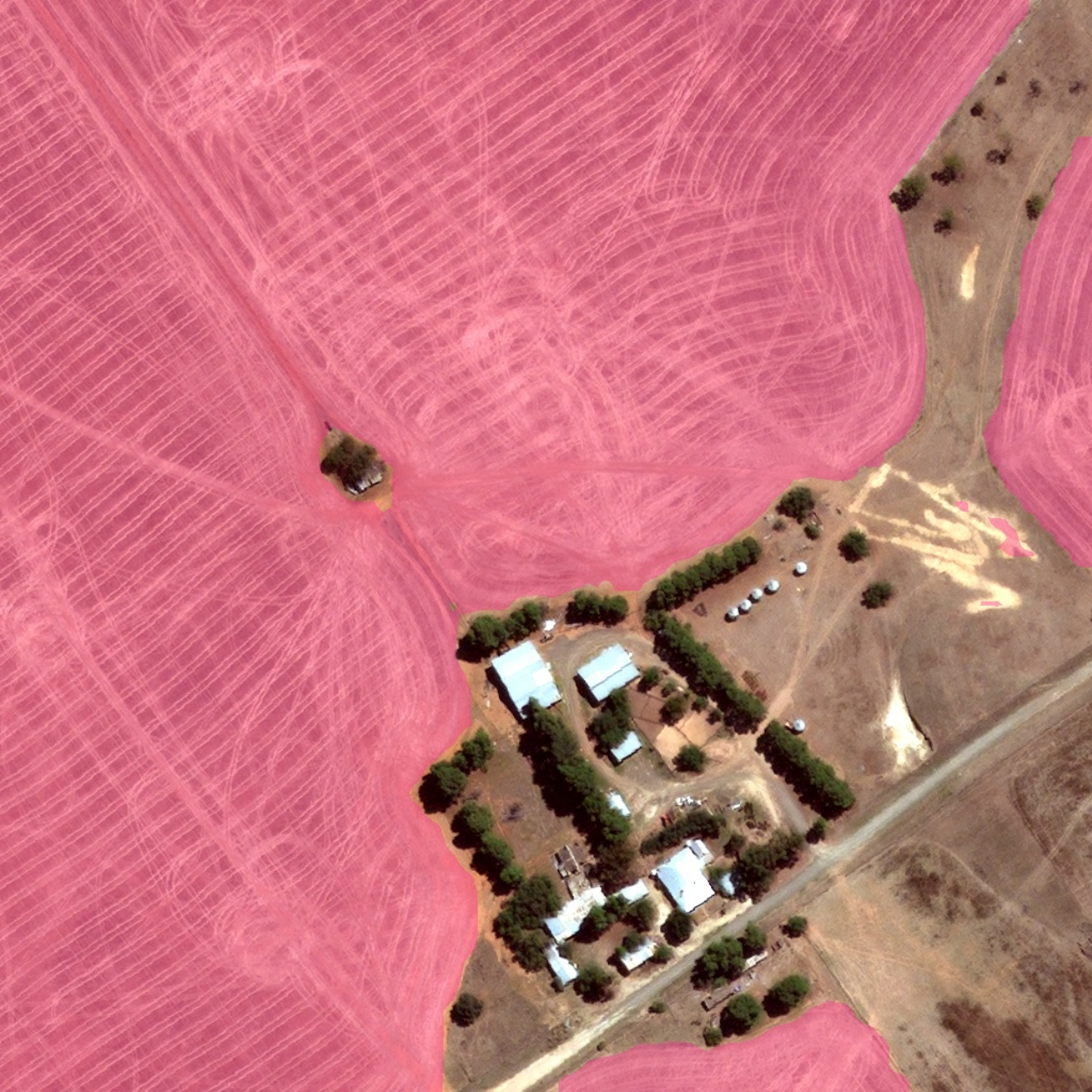}
  & \includegraphics[width=\panw]{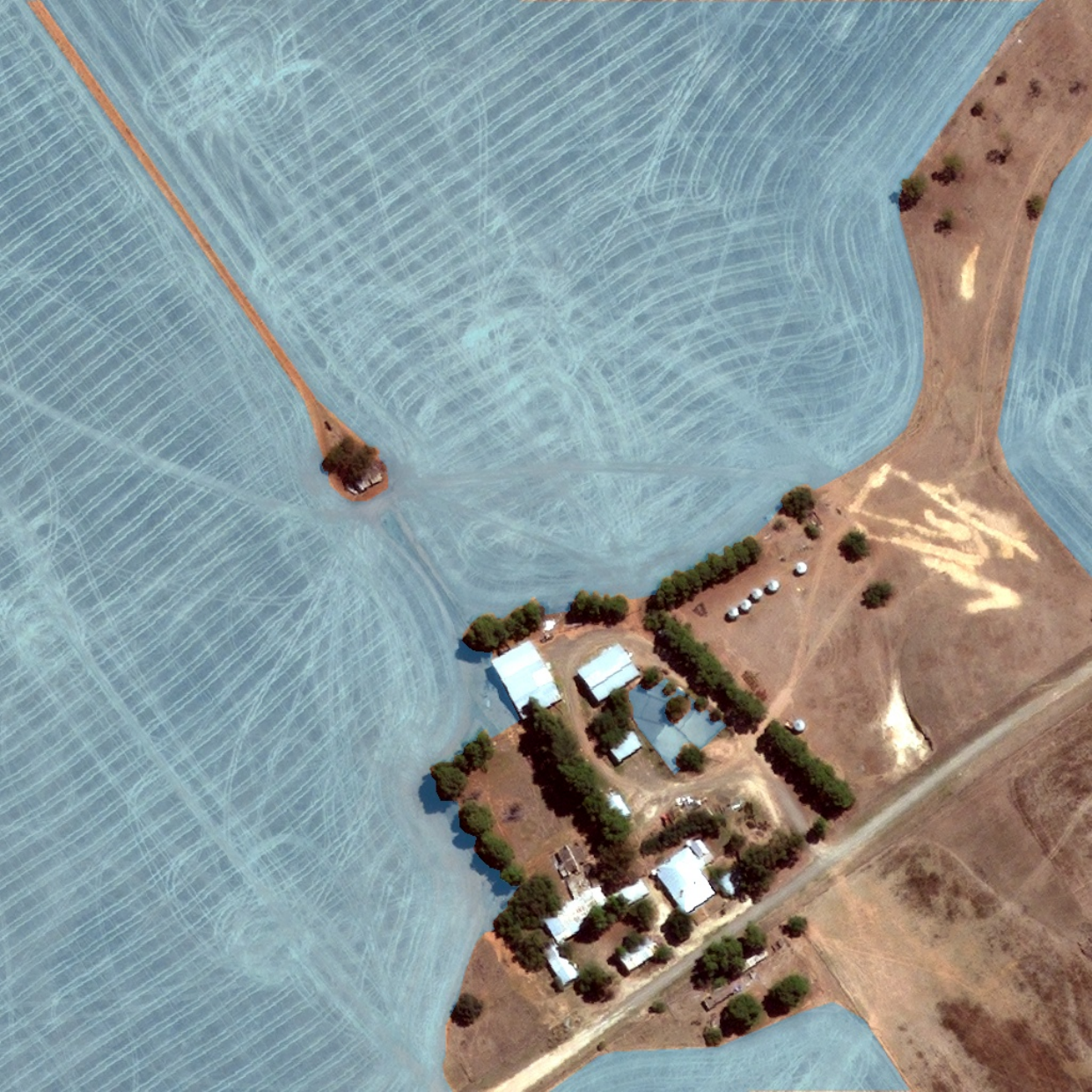} \\

\end{tabular}
&
% ========== 오른쪽 그룹: moving car / plane / ship ==========
\begin{tabular}{@{}m{1em} *{5}{>{\centering\arraybackslash}m{\panw}}@{}}
 & \scriptsize Reference & \scriptsize FSSDINO & \scriptsize INSID3 & \scriptsize FROST & \scriptsize GT \\
\rotatebox{90}{\scriptsize Moving car}
  & \includegraphics[width=\panw]{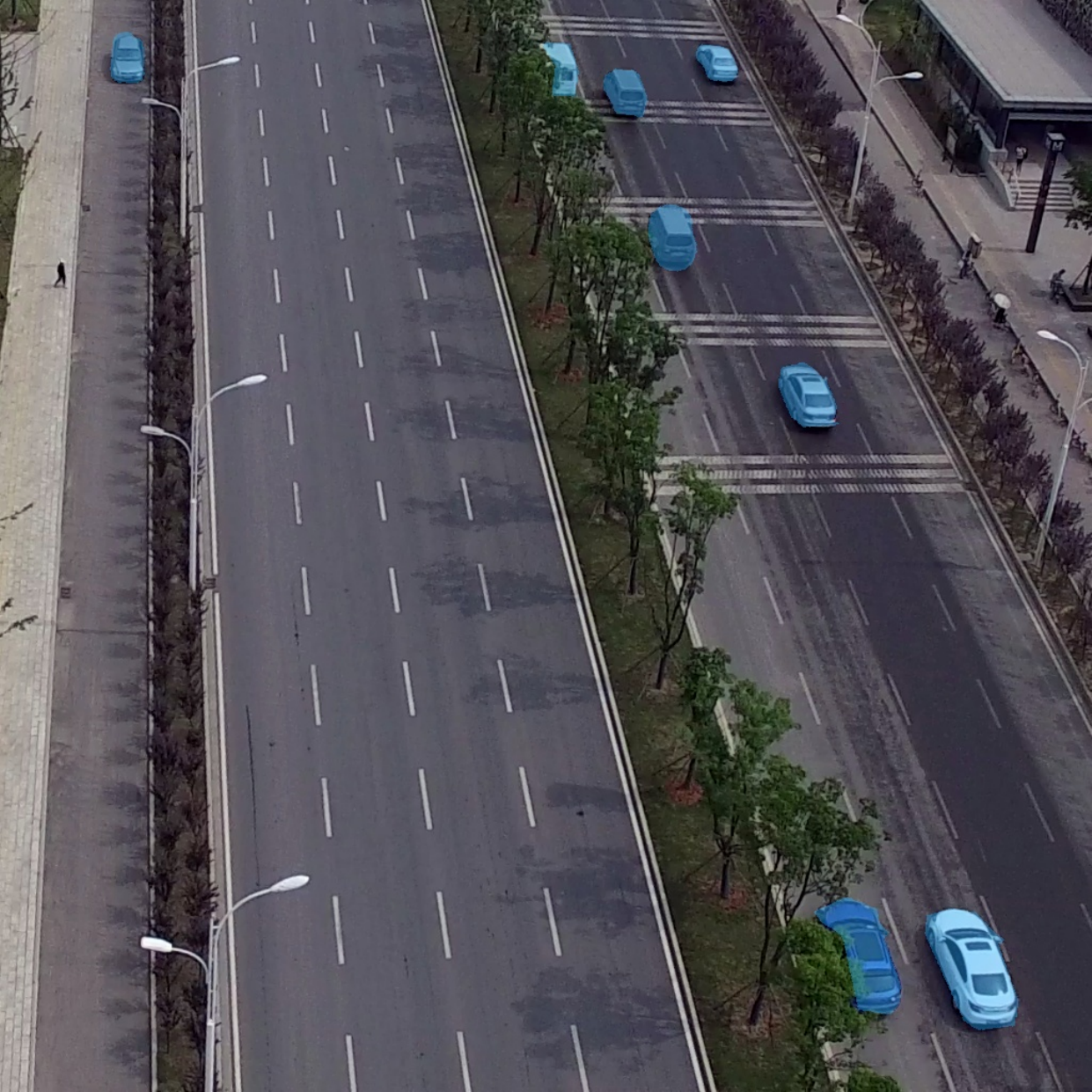}
  & \includegraphics[width=\panw]{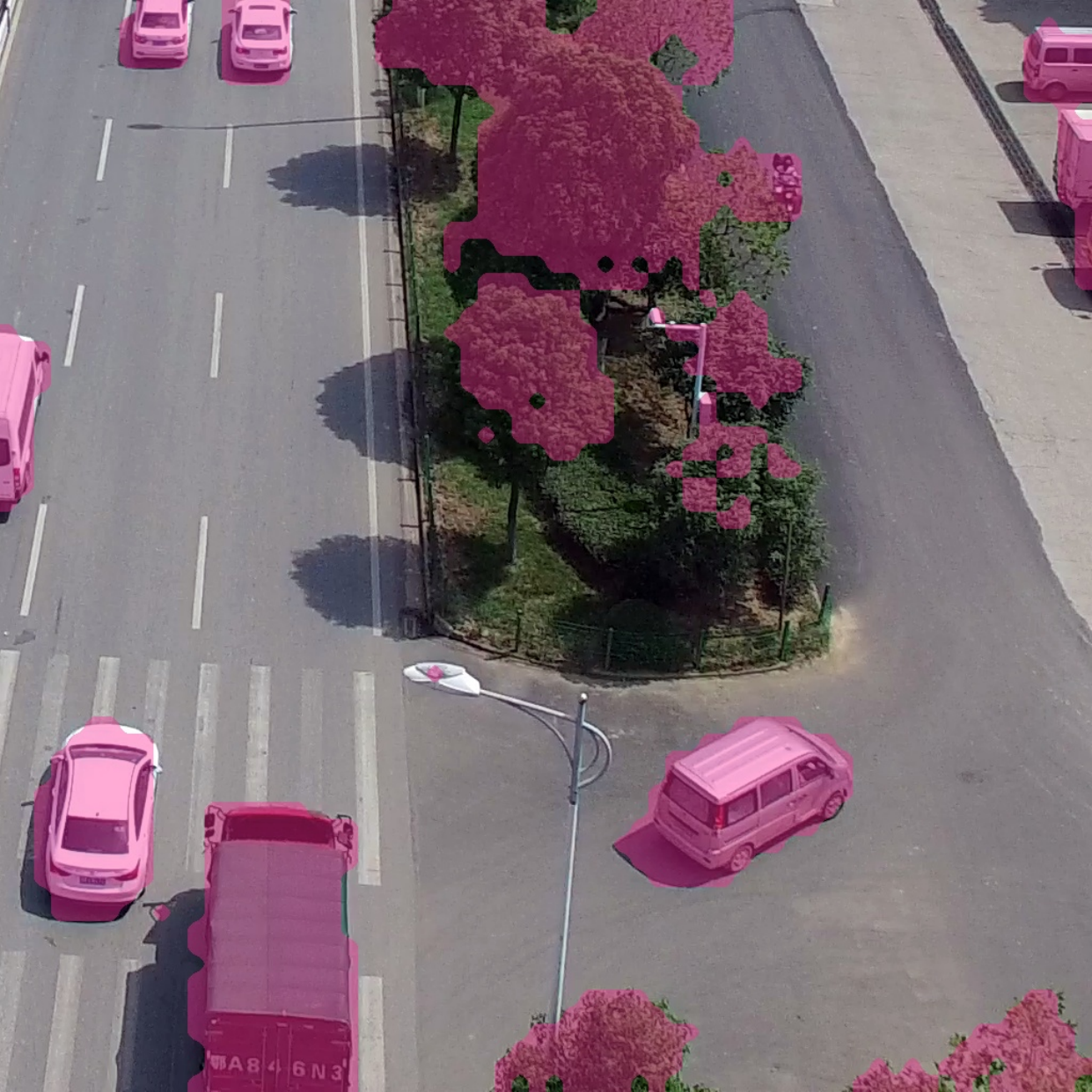}
  & \includegraphics[width=\panw]{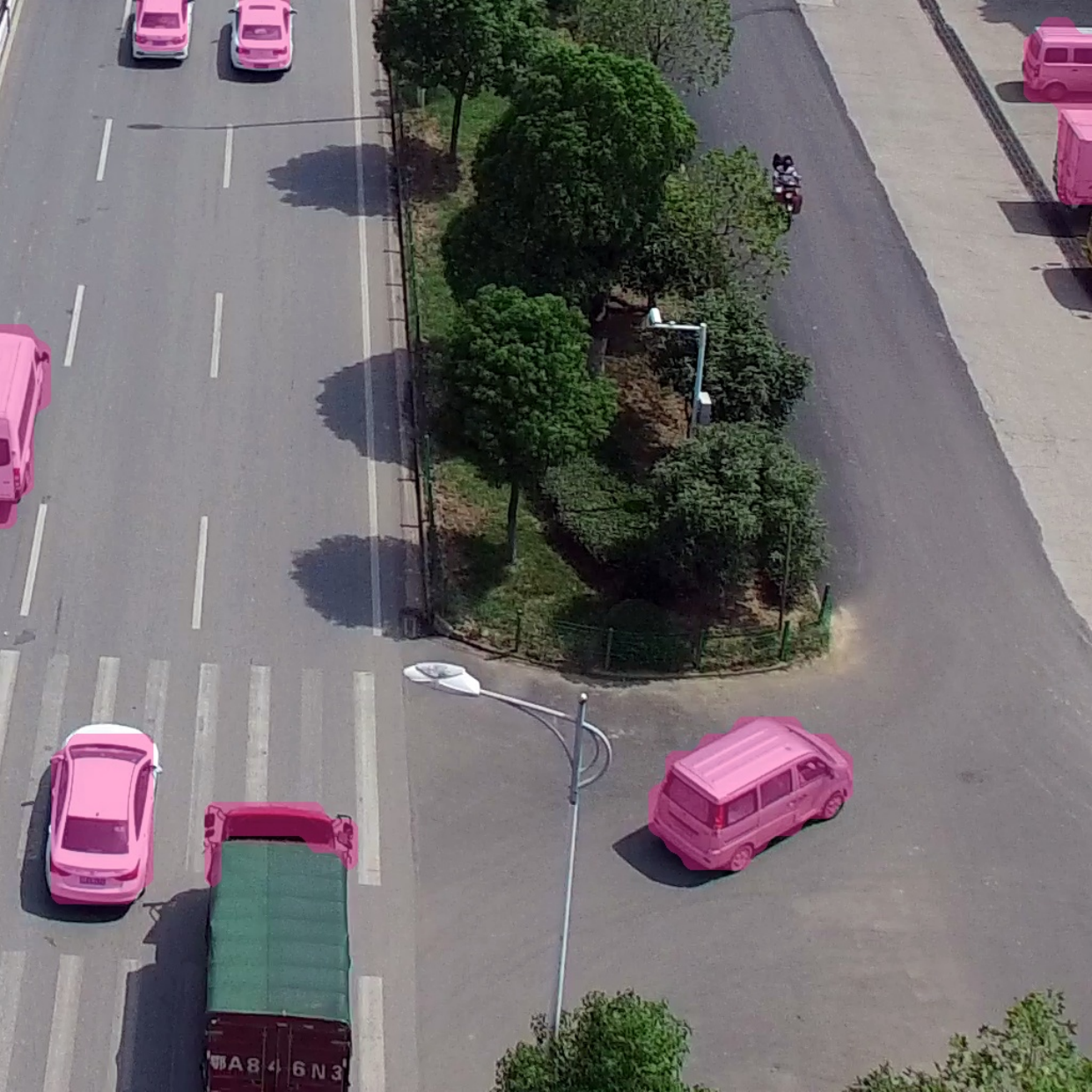}
  & \includegraphics[width=\panw]{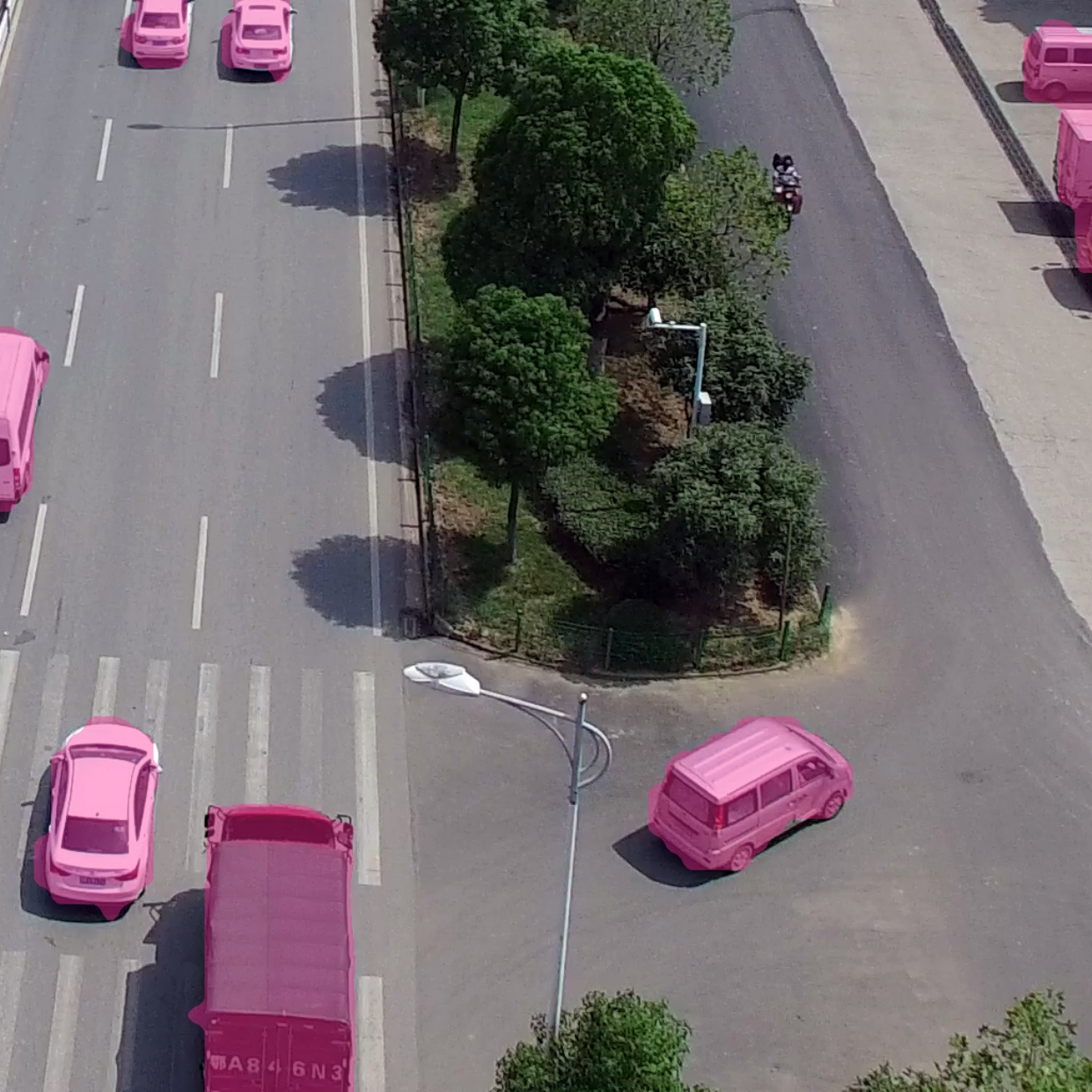}
  & \includegraphics[width=\panw]{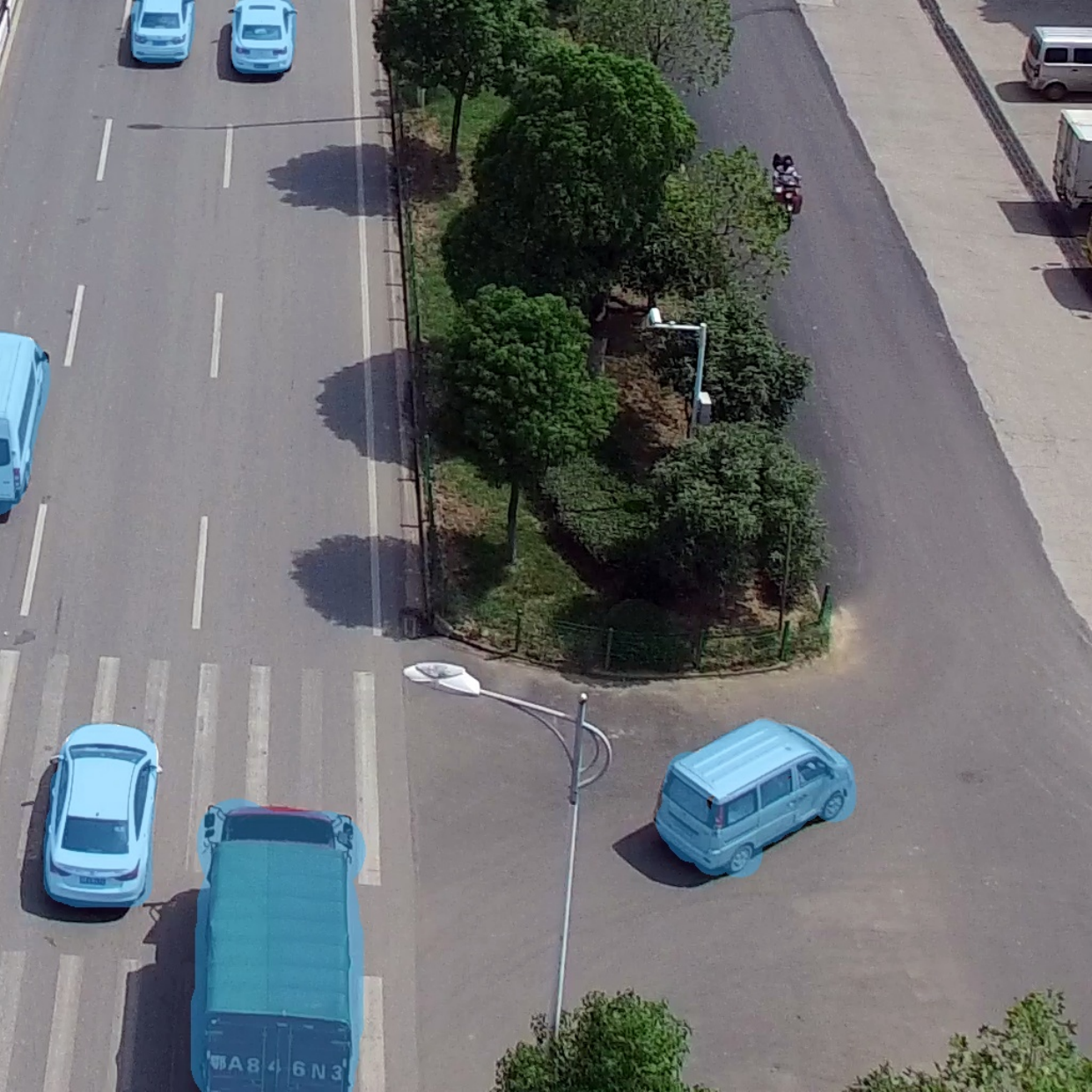} \\
\rotatebox{90}{\scriptsize Plane}
  & \includegraphics[width=\panw]{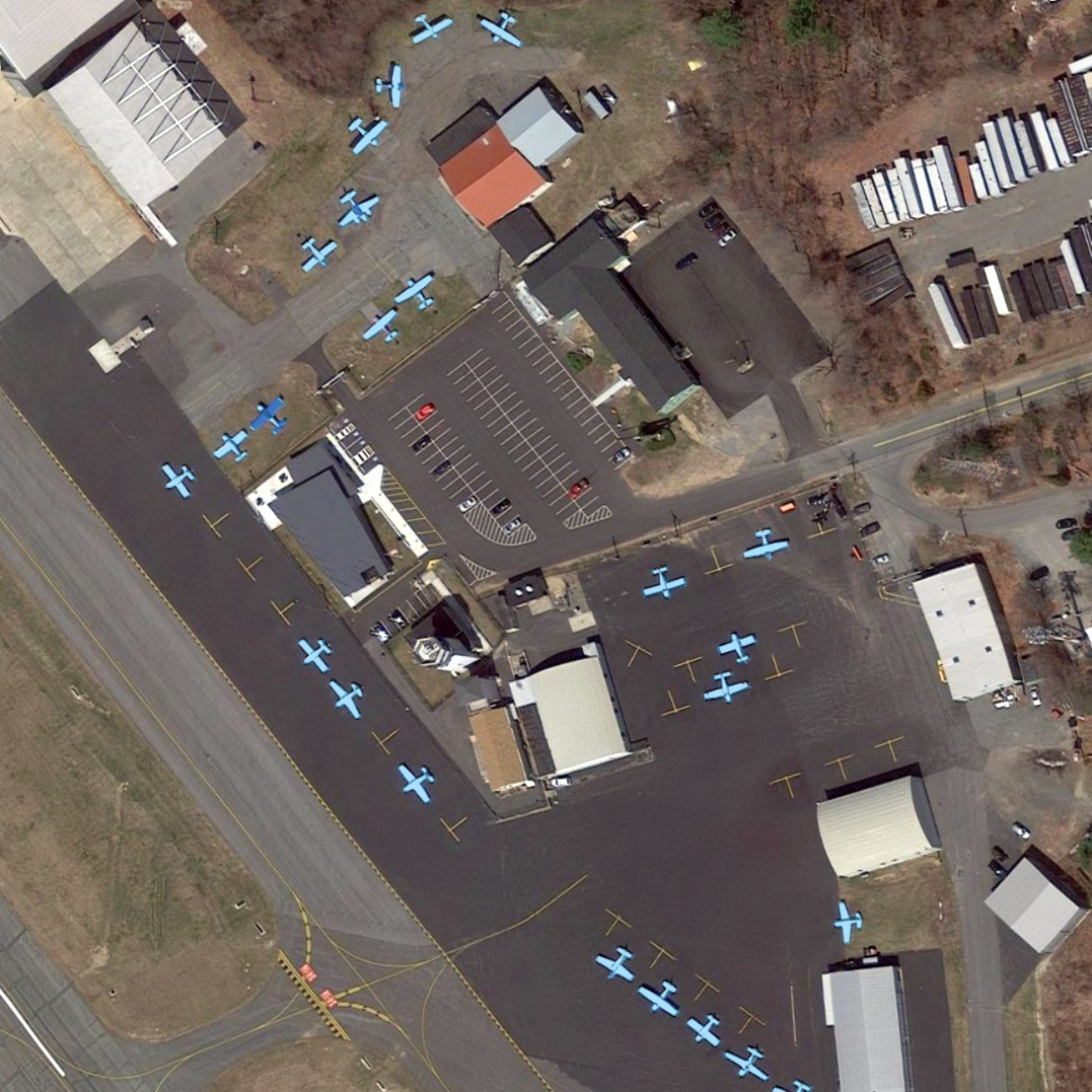}
  & \includegraphics[width=\panw]{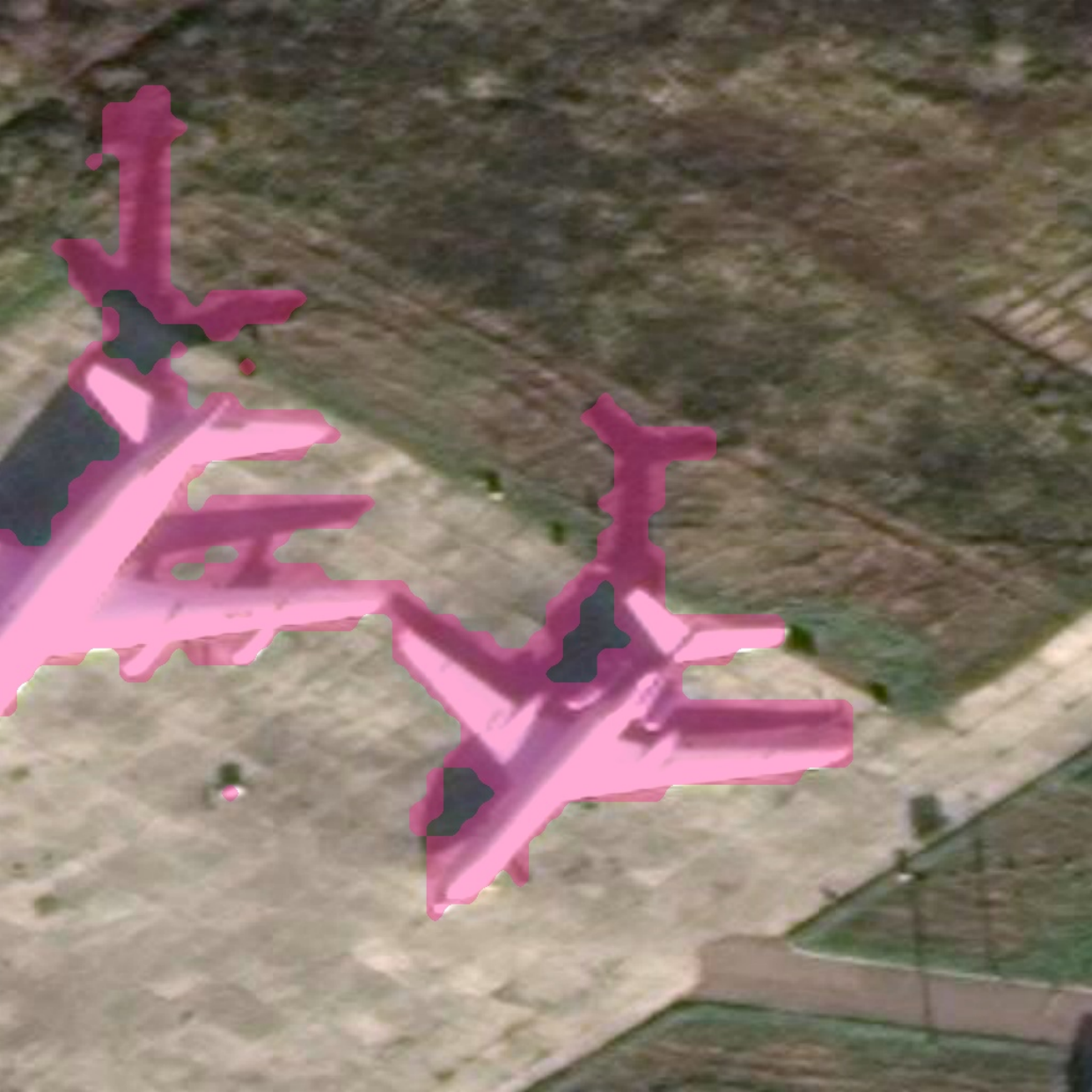}
  & \includegraphics[width=\panw]{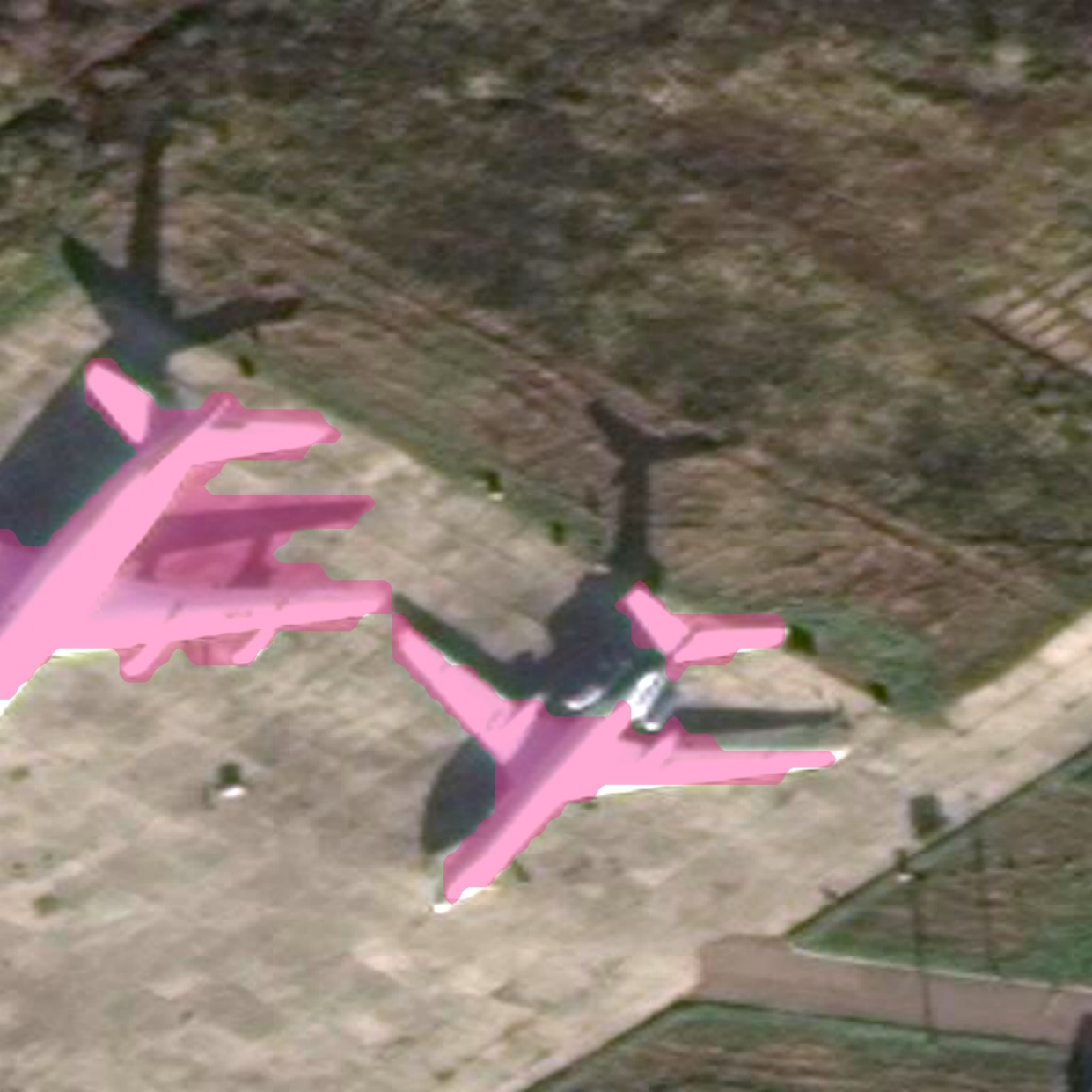}
  & \includegraphics[width=\panw]{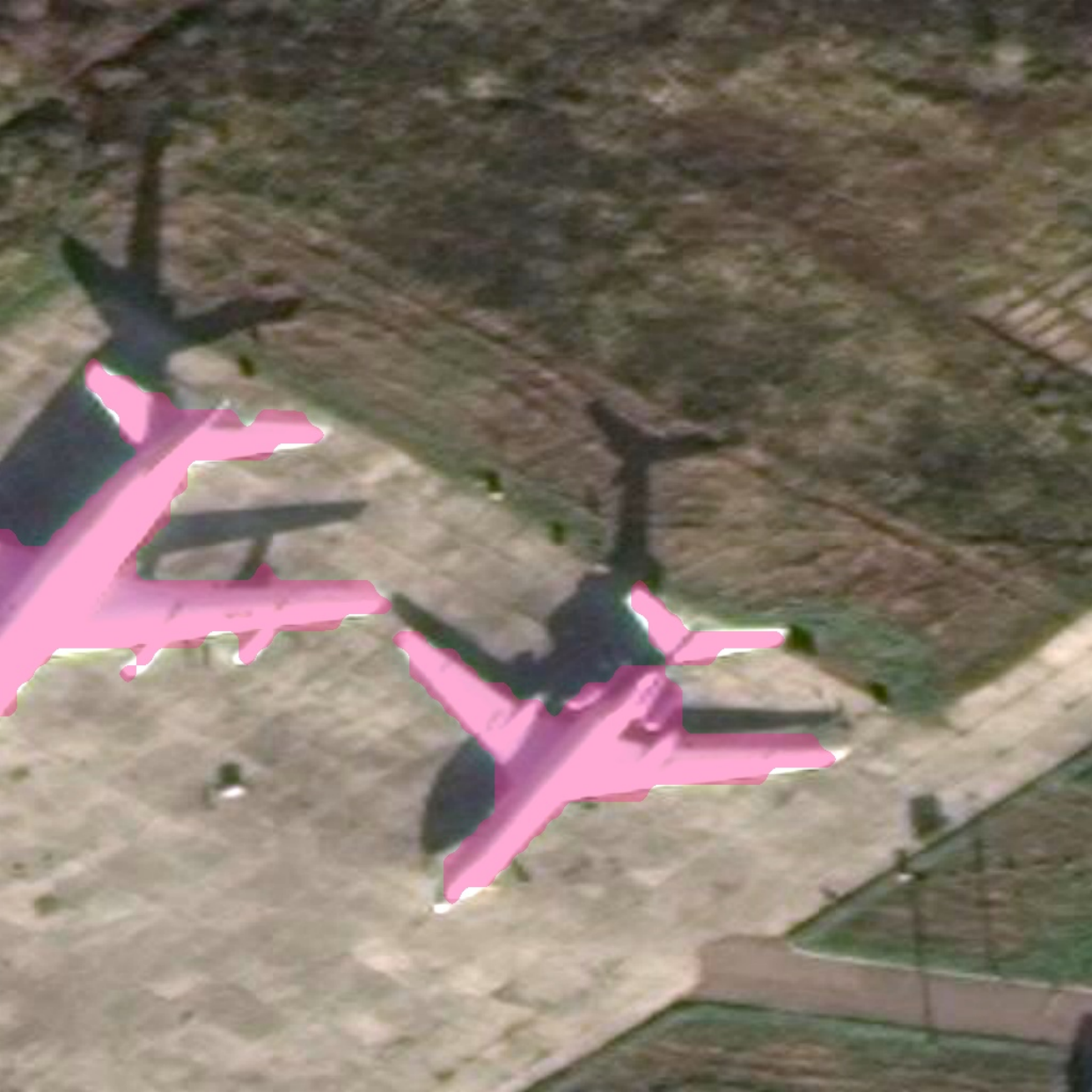}
  & \includegraphics[width=\panw]{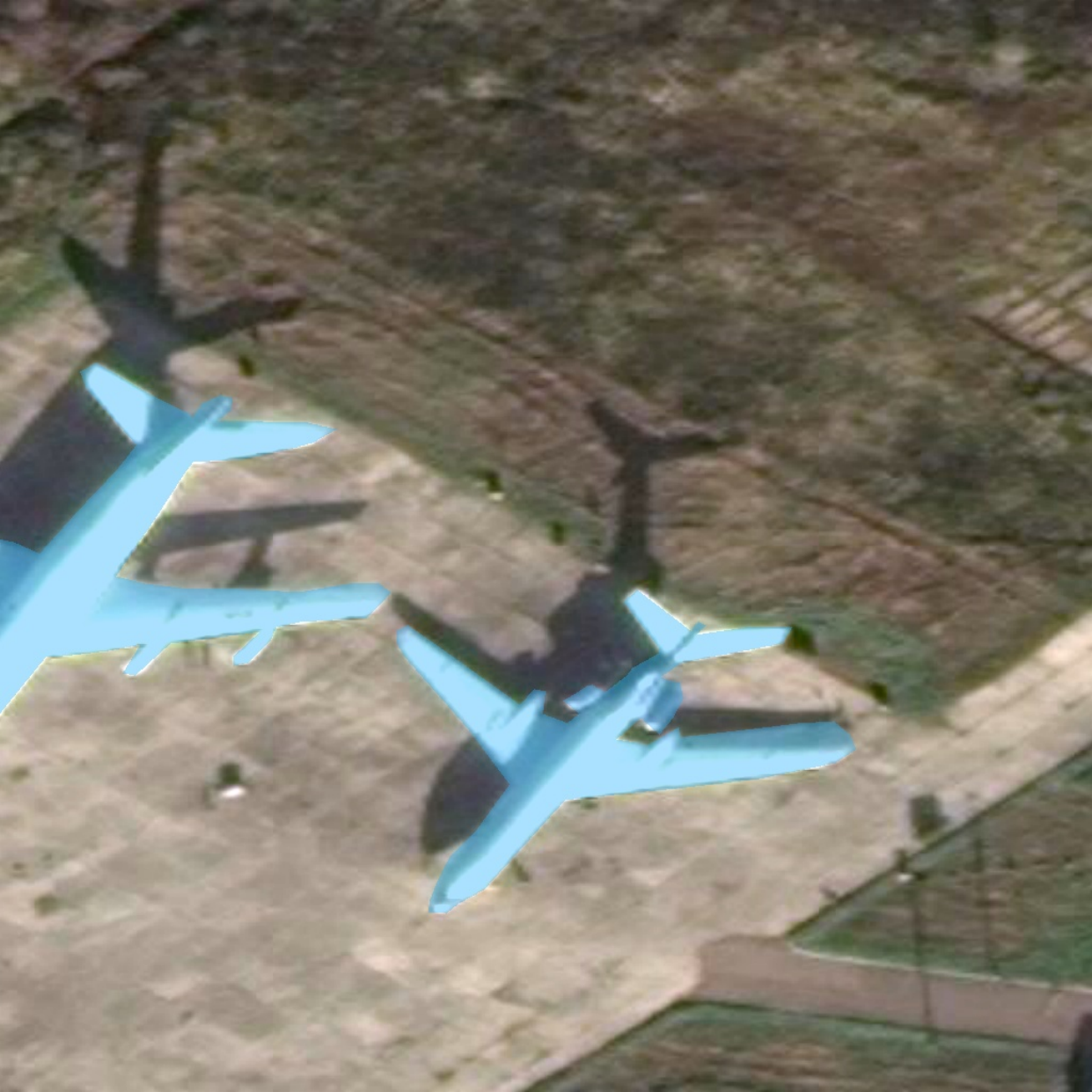} \\
\rotatebox{90}{\scriptsize Ship}
  & \includegraphics[width=\panw]{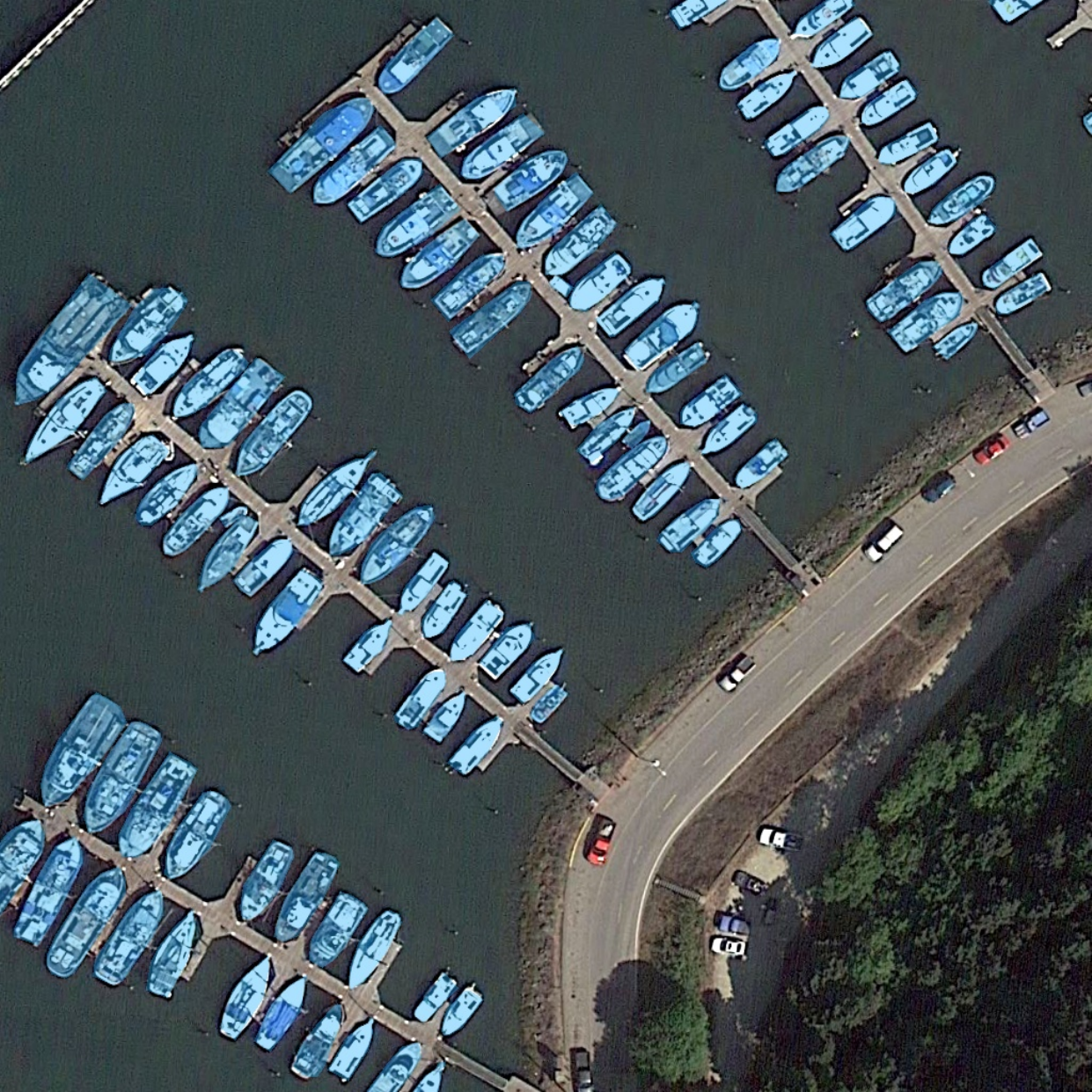}
  & \includegraphics[width=\panw]{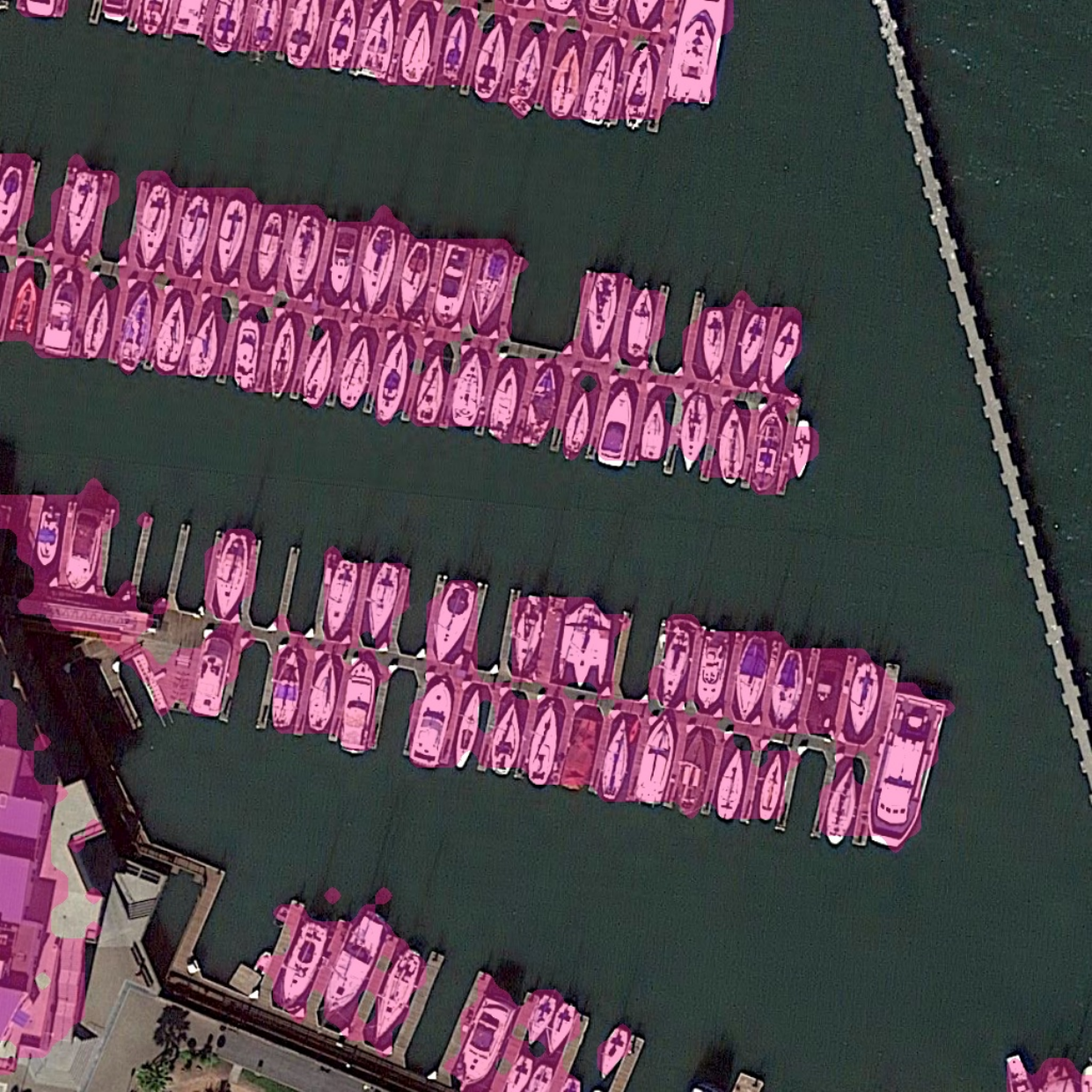}
  & \includegraphics[width=\panw]{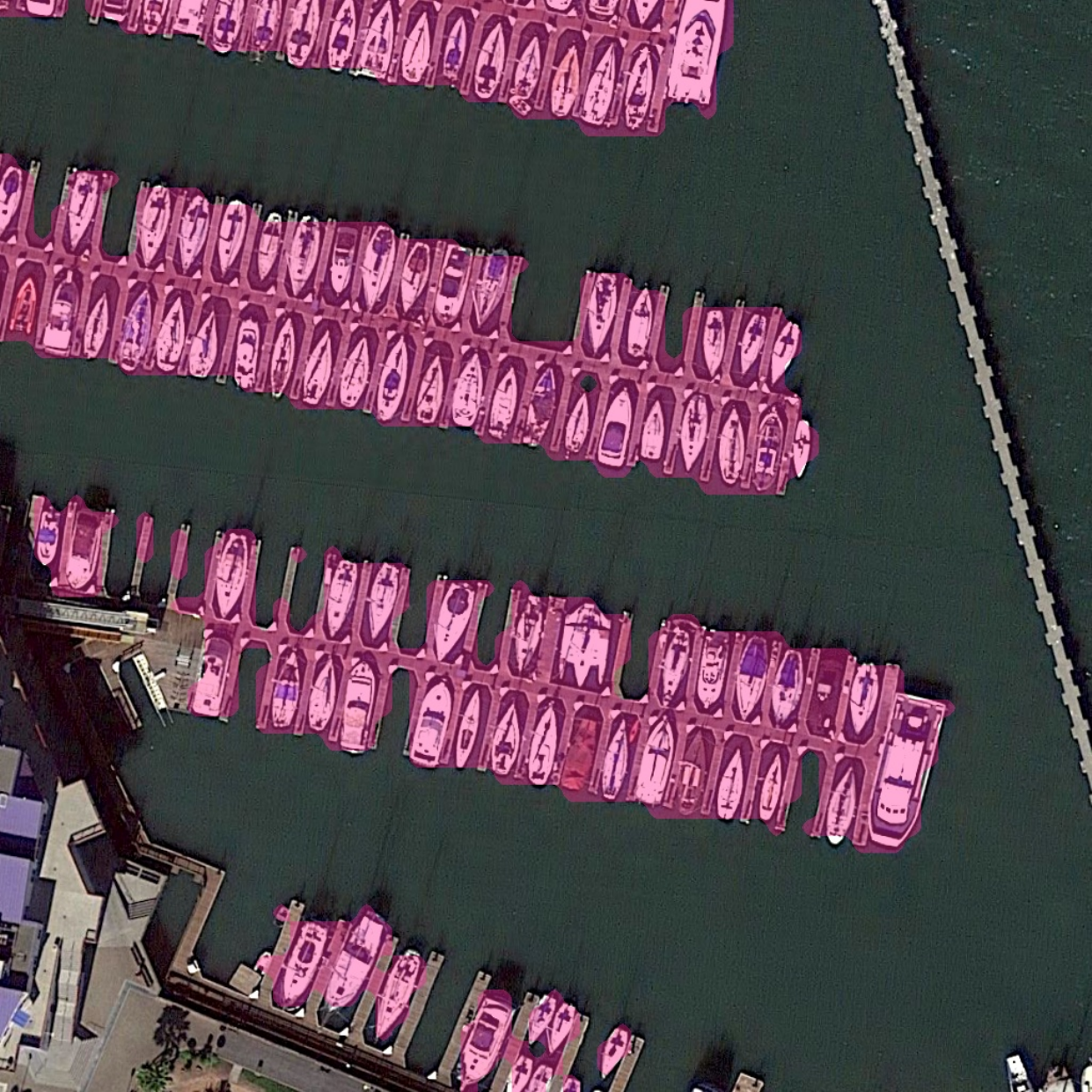}
  & \includegraphics[width=\panw]{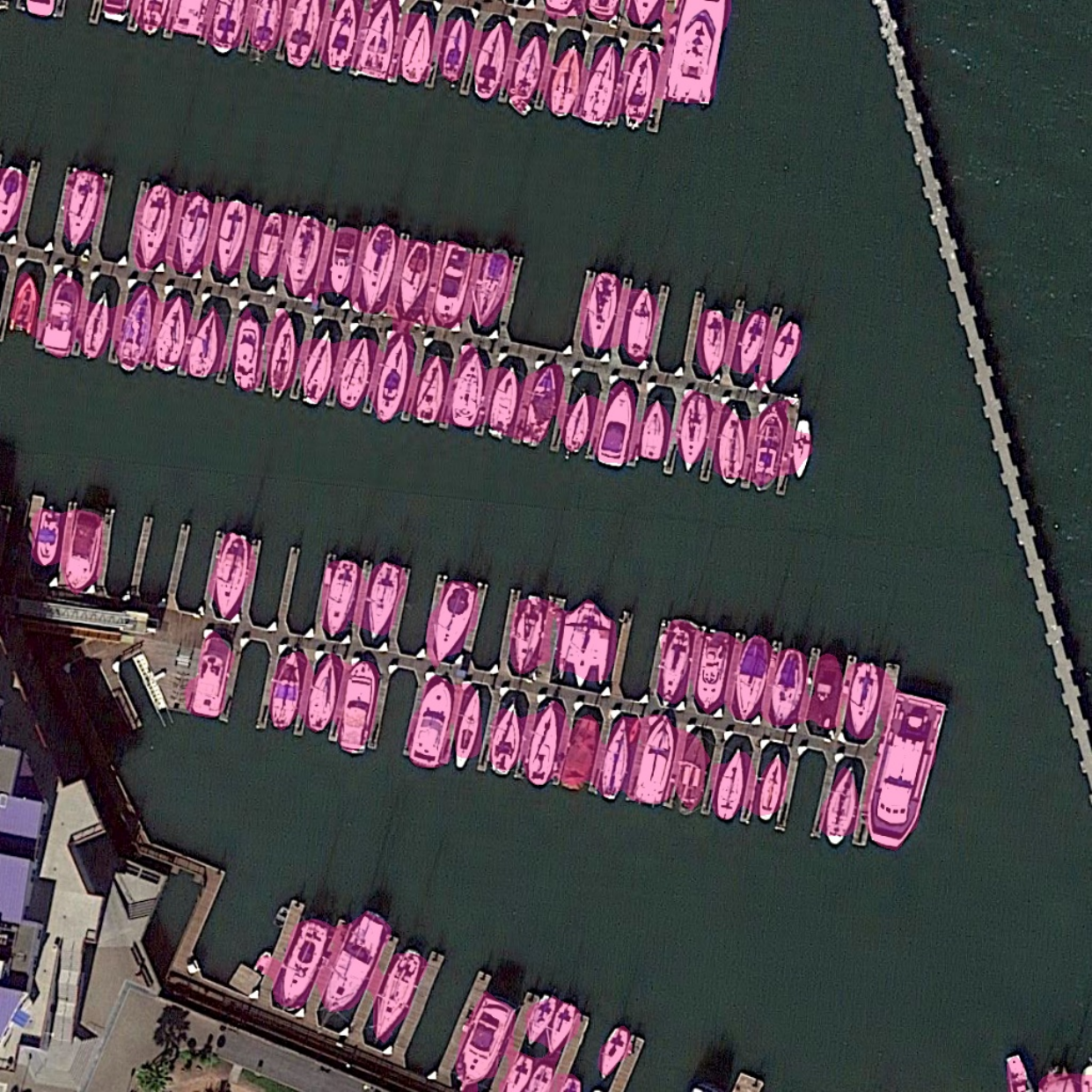}
  & \includegraphics[width=\panw]{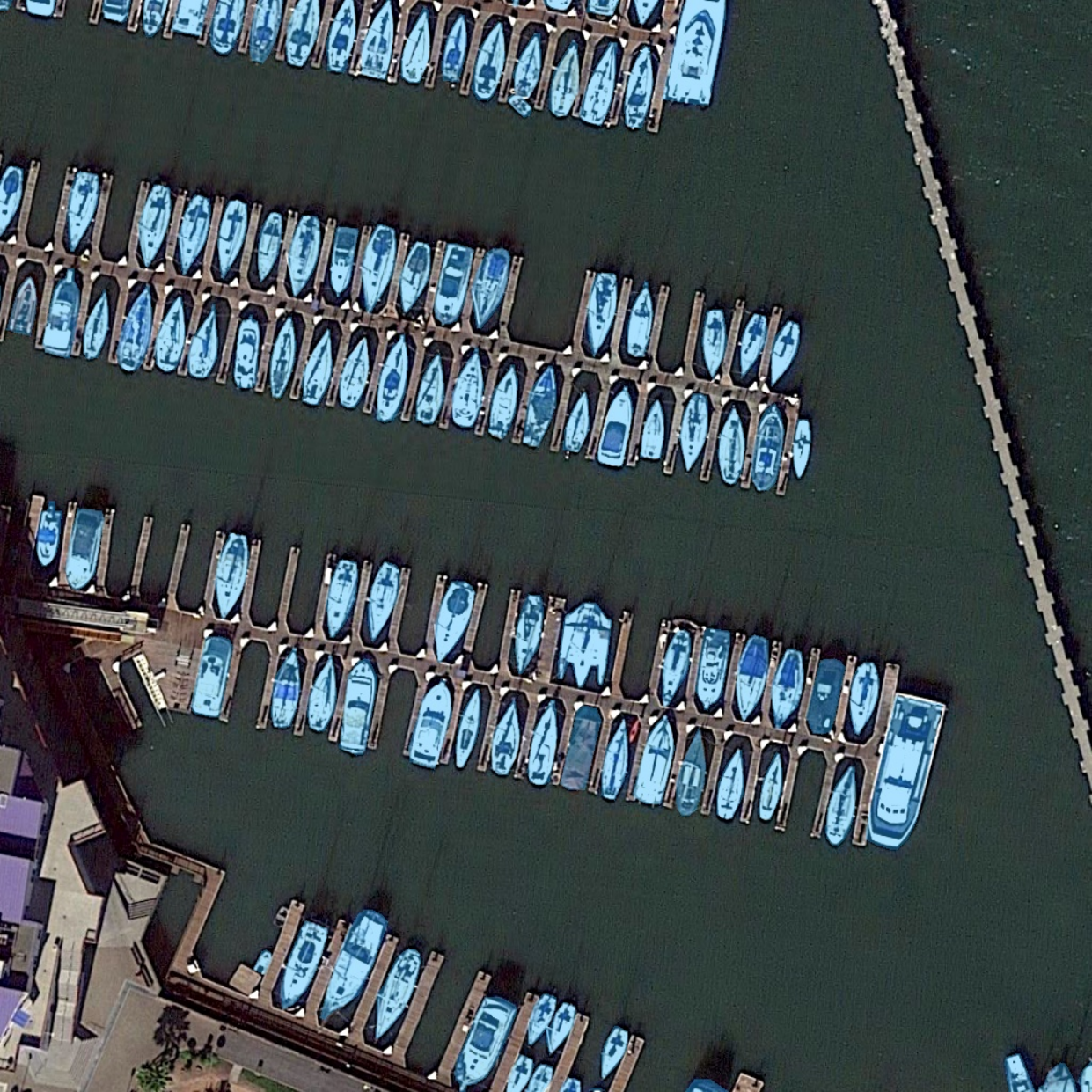} \\
\end{tabular}
\end{tabular}
}
\caption{Qualitative results on remote-sensing imagery, with building, agriculture, moving car, plane, and ship classes.}
\label{fig:qual}
\end{figure}

\subsection{Setup}
\label{sec:setup}

\paragraph{Benchmarks.} The benchmarks on which we develop and primarily evaluate FROST are seventeen remote-sensing segmentation benchmarks that span high-altitude satellite collections and lower-altitude drone views, and we group them by the task they pose. The building and footprint group contains WHU Building \citep{ji2018fully}, the two SpaceNet challenges \citep{van2018spacenet}, Massachusetts Buildings \citep{mnih2013machine}, and the damaged-building benchmark xBD \citep{gupta2019xbd}, where each target is a field of many small and visually varied structures. The land-cover group contains LoveDA \citep{wang2021loveda}, the ISPRS Potsdam and Vaihingen labelling sets \citep{rottensteiner2012isprs}, LandCover.ai \citep{boguszewski2021landcover}, OpenEarthMap \citep{xia2023openearthmap}, the multi-class benchmark DeepGlobe \citep{demir2018deepglobe} over whose promptable land-cover classes we evaluate in turn, the drone benchmarks UAVid \citep{lyu2020uavid} and UDD5 \citep{chen2018large}, iSAID \citep{waqas2019isaid}, DLRSD \citep{shao2018performance}, and WHDLD \citep{shao2018performance}. The flood group contains the drone benchmark FloodNet \citep{rahnemoonfar2021floodnet}, whose imagery captures inundated urban scenes. These benchmarks share the property that a class typically appears as a scatter of many small and dissimilar instances rather than a single dominant object, which is the regime FROST is designed for. To test how the same fixed configuration transfers beyond overhead imagery, we additionally evaluate on six standard few-shot segmentation benchmarks. Four are natural-image benchmarks, the object benchmarks PASCAL-$5^i$ \citep{shaban2017one}, COCO-$20^i$ \citep{nguyen2019feature}, and LVIS-$92^i$ \citep{liu2023matcher} together with the part benchmark PACO-Part \citep{liu2023matcher}, whose targets are typically large object-centric or part-centric instances rather than the scatter of small instances of the overhead regime. The remaining two are cross-domain benchmarks that depart from natural imagery entirely, the dermoscopy benchmark ISIC \citep{codella2019skin} and the underwater benchmark SUIM \citep{islam2020semantic}, which probe transfer to imaging conditions unseen during pretraining, and we report these results in Section~\ref{sec:natural}.

\begin{table}[t]
\centering
\caption{Foreground mean intersection over union (mIoU) on the remote-sensing building and footprint benchmarks together with the land-cover benchmark DeepGlobe and the flood benchmark FloodNet, with each entry the mean and the standard deviation over the four support sizes $k \in \{1, 3, 5, 10\}$, the best value in each column in bold and the second best underlined.}
\label{tab:buildingwater}
\resizebox{\linewidth}{!}{%
\begin{tabular}{l ccccccc}
\toprule
Method & WHU Building & SpaceNet-1 & SpaceNet-2 & Mass. Build. & xBD & DeepGlobe & FloodNet \\
\midrule
\multicolumn{8}{l}{\textit{Learning-based}} \\
SegGPT & 68.9 {\scriptsize $\pm$ 5.8} & 41.3 {\scriptsize $\pm$ 2.2} & 48.1 {\scriptsize $\pm$ 4.3} & 11.2 {\scriptsize $\pm$ 0.2} & \underline{35.5} {\scriptsize $\pm$ 5.0} & 20.6 {\scriptsize $\pm$ 1.8} & 45.9 {\scriptsize $\pm$ 1.9} \\
SegIC & 56.5 {\scriptsize $\pm$ 8.3} & 48.0 {\scriptsize $\pm$ 4.0} & 46.9 {\scriptsize $\pm$ 2.9} & 22.4 {\scriptsize $\pm$ 3.2} & 31.2 {\scriptsize $\pm$ 3.8} & 40.0 {\scriptsize $\pm$ 0.7} & 65.1 {\scriptsize $\pm$ 2.9} \\
DiffewS & 63.1 {\scriptsize $\pm$ 1.7} & 45.5 {\scriptsize $\pm$ 0.6} & 36.0 {\scriptsize $\pm$ 3.7} & 18.3 {\scriptsize $\pm$ 0.4} & 17.3 {\scriptsize $\pm$ 3.3} & 39.7 {\scriptsize $\pm$ 2.6} & 67.2 {\scriptsize $\pm$ 5.9} \\
SINE & 22.2 {\scriptsize $\pm$ 5.5} & 29.2 {\scriptsize $\pm$ 6.7} & 15.8 {\scriptsize $\pm$ 4.9} & 18.0 {\scriptsize $\pm$ 0.9} & 11.3 {\scriptsize $\pm$ 5.1} & 30.7 {\scriptsize $\pm$ 3.7} & 39.2 {\scriptsize $\pm$ 4.5} \\
\midrule
\multicolumn{8}{l}{\textit{Training-free}} \\
PerSAM & 2.6 {\scriptsize $\pm$ 0.1} & 2.4 {\scriptsize $\pm$ 1.7} & 2.9 {\scriptsize $\pm$ 0.3} & 0.6 {\scriptsize $\pm$ 0.5} & 2.3 {\scriptsize $\pm$ 0.5} & 22.9 {\scriptsize $\pm$ 0.3} & 34.0 {\scriptsize $\pm$ 3.9} \\
Matcher & 30.7 {\scriptsize $\pm$ 3.0} & 32.5 {\scriptsize $\pm$ 1.8} & 19.1 {\scriptsize $\pm$ 1.5} & 15.2 {\scriptsize $\pm$ 0.3} & 11.6 {\scriptsize $\pm$ 1.7} & 38.2 {\scriptsize $\pm$ 3.1} & 55.7 {\scriptsize $\pm$ 3.2} \\
GF-SAM & 67.2 {\scriptsize $\pm$ 1.8} & 42.1 {\scriptsize $\pm$ 0.6} & 39.5 {\scriptsize $\pm$ 3.8} & 11.6 {\scriptsize $\pm$ 0.3} & 35.4 {\scriptsize $\pm$ 4.9} & 43.0 {\scriptsize $\pm$ 4.7} & 65.6 {\scriptsize $\pm$ 7.3} \\
FSSDINO & 72.2 {\scriptsize $\pm$ 2.0} & 47.0 {\scriptsize $\pm$ 4.3} & \underline{49.4} {\scriptsize $\pm$ 4.6} & \underline{31.0} {\scriptsize $\pm$ 1.7} & 33.0 {\scriptsize $\pm$ 7.0} & 43.3 {\scriptsize $\pm$ 7.0} & 60.2 {\scriptsize $\pm$ 10.4} \\
INSID3 & \underline{73.2} {\scriptsize $\pm$ 0.4} & \underline{48.5} {\scriptsize $\pm$ 1.8} & 44.0 {\scriptsize $\pm$ 2.4} & 20.0 {\scriptsize $\pm$ 0.7} & 32.1 {\scriptsize $\pm$ 3.5} & \underline{44.1} {\scriptsize $\pm$ 4.8} & \underline{70.9} {\scriptsize $\pm$ 5.2} \\
FROST (ours) & \textbf{77.9} {\scriptsize $\pm$ 0.4} & \textbf{51.3} {\scriptsize $\pm$ 2.3} & \textbf{59.2} {\scriptsize $\pm$ 2.3} & \textbf{33.6} {\scriptsize $\pm$ 0.4} & \textbf{52.8} {\scriptsize $\pm$ 4.2} & \textbf{45.6} {\scriptsize $\pm$ 4.6} & \textbf{72.4} {\scriptsize $\pm$ 8.5} \\
\bottomrule
\end{tabular}
}
\end{table}

\begin{table}[t]
\centering
\caption{Foreground mIoU on the remote-sensing land-cover and drone benchmarks, with each entry the mean and standard deviation over the four support sizes $k \in \{1, 3, 5, 10\}$, the best in each column in bold and the second best underlined.}
\label{tab:landcover}
\resizebox{\linewidth}{!}{%
\begin{tabular}{l cccccccccc}
\toprule
Method & LoveDA & Potsdam & Vaihingen & LandCover.ai & OpenEarthMap & UAVid & UDD5 & iSAID & DLRSD & WHDLD \\
\midrule
\multicolumn{11}{l}{\textit{Learning-based}} \\
SegGPT & 45.3 {\scriptsize $\pm$ 3.5} & 39.2 {\scriptsize $\pm$ 0.5} & 50.7 {\scriptsize $\pm$ 2.7} & 47.0 {\scriptsize $\pm$ 1.5} & 20.6 {\scriptsize $\pm$ 0.9} & 31.4 {\scriptsize $\pm$ 1.4} & 27.4 {\scriptsize $\pm$ 1.3} & 22.2 {\scriptsize $\pm$ 1.6} & 20.1 {\scriptsize $\pm$ 0.7} & 24.8 {\scriptsize $\pm$ 2.1} \\
SegIC & 34.4 {\scriptsize $\pm$ 0.5} & 61.2 {\scriptsize $\pm$ 1.3} & 51.0 {\scriptsize $\pm$ 2.8} & 56.4 {\scriptsize $\pm$ 3.9} & 29.2 {\scriptsize $\pm$ 0.6} & 53.0 {\scriptsize $\pm$ 1.0} & 53.2 {\scriptsize $\pm$ 2.6} & 23.5 {\scriptsize $\pm$ 2.0} & 32.4 {\scriptsize $\pm$ 0.7} & 34.1 {\scriptsize $\pm$ 0.6} \\
DiffewS & 29.5 {\scriptsize $\pm$ 1.6} & 52.3 {\scriptsize $\pm$ 5.2} & 47.0 {\scriptsize $\pm$ 1.5} & 49.3 {\scriptsize $\pm$ 5.1} & 23.3 {\scriptsize $\pm$ 0.9} & \textbf{56.1} {\scriptsize $\pm$ 2.5} & 44.6 {\scriptsize $\pm$ 4.6} & 12.7 {\scriptsize $\pm$ 1.4} & 31.3 {\scriptsize $\pm$ 1.0} & 30.6 {\scriptsize $\pm$ 1.6} \\
SINE & 15.3 {\scriptsize $\pm$ 1.6} & 34.0 {\scriptsize $\pm$ 3.6} & 23.1 {\scriptsize $\pm$ 1.9} & 34.8 {\scriptsize $\pm$ 1.8} & 20.6 {\scriptsize $\pm$ 2.4} & 25.1 {\scriptsize $\pm$ 3.6} & 27.2 {\scriptsize $\pm$ 3.1} & 5.5 {\scriptsize $\pm$ 1.7} & 16.2 {\scriptsize $\pm$ 0.5} & 24.2 {\scriptsize $\pm$ 1.4} \\
\midrule
\multicolumn{11}{l}{\textit{Training-free}} \\
PerSAM & 14.3 {\scriptsize $\pm$ 0.4} & 21.4 {\scriptsize $\pm$ 1.8} & 9.2 {\scriptsize $\pm$ 0.5} & 36.4 {\scriptsize $\pm$ 1.2} & 6.8 {\scriptsize $\pm$ 1.0} & 16.4 {\scriptsize $\pm$ 1.5} & 17.9 {\scriptsize $\pm$ 1.5} & 7.4 {\scriptsize $\pm$ 1.4} & 20.7 {\scriptsize $\pm$ 0.4} & 19.2 {\scriptsize $\pm$ 0.6} \\
Matcher & 27.4 {\scriptsize $\pm$ 0.8} & 43.1 {\scriptsize $\pm$ 2.2} & 32.5 {\scriptsize $\pm$ 1.8} & 43.4 {\scriptsize $\pm$ 1.2} & 20.8 {\scriptsize $\pm$ 1.2} & 35.9 {\scriptsize $\pm$ 4.4} & 33.5 {\scriptsize $\pm$ 1.4} & 19.8 {\scriptsize $\pm$ 3.9} & 27.8 {\scriptsize $\pm$ 1.6} & 24.9 {\scriptsize $\pm$ 0.3} \\
GF-SAM & 44.7 {\scriptsize $\pm$ 4.4} & 58.5 {\scriptsize $\pm$ 2.3} & 51.6 {\scriptsize $\pm$ 1.9} & 53.7 {\scriptsize $\pm$ 3.6} & 32.2 {\scriptsize $\pm$ 3.7} & 44.3 {\scriptsize $\pm$ 3.9} & 47.1 {\scriptsize $\pm$ 5.3} & 20.4 {\scriptsize $\pm$ 3.0} & \underline{33.2} {\scriptsize $\pm$ 2.2} & 36.9 {\scriptsize $\pm$ 2.6} \\
FSSDINO & 42.5 {\scriptsize $\pm$ 6.4} & 63.0 {\scriptsize $\pm$ 7.5} & 55.0 {\scriptsize $\pm$ 2.5} & 52.2 {\scriptsize $\pm$ 11.6} & \underline{38.2} {\scriptsize $\pm$ 6.5} & 49.9 {\scriptsize $\pm$ 7.8} & 50.0 {\scriptsize $\pm$ 9.3} & 21.5 {\scriptsize $\pm$ 4.4} & 23.8 {\scriptsize $\pm$ 3.9} & \underline{40.6} {\scriptsize $\pm$ 5.2} \\
INSID3 & \underline{48.4} {\scriptsize $\pm$ 5.0} & \underline{64.2} {\scriptsize $\pm$ 3.4} & \underline{57.7} {\scriptsize $\pm$ 1.4} & \underline{58.7} {\scriptsize $\pm$ 4.7} & 36.6 {\scriptsize $\pm$ 5.8} & 47.5 {\scriptsize $\pm$ 4.2} & \underline{54.5} {\scriptsize $\pm$ 7.3} & \underline{32.0} {\scriptsize $\pm$ 4.8} & 31.6 {\scriptsize $\pm$ 2.3} & 35.5 {\scriptsize $\pm$ 2.8} \\
FROST (ours) & \textbf{49.7} {\scriptsize $\pm$ 3.3} & \textbf{71.2} {\scriptsize $\pm$ 4.6} & \textbf{64.5} {\scriptsize $\pm$ 1.9} & \textbf{63.7} {\scriptsize $\pm$ 2.6} & \textbf{41.5} {\scriptsize $\pm$ 5.1} & \underline{54.0} {\scriptsize $\pm$ 3.9} & \textbf{60.3} {\scriptsize $\pm$ 7.7} & \textbf{37.3} {\scriptsize $\pm$ 6.4} & \textbf{35.3} {\scriptsize $\pm$ 3.7} & \textbf{42.5} {\scriptsize $\pm$ 3.3} \\
\bottomrule
\end{tabular}
}
\end{table}

\paragraph{Baselines.} We compare against training-free methods, namely the frozen-feature DINOv3 methods INSID3 \citep{cuttano2026insid3} and FSSDINO \citep{zakir2026revealing} together with the SAM-prompting methods PerSAM \citep{zhang2024personalize}, Matcher \citep{liu2023matcher}, and GF-SAM \citep{zhang2024bridge}, and against learning-based methods, namely SegIC \citep{meng2024segic}, SINE \citep{liu2024simple}, DiffewS \citep{zhu2024unleashing}, and the in-context generalist SegGPT \citep{wang2023seggpt}. FROST shares the frozen DINOv3 ViT-L backbone of INSID3 and FSSDINO, so their comparison isolates the decision rule, and Table~\ref{tab:backbones} lists the backbone and the regime of every method.

\paragraph{Implementation details.} We use a frozen DINOv3 ViT-L/16 backbone at $1024 \times 1024$ input with support sizes $k \in \{1, 3, 5, 10\}$, hold every hyperparameter fixed across the seventeen benchmarks, and run inference forward only with no test-time optimization. Feature refinement uses a positional debiasing rank $r = 250$ and a shrinkage intensity $\lambda = 0.95$, the kernel bandwidth $\sigma$ is chosen per episode by the leave-one-out margin of Section~\ref{sec:kde}, and the density ratio is thresholded at $\tau = 0$. The bilateral propagation runs $T = 10$ steps with mixing weight $\beta = 0.70$, feature and colour temperatures $\gamma_f = 0.20$ and $\gamma_c = 0.05$, and window radius $d_{\max} = 16$, and the candidate gating uses a backward neighbour count $k_b = 3$ and a maximal gate radius $\rho_{\max} = 4$.

\subsection{Results on remote-sensing benchmarks}
\paragraph{Overall comparison.}
\label{sec:main}
Table~\ref{tab:buildingwater} and Table~\ref{tab:landcover} report the foreground mIoU of every method, with each entry the mean and standard deviation over the four support sizes $k \in \{1, 3, 5, 10\}$, the best value in each column in bold and the second best underlined, the column Mass. Build. abbreviating Massachusetts Buildings, and mIoU the mean intersection over union. The spread reported beside each mean measures the change in accuracy across support sizes rather than run-to-run noise, so a large spread coincides with strong scaling, as on FloodNet where it rises from 60.2 at one shot to 79.5 at ten shots. FROST attains the best value on sixteen of the seventeen benchmarks and on every benchmark of the building and footprint group, and averaged over the benchmarks and the support sizes it reaches 53.7 mIoU, ahead of the strongest training-free baseline INSID3 by 6.7 mIoU and the strongest learning-based baseline SegIC by 10.3 mIoU. The single benchmark on which FROST is not first is UAVid, where the learning-based DiffewS leads by a small margin while FROST stays the strongest training-free method. The advantage is largest on building and footprint benchmarks such as xBD and SpaceNet-2, where the foreground is many small and varied structures that a single prototype blurs across, which is the regime the density ratio is built for. Figure~\ref{fig:qual} compares the masks of FROST with FSSDINO and INSID3 on representative building, agriculture, moving car, plane, and ship episodes, where FROST recovers the separate instances that the competing summaries merge. The per-support-size results that underlie these tables appear in Appendix~\ref{app:perk}.

\paragraph{Scaling with the support set.}
\label{sec:scaling}
Figure~\ref{fig:overview} reports the foreground mIoU averaged over the seventeen benchmarks at each support size for FROST and the strongest training-free methods, and FROST rises from 48.3 mIoU at one shot to 56.8 mIoU at ten shots while every competing training-free method improves more slowly, which follows the variance argument of Section~\ref{sec:kde}, since a larger support set lowers that variance. The per-benchmark behaviour appears in Figure~\ref{fig:radar}, where the FROST contour is the outermost at every support size, with INSID3 the only training-free method to move ahead, on LoveDA at five and ten shots. The margin of FROST over the strongest competing method at the same support size grows from 6.2 mIoU at one shot to 7.0 mIoU at five shots and stands at 6.5 mIoU at ten shots, and the number of benchmarks on which FROST records the best result grows monotonically from 13 of seventeen at one shot to 16 of seventeen at ten shots, so its lead broadens as references accumulate. The fine-tuned SegIC does not share this behaviour, peaking at three shots and then declining as further references conflict with its trained priors.

\subsection{Generalization beyond overhead imagery}
\label{sec:natural}

\begin{figure}[t]
\centering
\begin{minipage}[c]{0.48\textwidth}
\centering
\setlength{\tabcolsep}{3pt}
\renewcommand{\arraystretch}{0.95}
\captionof{table}{Foreground mIoU beyond overhead imagery at $k=1$ and $k=10$, the best in each column in bold and the second best underlined}
\label{tab:natural}
\resizebox{\linewidth}{!}{%
\begin{tabular}{l cccccc}
\toprule
Method & PASCAL-$5^i$ & COCO-$20^i$ & LVIS-$92^i$ & PACO-Part & ISIC & SUIM \\
\midrule
\multicolumn{7}{l}{\textit{1-shot}} \\
PerSAM & 45.6 & 21.8 & 13.3 & 20.8 & 22.7 & 29.6 \\
Matcher & 68.2 & 54.6 & 36.5 & 34.3 & 33.4 & 46.7 \\
GF-SAM & \textbf{71.9} & \textbf{59.2} & \underline{37.9} & 38.1 & 48.9 & 55.0 \\
FSSDINO & 65.5 & 39.9 & 22.2 & 34.3 & \underline{52.3} & 51.3 \\
INSID3 & 69.9 & \underline{56.5} & \textbf{40.9} & \textbf{39.6} & 50.5 & \underline{56.2} \\
FROST (ours) & \underline{70.1} & 48.4 & 32.8 & \underline{39.1} & \textbf{52.6} & \textbf{58.0} \\
\midrule
\multicolumn{7}{l}{\textit{10-shot}} \\
PerSAM & 55.0 & 29.9 & 16.4 & 27.1 & 26.9 & 35.9 \\
Matcher & 75.4 & 61.7 & 39.4 & 34.3 & 36.2 & 48.8 \\
GF-SAM & \textbf{83.9} & \textbf{68.1} & \underline{48.1} & 45.4 & 60.7 & 63.2 \\
FSSDINO & 80.2 & 56.7 & 35.3 & \underline{50.6} & 65.4 & 63.2 \\
INSID3 & 81.5 & 64.0 & 46.8 & 50.3 & \underline{67.0} & \underline{67.6} \\
FROST (ours) & \underline{83.7} & \underline{64.3} & \textbf{51.9} & \textbf{57.0} & \textbf{70.4} & \textbf{70.1} \\
\bottomrule
\end{tabular}
}
\end{minipage}
\hfill
\begin{minipage}[c]{0.5\textwidth}
\centering
\providecommand{\cwn}{0.155\linewidth}
\setlength{\tabcolsep}{1.2pt}
\renewcommand{\arraystretch}{0}
\begin{tabular}{@{}*{5}{c}@{}}
{\scriptsize Reference} & {\scriptsize FSSDINO} & {\scriptsize INSID3} & {\scriptsize FROST} & {\scriptsize GT} \\[2pt]
\includegraphics[width=\cwn]{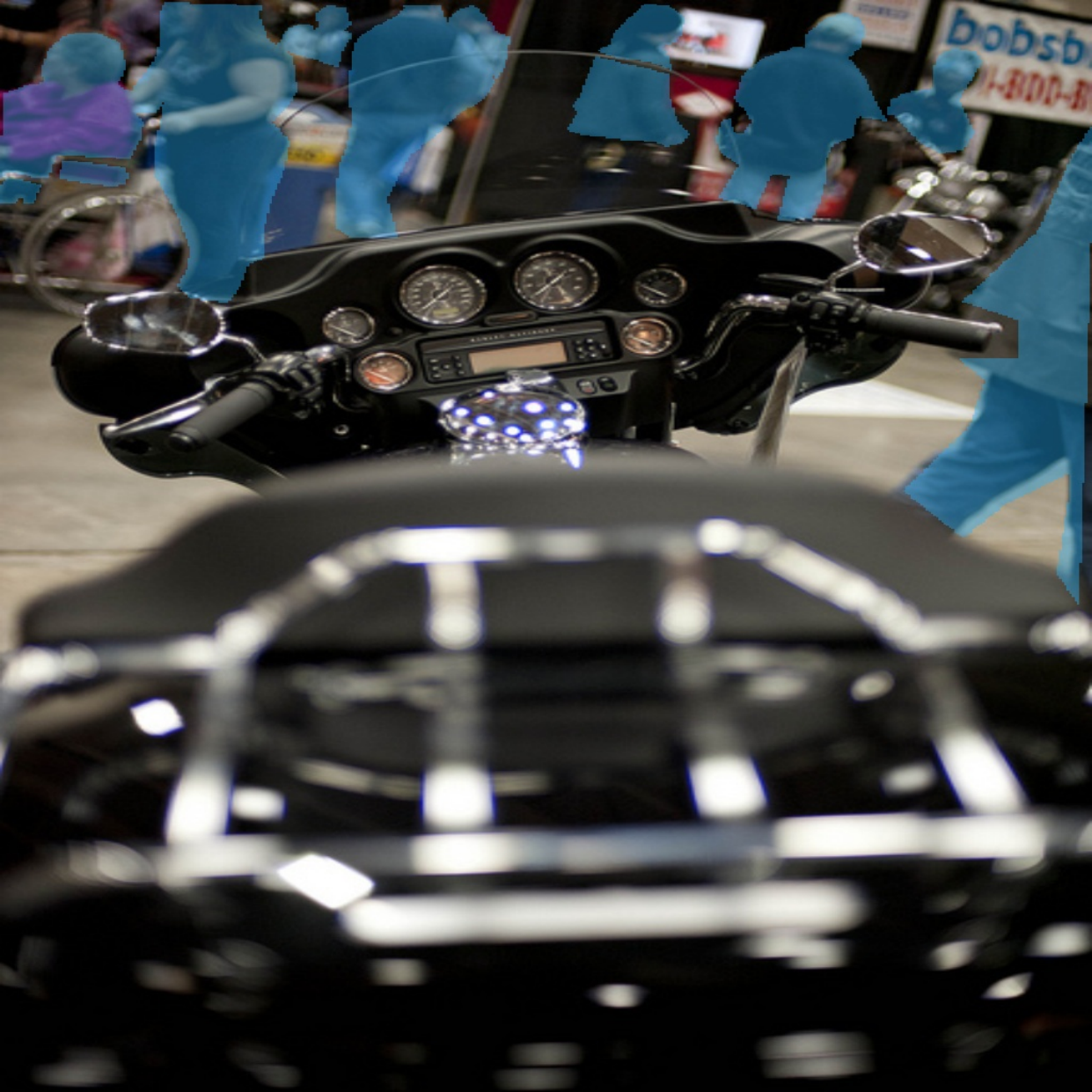} &
\includegraphics[width=\cwn]{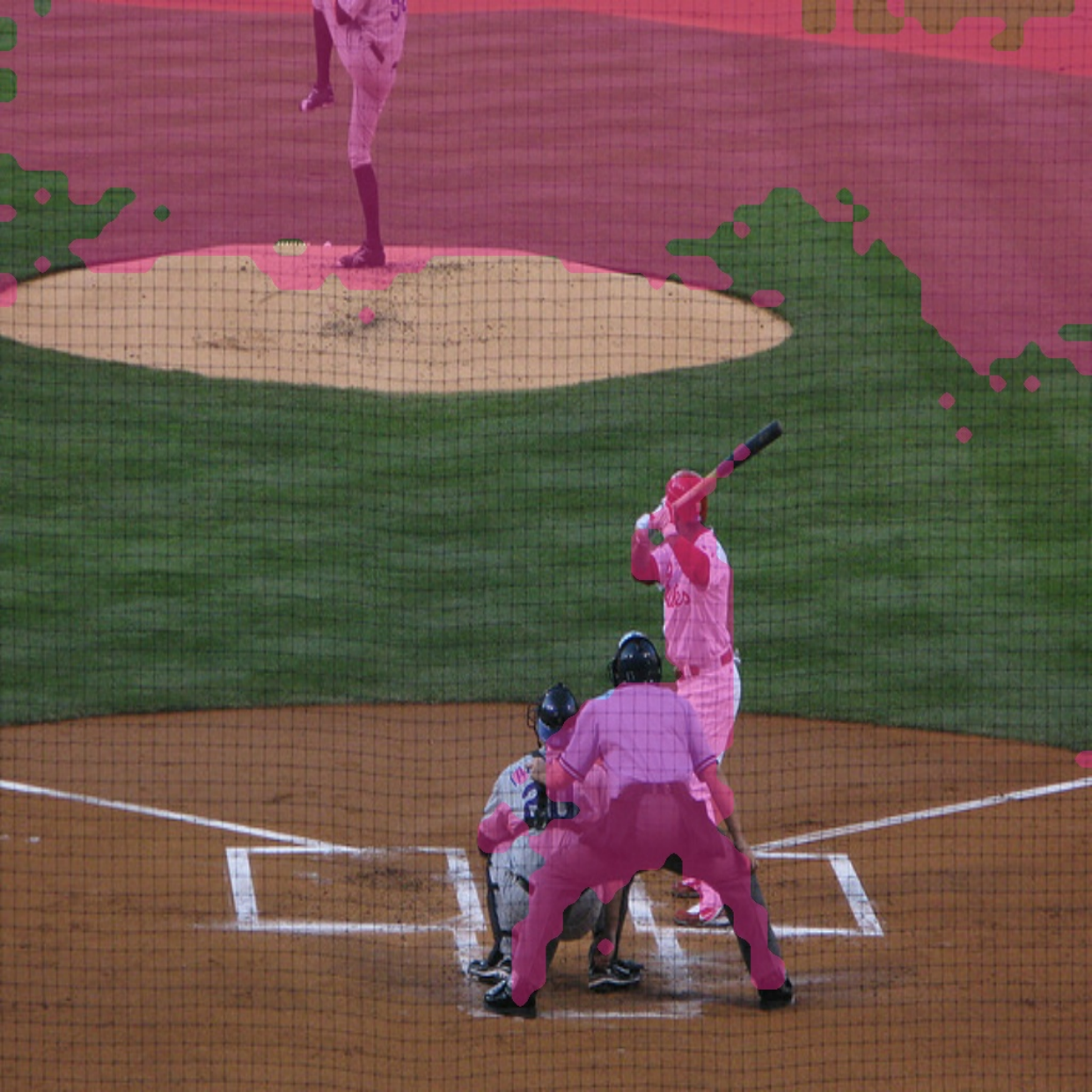} &
\includegraphics[width=\cwn]{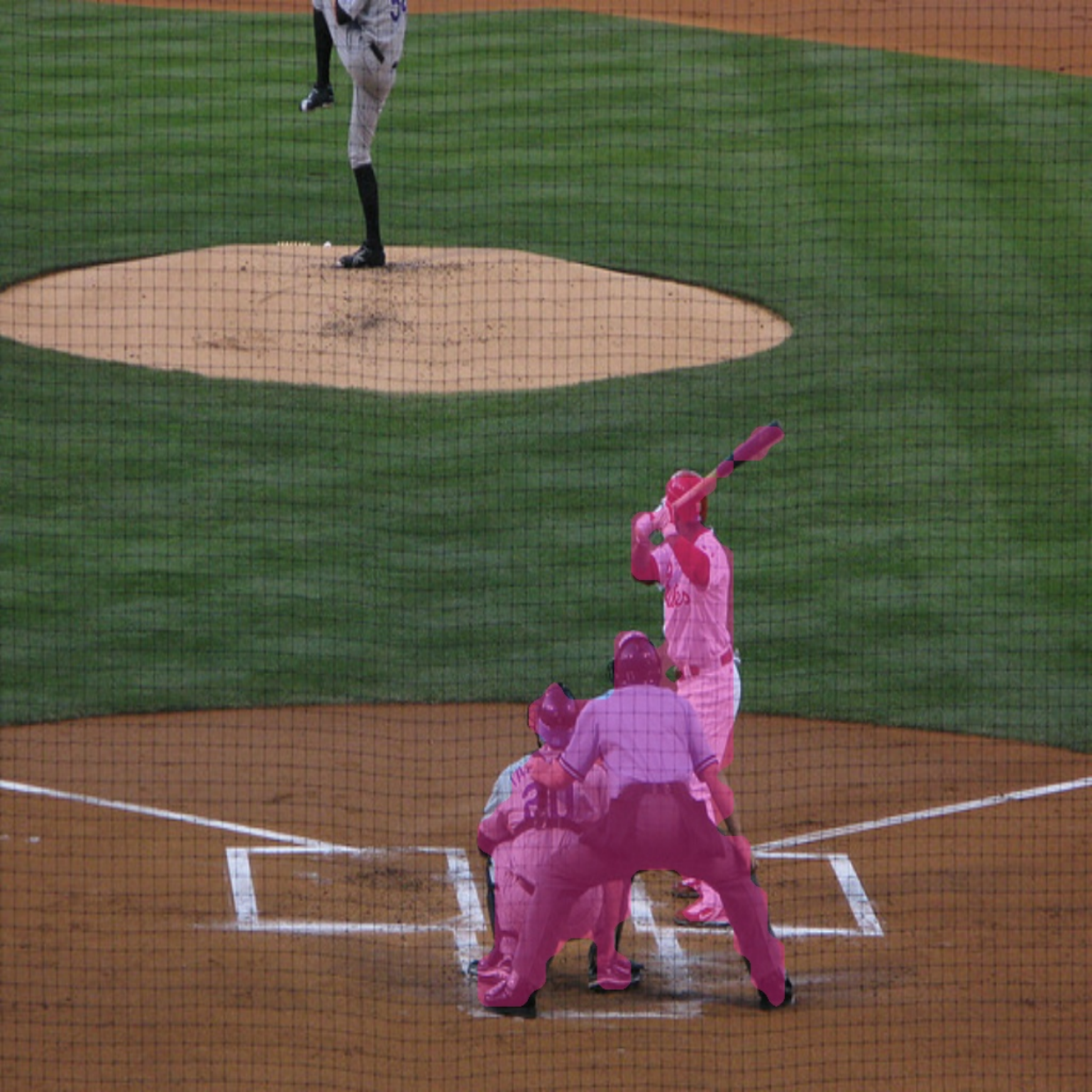} &
\includegraphics[width=\cwn]{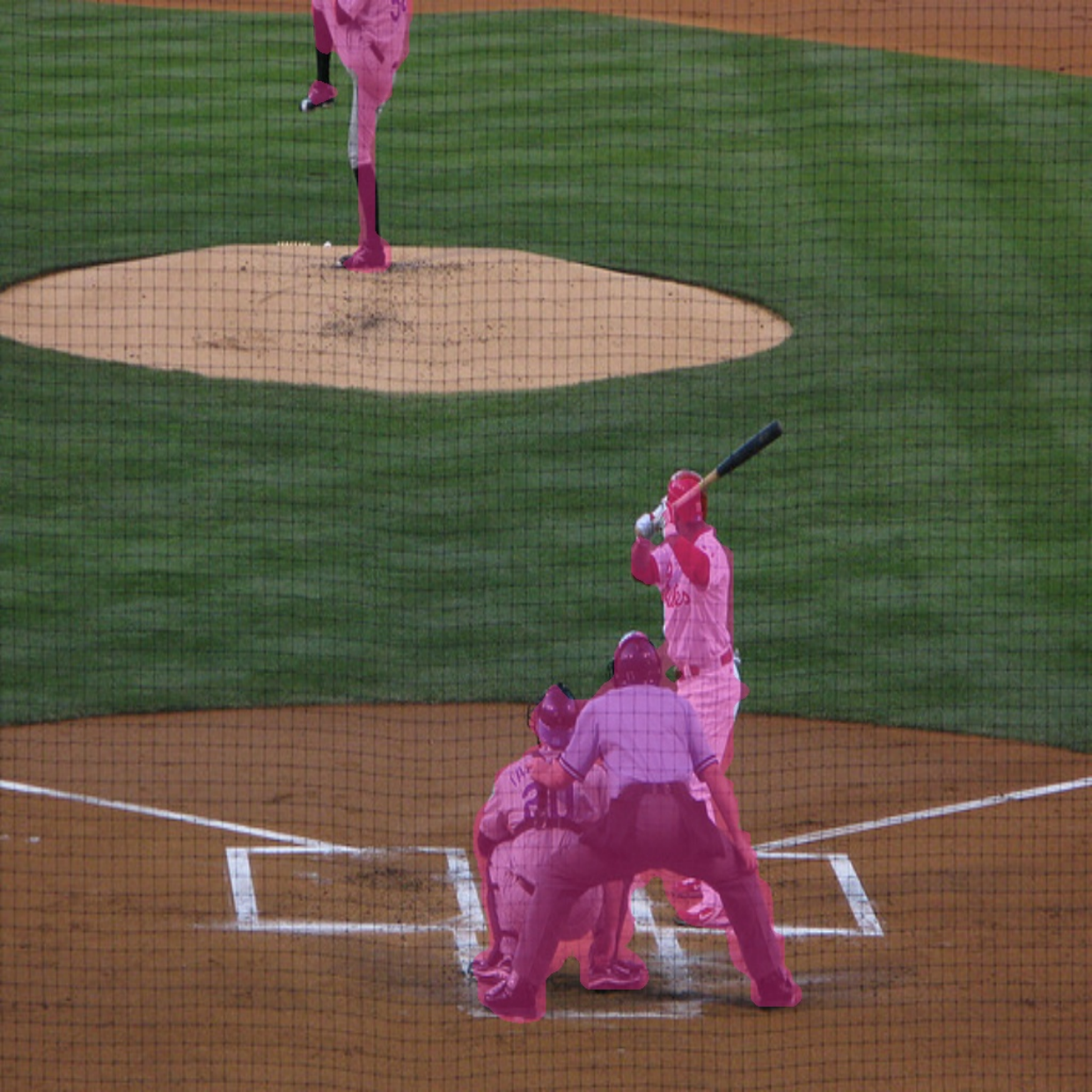} &
\includegraphics[width=\cwn]{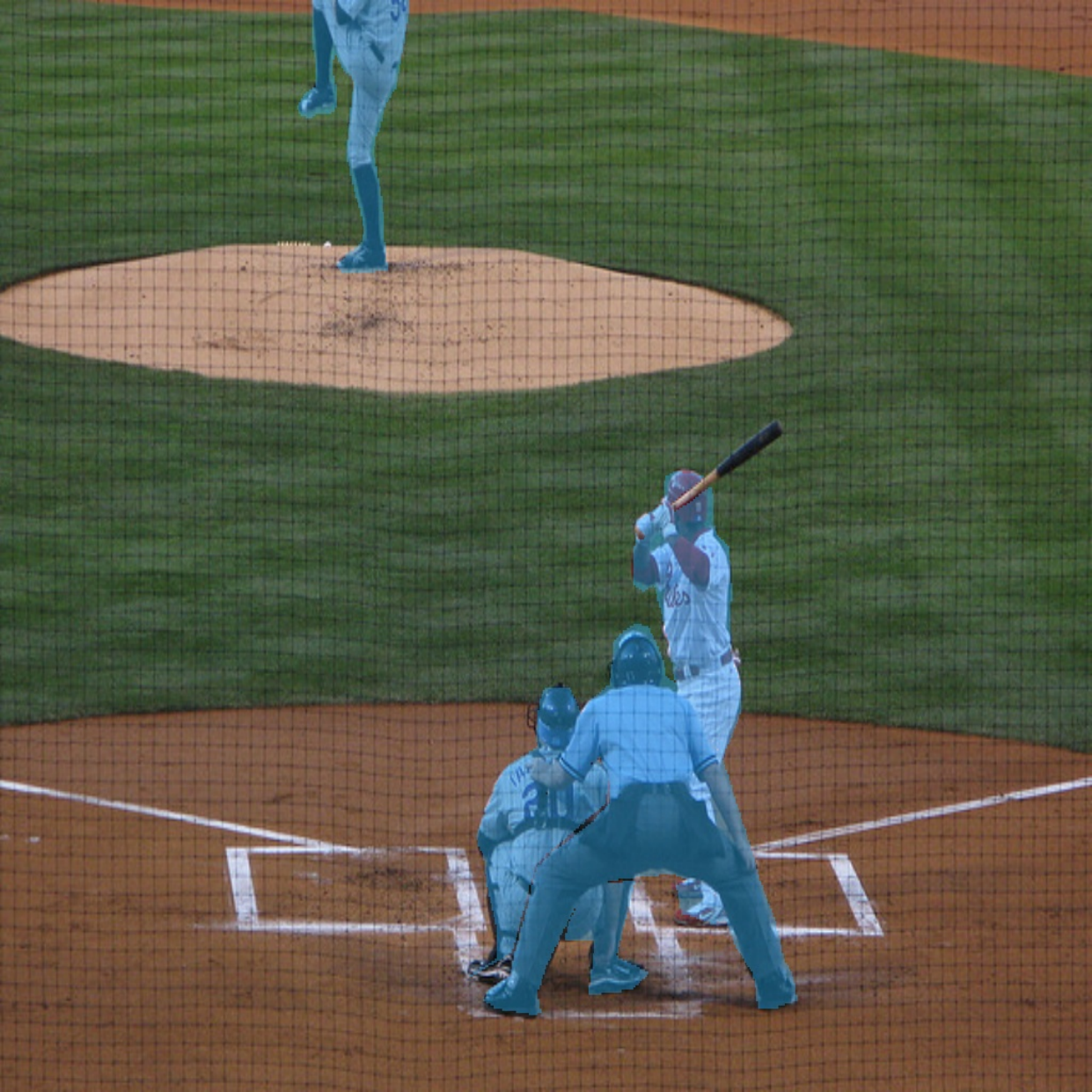} \\[2pt]
\includegraphics[width=\cwn]{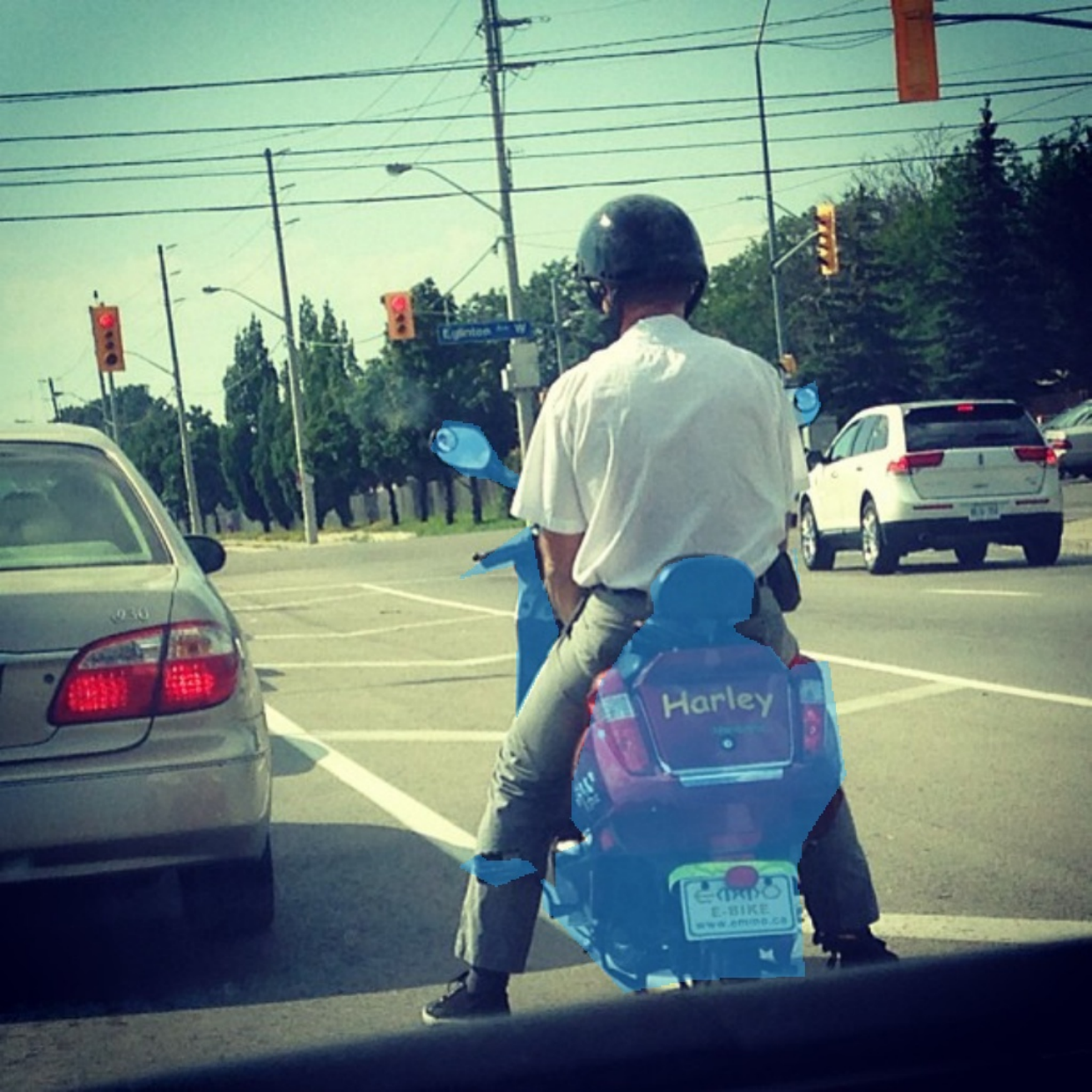} &
\includegraphics[width=\cwn]{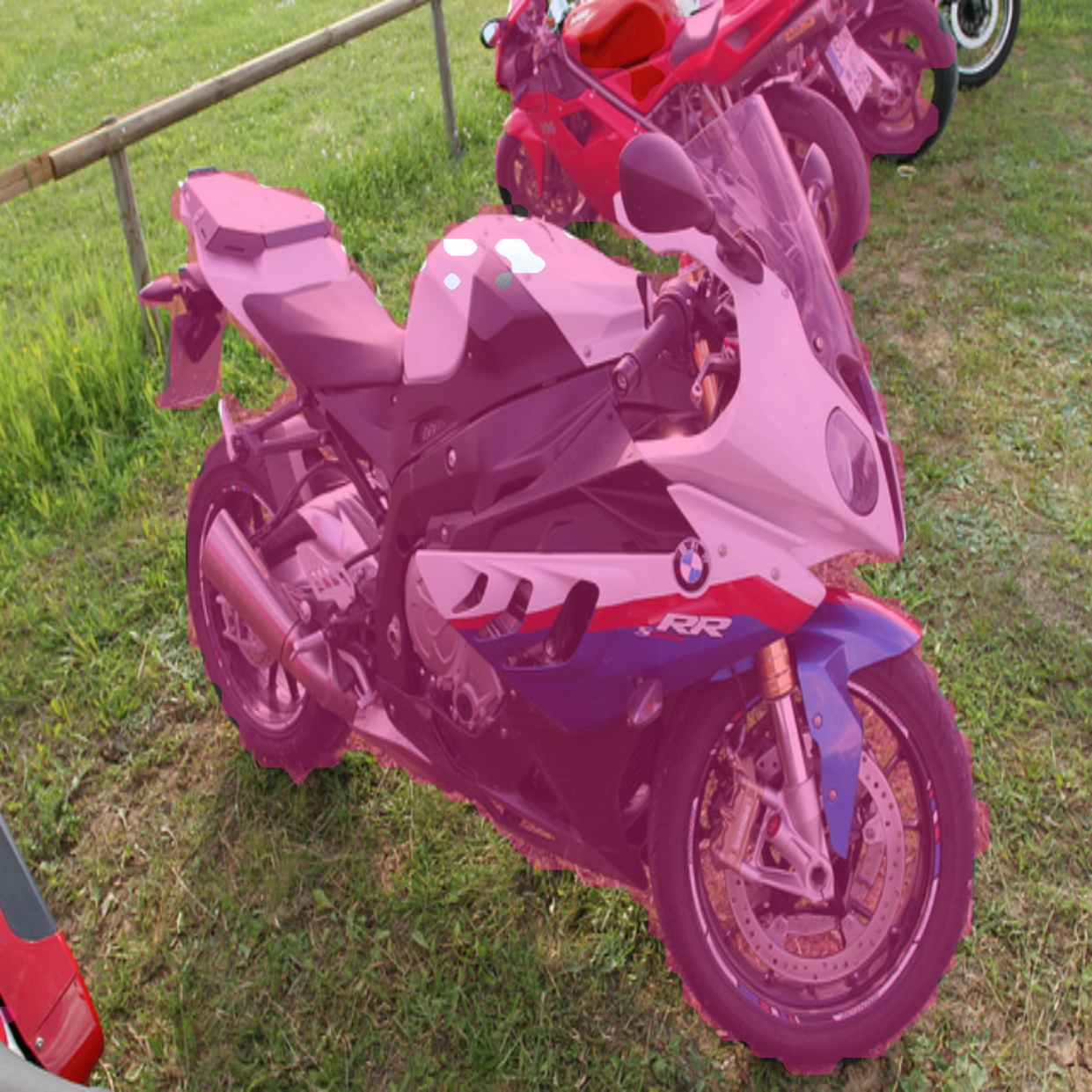} &
\includegraphics[width=\cwn]{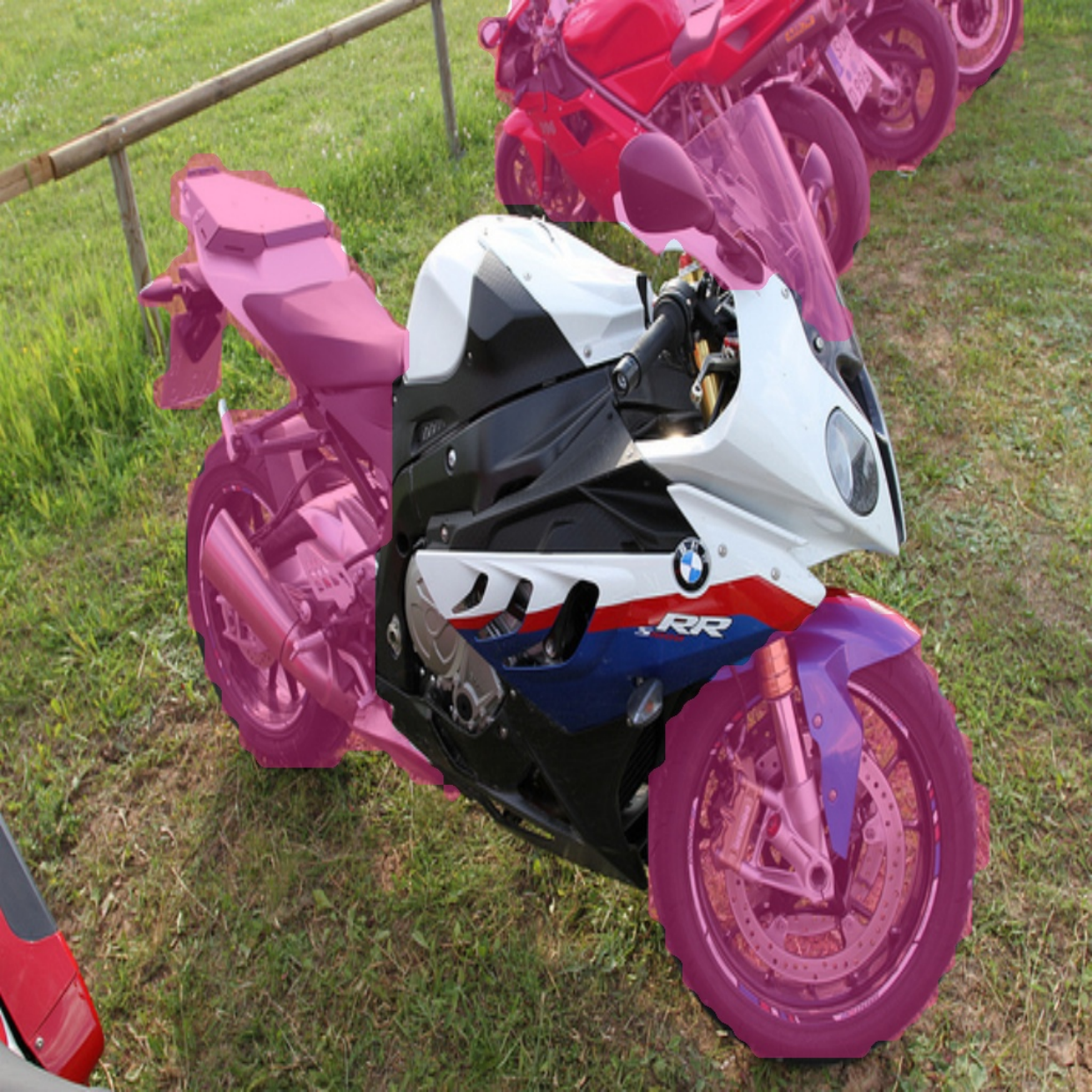} &
\includegraphics[width=\cwn]{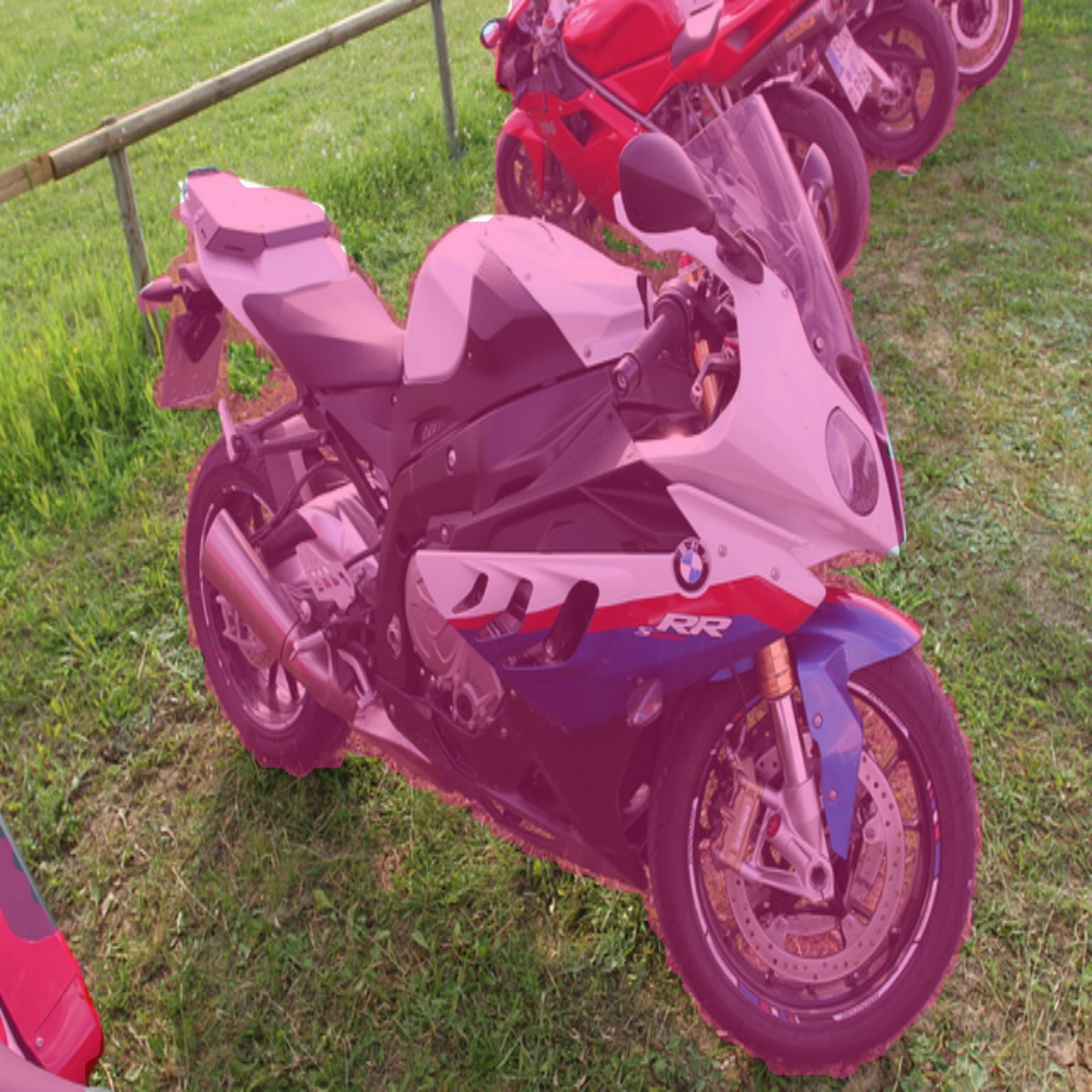} &
\includegraphics[width=\cwn]{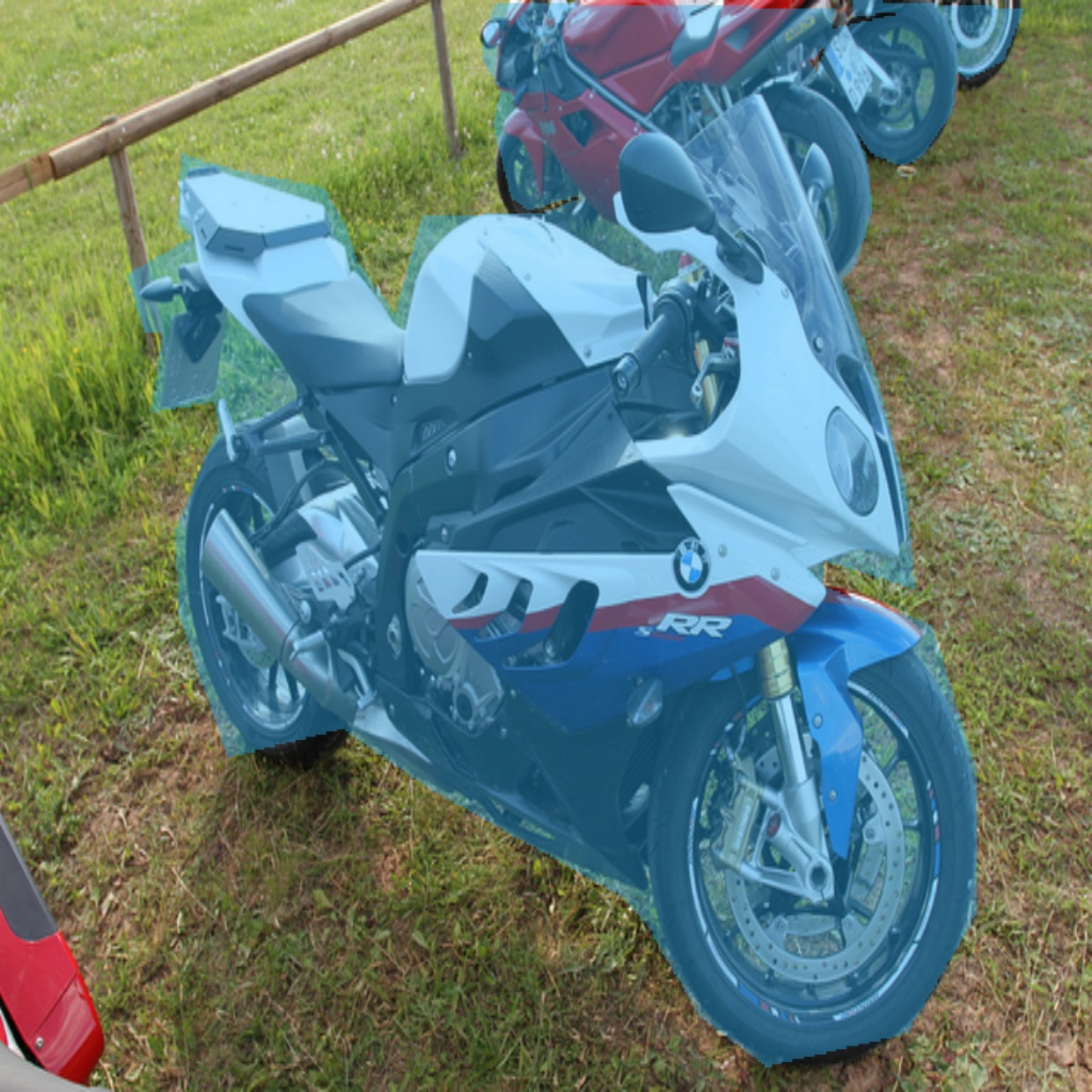} \\[2pt]
\includegraphics[width=\cwn]{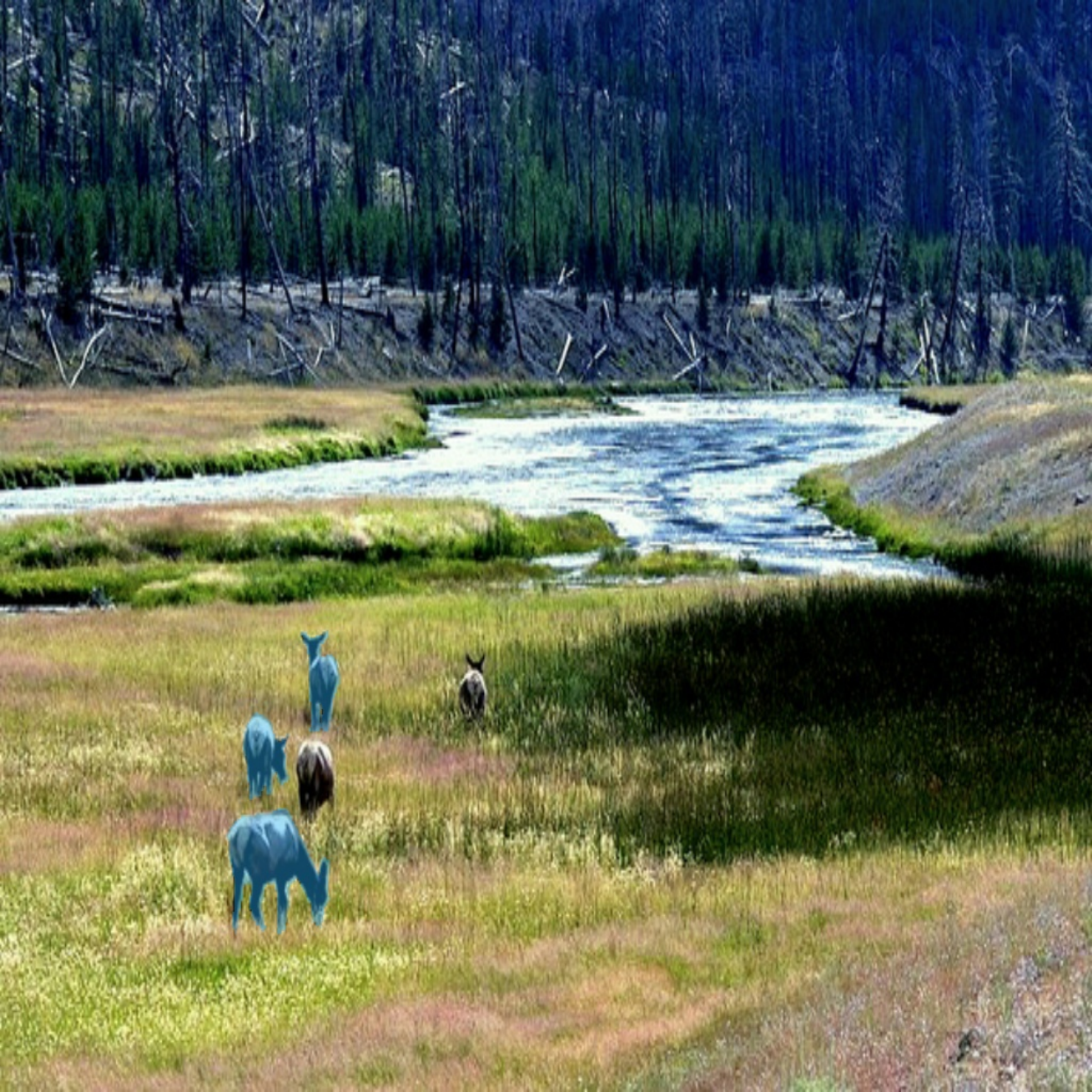} &
\includegraphics[width=\cwn]{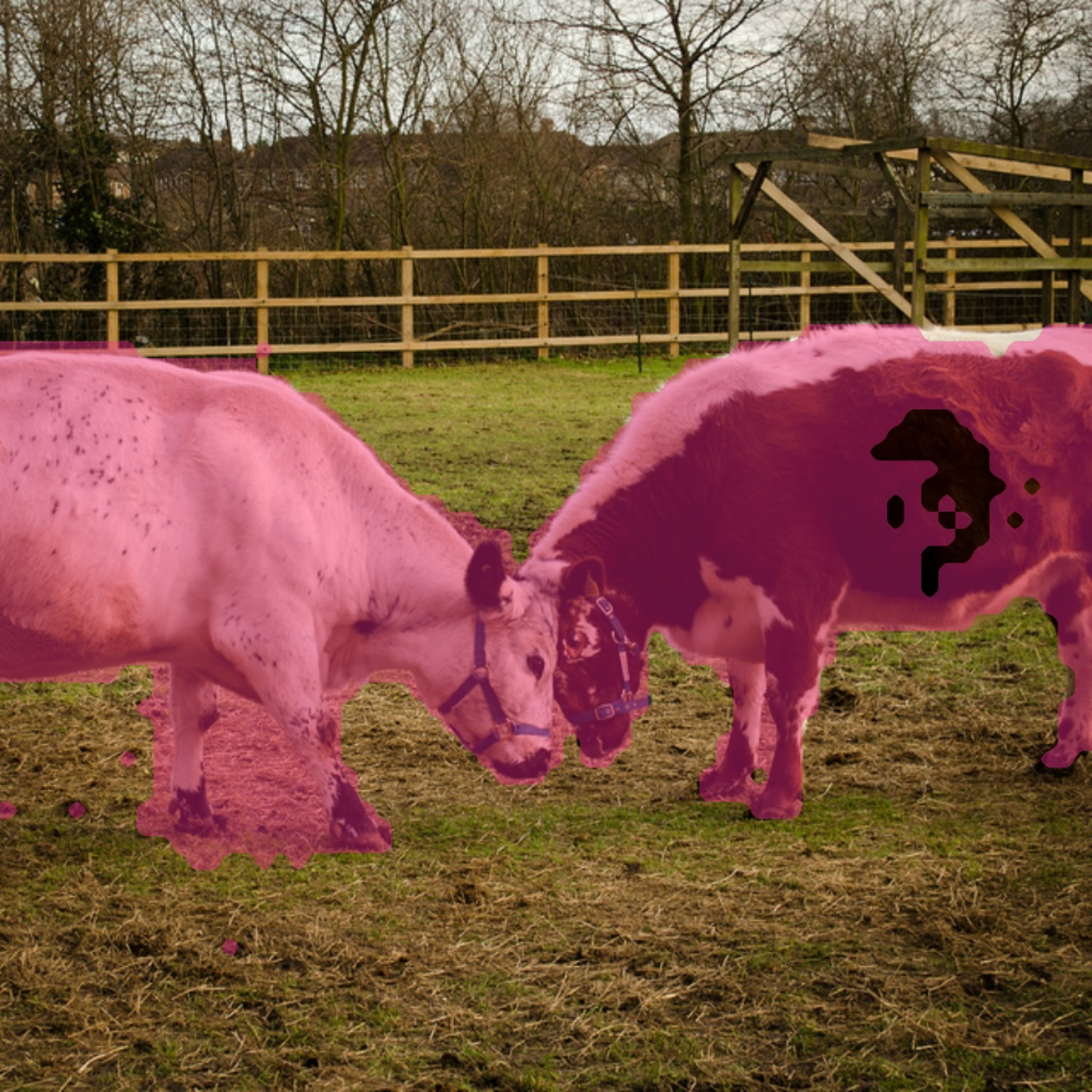} &
\includegraphics[width=\cwn]{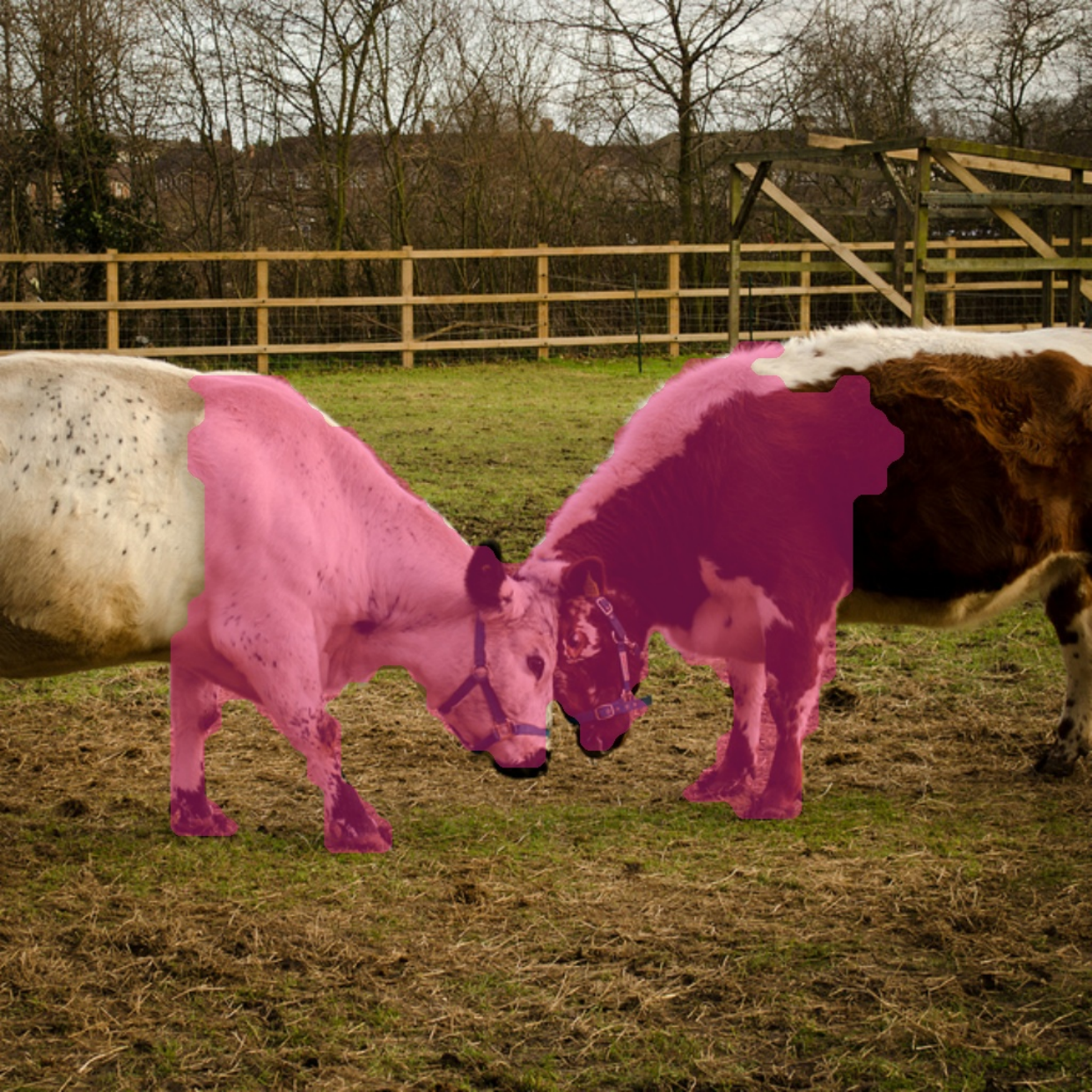} &
\includegraphics[width=\cwn]{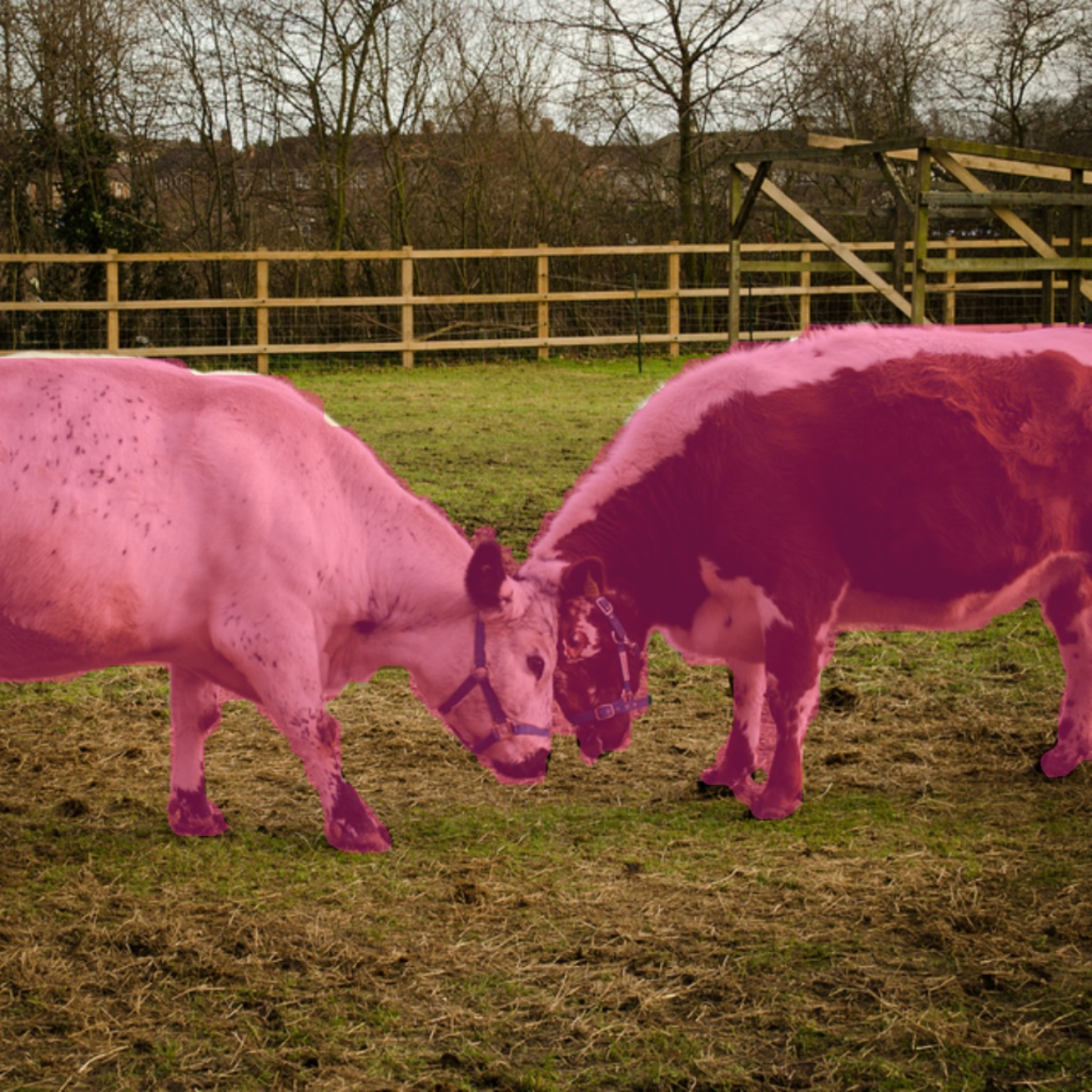} &
\includegraphics[width=\cwn]{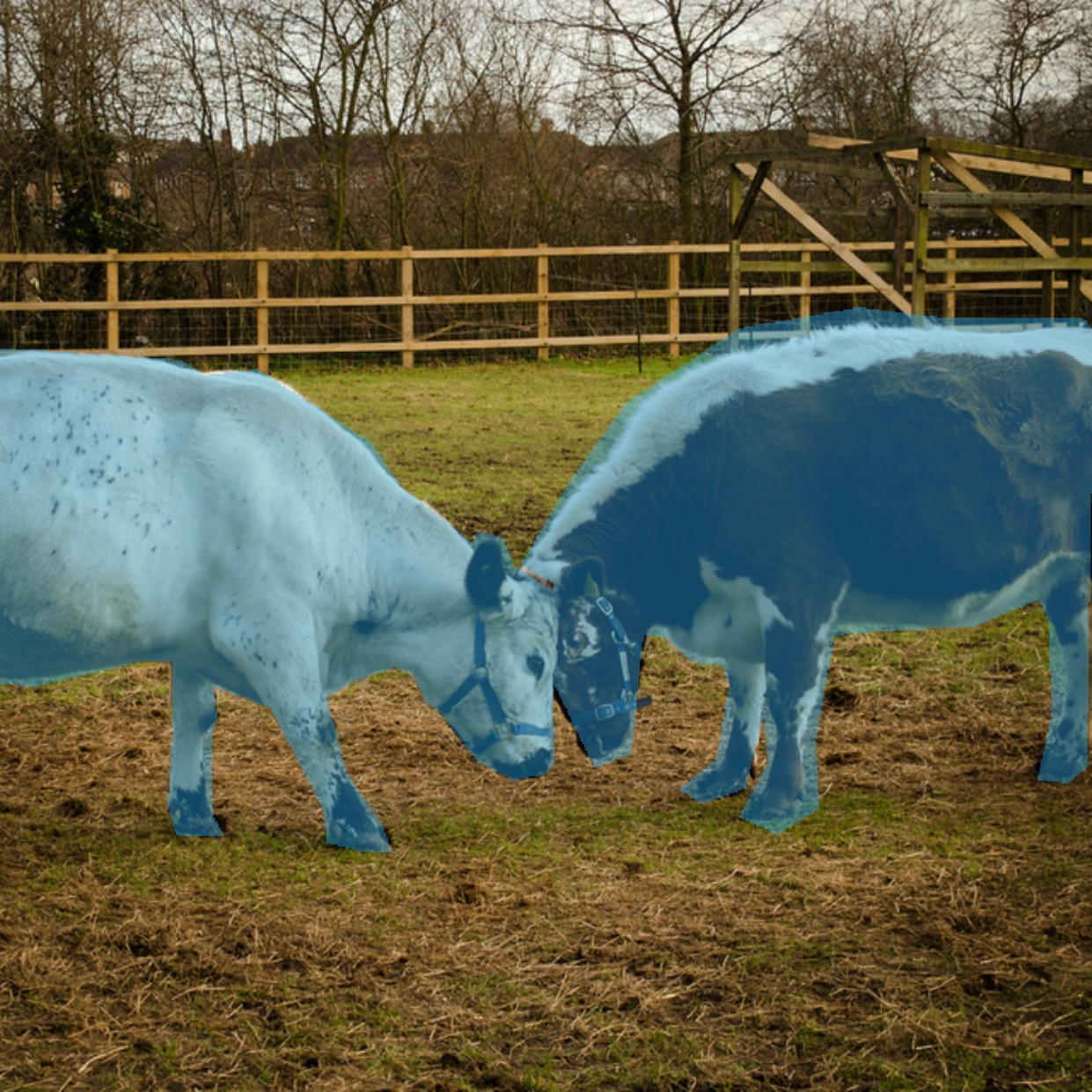} \\
\end{tabular}
\captionof{figure}{One-shot COCO-$20^i$ comparison.}
\label{fig:qual_natural}
\end{minipage}
\end{figure}

To check that the same fixed configuration is not specific to overhead imagery, we apply it without change to the six benchmarks of Section~\ref{sec:setup}, the four natural-image benchmarks PASCAL-$5^i$, COCO-$20^i$, LVIS-$92^i$, and PACO-Part and the two cross-domain benchmarks ISIC and SUIM. Table~\ref{tab:natural} reports the training-free methods at the smallest and the largest support size, the best in each column in bold and the second best underlined, and defers the learning-based methods and the support sizes $k \in \{3, 5\}$ to Appendix~\ref{app:natural}. Averaged over the six benchmarks FROST trails the strongest competitor by 2.1 mIoU at one shot and leads by 3.4 mIoU at ten shots, the variance-driven scaling of Section~\ref{sec:scaling}, far smaller than the 6.7 mIoU lead over INSID3 on the seventeen remote-sensing benchmarks. The gain is largest on the cross-domain benchmarks ISIC and SUIM, where no pretraining prior transfers and FROST leads at both support sizes, extending to PACO-Part and LVIS-$92^i$ by ten shots, while the canonical PASCAL-$5^i$ and COCO-$20^i$ stay with methods that carry natural-image object priors. Figure~\ref{fig:qual_natural} reads this off one-shot COCO-$20^i$ episodes, where FROST recovers the target with boundaries that follow the object, whereas FSSDINO spills into the background and INSID3 leaves parts unlabelled, so the residual gap on these object-centric classes comes from harder scenes with several co-occurring instances rather than gross failures.

\subsection{Ablation}
\label{sec:ablation}
\begin{wraptable}{r}{0.45\textwidth}
\centering
\vspace{-\baselineskip}%   제목 줄과의 위 여백 제거
\small
\setlength{\tabcolsep}{4pt}
\caption{One-shot component ablation on the seventeen remote-sensing benchmarks.}
\label{tab:ablation}
\begin{tabular}{lcc}
\toprule
Variant & Mean mIoU & $\Delta$ \\
\midrule
FROST (full) & 48.3 & --- \\
$-$ bilateral propagation & 48.0 & $-0.3$ \\
$-$ whitening & 47.5 & $-0.8$ \\
$-$ support flip & 47.2 & $-1.1$ \\
$-$ positional debiasing & 46.2 & $-2.1$ \\
$-$ feature refinement & 43.7 & $-4.6$ \\
\bottomrule
\end{tabular}
\end{wraptable}

We remove each fixed component of FROST in turn and report the one-shot mean of Table~\ref{tab:ablation}, holding every other choice at its default, where the feature-refinement variant removes the positional debiasing and the shrinkage whitening at once. Every component contributes and none harms accuracy. The feature refinement is the largest single source, and removing both parts together costs 4.6 mIoU, more than the 2.1 from the positional debiasing and the 0.8 from the whitening alone, so the two are complementary. The horizontal flip adds 1.1 mIoU by doubling the anchors and lowering the variance of the estimate, the whitening adds 0.8 by reweighting class-discriminative directions, and the bilateral propagation adds a smaller 0.3 that sharpens boundaries at negligible cost, an ordering that matches the design, since the steps shaping the feature geometry move accuracy most and the spatial smoothing moves it least. We do not ablate the initial $L^2$ normalization, which is a premise of the pipeline rather than an additive component, because it conditions the feature geometry the later stages operate on. Without it the token covariance behind the positional debiasing is dominated by token magnitude rather than content, so the directions it projects out no longer correspond to position, whereas the kernel compares directions through the cosine similarity and is unaffected by token scale, so the normalization matters for that geometry rather than for the kernel value.

\section{Conclusion}
\label{sec:conclusion}
We presented FROST, a training-free few-shot segmenter that labels every query token by a nonparametric density ratio over the frozen features of a single DINOv3 model, with a threshold the Bayes rule fixes at equal priors and a bandwidth and a spatial gate read from the support set. Describing a class by a distribution rather than a lossy summary turns each added reference into a tighter estimate, so the decision sharpens as annotations accumulate, and across seventeen remote-sensing benchmarks FROST surpasses both training-free and learning-based methods while remaining among the smallest models compared. The same fixed configuration also holds beyond overhead imagery on natural-image and cross-domain benchmarks, where it overtakes the strongest training-free methods as the support set grows, and its advantage is largest in the overhead regime its design targets.

\paragraph{Limitations.} FROST rests on assumptions that bound where it helps. Its decision reads only from the support set, so a single unrepresentative reference gives a poorly estimated density, which is where its remaining failures concentrate and why it trails the strongest competitor at one shot before overtaking it as references accumulate, and since the prediction is no better than the frozen DINOv3 features it reads from, a domain the backbone encodes poorly would limit the density ratio with no training signal to correct it. The equal-prior threshold and the multimodal density are matched to overhead tiles, so on canonical object benchmarks such as PASCAL-$5^i$ and COCO-$20^i$, where a single object occupies a small fraction of the frame, methods that carry natural-image object priors retain a lead. Finally, FROST treats a class as one foreground density against one background density and segments a single class per episode, so multi-class scenes are handled one class at a time, and extending the density ratio to a joint multi-class decision is left to future work.

\clearpage

\bibliographystyle{iclr2026_conference}
\bibliography{iclr2026_conference.bib}

\clearpage

\appendix

\section{Implementation Details and Environment Setup}
\label{app:impl}

\subsection{Hardware and software environment}
All experiments run on a single workstation with $4\times$ NVIDIA RTX~A6000 GPUs (46\,GB each), an AMD Ryzen Threadripper PRO 5975WX (32 cores and 64 threads), and 503\,GB of system memory, under Ubuntu~20.04.6 LTS with NVIDIA driver 535.183.01. Inference for a single method fits on one GPU, and the four GPUs are used only to evaluate different datasets and methods in parallel. The implementation uses Python~3.12, PyTorch~2.7.1 with CUDA~12.6 and cuDNN~9.0.5, and torchvision~0.22.1, together with NumPy~2.2.6, SciPy~1.16.0, Pillow~11.3.0, \texttt{timm}~1.0.26, \texttt{einops}~0.8.2, and scikit-learn~1.8.0. All models run in full \texttt{float32}, with no test-time augmentation, no mixed precision, and no \texttt{torch.compile}.

\subsection{Backbone and features}
The backbone is a frozen general-domain DINOv3 ViT-L/16 pretrained on LVD-1689M, with patch size $16$ and about $304$\,M parameters. Every image is processed at a single $1024\times1024$ scale, which yields a $64\times64$ grid of $1024$-dimensional last-layer patch tokens. Each reference is encoded under two views, the identity and its horizontal flip, with the same flip applied to the image and its mask, which doubles the anchor set to $2k$ views at negligible cost. Patch tokens are $\ell_2$ normalized before all subsequent operations. As summarized in Table~\ref{tab:backbones}, the same frozen DINOv3 ViT-L backbone, with identical settings, is shared by the two training-free baselines on frozen DINOv3 features, INSID3 and FSSDINO, so that the comparison among these three methods isolates the decision rule rather than the backbone, and FROST is among the smallest models in the comparison.

\subsection{Method hyperparameters}
Every hyperparameter is held fixed across all seventeen benchmarks, with no per-dataset tuning and no test-time optimization. Feature refinement applies a positional debiasing that projects out the top $r=250$ singular directions, followed by a within-class Mahalanobis whitening whose within-class scatter is shrunk toward a scaled identity with shrinkage intensity $\lambda=0.95$, so that the empirical scatter retains the weight $1-\lambda=0.05$, and a small floor $\epsilon=10^{-4}$ is added to the diagonal before inversion for numerical stability. The whitened tokens keep their scale, with the bilateral term reading them directly and the density ratio comparing their directions through cosine similarity. The class-conditional density is a mixture of von Mises--Fisher kernels over all foreground or background anchors of the support set, the bandwidth $\sigma$ is selected per episode by the leave-one-out class margin over the grid $\{0.05, 0.1, 0.2, 0.5, 1.0\}$, and the log-density ratio is thresholded at $\tau=0$ under equal class priors. Bilateral label propagation on the $64\times64$ grid runs $T=10$ iterations with mixing weight $\beta=0.70$, feature and colour temperatures $\gamma_f=0.20$ and $\gamma_c=0.05$, and a spatial window radius $d_{\max}=16$ grid positions, with the class-prototype term disabled. Candidate gating intersects a forward cosine test, which admits a position with positive cosine similarity to the foreground prototype, with a backward vote built from the three nearest reference patches per view, and the resulting region is dilated by a support-adaptive radius $\rho = \lfloor \rho_{\max}\, \pi_{\mathrm{FG}} \rceil$ with $\rho_{\max}=4$, where $\pi_{\mathrm{FG}}=N_{\mathrm{FG}}/(N_{\mathrm{FG}}+N_{\mathrm{BG}})$ is the foreground fraction of the reference masks. Finally, the continuous propagated field is bilinearly upsampled to the image resolution and thresholded there, and the result is intersected with the candidate region carried to the same resolution, which recovers sub-patch boundaries.

\begin{table}[t]
\centering
\caption{Backbone and parameter count of each method, where M denotes millions of parameters from the released checkpoint. The upper block lists the learning-based methods and the lower block lists the training-free methods, each ordered by publication date. FROST, INSID3, and FSSDINO share the same frozen DINOv3 ViT-L backbone, so that the comparison among these three isolates the decision rule rather than the backbone, and FROST is among the smallest models. Every method runs forward only at inference with no test-time optimization. For the learning-based methods the count includes the trained components on top of the backbone, and for Matcher and GF-SAM it is the sum of a DINOv2 ViT-L and a SAM ViT-H encoder.}
\label{tab:backbones}
\begin{tabular}{llc}
\toprule
Method & Backbone & Parameters \\
\midrule
\multicolumn{3}{l}{\textit{Learning-based}} \\
SegGPT & ViT-L & 371M \\
SegIC & DINOv2 ViT-L & 310M \\
DiffewS & Stable Diffusion 2.1 & 950M \\
SINE & DINOv2 ViT-L & 323M \\
\midrule
\multicolumn{3}{l}{\textit{Training-free}} \\
PerSAM & SAM ViT-H & 641M \\
Matcher & DINOv2 ViT-L and SAM ViT-H & 946M \\
GF-SAM & DINOv2 ViT-L and SAM ViT-H & 946M \\
INSID3 & DINOv3 ViT-L & 304M \\
FSSDINO & DINOv3 ViT-L & 304M \\
FROST (ours) & DINOv3 ViT-L & 304M \\
\bottomrule
\end{tabular}
\end{table}

\section{Per-support-size results}
\label{app:perk}
\begin{figure}[t]
\centering
\includegraphics[width=0.95\linewidth]{figure/radar_legend.png}\\[3pt]
\begin{subfigure}[b]{0.24\linewidth}\centering
  \includegraphics[width=\linewidth]{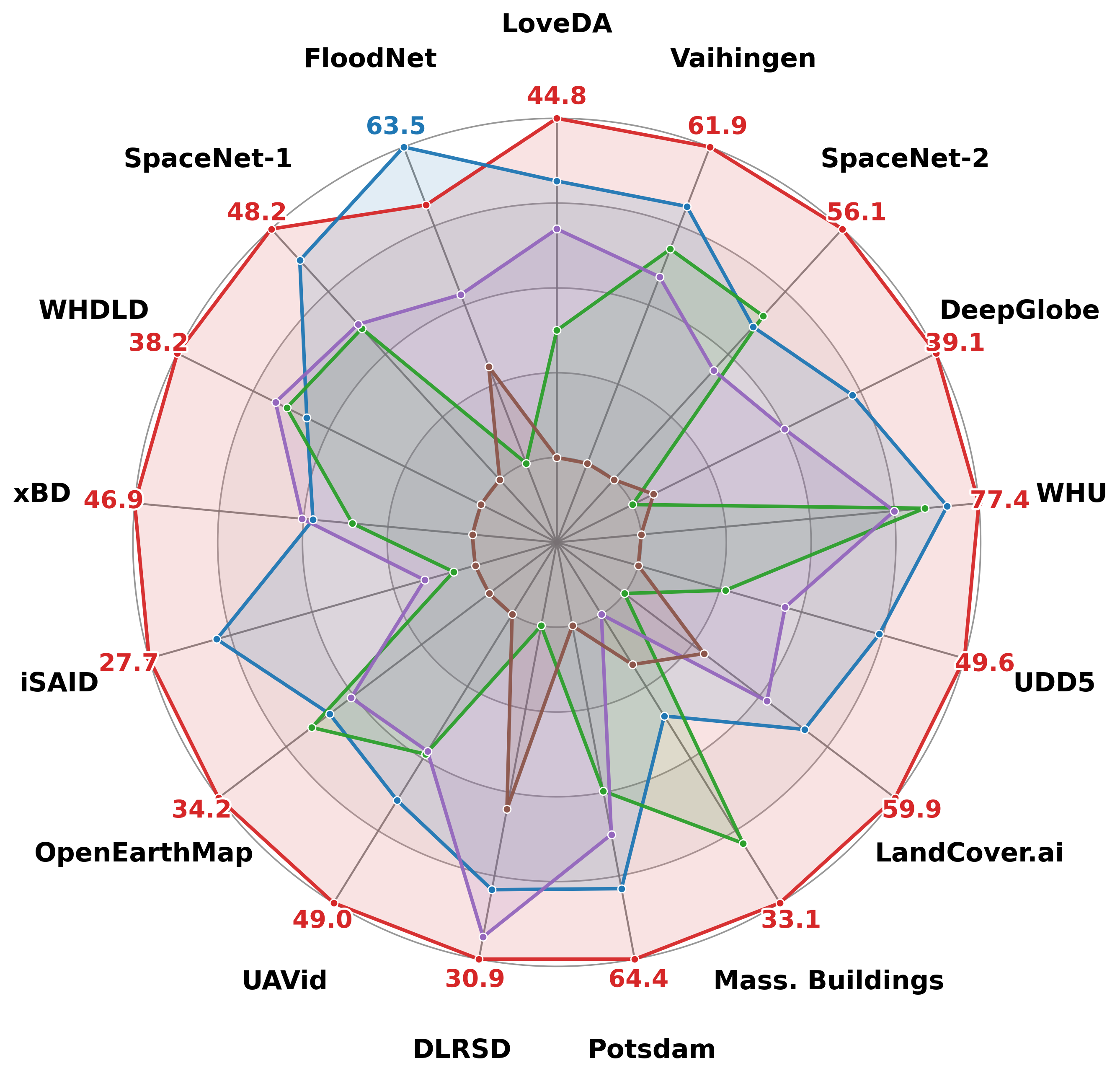}
  \caption{1-shot}\label{fig:radar1}
\end{subfigure}\hfill
\begin{subfigure}[b]{0.24\linewidth}\centering
  \includegraphics[width=\linewidth]{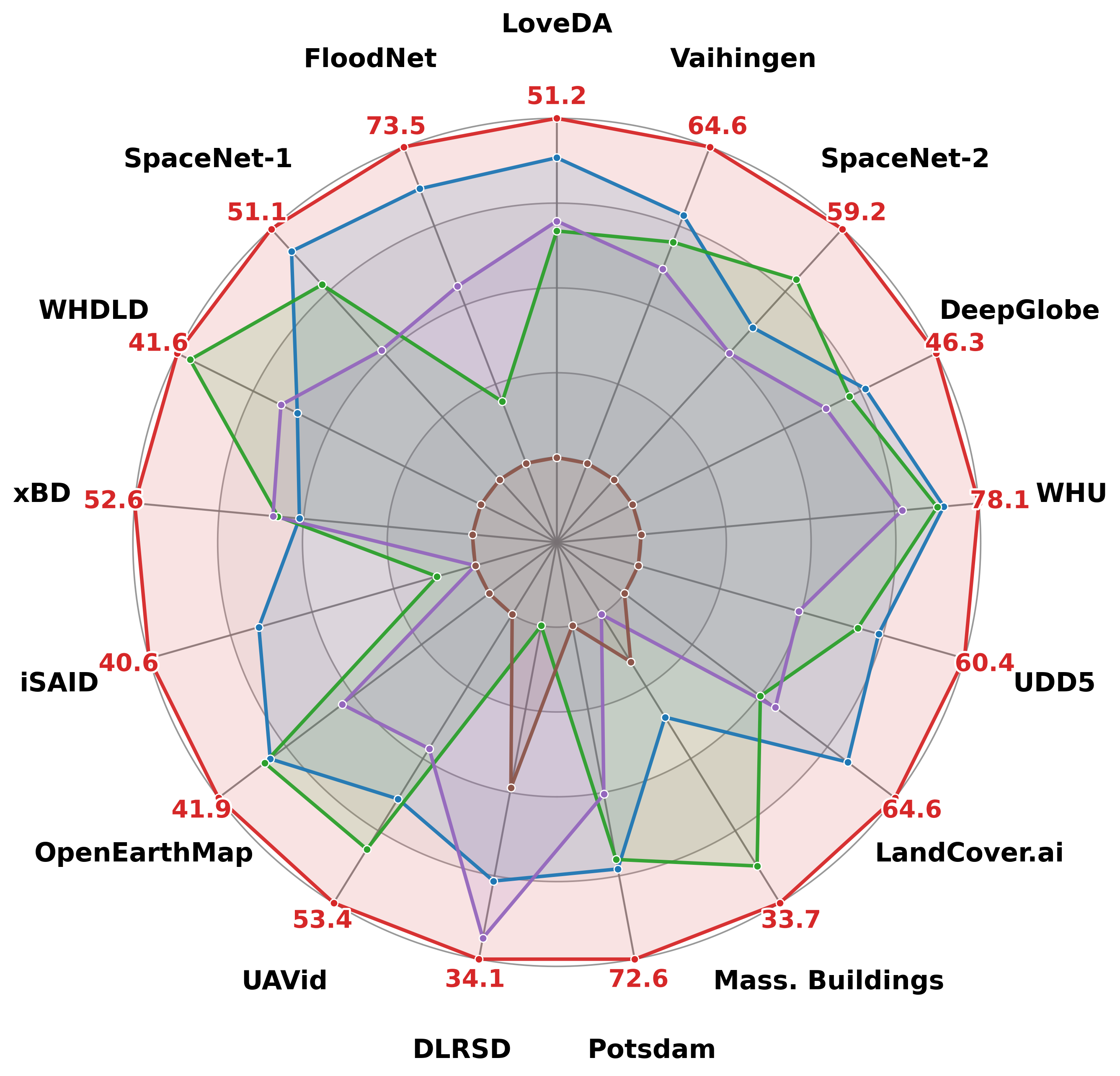}
  \caption{3-shot}\label{fig:radar3}
\end{subfigure}\hfill
\begin{subfigure}[b]{0.24\linewidth}\centering
  \includegraphics[width=\linewidth]{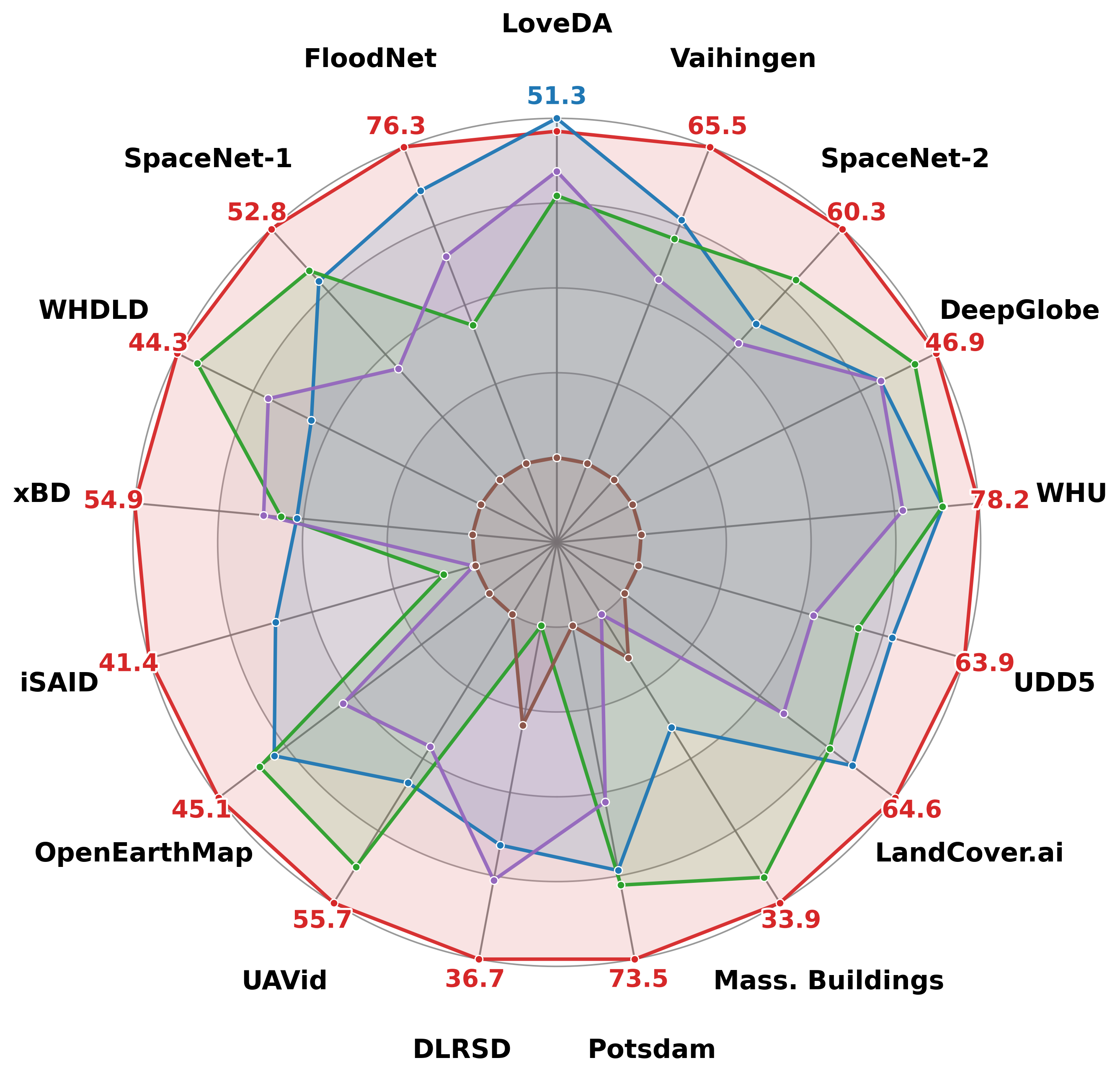}
  \caption{5-shot}\label{fig:radar5}
\end{subfigure}\hfill
\begin{subfigure}[b]{0.24\linewidth}\centering
  \includegraphics[width=\linewidth]{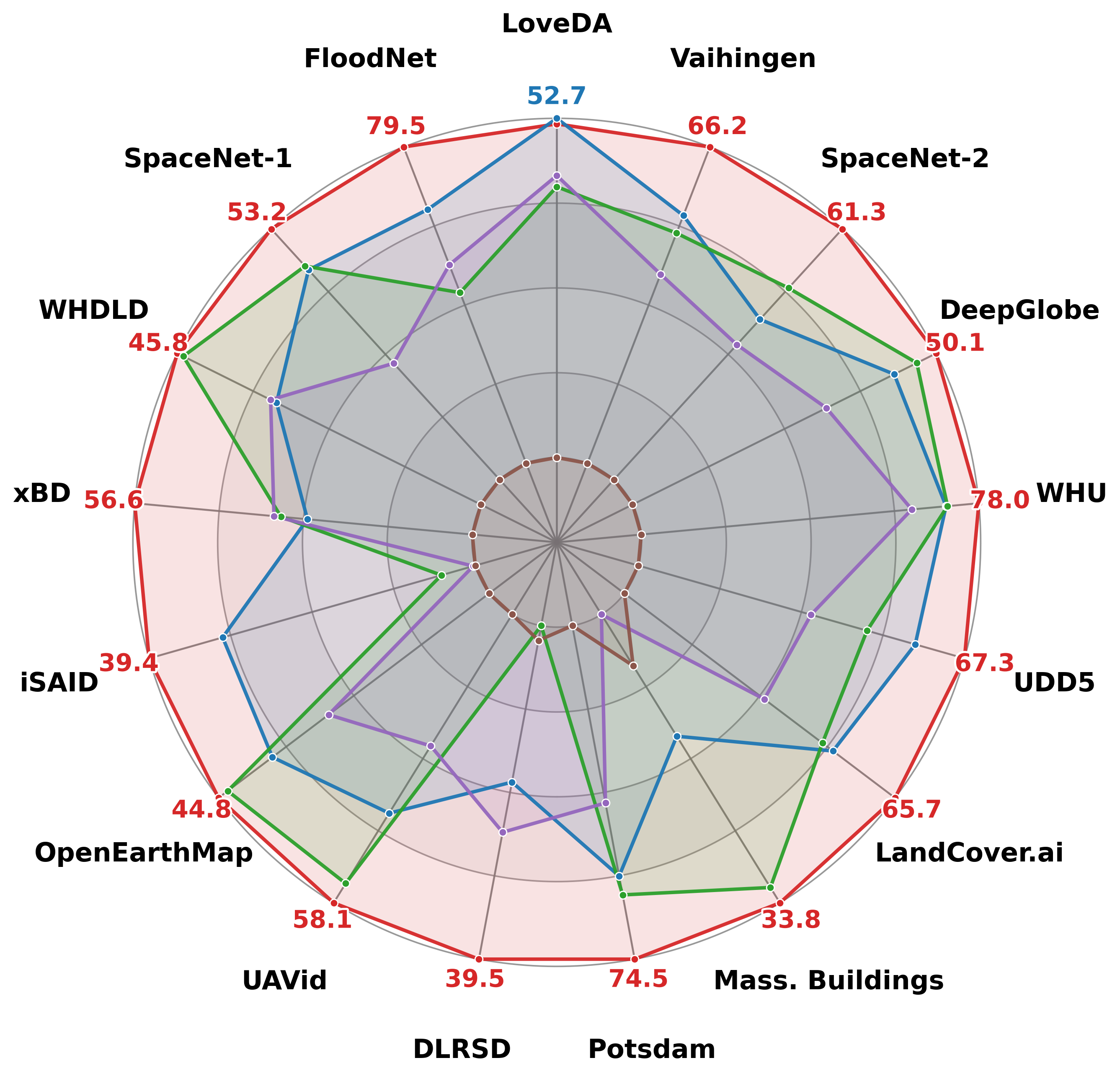}
  \caption{10-shot}\label{fig:radar10}
\end{subfigure}
\caption{Per-benchmark foreground mIoU on the remote-sensing benchmarks for FROST and the four strongest training-free methods at each support size $k \in \{1, 3, 5, 10\}$, with the shared legend at the top.}
\label{fig:radar}
\end{figure}

Figure~\ref{fig:radar} together with Table~\ref{tab:appbw} and Table~\ref{tab:applc} give the foreground mIoU of every method at each support size, the values that the main tables average, with the best value in each column in bold and the second best underlined within each support size. The overall accuracy of FROST increases monotonically with the number of reference masks, from 48.3 mIoU at one shot to 54.1, 55.6, and 56.8 mIoU at three, five, and ten shots, whereas the fine-tuned SegIC peaks at three shots and then declines. FROST records the best result of any method on 13 of the seventeen benchmarks at one shot and on 16 at ten shots, so its lead widens rather than narrows as references accumulate.

\begin{table}[t]
\centering
\caption{Foreground mIoU on the remote-sensing building and footprint benchmarks together with DeepGlobe and FloodNet, reported at each of the four support sizes $k \in \{1, 3, 5, 10\}$, with the best value in each column in bold and the second best underlined within each support size.}
\label{tab:appbw}
\resizebox{\linewidth}{!}{%
\begin{tabular}{l ccccccc}
\toprule
Method & WHU Building & SpaceNet-1 & SpaceNet-2 & Mass. Build. & xBD & DeepGlobe & FloodNet \\
\midrule
\multicolumn{8}{l}{\textit{1-shot}} \\
SegGPT & 60.2 & 38.1 & 41.7 & 11.5 & 28.1 & 23.3 & 44.4 \\
SegIC & 65.3 & \textbf{49.1} & \underline{45.9} & 26.0 & 25.6 & \underline{39.0} & \textbf{63.7} \\
DiffewS & 61.2 & 46.1 & 30.7 & 17.7 & 12.6 & 35.9 & 58.5 \\
SINE & 14.1 & 20.0 & 8.8 & 16.7 & 4.2 & 25.3 & 32.5 \\
PerSAM & 2.7 & 4.9 & 2.6 & 1.3 & 1.5 & 22.6 & 28.5 \\
Matcher & 26.3 & 29.8 & 16.9 & 15.5 & 9.2 & 33.7 & 51.0 \\
GF-SAM & 64.6 & 41.2 & 34.0 & 11.8 & \underline{28.2} & 36.2 & 55.1 \\
FSSDINO & 69.2 & 40.9 & 42.5 & \underline{28.7} & 22.6 & 33.3 & 45.5 \\
INSID3 & \underline{72.6} & 45.9 & 40.8 & 19.3 & 27.0 & 37.5 & \underline{63.5} \\
FROST (ours) & \textbf{77.4} & \underline{48.2} & \textbf{56.1} & \textbf{33.1} & \textbf{46.9} & \textbf{39.1} & 60.2 \\
\midrule
\multicolumn{8}{l}{\textit{3-shot}} \\
SegGPT & 70.6 & 42.1 & 49.4 & 11.2 & \underline{36.8} & 19.7 & 48.6 \\
SegIC & 60.0 & \textbf{51.9} & 49.8 & 23.7 & 33.3 & 40.4 & 68.5 \\
DiffewS & 64.9 & 45.9 & 36.4 & 18.5 & 19.8 & 41.0 & 68.8 \\
SINE & 23.5 & 28.5 & 16.3 & 18.2 & 10.8 & 31.1 & 41.0 \\
PerSAM & 2.4 & 1.4 & 3.1 & 0.5 & 2.3 & 23.2 & 34.4 \\
Matcher & 31.9 & 33.1 & 20.0 & 15.4 & 11.8 & 38.6 & 57.6 \\
GF-SAM & 67.6 & 42.4 & 39.8 & 11.8 & 35.9 & 43.5 & 66.5 \\
FSSDINO & 72.4 & 47.1 & \underline{51.3} & \underline{30.9} & 35.3 & 44.1 & 60.7 \\
INSID3 & \underline{73.3} & 49.5 & 43.8 & 19.6 & 32.7 & \underline{44.5} & \underline{71.4} \\
FROST (ours) & \textbf{78.1} & \underline{51.1} & \textbf{59.2} & \textbf{33.7} & \textbf{52.6} & \textbf{46.3} & \textbf{73.5} \\
\midrule
\multicolumn{8}{l}{\textit{5-shot}} \\
SegGPT & 72.1 & 42.5 & 50.7 & 11.3 & 38.1 & 19.3 & 44.6 \\
SegIC & 54.7 & 48.8 & 48.5 & 21.2 & 34.0 & 40.7 & 66.4 \\
DiffewS & 63.9 & 45.4 & 38.5 & 18.6 & 19.6 & 40.8 & 71.3 \\
SINE & 25.3 & 33.5 & 18.5 & 18.4 & 14.7 & 32.8 & 41.5 \\
PerSAM & 2.7 & 1.4 & 3.2 & 0.4 & 2.7 & 23.1 & 36.0 \\
Matcher & 32.9 & 33.6 & 19.9 & 14.8 & 12.2 & 39.8 & 57.5 \\
GF-SAM & 68.0 & 42.1 & 41.9 & 11.4 & \underline{38.6} & 45.6 & 69.8 \\
FSSDINO & 73.3 & \underline{49.6} & \underline{52.1} & \underline{31.9} & 36.4 & \underline{46.4} & 65.7 \\
INSID3 & \underline{73.4} & 48.8 & 45.0 & 20.2 & 34.4 & 45.6 & \underline{73.7} \\
FROST (ours) & \textbf{78.2} & \textbf{52.8} & \textbf{60.3} & \textbf{33.9} & \textbf{54.9} & \textbf{46.9} & \textbf{76.3} \\
\midrule
\multicolumn{8}{l}{\textit{10-shot}} \\
SegGPT & 72.6 & 42.7 & 50.7 & 11.0 & \underline{39.1} & 20.0 & 45.8 \\
SegIC & 45.9 & 42.4 & 43.2 & 18.6 & 31.8 & 39.9 & 61.9 \\
DiffewS & 62.2 & 44.7 & 38.6 & 18.4 & 17.4 & 41.2 & 70.4 \\
SINE & 25.8 & 34.9 & 19.7 & 18.6 & 15.4 & 33.5 & 42.0 \\
PerSAM & 2.5 & 1.9 & 2.9 & 0.2 & 2.6 & 22.6 & 37.2 \\
Matcher & 31.7 & 33.6 & 19.7 & 15.3 & 13.2 & 40.7 & 56.7 \\
GF-SAM & 68.8 & 42.7 & 42.1 & 11.3 & 38.7 & 46.7 & 71.0 \\
FSSDINO & \underline{73.7} & \underline{50.3} & \underline{51.5} & \underline{32.6} & 37.8 & \underline{49.5} & 69.0 \\
INSID3 & 73.5 & 50.0 & 46.3 & 20.8 & 34.4 & 48.8 & \underline{75.0} \\
FROST (ours) & \textbf{78.0} & \textbf{53.2} & \textbf{61.3} & \textbf{33.8} & \textbf{56.6} & \textbf{50.1} & \textbf{79.5} \\
\bottomrule
\end{tabular}
}
\end{table}

\begin{table}[t]
\centering
\caption{Foreground mIoU on the remote-sensing land-cover and drone benchmarks, reported at each of the four support sizes $k \in \{1, 3, 5, 10\}$, with the best value in each column in bold and the second best underlined within each support size.}
\label{tab:applc}
\resizebox{\linewidth}{!}{%
\begin{tabular}{l cccccccccc}
\toprule
Method & LoveDA & Potsdam & Vaihingen & LandCover.ai & OpenEarthMap & UAVid & UDD5 & iSAID & DLRSD & WHDLD \\
\midrule
\multicolumn{11}{l}{\textit{1-shot}} \\
SegGPT & 40.3 & 38.5 & 46.8 & 45.0 & 21.1 & 33.4 & 29.3 & 21.9 & 20.8 & 27.5 \\
SegIC & 34.1 & \underline{59.6} & 52.9 & 50.6 & 28.3 & \underline{51.7} & \underline{49.4} & 20.7 & \textbf{31.4} & 33.7 \\
DiffewS & 27.4 & 44.6 & 45.6 & 45.7 & 22.4 & \textbf{52.4} & 37.8 & 10.7 & 29.9 & 28.2 \\
SINE & 13.5 & 28.6 & 20.3 & 32.2 & 17.1 & 20.0 & 22.5 & 3.2 & 15.6 & 22.2 \\
PerSAM & 13.9 & 19.3 & 9.5 & 34.7 & 6.8 & 14.2 & 15.7 & 5.6 & 20.5 & 18.7 \\
Matcher & 26.4 & 39.8 & 30.0 & 42.6 & 19.1 & 29.6 & 31.6 & 14.2 & 25.5 & 24.6 \\
GF-SAM & 38.8 & 55.2 & 48.8 & 48.3 & 26.8 & 38.8 & 39.7 & 16.3 & 30.1 & \underline{33.8} \\
FSSDINO & 33.3 & 52.0 & 51.6 & 35.4 & \underline{29.0} & 39.0 & 36.4 & 15.1 & 18.9 & 33.3 \\
INSID3 & \underline{41.4} & 59.2 & \underline{55.9} & \underline{51.7} & 28.0 & 42.1 & 44.9 & \underline{24.9} & 28.4 & 32.4 \\
FROST (ours) & \textbf{44.8} & \textbf{64.4} & \textbf{61.9} & \textbf{59.9} & \textbf{34.2} & 49.0 & \textbf{49.6} & \textbf{27.7} & \underline{30.9} & \textbf{38.2} \\
\midrule
\multicolumn{11}{l}{\textit{3-shot}} \\
SegGPT & 45.7 & 39.5 & 51.2 & 47.0 & 21.6 & 31.2 & 27.3 & 22.7 & 20.5 & 25.2 \\
SegIC & 35.1 & 62.2 & 53.2 & 58.8 & 29.7 & \underline{54.1} & \underline{54.9} & 24.6 & 33.0 & 34.4 \\
DiffewS & 30.9 & 55.5 & 48.7 & 56.5 & 24.5 & \textbf{57.4} & 46.6 & 12.8 & 32.1 & 30.9 \\
SINE & 14.9 & 35.3 & 23.6 & 35.3 & 21.2 & 25.2 & 28.4 & 5.4 & 16.6 & 24.4 \\
PerSAM & 14.1 & 21.1 & 9.5 & 37.1 & 7.8 & 16.9 & 18.2 & 8.4 & 20.2 & 18.7 \\
Matcher & 27.1 & 44.5 & 32.7 & 42.9 & 22.0 & 36.2 & 33.8 & 20.1 & 28.4 & 24.9 \\
GF-SAM & 43.9 & 58.7 & 52.3 & 55.0 & 32.8 & 44.2 & 46.9 & 20.1 & \underline{33.4} & 35.9 \\
FSSDINO & 43.2 & 64.2 & 55.0 & 53.8 & \underline{38.5} & 50.2 & 51.7 & 22.5 & 23.0 & \underline{40.9} \\
INSID3 & \underline{48.4} & \underline{65.0} & \underline{57.7} & \underline{60.8} & 38.1 & 47.2 & 53.4 & \underline{33.7} & 31.5 & 35.0 \\
FROST (ours) & \textbf{51.2} & \textbf{72.6} & \textbf{64.6} & \textbf{64.6} & \textbf{41.9} & 53.4 & \textbf{60.4} & \textbf{40.6} & \textbf{34.1} & \textbf{41.6} \\
\midrule
\multicolumn{11}{l}{\textit{5-shot}} \\
SegGPT & 47.0 & 39.4 & 51.8 & 47.4 & 20.3 & 30.3 & 26.6 & 23.9 & 20.1 & 23.7 \\
SegIC & 34.1 & 62.3 & 50.7 & 58.5 & 29.4 & 53.6 & 54.9 & 25.1 & 32.9 & 34.8 \\
DiffewS & 30.7 & 55.1 & 47.8 & 49.4 & 23.5 & \textbf{57.7} & 47.6 & 13.1 & 31.9 & 32.0 \\
SINE & 15.6 & 35.9 & 24.2 & 35.4 & 21.8 & 26.9 & 29.0 & 6.2 & 16.7 & 25.0 \\
PerSAM & 14.8 & 21.7 & 9.2 & 36.9 & 5.5 & 17.4 & 18.3 & 8.6 & 20.8 & 19.7 \\
Matcher & 27.6 & 43.8 & 33.5 & 43.0 & 20.8 & 38.9 & 35.0 & 21.8 & 28.7 & 24.6 \\
GF-SAM & 47.6 & 59.5 & 52.1 & 55.7 & 33.9 & 46.6 & 50.5 & 21.9 & \underline{34.0} & 38.4 \\
FSSDINO & 45.9 & \underline{66.9} & 56.2 & 59.4 & \underline{41.4} & 53.6 & 54.5 & 23.7 & 25.3 & \underline{43.0} \\
INSID3 & \textbf{51.3} & 65.6 & \underline{58.1} & \underline{61.2} & 40.1 & 48.7 & \underline{57.5} & \underline{33.8} & 32.8 & 35.6 \\
FROST (ours) & \underline{50.4} & \textbf{73.5} & \textbf{65.5} & \textbf{64.6} & \textbf{45.1} & \underline{55.7} & \textbf{63.9} & \textbf{41.4} & \textbf{36.7} & \textbf{44.3} \\
\midrule
\multicolumn{11}{l}{\textit{10-shot}} \\
SegGPT & 48.3 & 39.5 & 52.9 & 48.5 & 19.5 & 30.6 & 26.3 & 20.2 & 19.1 & 22.6 \\
SegIC & 34.3 & 60.6 & 47.2 & 57.8 & 29.3 & 52.8 & 53.6 & 23.5 & 32.3 & 33.5 \\
DiffewS & 29.1 & 53.9 & 45.8 & 45.7 & 22.7 & \underline{56.9} & 46.6 & 14.0 & 31.3 & 31.1 \\
SINE & 17.4 & 36.2 & 24.3 & 36.4 & 22.4 & 28.3 & 28.8 & 7.1 & 16.0 & 25.2 \\
PerSAM & 14.3 & 23.6 & 8.5 & 37.0 & 7.0 & 17.3 & 19.3 & 7.0 & 21.2 & 19.6 \\
Matcher & 28.4 & 44.4 & 33.9 & 45.2 & 21.2 & 38.8 & 33.5 & 23.0 & 28.7 & 25.3 \\
GF-SAM & 48.6 & 60.4 & 53.2 & 55.8 & 35.2 & 47.6 & 51.4 & 23.1 & \underline{35.2} & 39.5 \\
FSSDINO & 47.8 & \underline{68.7} & 57.4 & 60.2 & \underline{44.0} & 56.8 & 57.2 & 24.7 & 28.2 & \underline{45.4} \\
INSID3 & \textbf{52.7} & 67.0 & \underline{59.2} & \underline{61.0} & 40.1 & 52.1 & \underline{62.2} & \underline{35.7} & 33.5 & 39.1 \\
FROST (ours) & \underline{52.3} & \textbf{74.5} & \textbf{66.2} & \textbf{65.7} & \textbf{44.8} & \textbf{58.1} & \textbf{67.3} & \textbf{39.4} & \textbf{39.5} & \textbf{45.8} \\
\bottomrule
\end{tabular}
}
\end{table}

\section{Generalization beyond overhead imagery}
\label{app:natural}
Table~\ref{tab:naturalk} gives the foreground mIoU of every method on the six benchmarks at each support size, the values that the main comparison in Table~\ref{tab:natural} draws from, with the best value in each column in bold and the second best underlined within each support size. The accuracy of FROST rises with the number of reference masks as it does on the remote-sensing benchmarks, and among the training-free methods FROST moves from below the strongest competitor at one shot to the best overall accuracy at ten shots, while on the canonical object benchmarks PASCAL-$5^i$ and COCO-$20^i$ the lead stays with methods that carry natural-image object priors at every support size.

\begin{table}[t]
\centering
\caption{Foreground mIoU on the six benchmarks beyond overhead imagery at each of the four support sizes $k \in \{1, 3, 5, 10\}$, with the best value in each column in bold and the second best underlined within each support size. The learning-based methods SegGPT, SegIC, DiffewS, and SINE are trained on segmentation data that overlaps the natural-image benchmarks, so they are shown in grey and excluded from the best and second-best marking, which is computed over the training-free methods alone.}
\label{tab:naturalk}
\resizebox{\linewidth}{!}{%
\begin{tabular}{l cccccc}
\toprule
Method & PASCAL-$5^i$ & COCO-$20^i$ & LVIS-$92^i$ & PACO-Part & ISIC & SUIM \\
\midrule
\multicolumn{7}{l}{\textit{1-shot}} \\
\textcolor{gray}{SegGPT} & \textcolor{gray}{81.5} & \textcolor{gray}{53.9} & \textcolor{gray}{18.3} & \textcolor{gray}{17.7} & \textcolor{gray}{32.4} & \textcolor{gray}{38.7} \\
\textcolor{gray}{SegIC} & \textcolor{gray}{85.6} & \textcolor{gray}{71.1} & \textcolor{gray}{41.4} & \textcolor{gray}{27.3} & \textcolor{gray}{27.1} & \textcolor{gray}{59.8} \\
\textcolor{gray}{DiffewS} & \textcolor{gray}{88.2} & \textcolor{gray}{50.7} & \textcolor{gray}{31.5} & \textcolor{gray}{22.9} & \textcolor{gray}{27.4} & \textcolor{gray}{51.9} \\
\textcolor{gray}{SINE} & \textcolor{gray}{33.5} & \textcolor{gray}{18.9} & \textcolor{gray}{6.8} & \textcolor{gray}{14.9} & \textcolor{gray}{28.5} & \textcolor{gray}{28.1} \\
PerSAM & 45.6 & 21.8 & 13.3 & 20.8 & 22.7 & 29.6 \\
Matcher & 68.2 & 54.6 & 36.5 & 34.3 & 33.4 & 46.7 \\
GF-SAM & \textbf{71.9} & \textbf{59.2} & \underline{37.9} & 38.1 & 48.9 & 55.0 \\
FSSDINO & 65.5 & 39.9 & 22.2 & 34.3 & \underline{52.3} & 51.3 \\
INSID3 & 69.9 & \underline{56.5} & \textbf{40.9} & \textbf{39.6} & 50.5 & \underline{56.2} \\
FROST (ours) & \underline{70.1} & 48.4 & 32.8 & \underline{39.1} & \textbf{52.6} & \textbf{58.0} \\
\midrule
\multicolumn{7}{l}{\textit{3-shot}} \\
\textcolor{gray}{SegGPT} & \textcolor{gray}{84.7} & \textcolor{gray}{55.2} & \textcolor{gray}{19.1} & \textcolor{gray}{15.0} & \textcolor{gray}{43.5} & \textcolor{gray}{36.1} \\
\textcolor{gray}{SegIC} & \textcolor{gray}{87.0} & \textcolor{gray}{70.3} & \textcolor{gray}{41.4} & \textcolor{gray}{29.7} & \textcolor{gray}{24.6} & \textcolor{gray}{61.6} \\
\textcolor{gray}{DiffewS} & \textcolor{gray}{89.0} & \textcolor{gray}{57.8} & \textcolor{gray}{33.1} & \textcolor{gray}{24.8} & \textcolor{gray}{25.8} & \textcolor{gray}{62.5} \\
\textcolor{gray}{SINE} & \textcolor{gray}{36.0} & \textcolor{gray}{23.5} & \textcolor{gray}{7.9} & \textcolor{gray}{16.1} & \textcolor{gray}{31.1} & \textcolor{gray}{36.6} \\
PerSAM & 51.0 & 30.0 & 16.1 & 23.9 & 23.6 & 36.6 \\
Matcher & 72.7 & 59.3 & 41.6 & 35.7 & 33.4 & 49.5 \\
GF-SAM & \textbf{80.8} & \textbf{65.5} & \underline{44.6} & 41.7 & 53.1 & \textbf{66.8} \\
FSSDINO & 75.9 & 52.5 & 29.4 & 41.9 & 60.3 & 63.0 \\
INSID3 & 77.9 & \underline{64.3} & \textbf{45.2} & \underline{44.4} & \underline{62.4} & \textbf{66.8} \\
FROST (ours) & \underline{80.5} & 59.5 & 43.5 & \textbf{47.9} & \textbf{66.0} & \underline{66.2} \\
\midrule
\multicolumn{7}{l}{\textit{5-shot}} \\
\textcolor{gray}{SegGPT} & \textcolor{gray}{85.6} & \textcolor{gray}{55.6} & \textcolor{gray}{19.7} & \textcolor{gray}{14.6} & \textcolor{gray}{43.7} & \textcolor{gray}{40.8} \\
\textcolor{gray}{SegIC} & \textcolor{gray}{86.2} & \textcolor{gray}{68.0} & \textcolor{gray}{39.7} & \textcolor{gray}{30.5} & \textcolor{gray}{26.2} & \textcolor{gray}{60.7} \\
\textcolor{gray}{DiffewS} & \textcolor{gray}{88.6} & \textcolor{gray}{58.1} & \textcolor{gray}{34.7} & \textcolor{gray}{25.5} & \textcolor{gray}{28.8} & \textcolor{gray}{62.6} \\
\textcolor{gray}{SINE} & \textcolor{gray}{36.7} & \textcolor{gray}{22.9} & \textcolor{gray}{8.8} & \textcolor{gray}{17.3} & \textcolor{gray}{29.3} & \textcolor{gray}{37.1} \\
PerSAM & 55.0 & 31.3 & 17.3 & 25.0 & 26.3 & 34.0 \\
Matcher & 73.9 & 57.1 & 41.2 & 33.3 & 34.7 & 53.2 \\
GF-SAM & \textbf{83.1} & \textbf{65.8} & \underline{46.5} & 42.2 & 55.3 & \underline{65.1} \\
FSSDINO & 78.9 & 53.8 & 32.4 & 45.6 & 62.8 & 62.7 \\
INSID3 & 81.0 & \underline{63.0} & 46.0 & \underline{47.2} & \underline{63.8} & 64.1 \\
FROST (ours) & \underline{82.5} & 60.5 & \textbf{47.1} & \textbf{52.1} & \textbf{66.3} & \textbf{68.8} \\
\midrule
\multicolumn{7}{l}{\textit{10-shot}} \\
\textcolor{gray}{SegGPT} & \textcolor{gray}{86.6} & \textcolor{gray}{59.1} & \textcolor{gray}{20.2} & \textcolor{gray}{16.4} & \textcolor{gray}{46.1} & \textcolor{gray}{36.6} \\
\textcolor{gray}{SegIC} & \textcolor{gray}{85.9} & \textcolor{gray}{66.6} & \textcolor{gray}{37.7} & \textcolor{gray}{31.3} & \textcolor{gray}{26.2} & \textcolor{gray}{56.4} \\
\textcolor{gray}{DiffewS} & \textcolor{gray}{86.8} & \textcolor{gray}{60.4} & \textcolor{gray}{33.4} & \textcolor{gray}{27.0} & \textcolor{gray}{29.6} & \textcolor{gray}{57.8} \\
\textcolor{gray}{SINE} & \textcolor{gray}{37.4} & \textcolor{gray}{21.6} & \textcolor{gray}{10.6} & \textcolor{gray}{18.5} & \textcolor{gray}{32.8} & \textcolor{gray}{37.3} \\
PerSAM & 55.0 & 29.9 & 16.4 & 27.1 & 26.9 & 35.9 \\
Matcher & 75.4 & 61.7 & 39.4 & 34.3 & 36.2 & 48.8 \\
GF-SAM & \textbf{83.9} & \textbf{68.1} & \underline{48.1} & 45.4 & 60.7 & 63.2 \\
FSSDINO & 80.2 & 56.7 & 35.3 & \underline{50.6} & 65.4 & 63.2 \\
INSID3 & 81.5 & 64.0 & 46.8 & 50.3 & \underline{67.0} & \underline{67.6} \\
FROST (ours) & \underline{83.7} & \underline{64.3} & \textbf{51.9} & \textbf{57.0} & \textbf{70.4} & \textbf{70.1} \\
\bottomrule
\end{tabular}
}
\end{table}

\section{Qualitative results}
\label{app:qualitative}

\begin{figure}[t]
\centering
\setlength{\tabcolsep}{1.5pt}
\renewcommand{\arraystretch}{0.4}
\begin{tabular}{ccc@{\hspace{10pt}}ccc}
\footnotesize Reference & \footnotesize GT & \footnotesize Ours &
\footnotesize Reference & \footnotesize GT & \footnotesize Ours \\[2pt]
\includegraphics[width=0.155\linewidth]{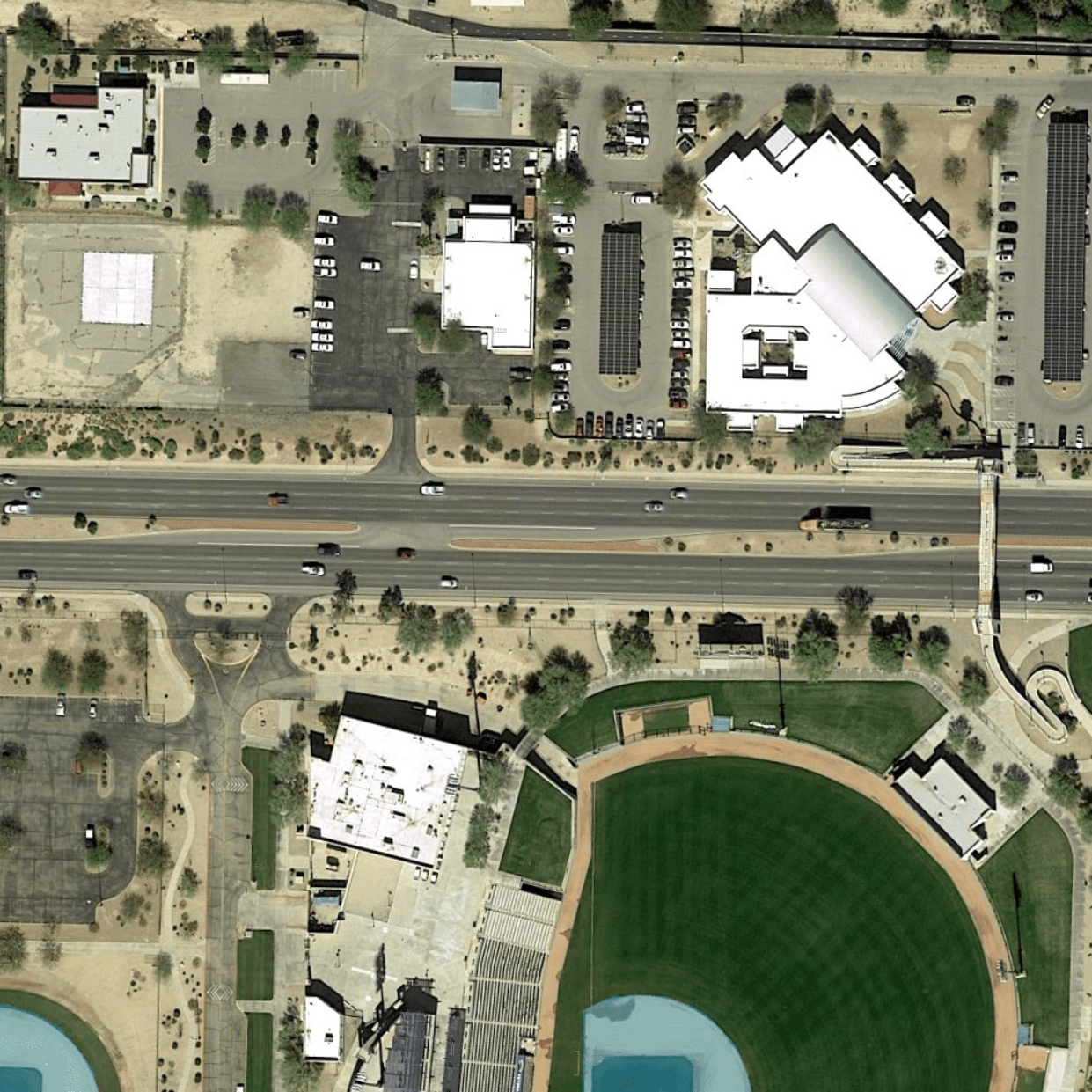} &
\includegraphics[width=0.155\linewidth]{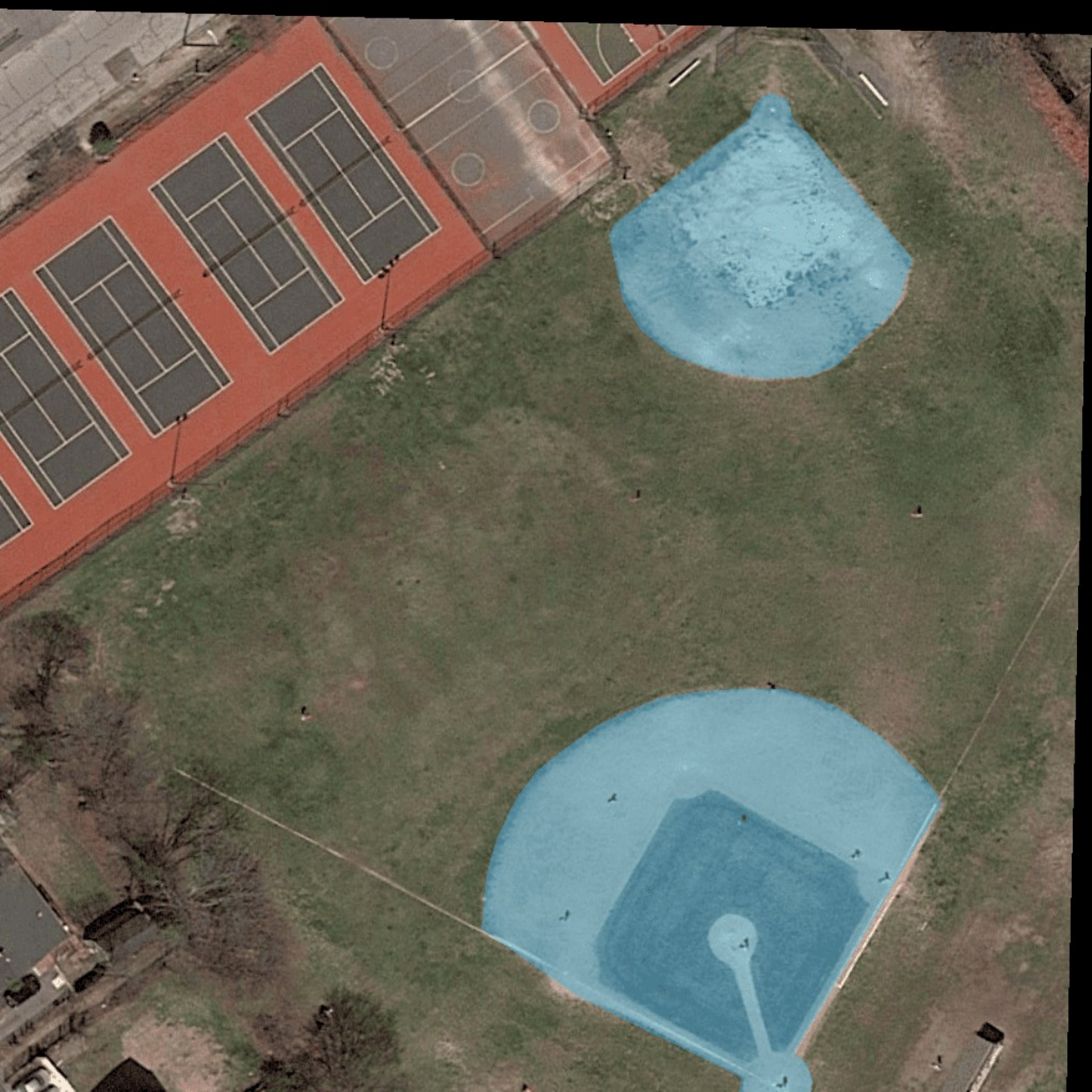} &
\includegraphics[width=0.155\linewidth]{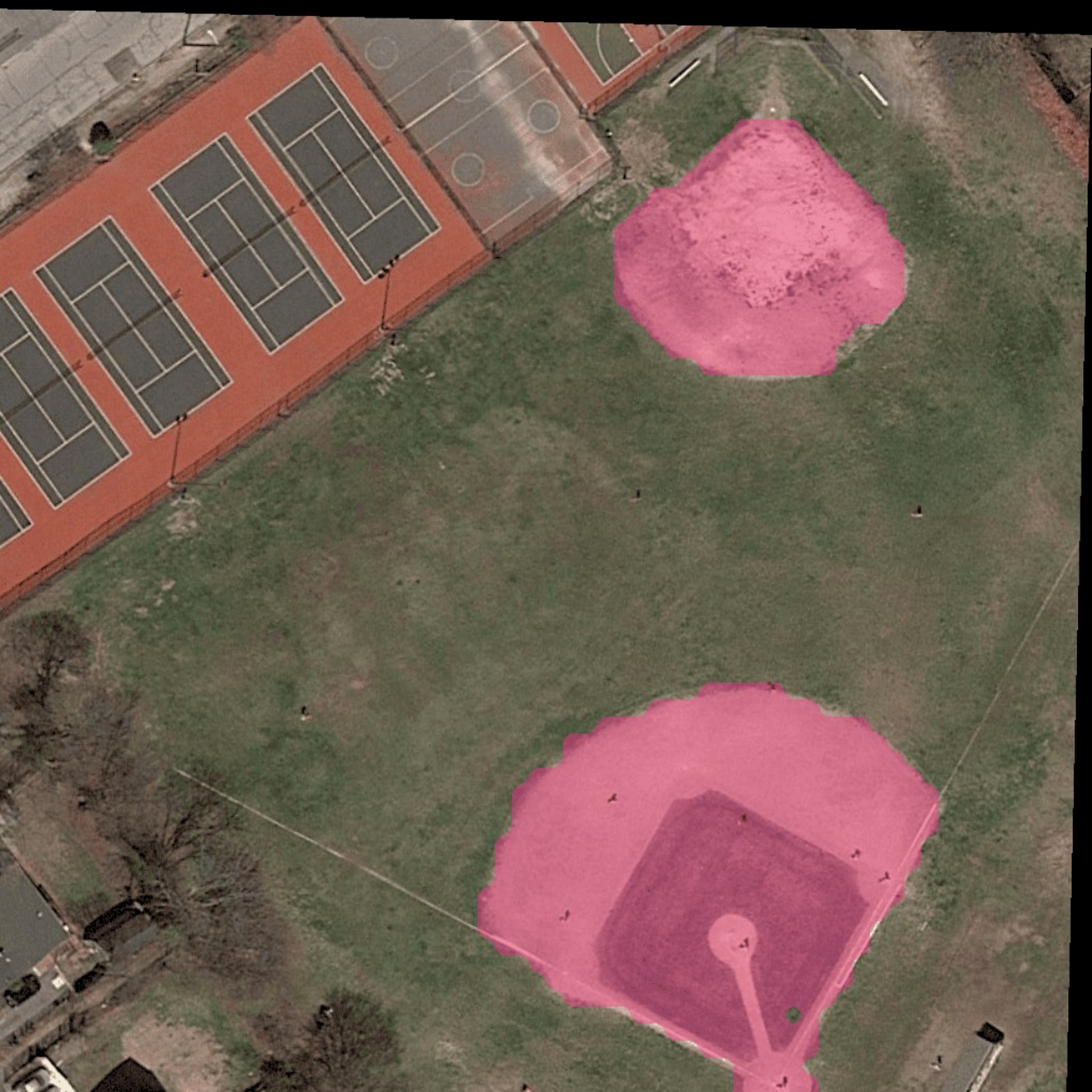} &
\includegraphics[width=0.155\linewidth]{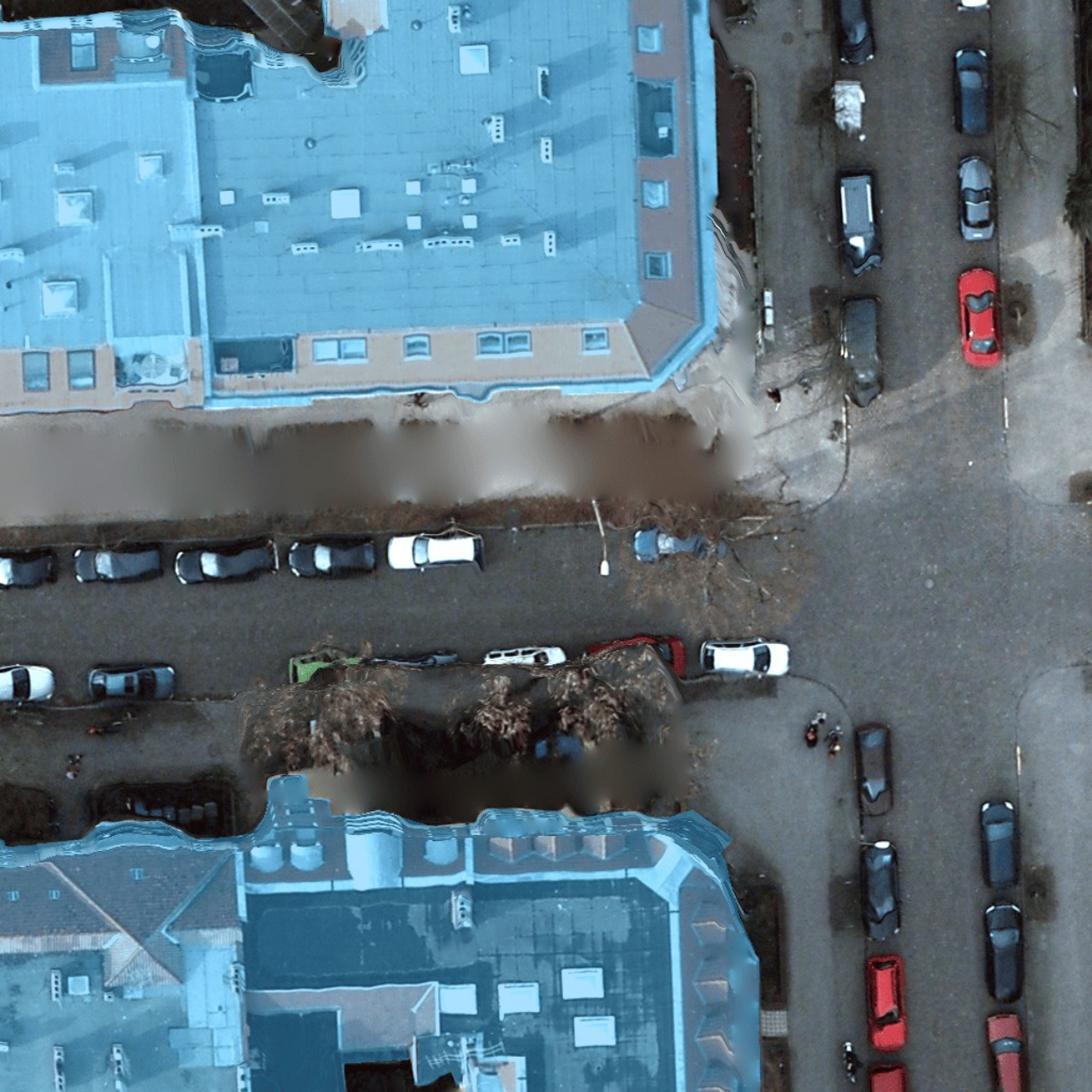} &
\includegraphics[width=0.155\linewidth]{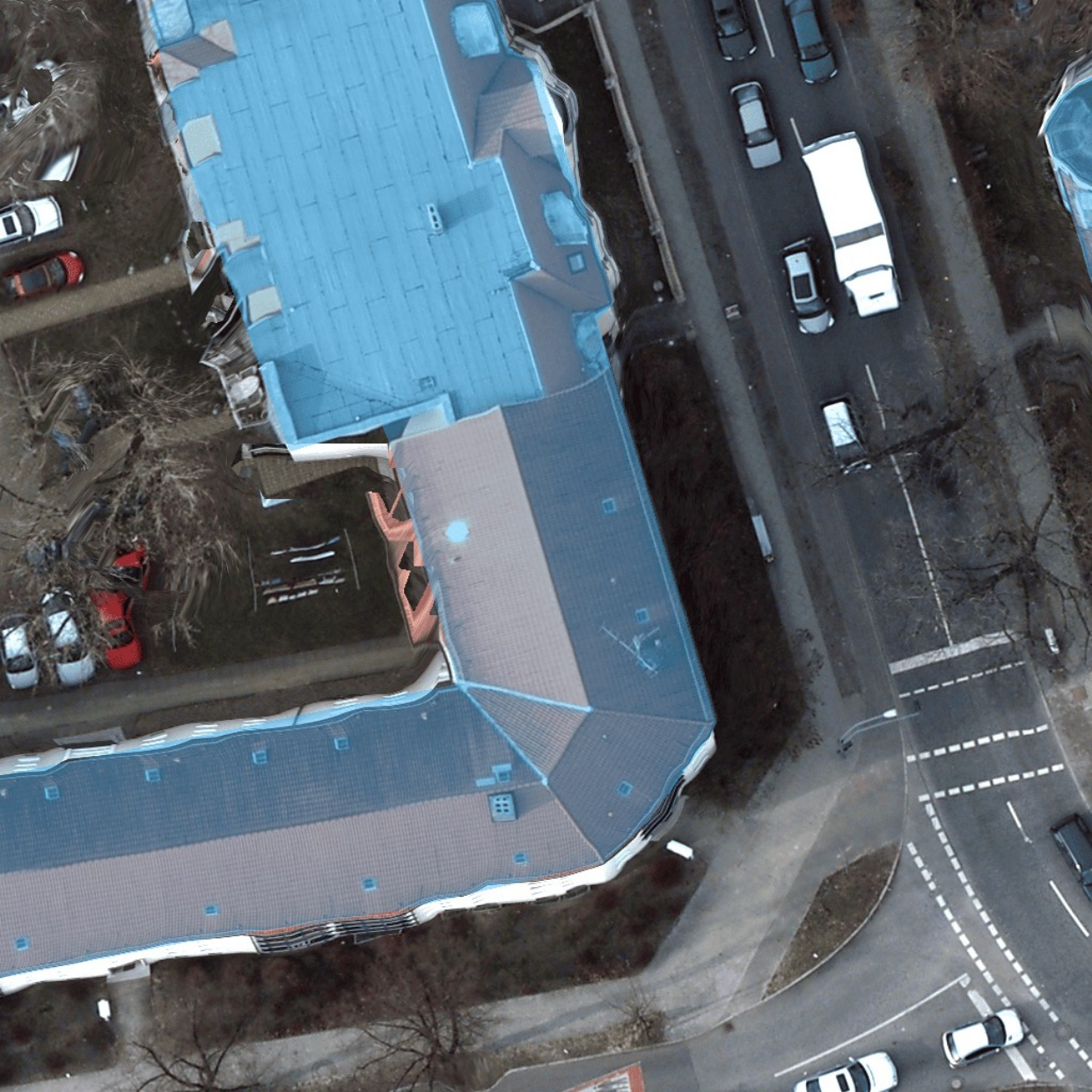} &
\includegraphics[width=0.155\linewidth]{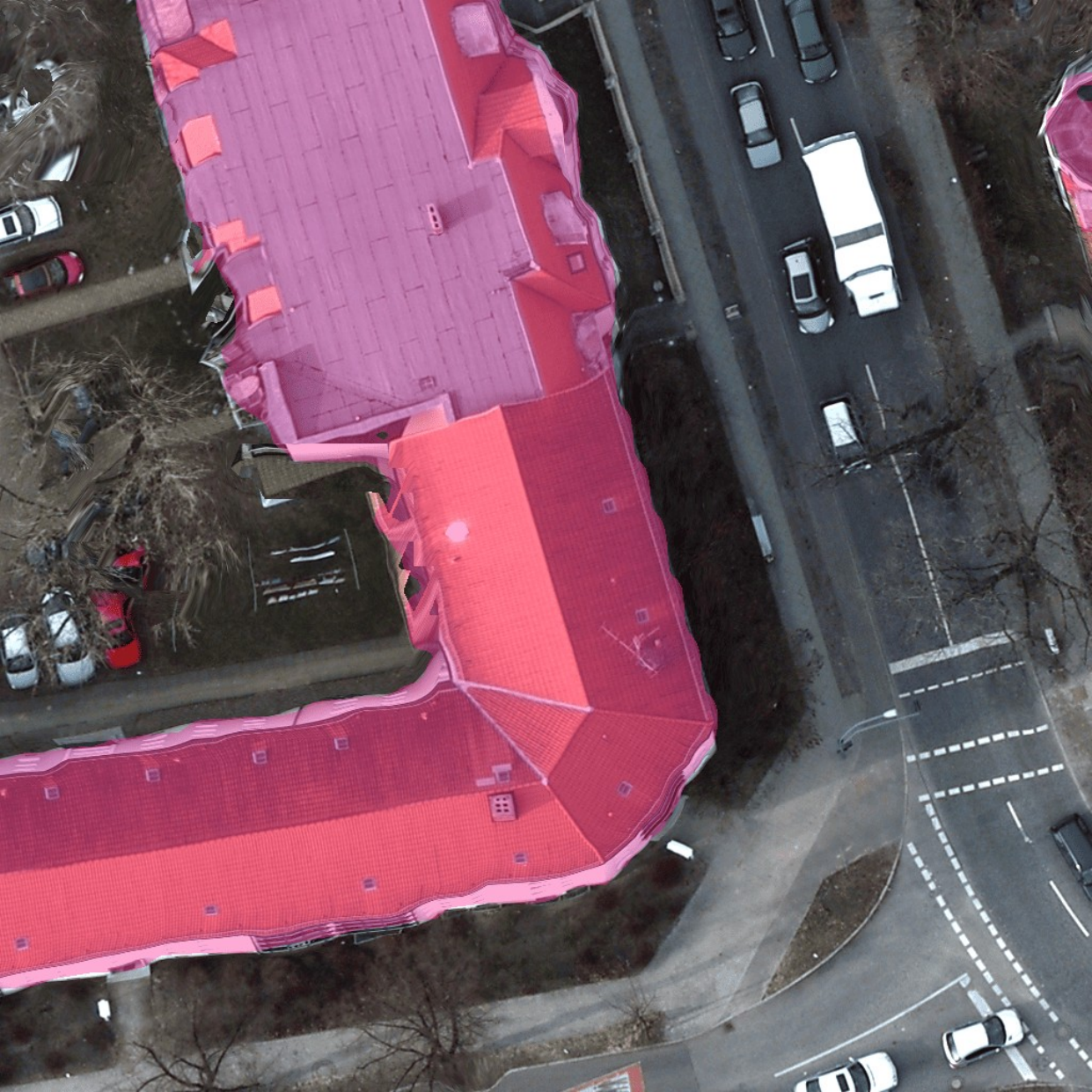} \\
\includegraphics[width=0.155\linewidth]{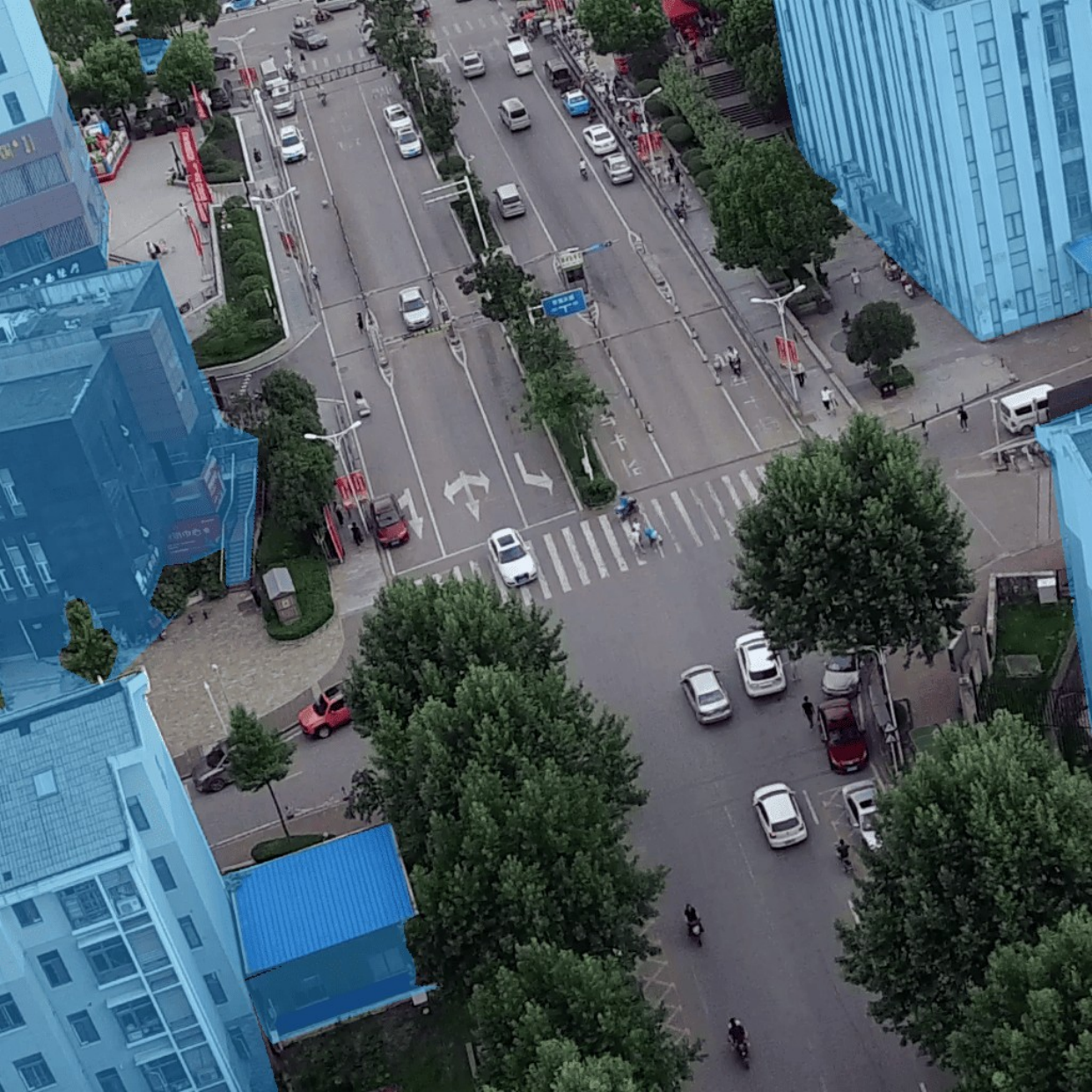} &
\includegraphics[width=0.155\linewidth]{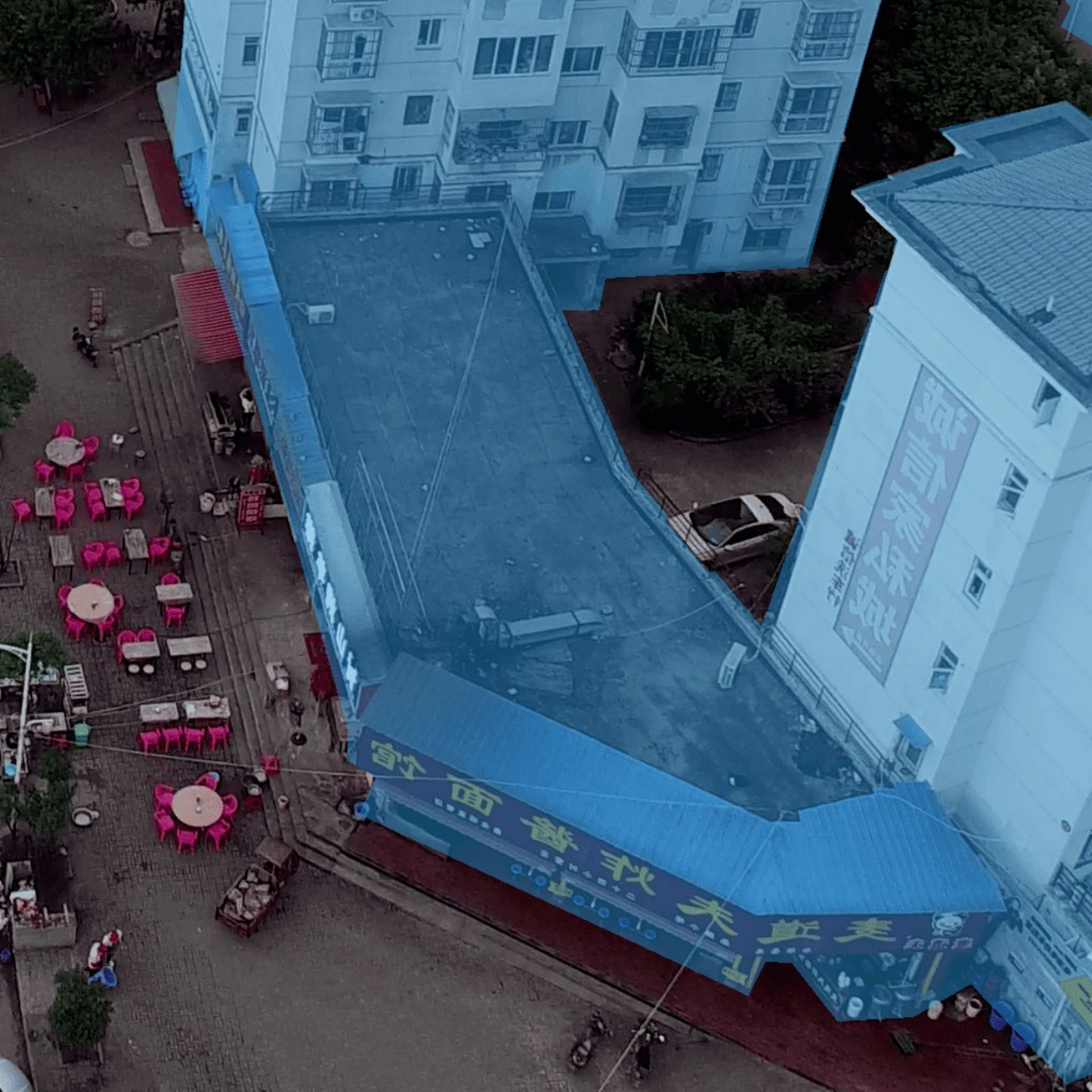} &
\includegraphics[width=0.155\linewidth]{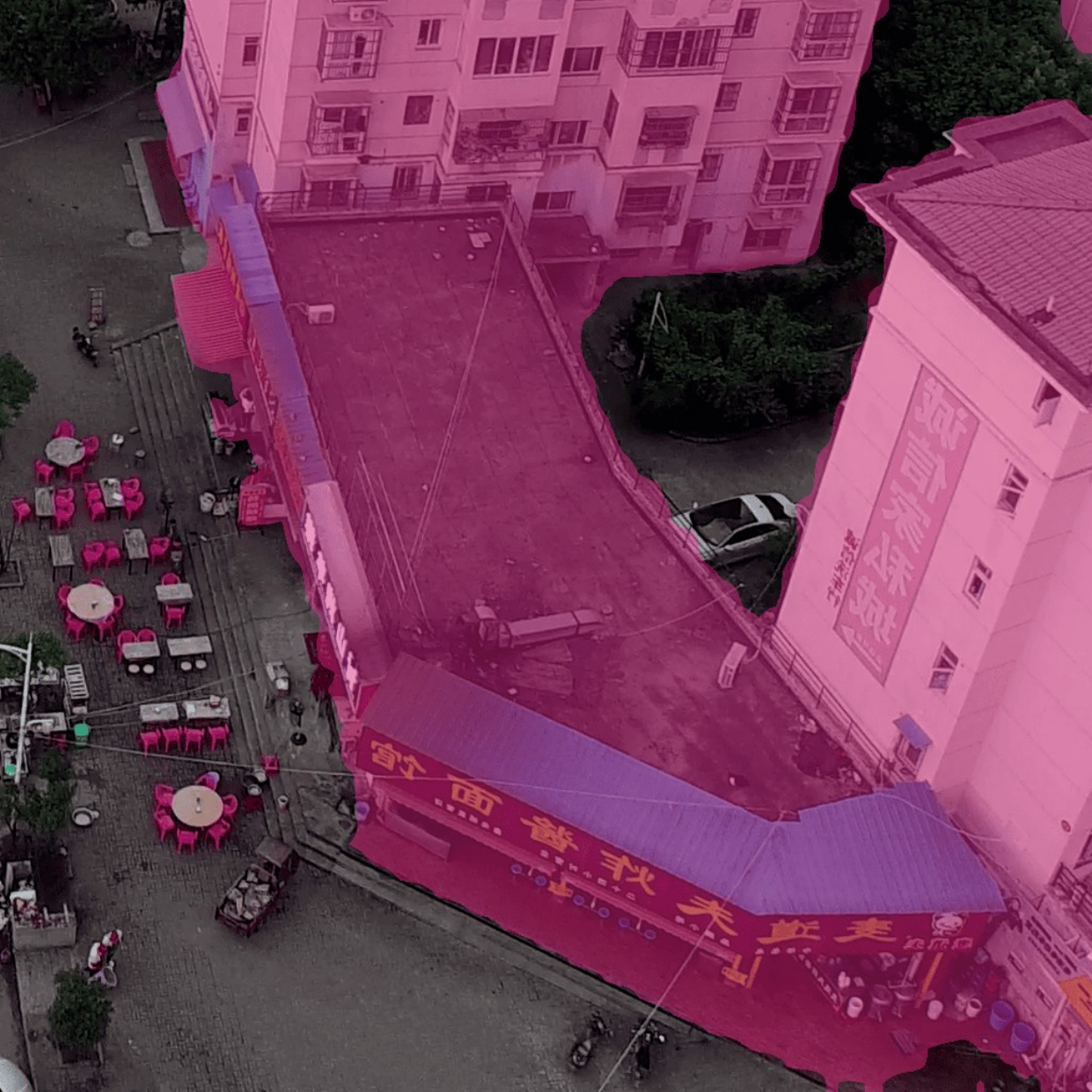} &
\includegraphics[width=0.155\linewidth]{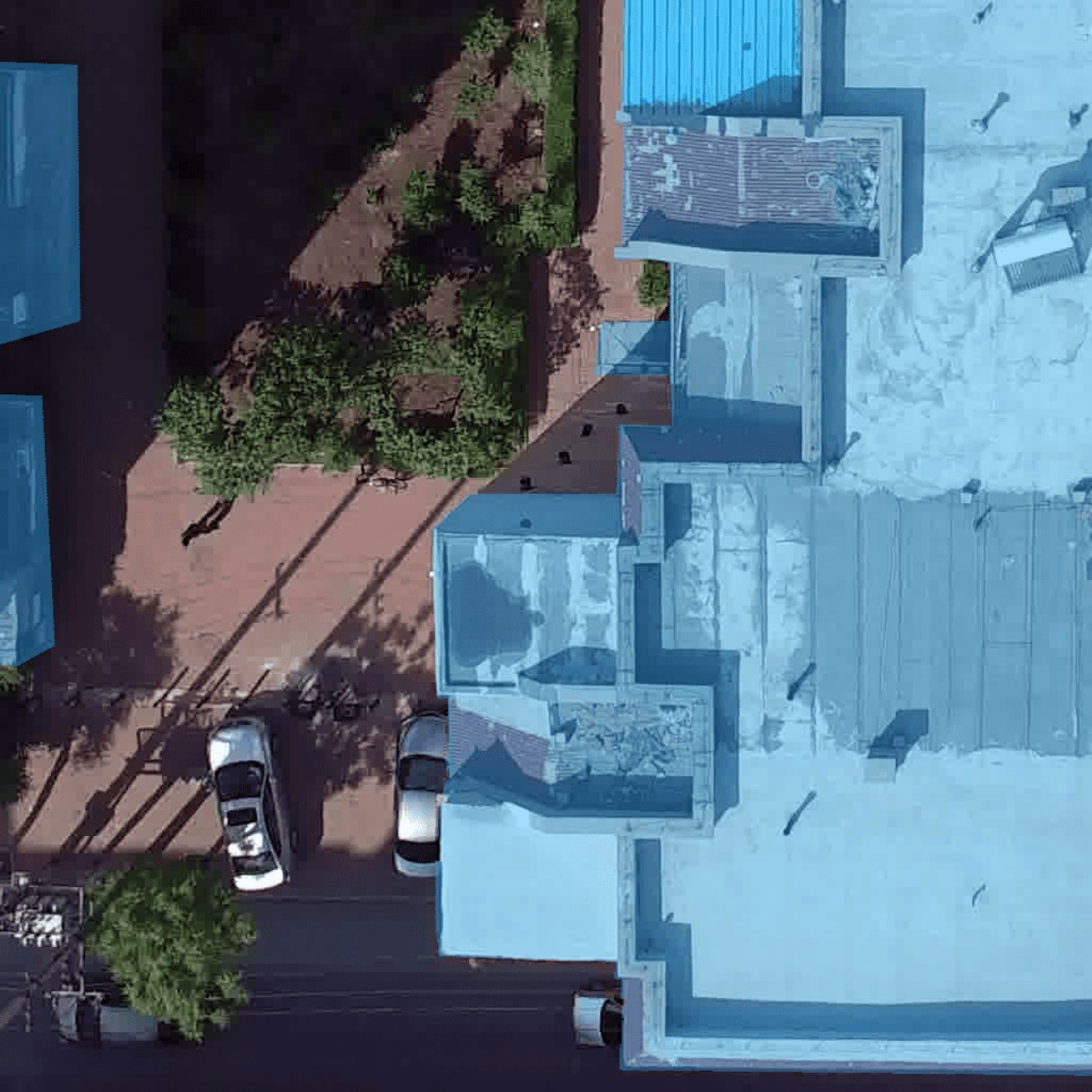} &
\includegraphics[width=0.155\linewidth]{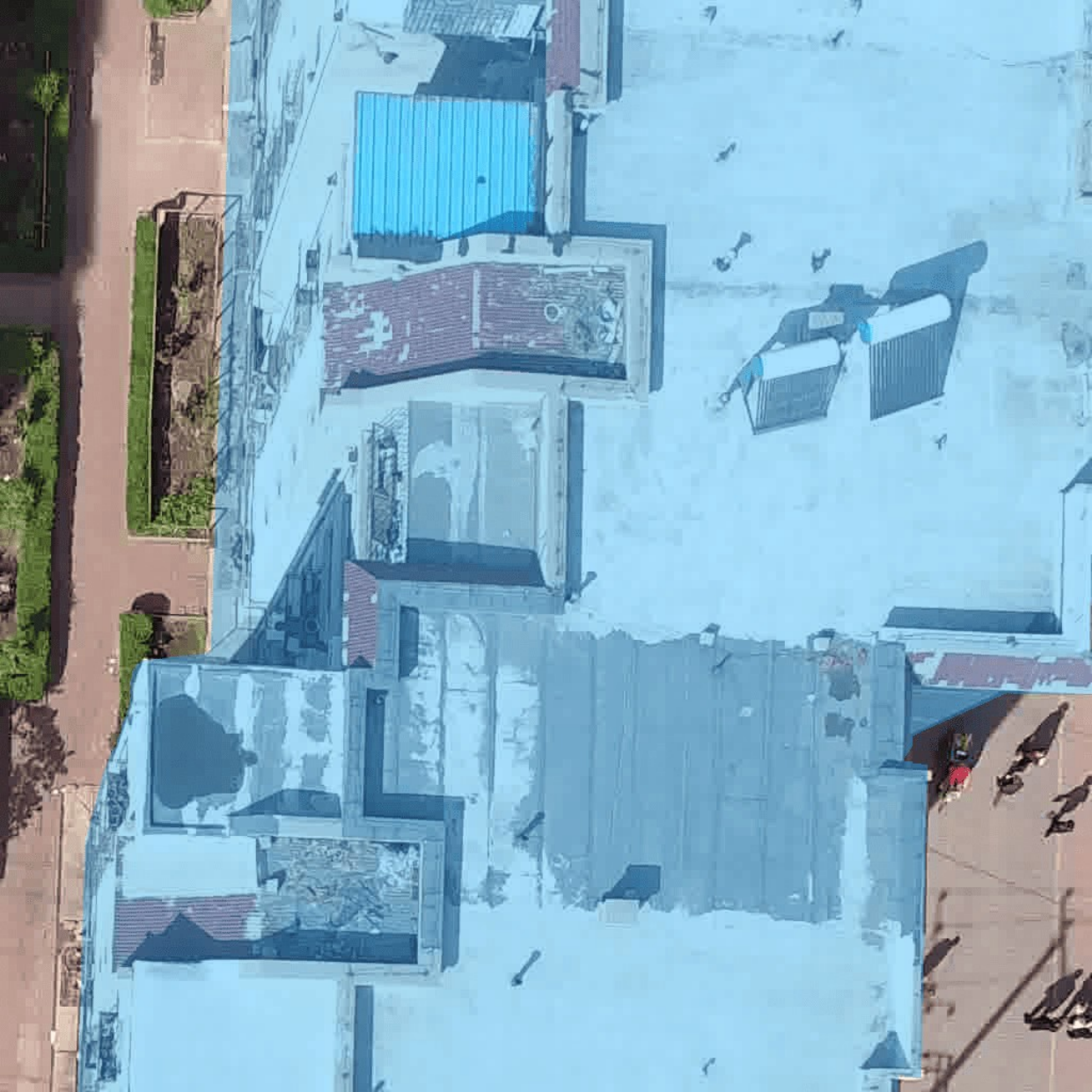} &
\includegraphics[width=0.155\linewidth]{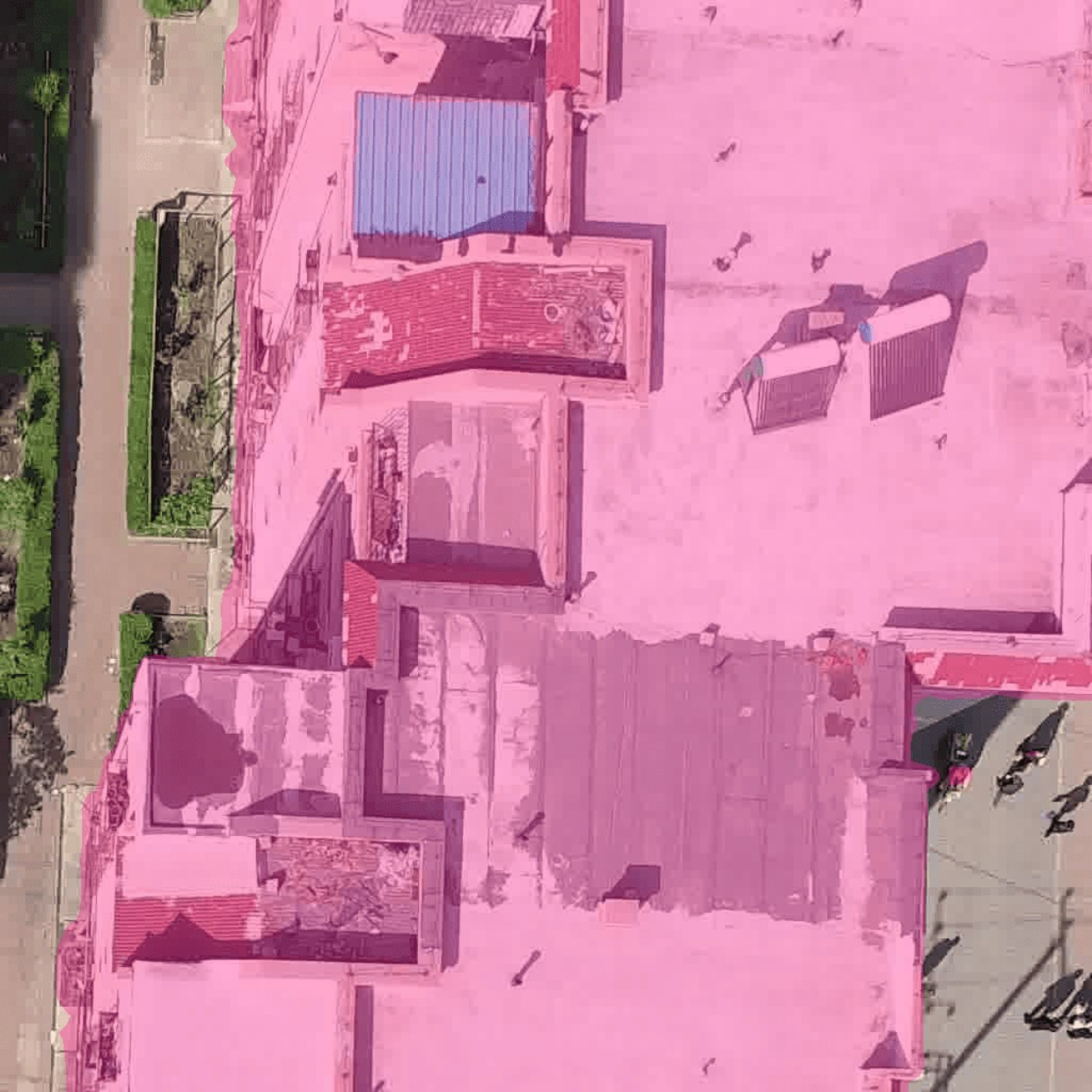} \\
\includegraphics[width=0.155\linewidth]{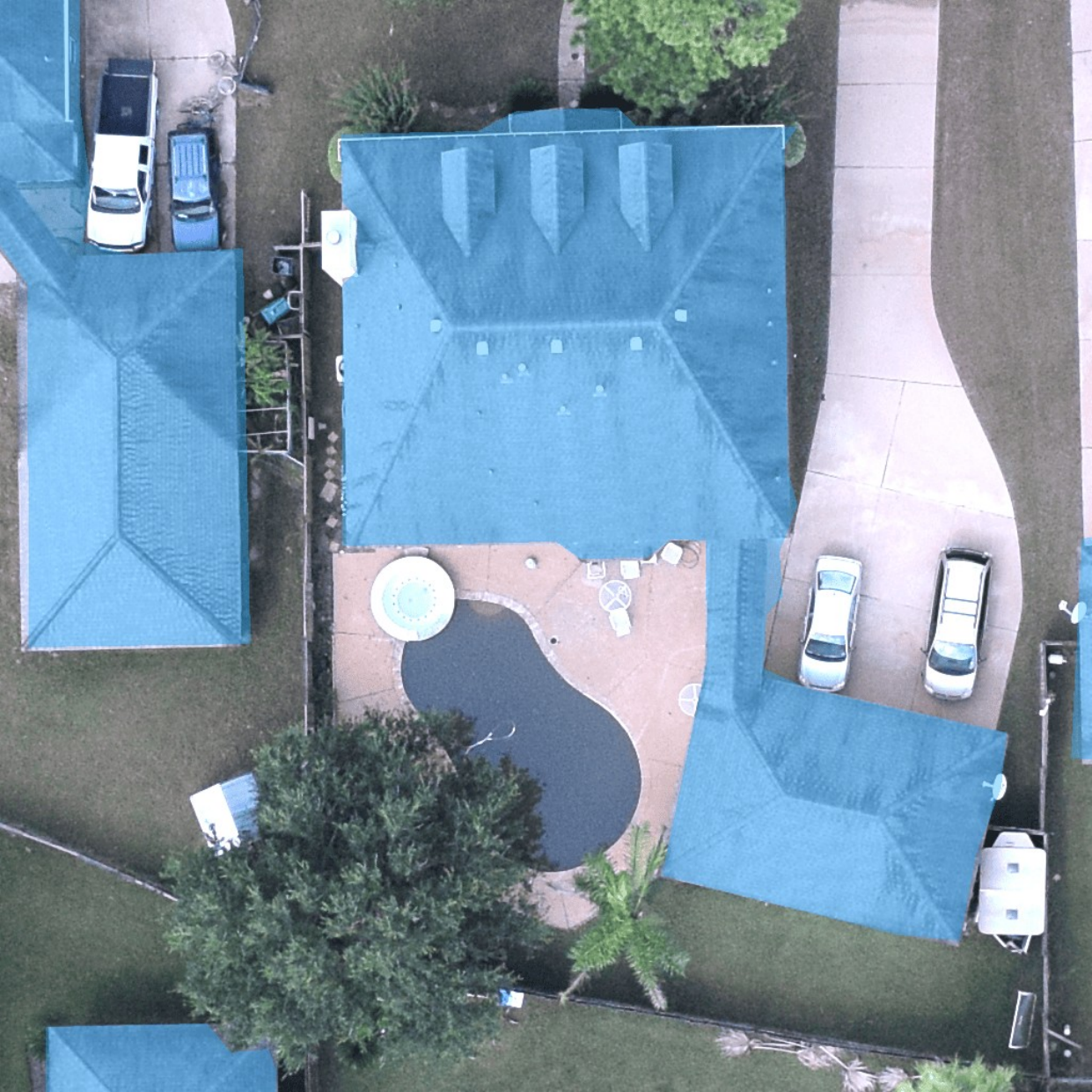} &
\includegraphics[width=0.155\linewidth]{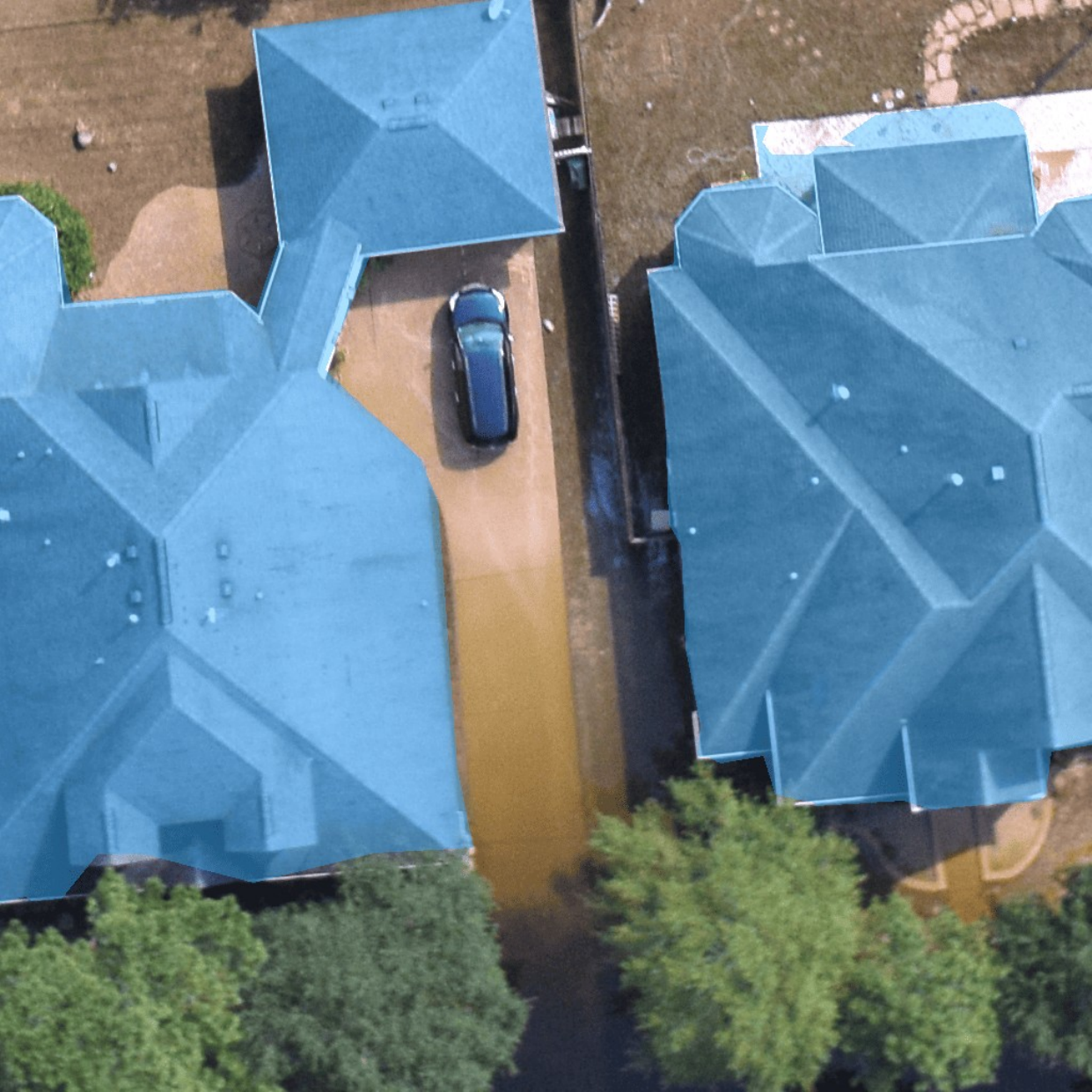} &
\includegraphics[width=0.155\linewidth]{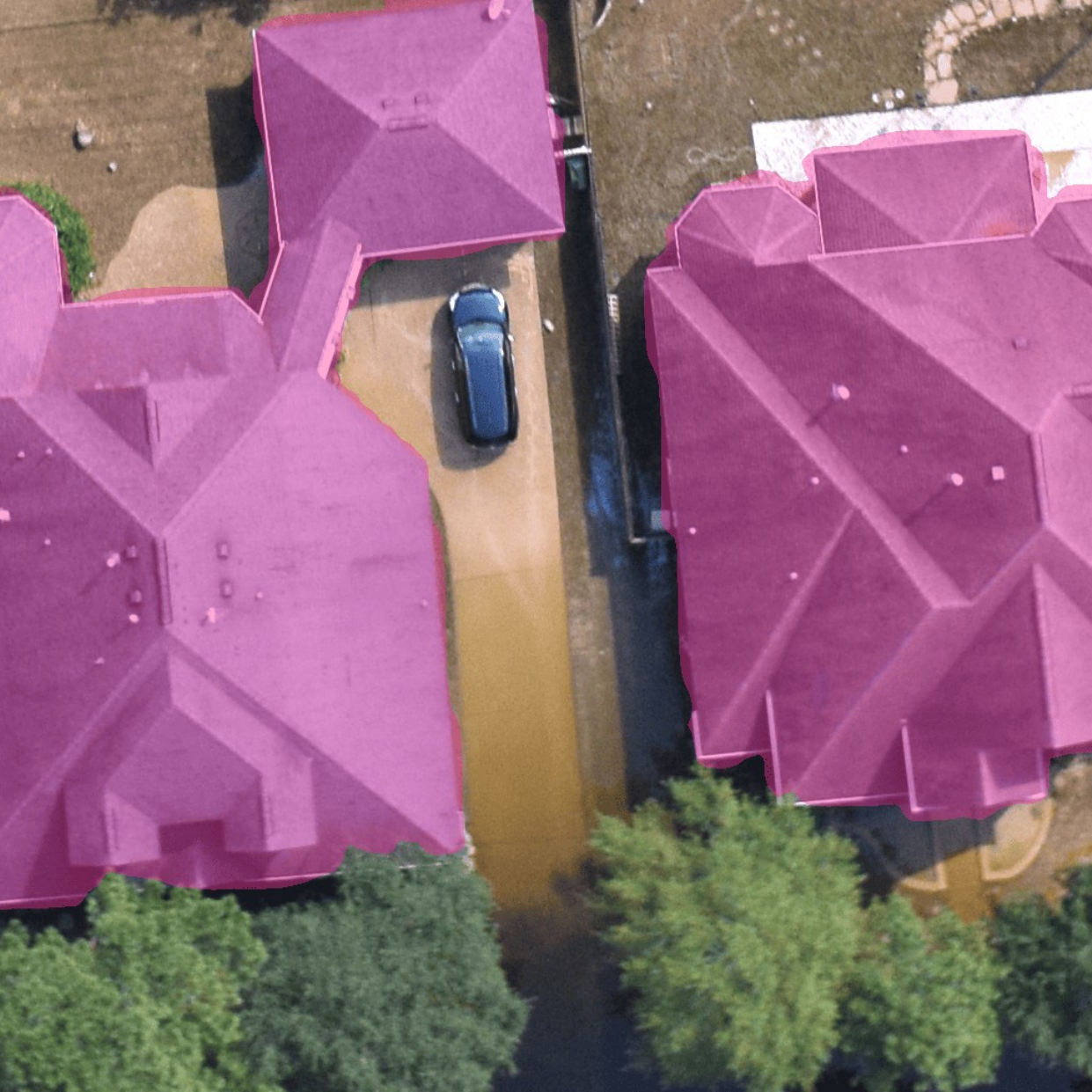} &
\includegraphics[width=0.155\linewidth]{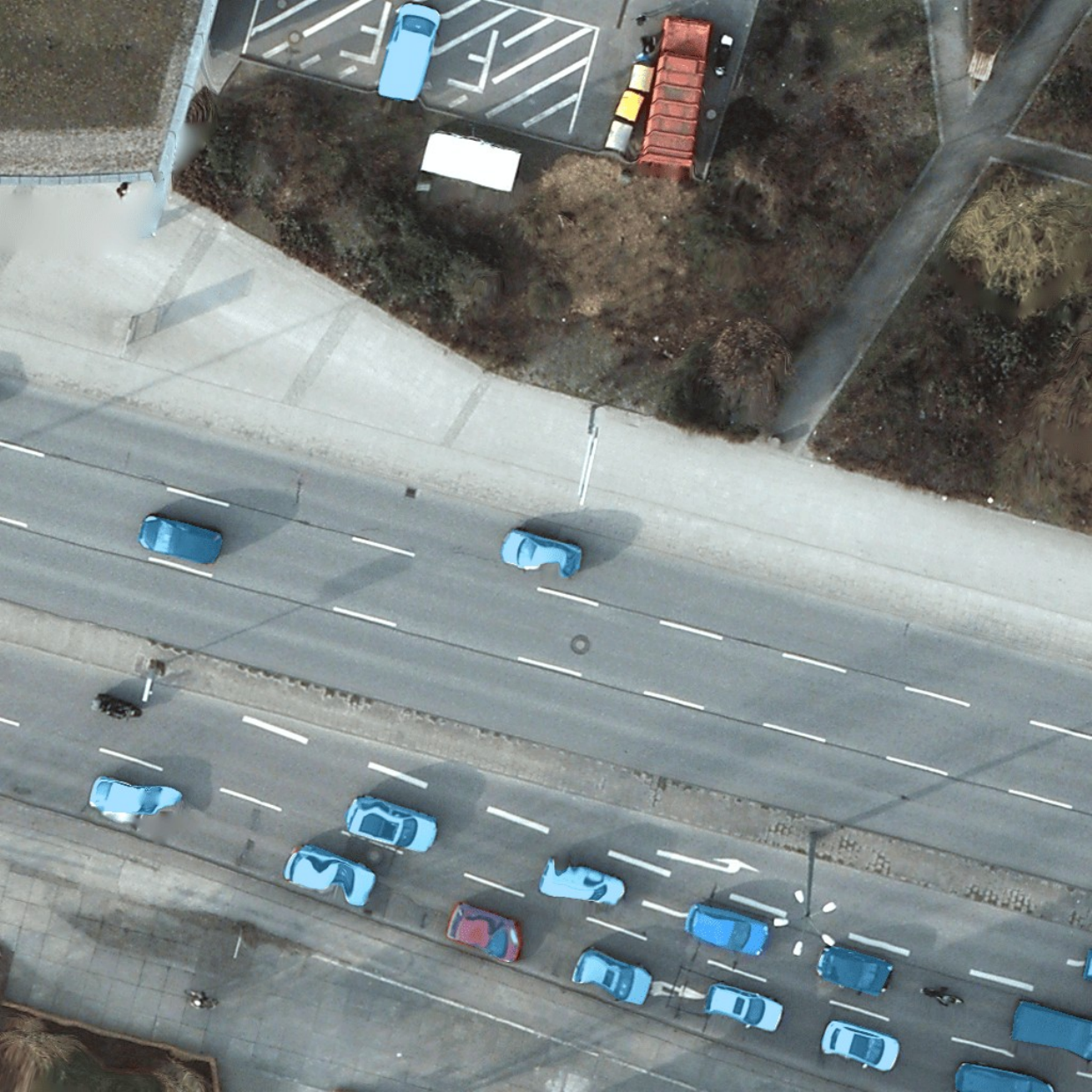} &
\includegraphics[width=0.155\linewidth]{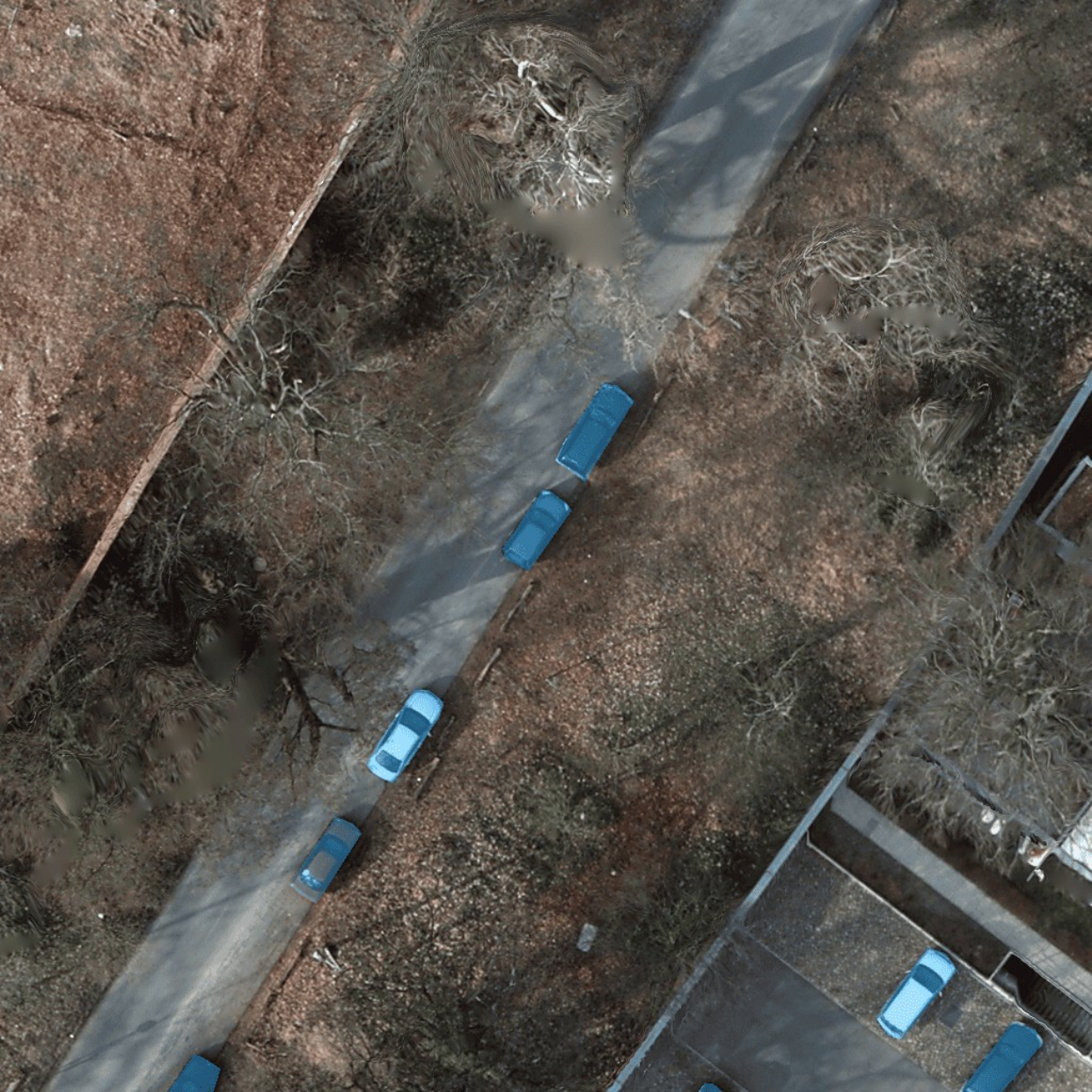} &
\includegraphics[width=0.155\linewidth]{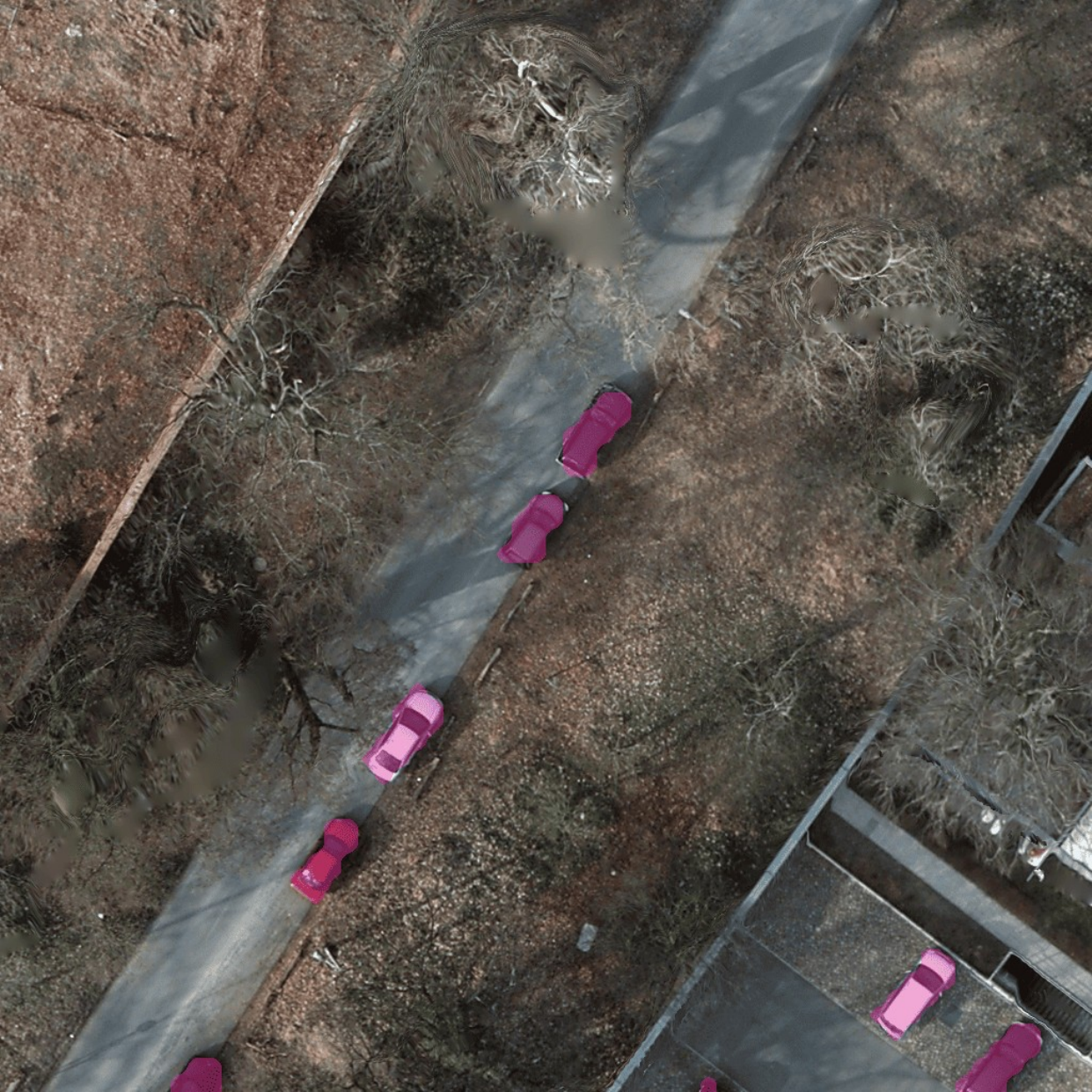} \\
\includegraphics[width=0.155\linewidth]{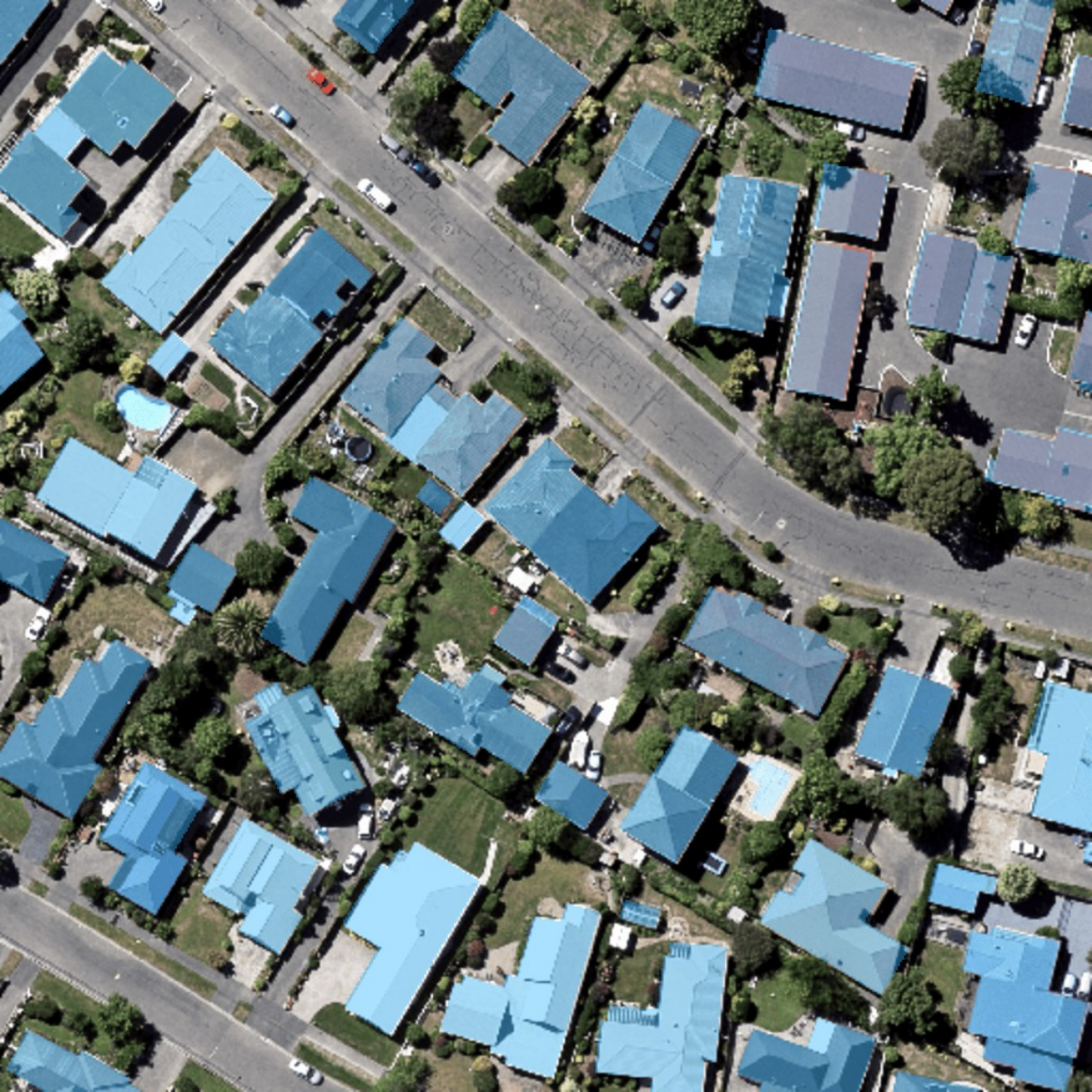} &
\includegraphics[width=0.155\linewidth]{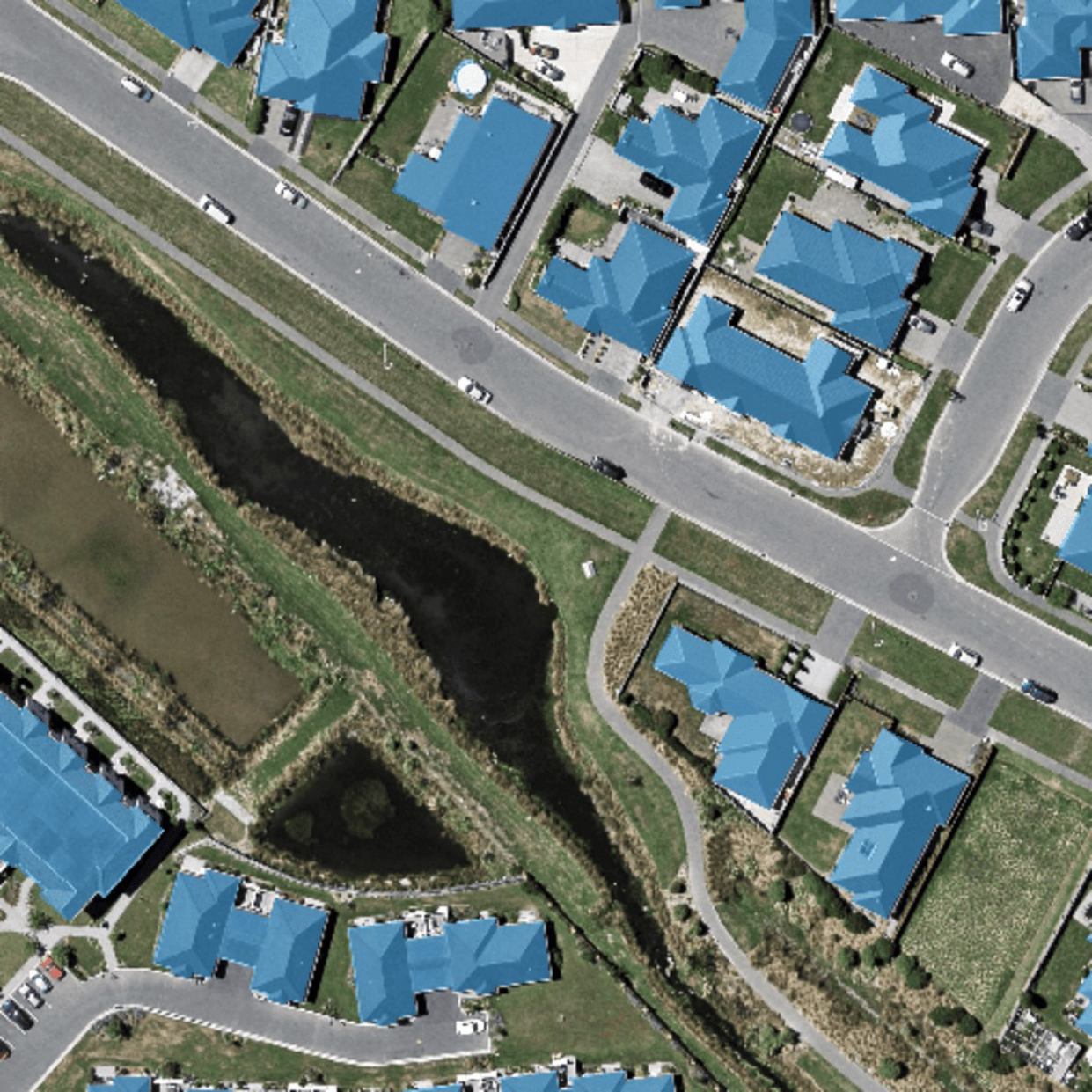} &
\includegraphics[width=0.155\linewidth]{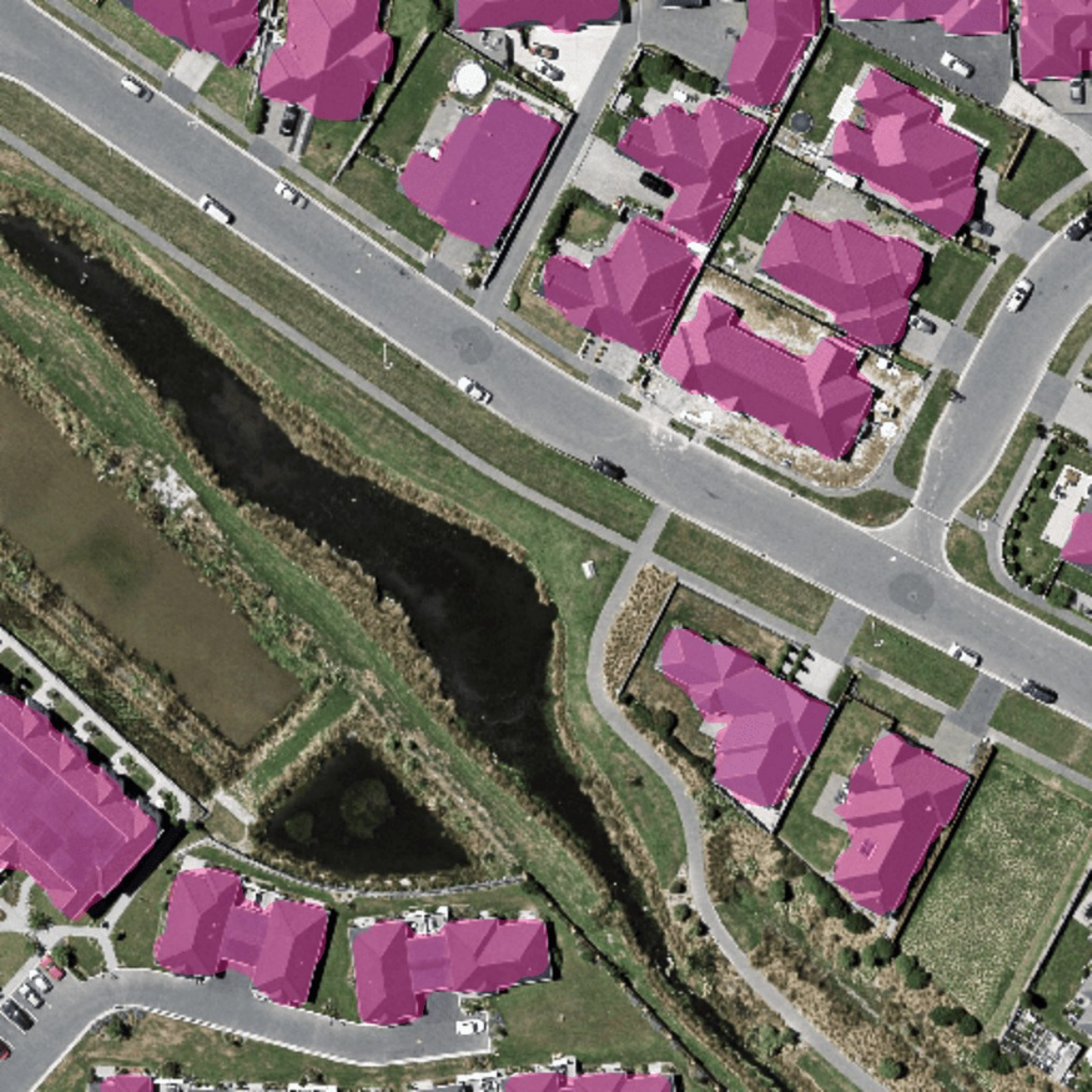} &
\includegraphics[width=0.155\linewidth]{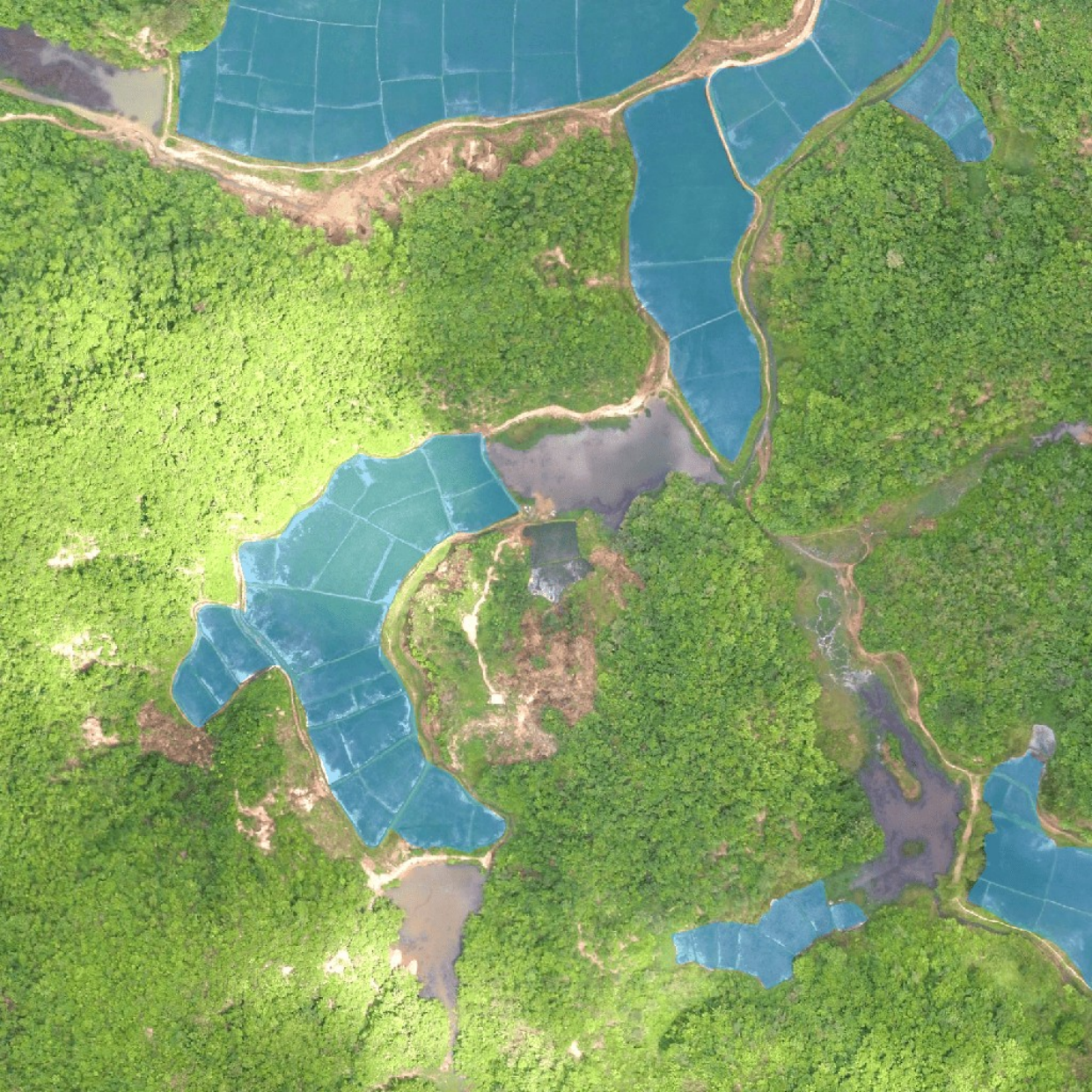} &
\includegraphics[width=0.155\linewidth]{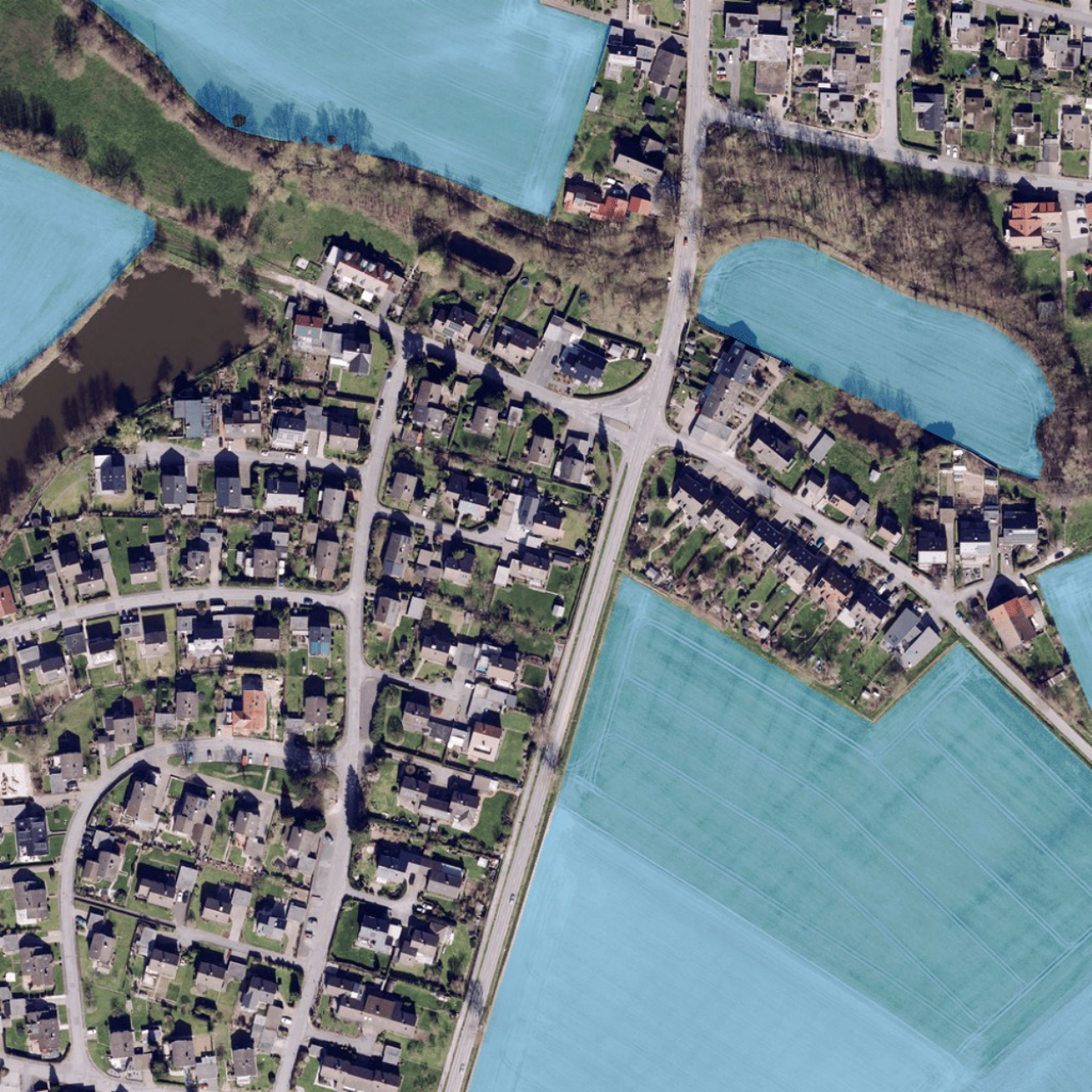} &
\includegraphics[width=0.155\linewidth]{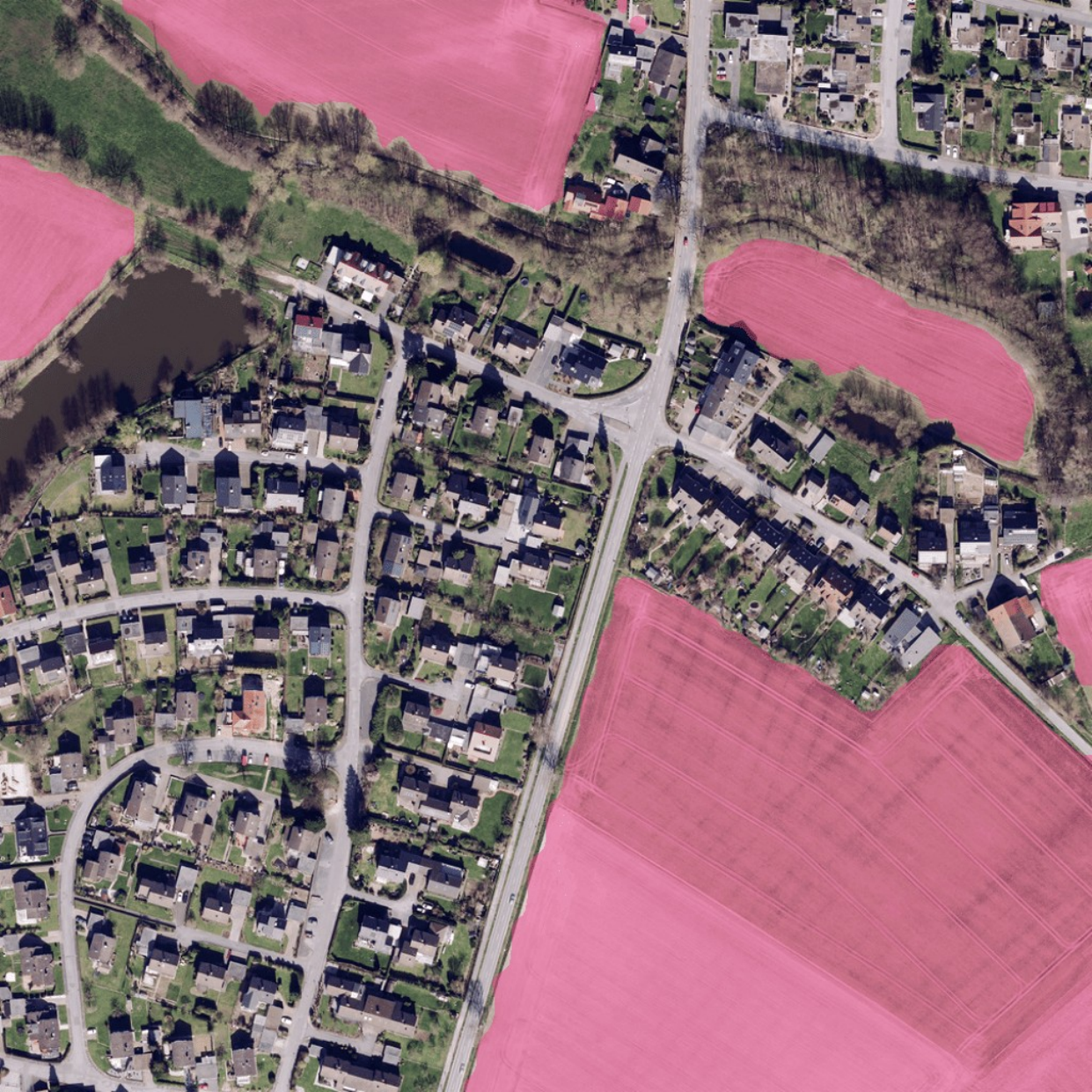} \\
\includegraphics[width=0.155\linewidth]{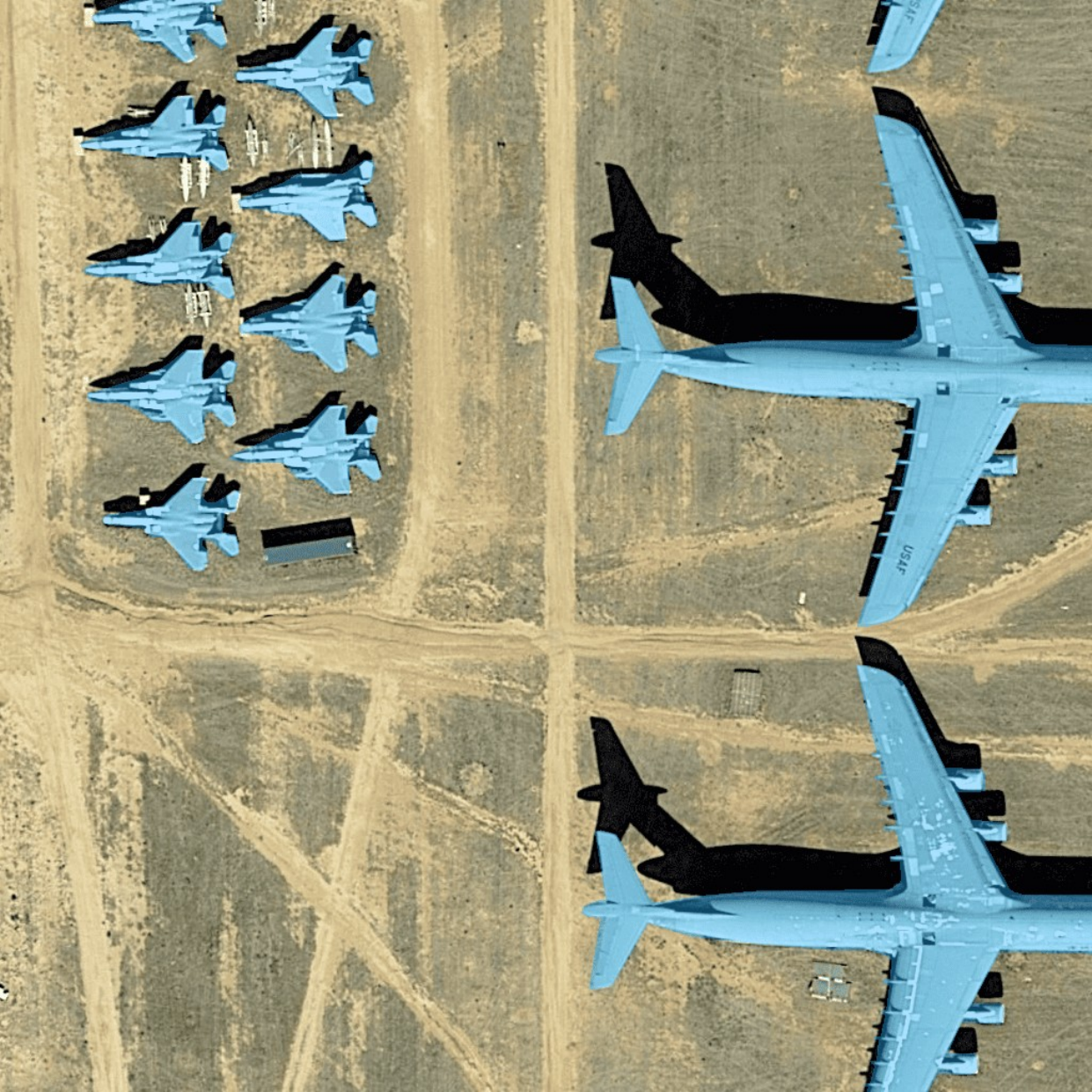} &
\includegraphics[width=0.155\linewidth]{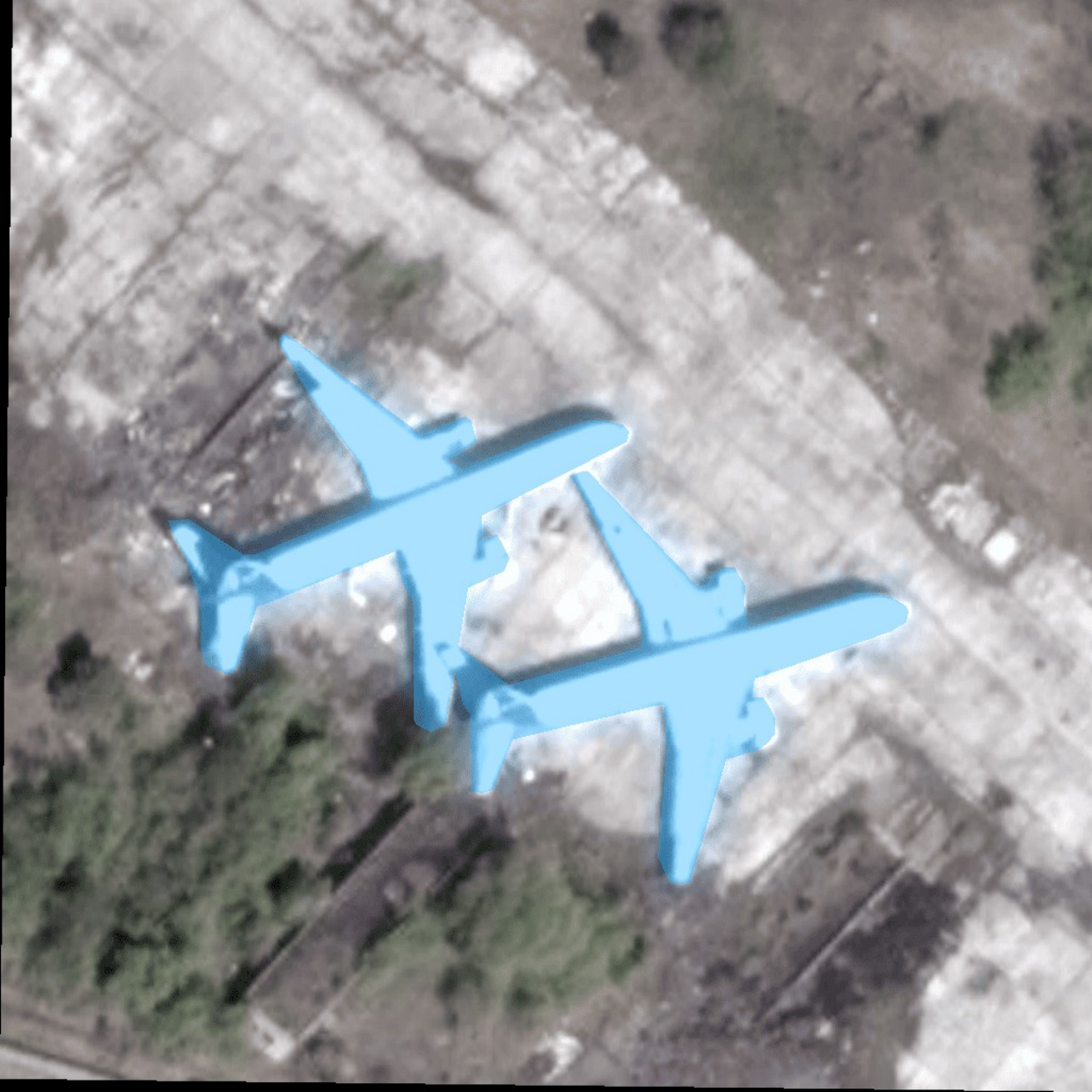} &
\includegraphics[width=0.155\linewidth]{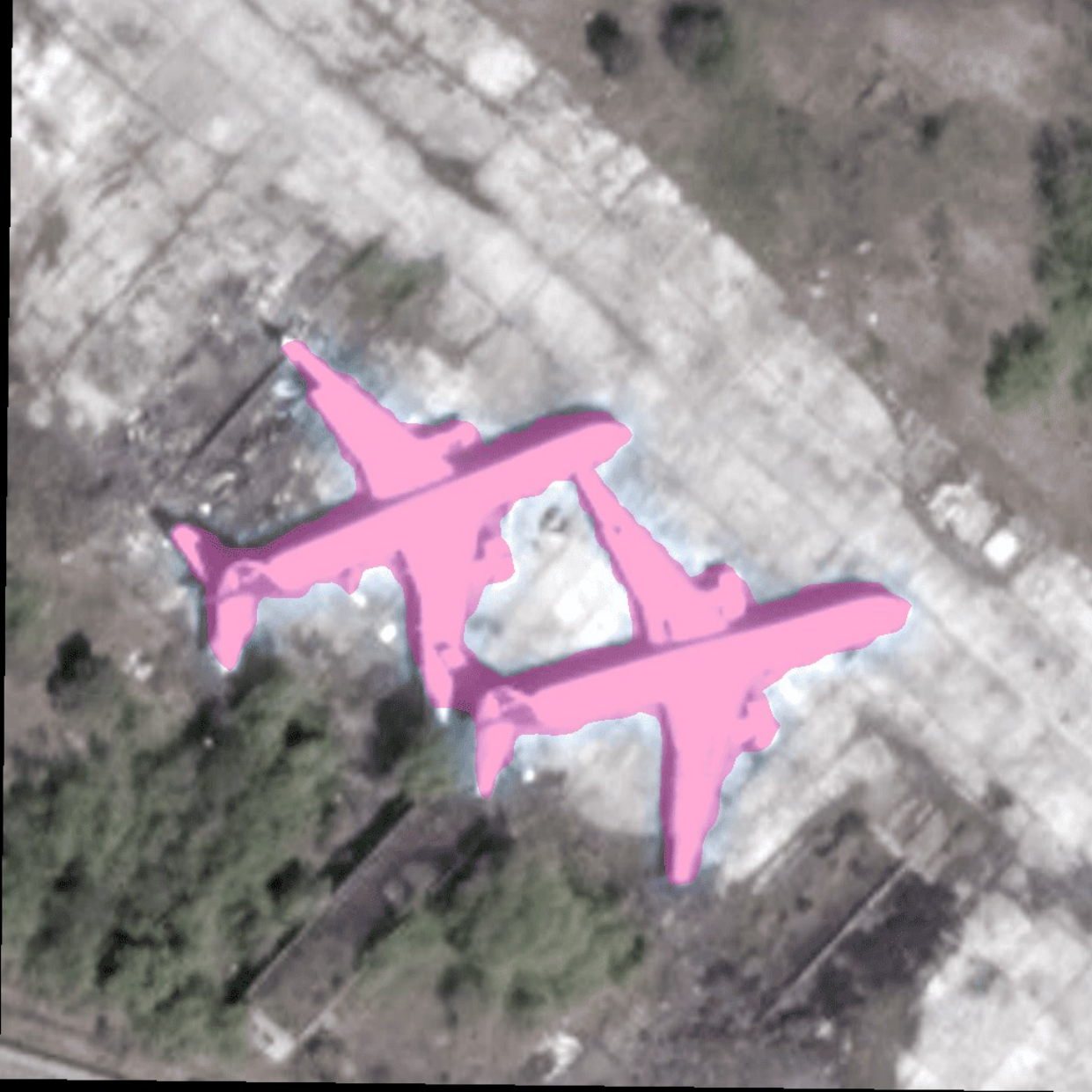} &
\includegraphics[width=0.155\linewidth]{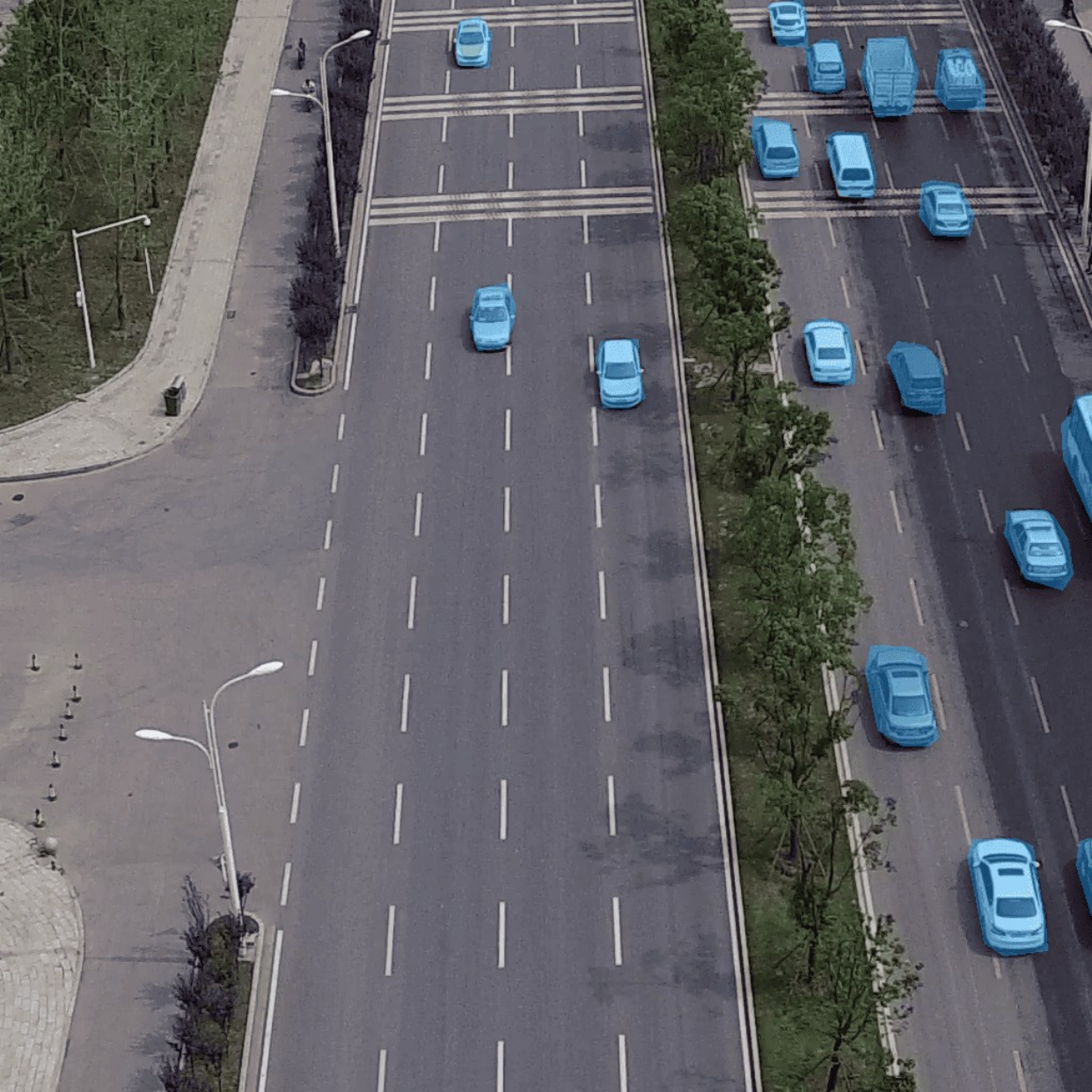} &
\includegraphics[width=0.155\linewidth]{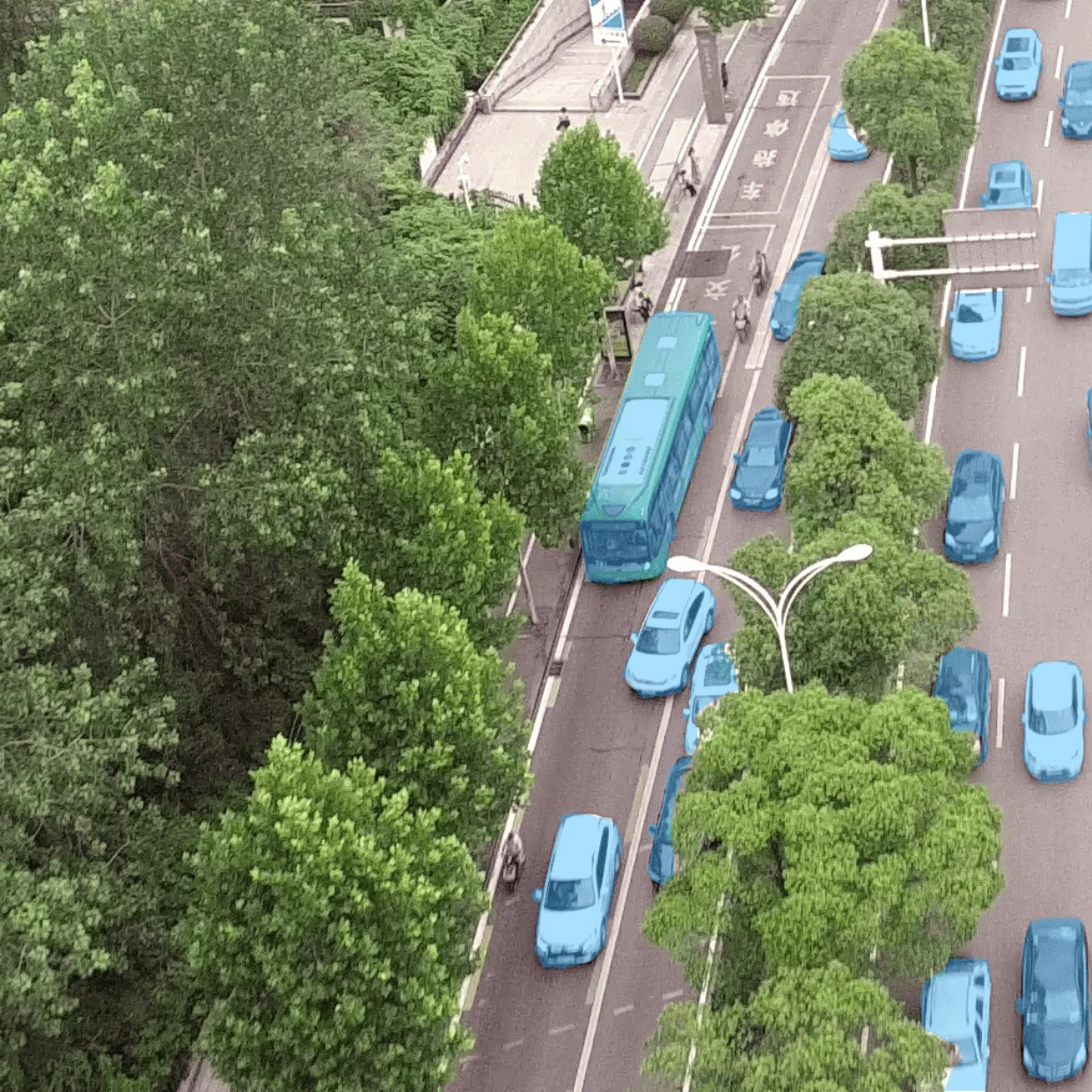} &
\includegraphics[width=0.155\linewidth]{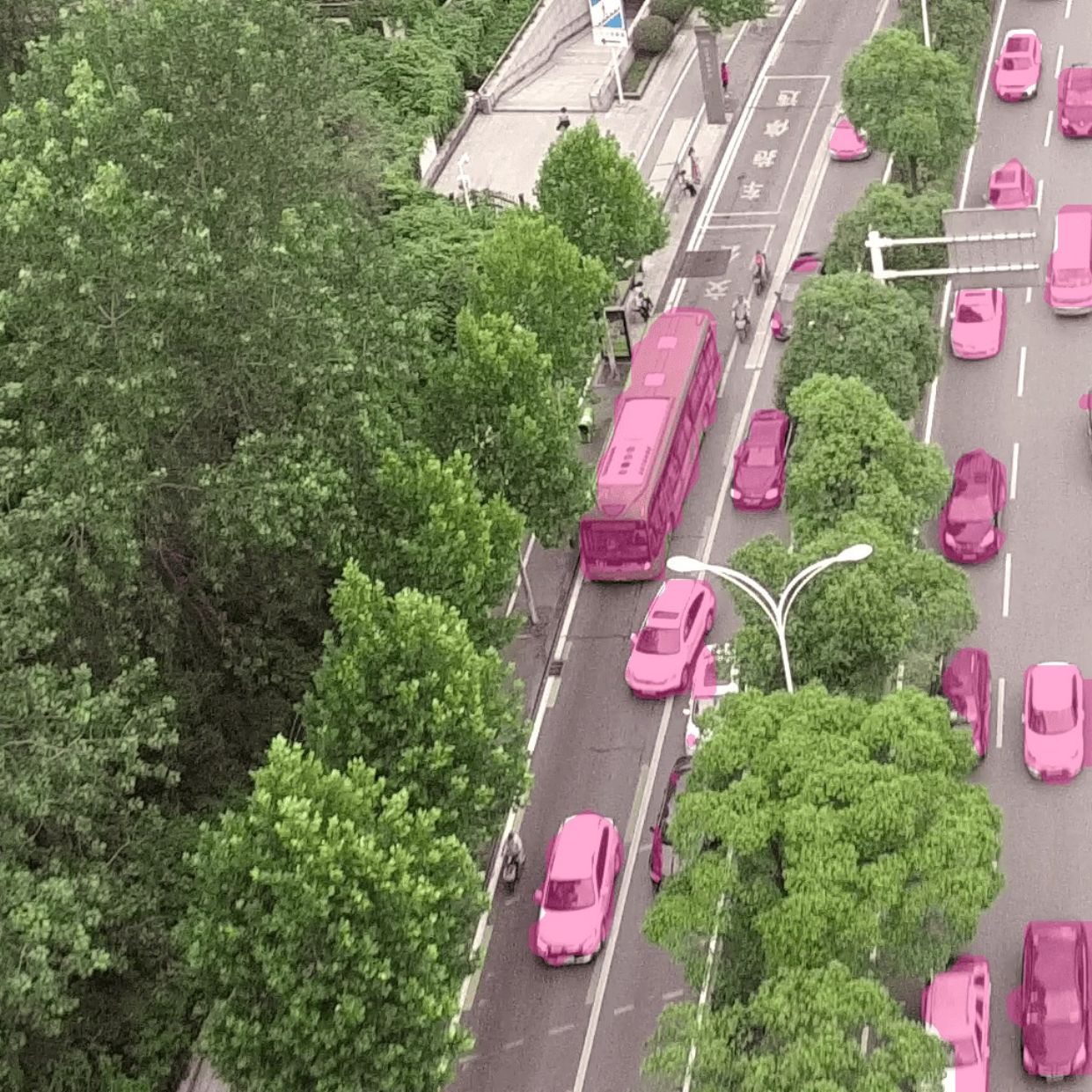} \\
\includegraphics[width=0.155\linewidth]{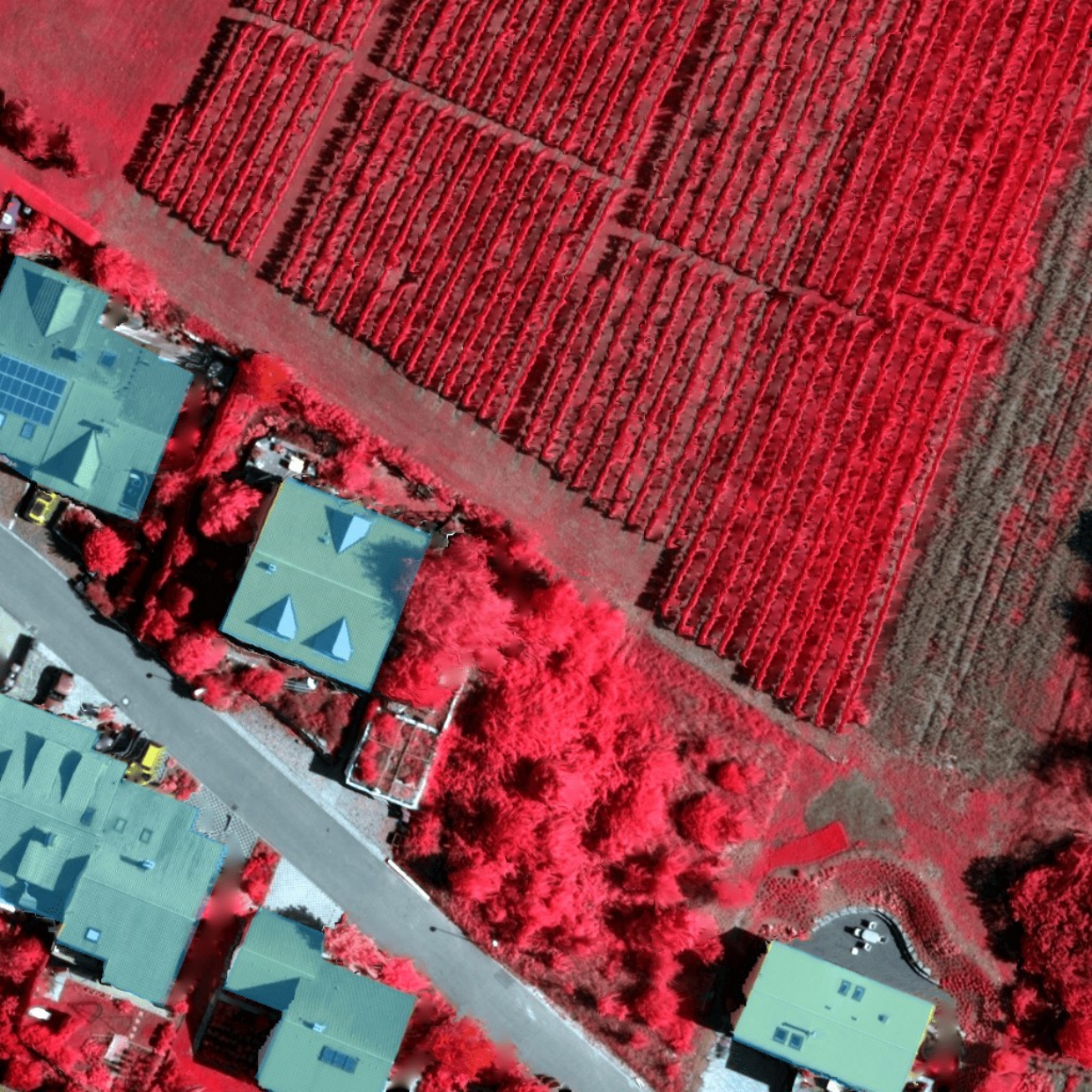} &
\includegraphics[width=0.155\linewidth]{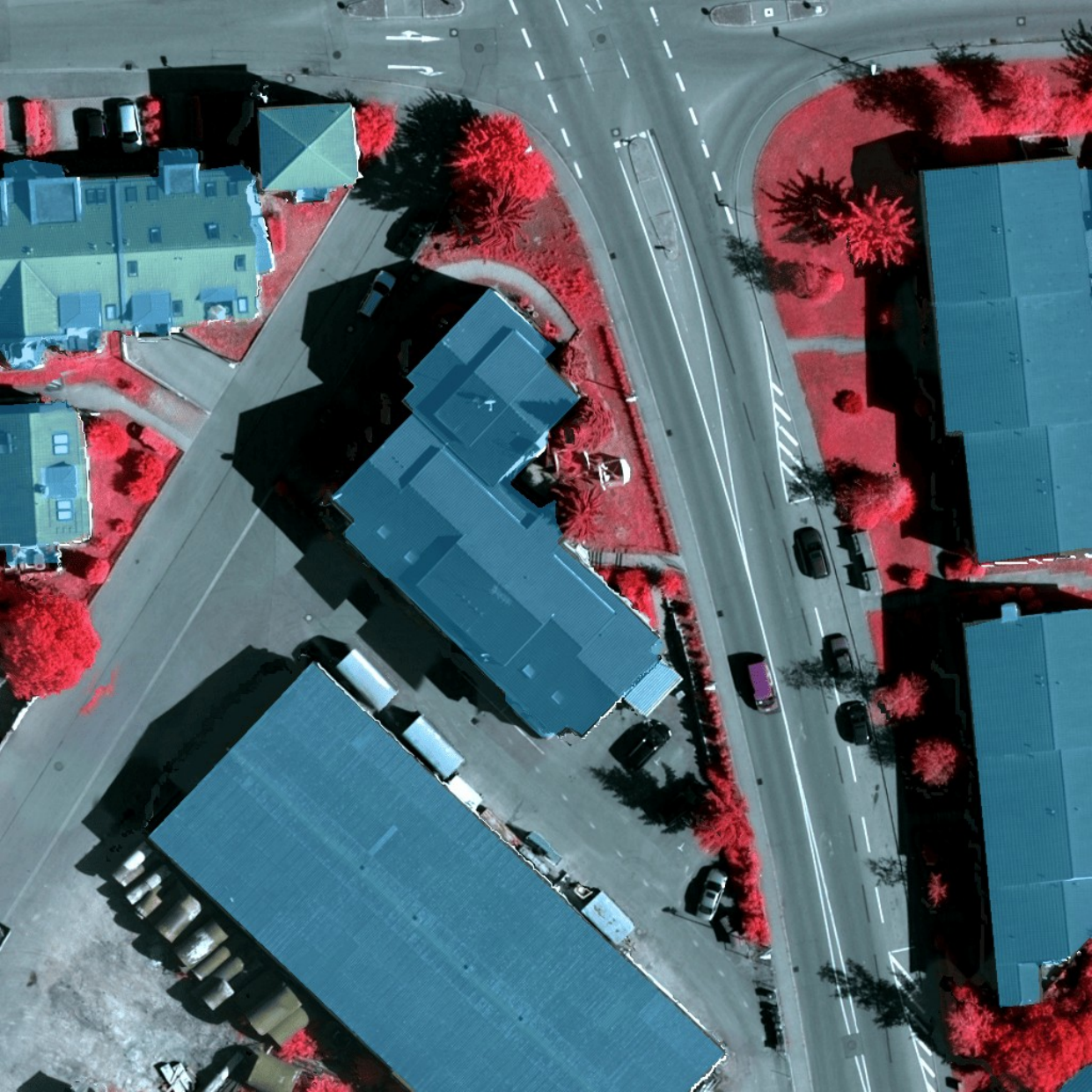} &
\includegraphics[width=0.155\linewidth]{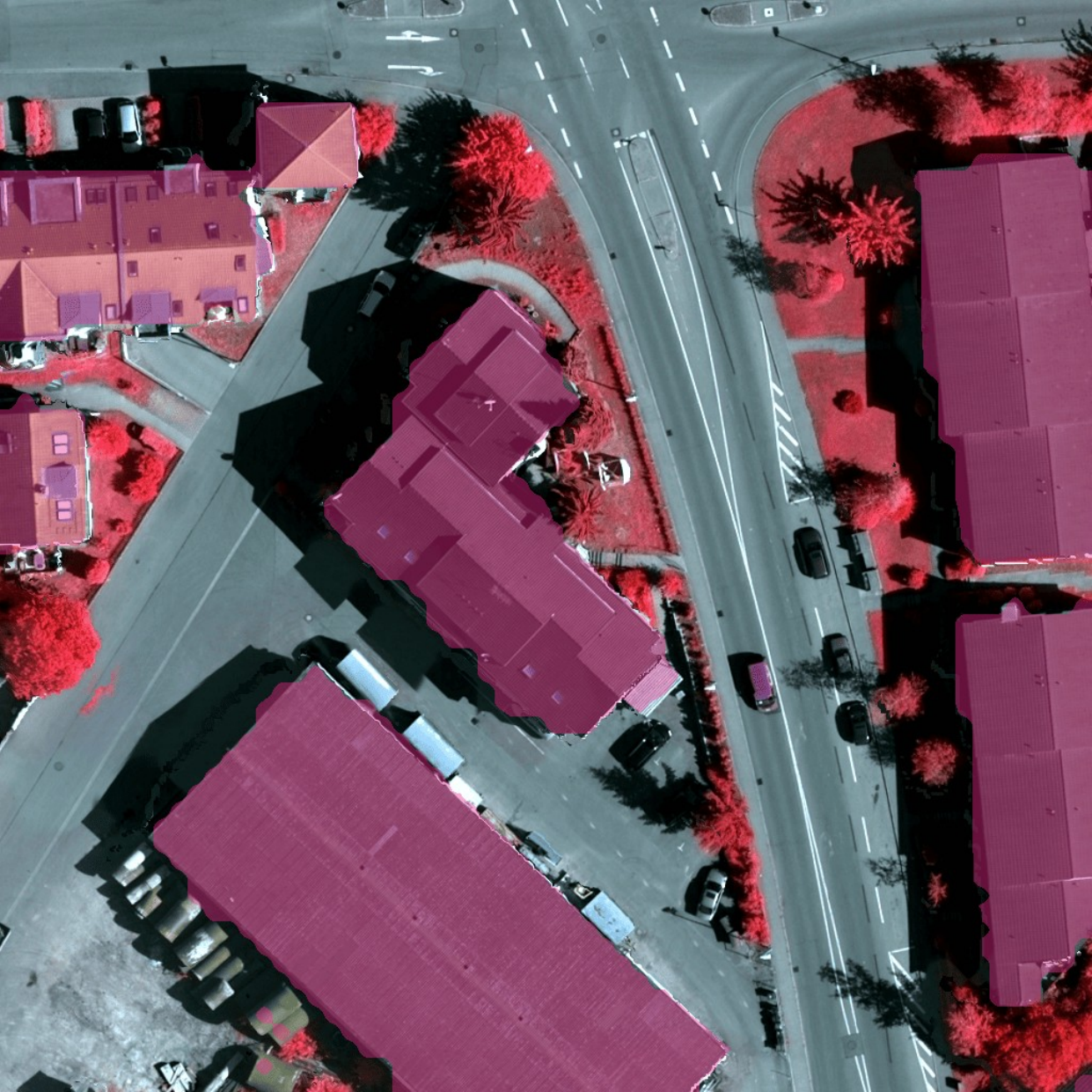} &
\includegraphics[width=0.155\linewidth]{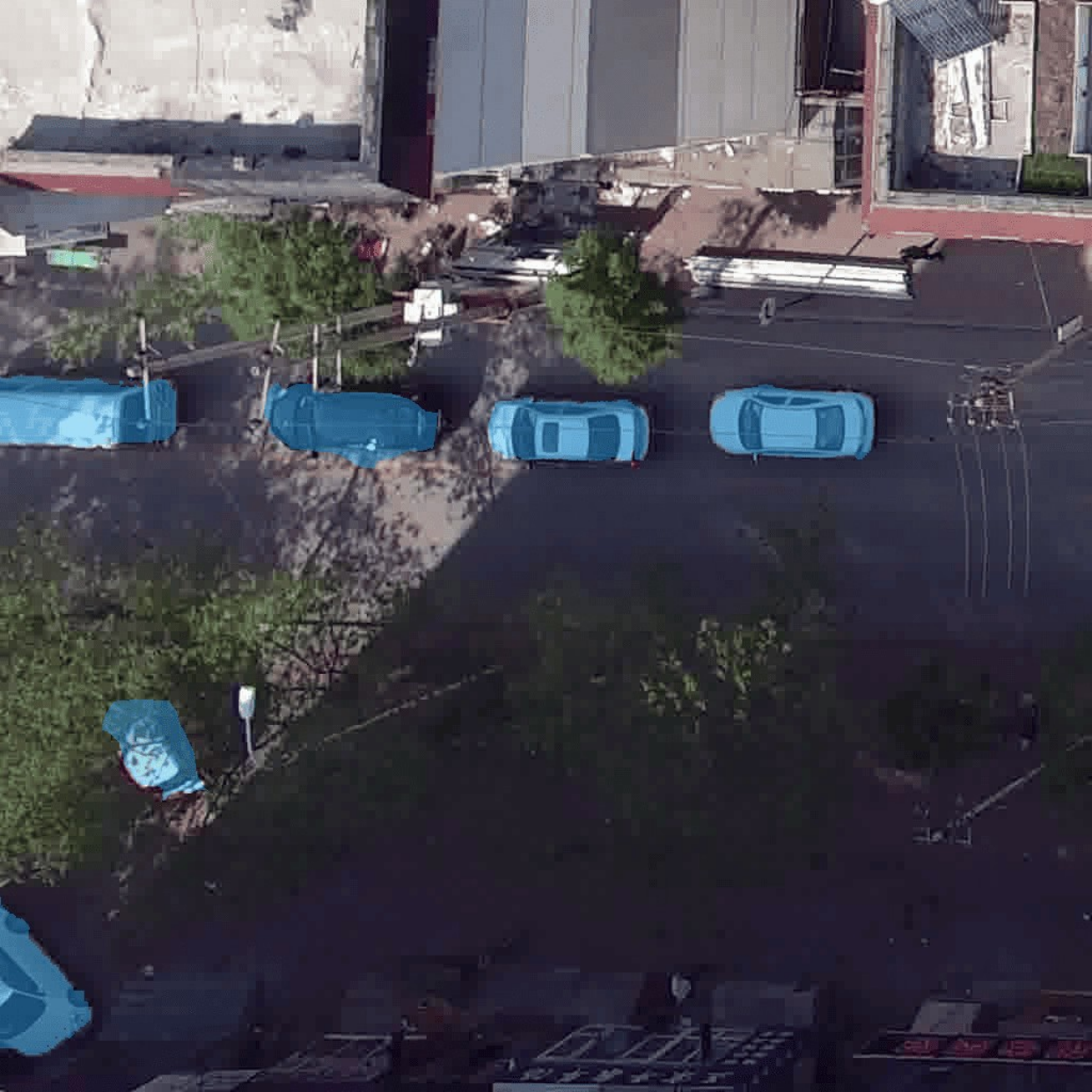} &
\includegraphics[width=0.155\linewidth]{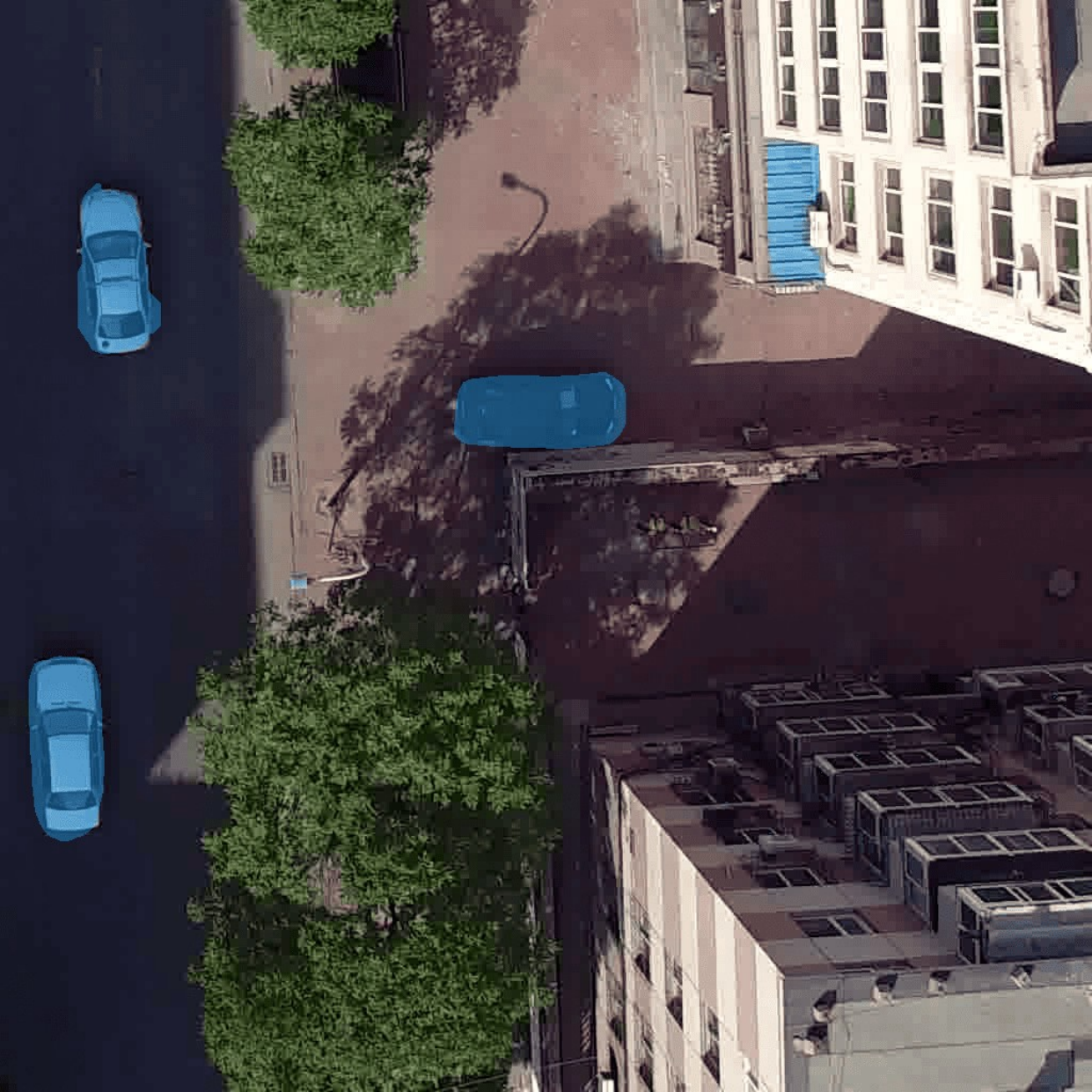} &
\includegraphics[width=0.155\linewidth]{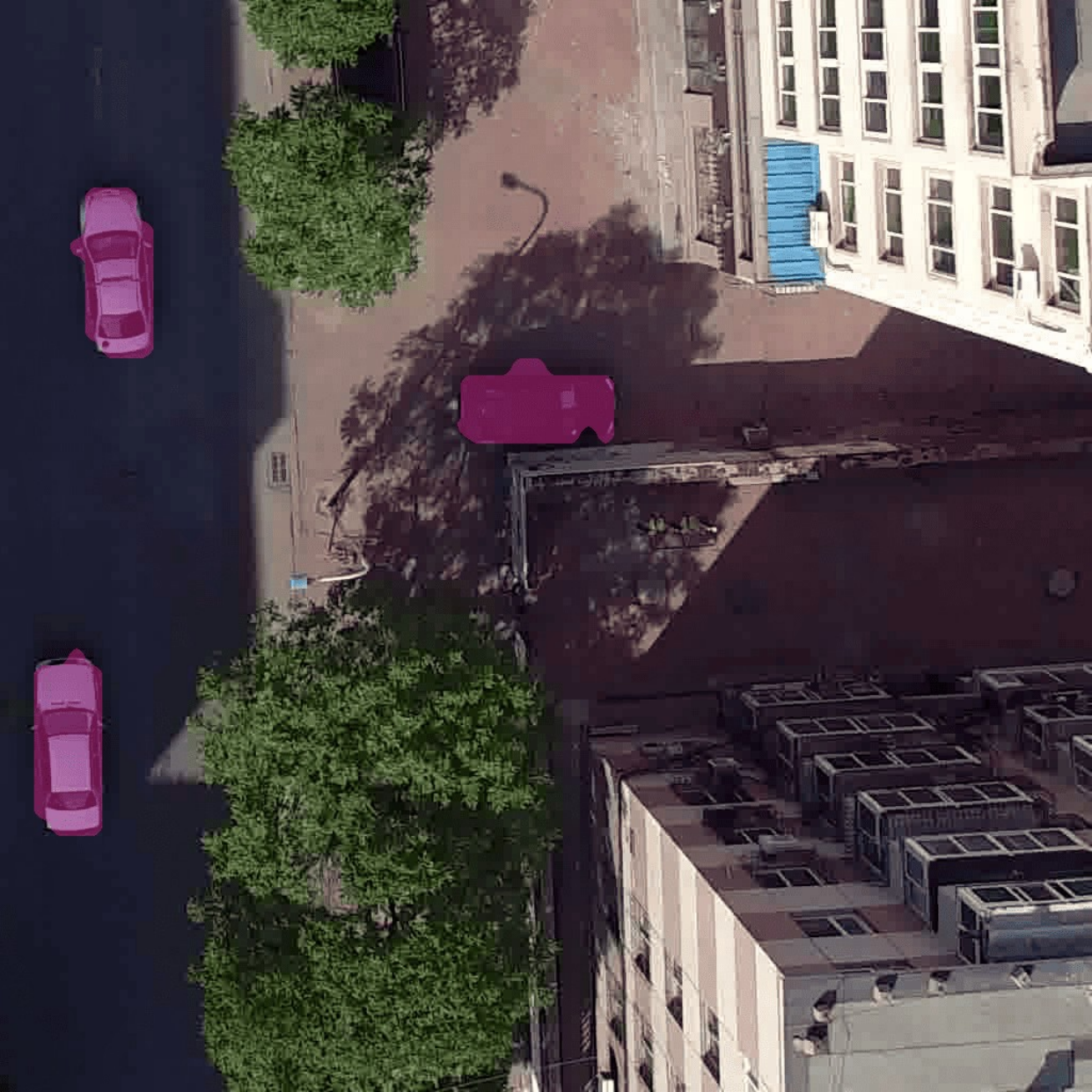} \\
\includegraphics[width=0.155\linewidth]{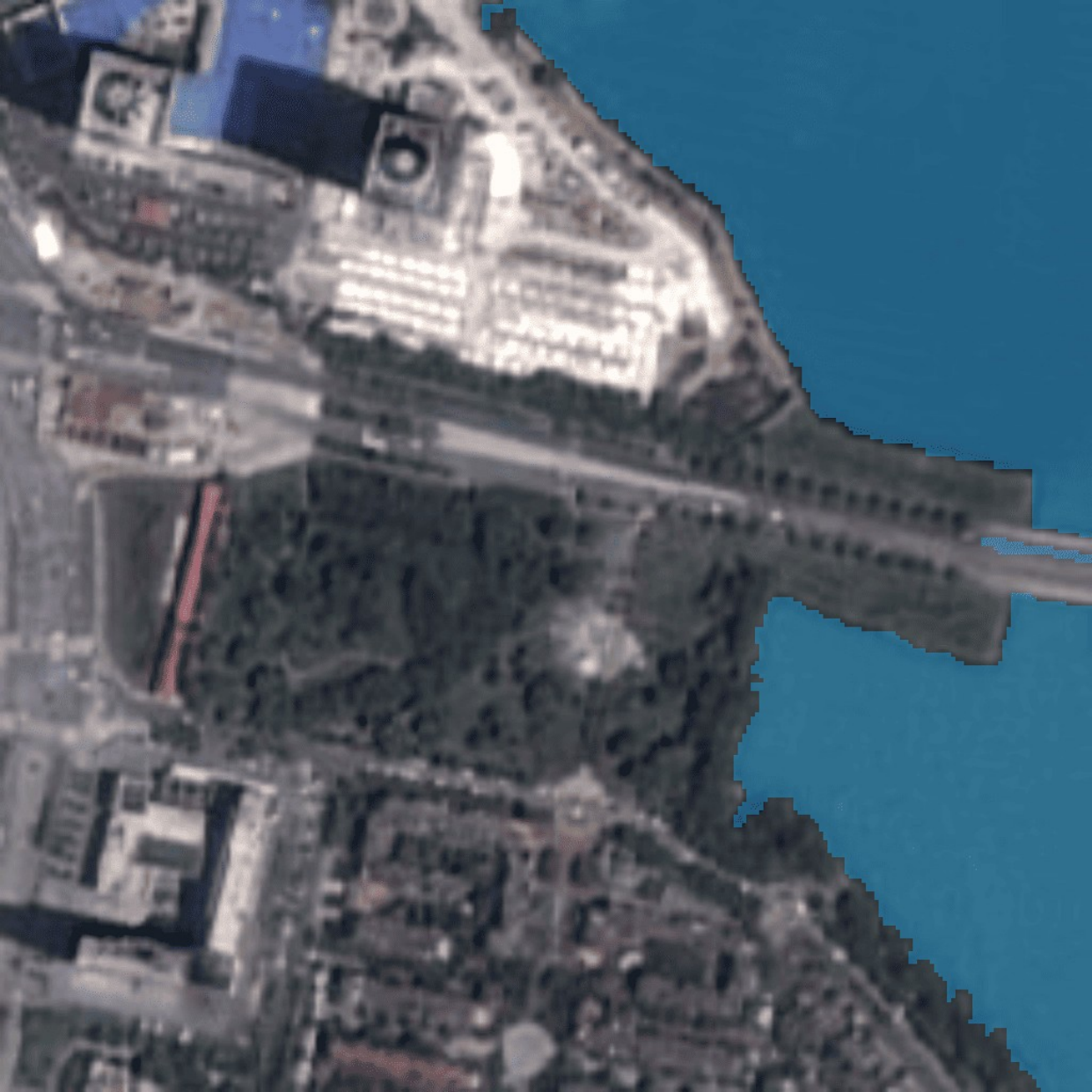} &
\includegraphics[width=0.155\linewidth]{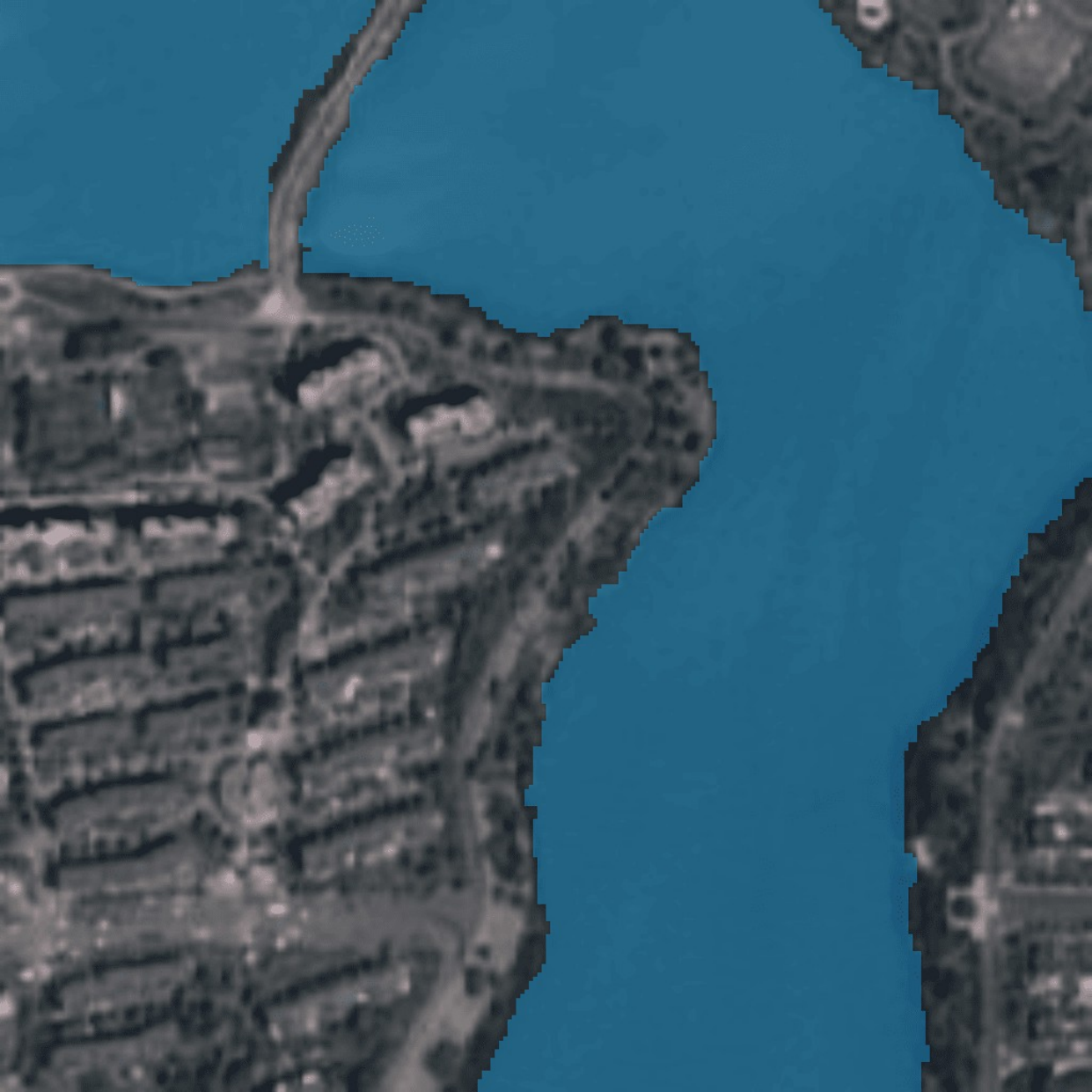} &
\includegraphics[width=0.155\linewidth]{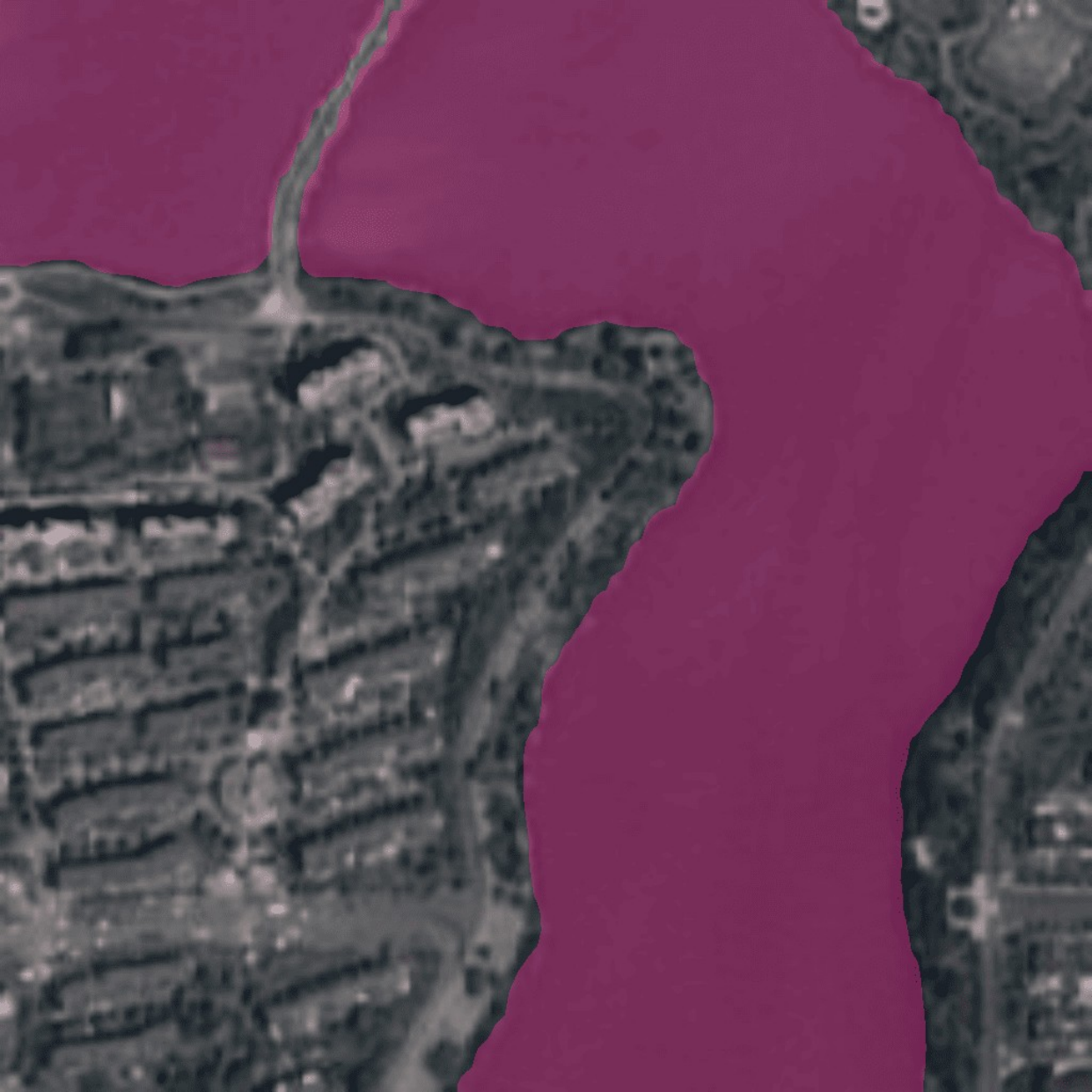} &
\includegraphics[width=0.155\linewidth]{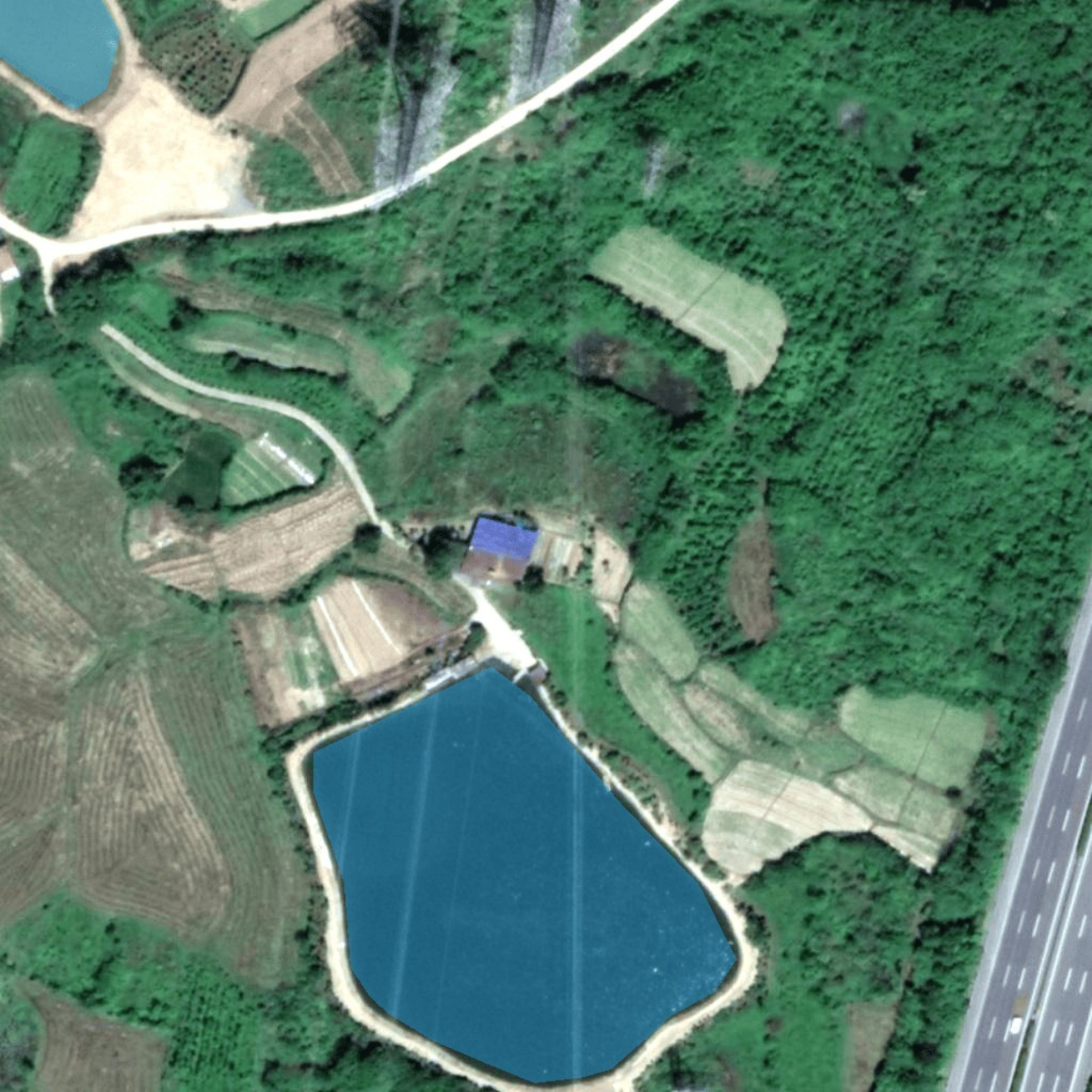} &
\includegraphics[width=0.155\linewidth]{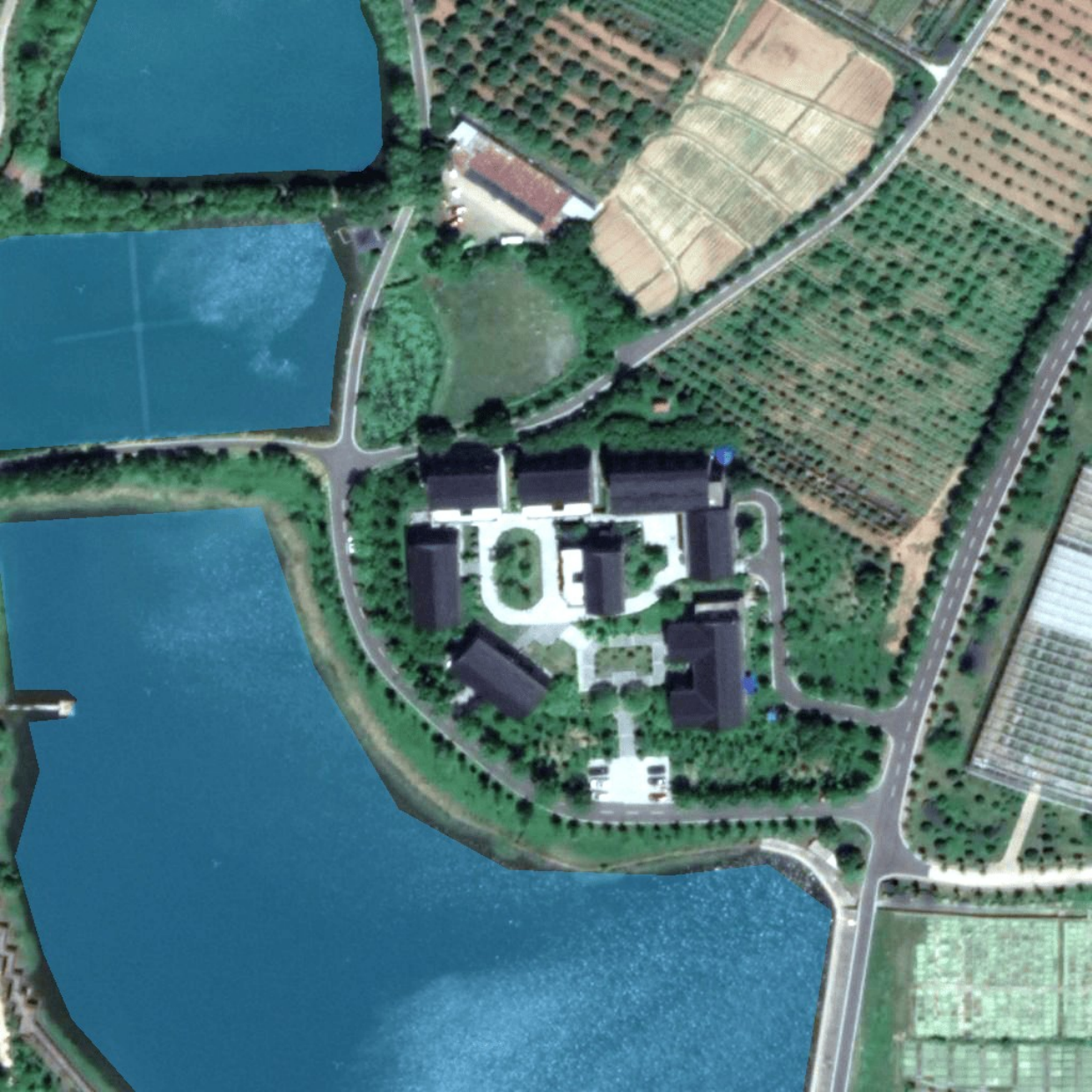} &
\includegraphics[width=0.155\linewidth]{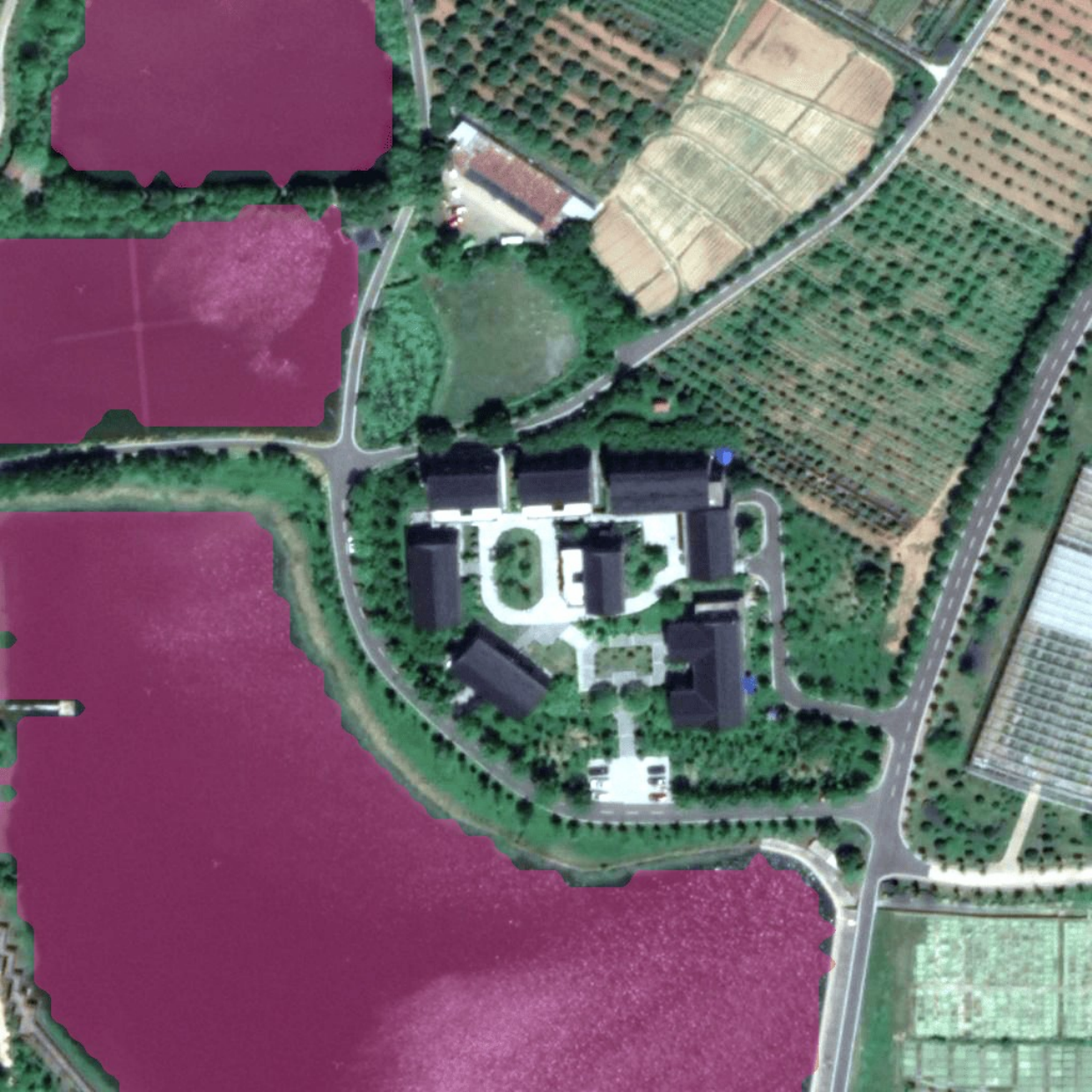} \\
\includegraphics[width=0.155\linewidth]{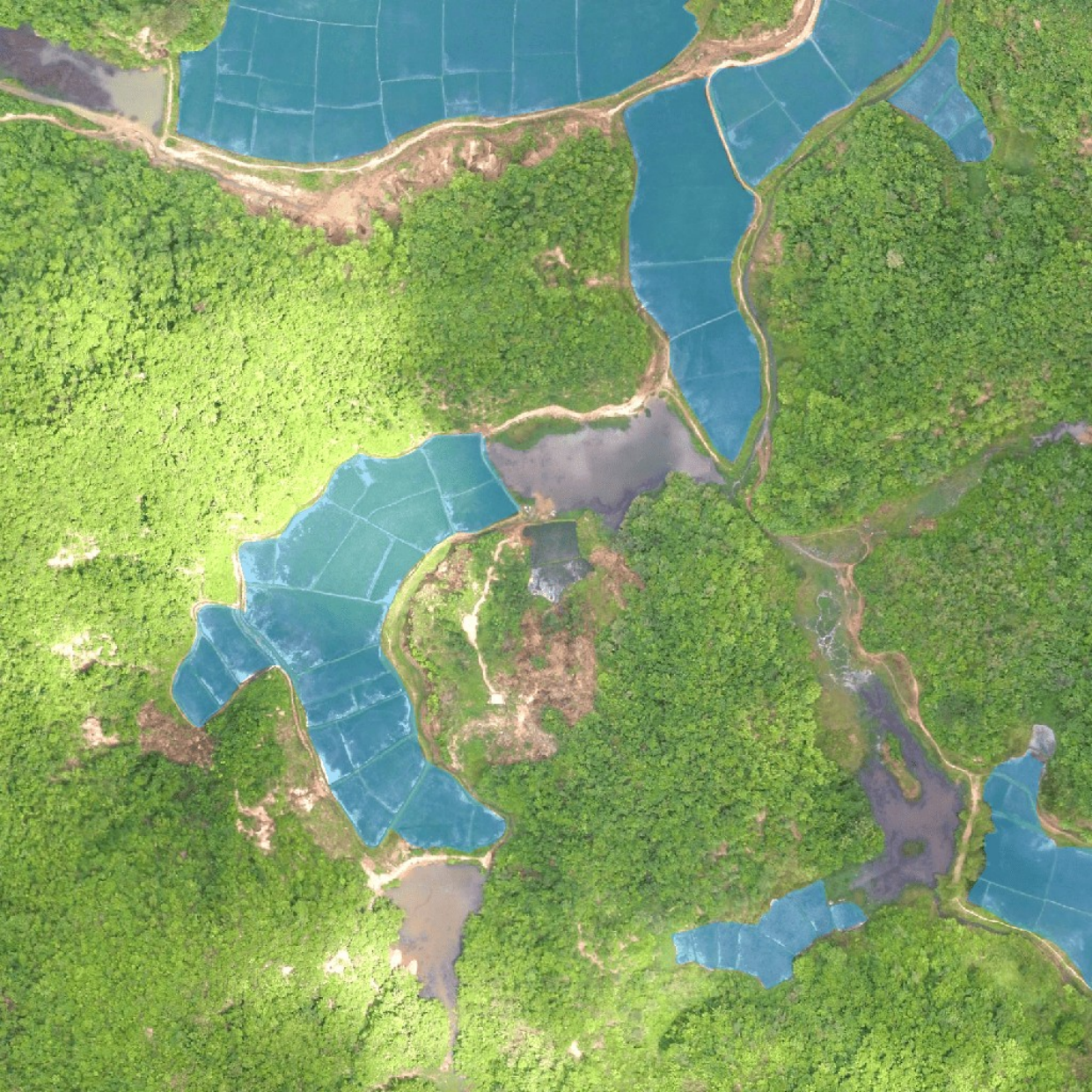} &
\includegraphics[width=0.155\linewidth]{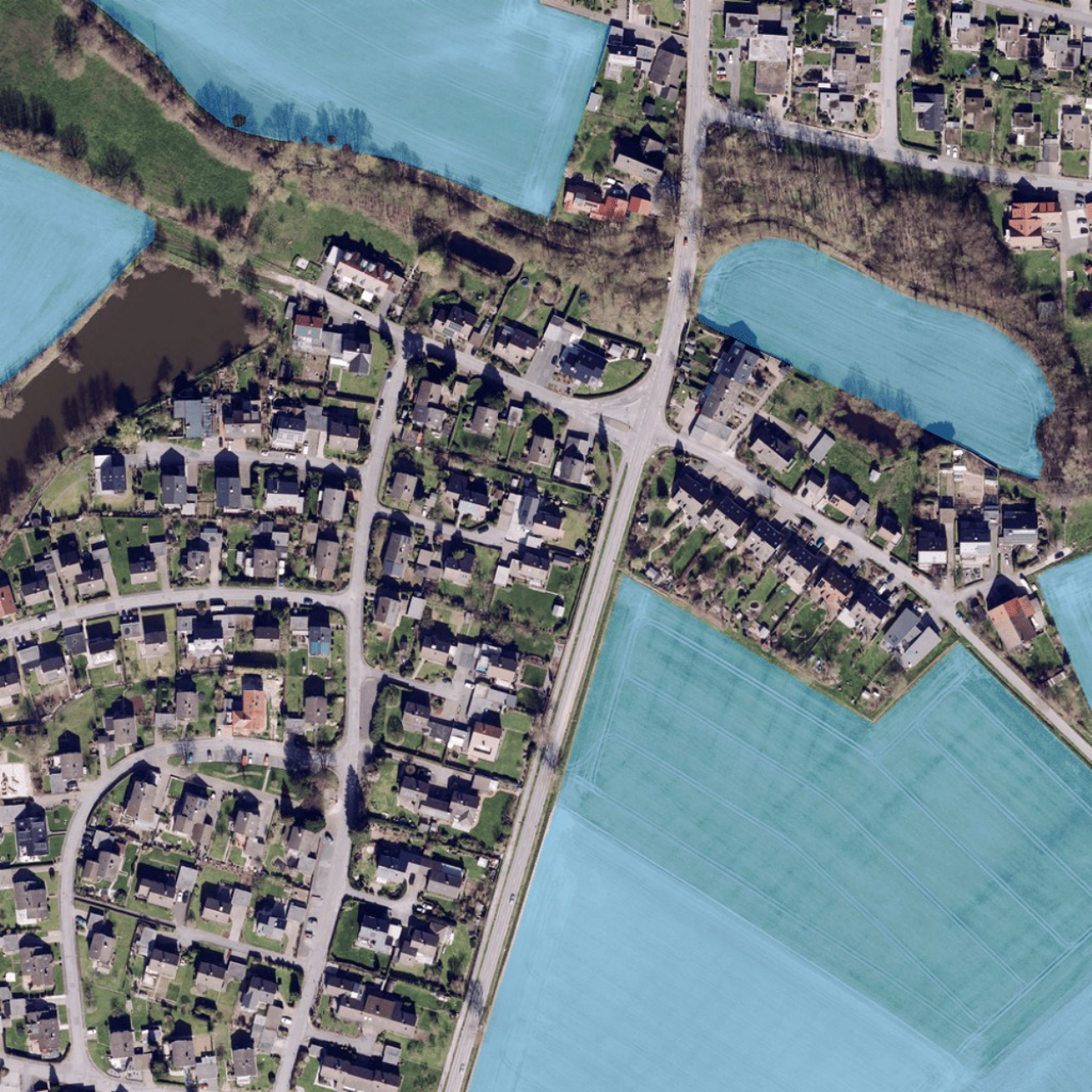} &
\includegraphics[width=0.155\linewidth]{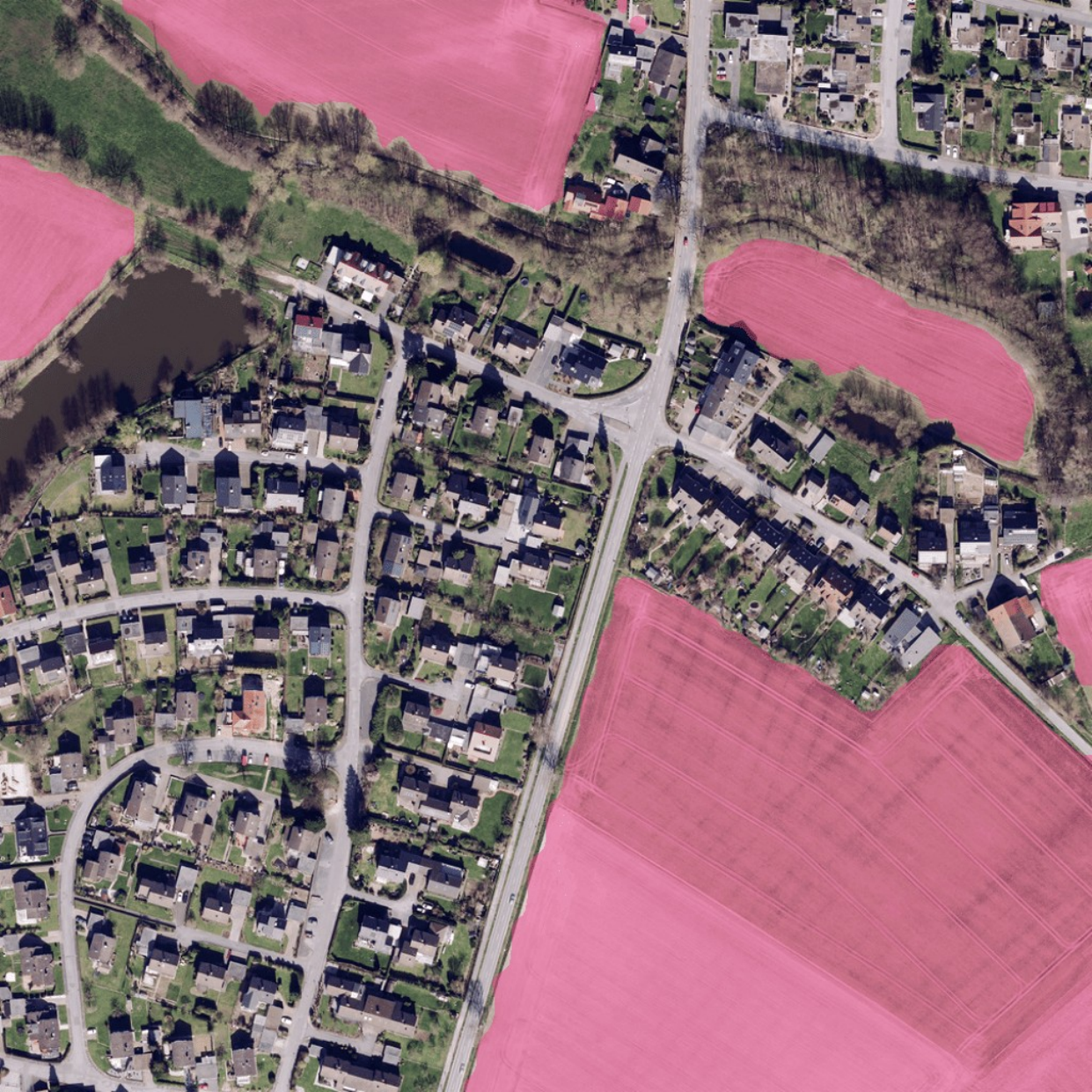} &
\includegraphics[width=0.155\linewidth]{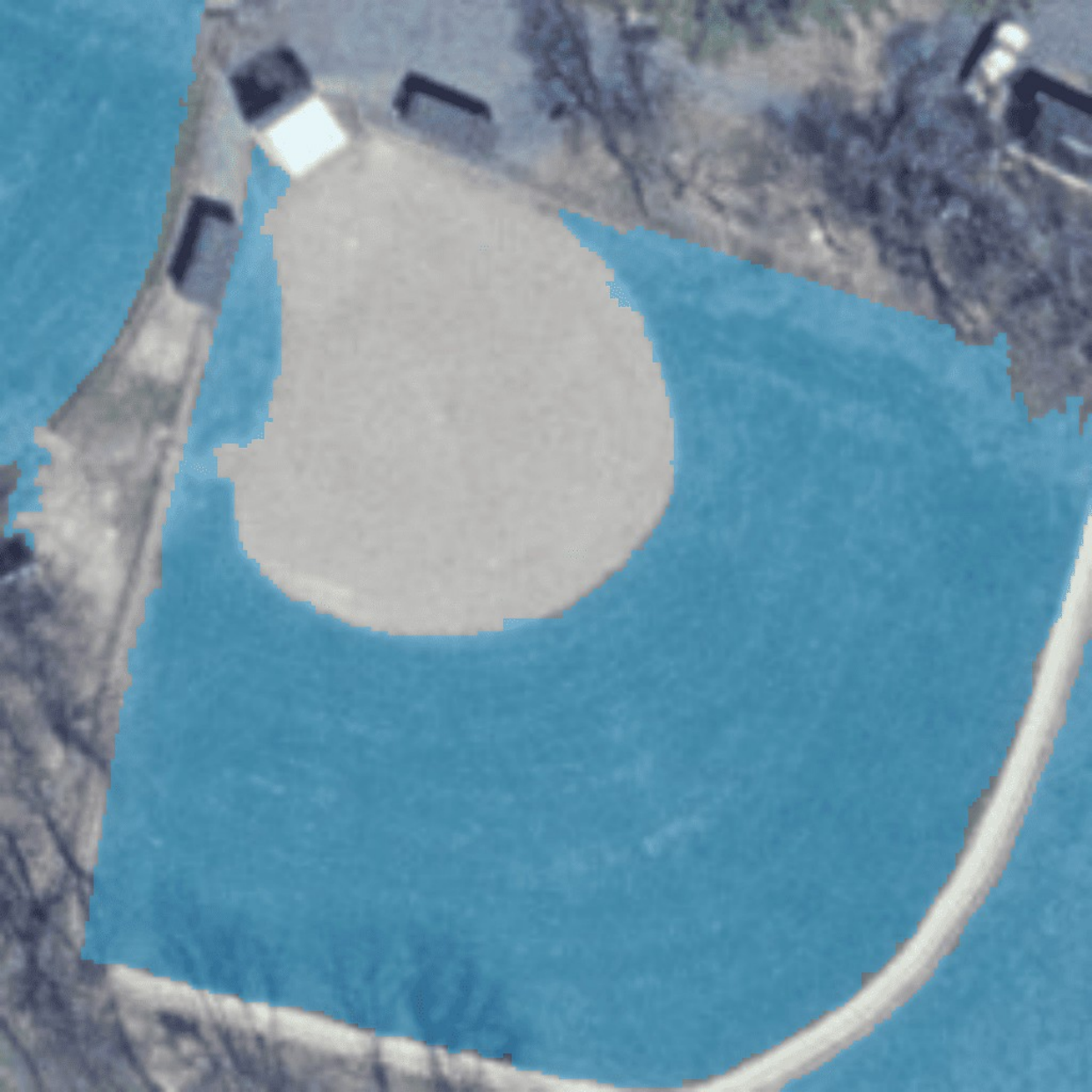} &
\includegraphics[width=0.155\linewidth]{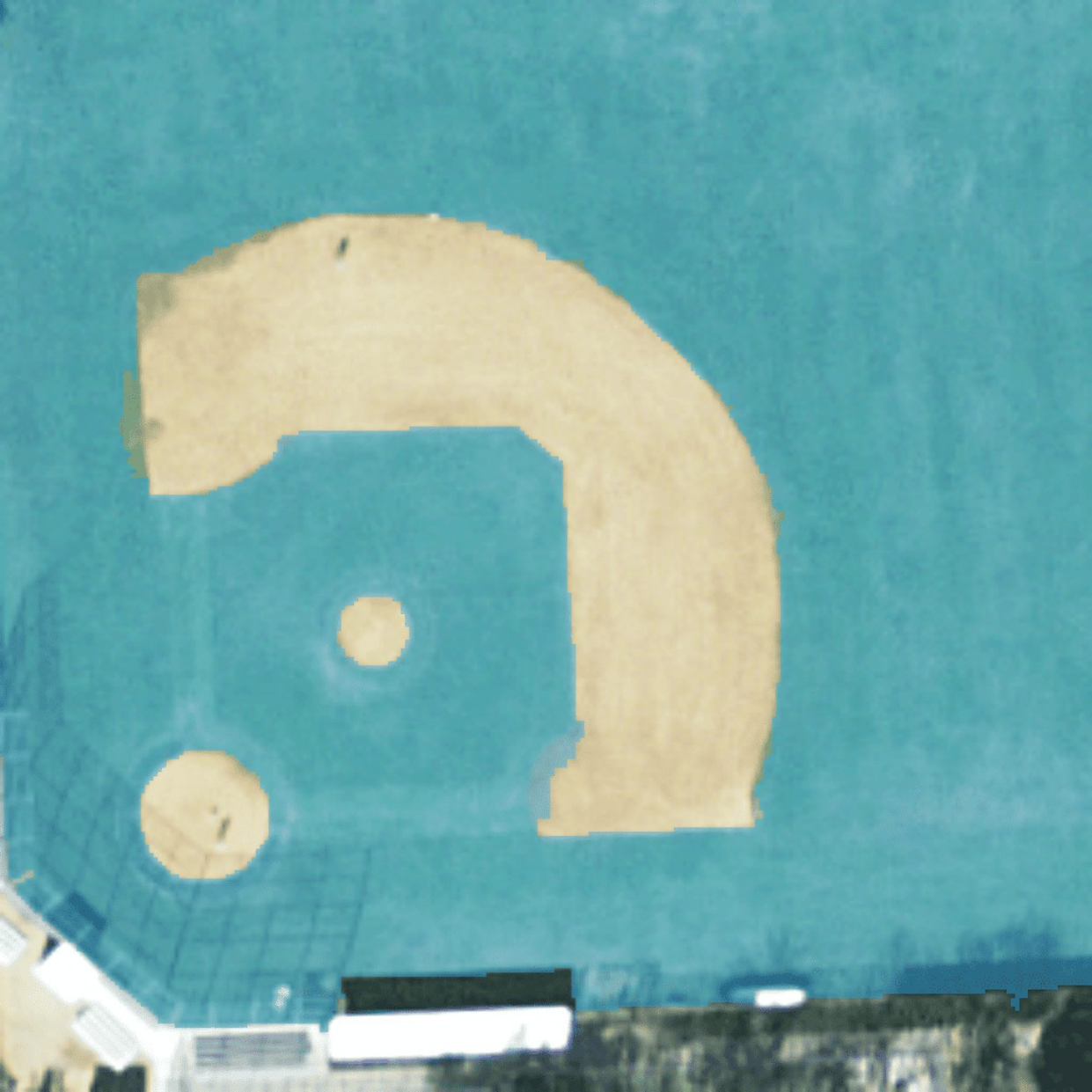} &
\includegraphics[width=0.155\linewidth]{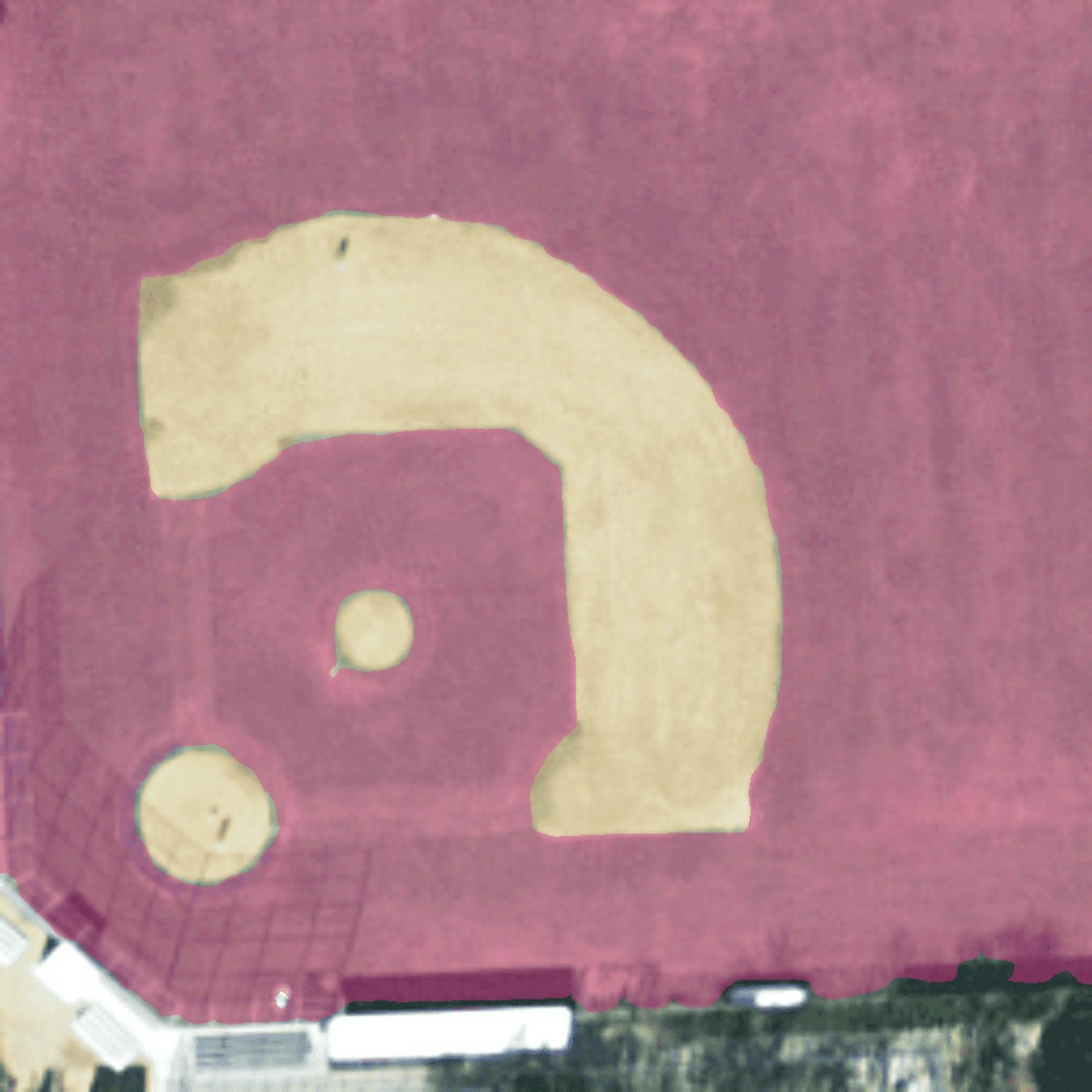} \\
\end{tabular}
\caption{Qualitative results of FROST on the remote-sensing datasets. Each triplet shows, from left to right, the reference image with its mask, the query image with its mask, and the prediction of FROST, with two triplets per row separated by a gap. All examples are one-shot predictions obtained from a frozen DINOv3 backbone without any training.}
\label{fig:qual_rs}
\end{figure}

This appendix complements the quantitative comparison with a visual inspection of the masks that FROST produces. Figure~\ref{fig:qual_rs} collects sixteen one-shot examples drawn from the remote-sensing benchmarks, spanning the building and footprint, land-cover and urban-scene, and flood groups, so that the reader can judge segmentation quality across the range of targets that the benchmarks cover. Every example is a one-shot prediction obtained from a frozen DINOv3 backbone with no training, and the prediction is overlaid on the query image in the same colour as the reference mask. The reference column shows the one annotated image and mask pair that defines the target class, the ground-truth column shows the mask of the query that the prediction is compared against, and the column labelled ours shows the FROST prediction.

The examples illustrate the behaviour that the quantitative results summarize. On targets that recur as many small and visually distinct instances, such as the buildings and vehicles of the building and urban-scene benchmarks, FROST recovers the separate instances rather than merging them into a single blob, which is the regime in which the density ratio is expected to improve on a single prototype. On the large and continuous regions of the land-cover and flood benchmarks the predicted boundaries follow the query edges closely, a consequence of the bilateral propagation and the upsampling before thresholding. The failure cases that remain are concentrated on classes whose appearance in the query departs strongly from the single reference, where the reference statistics are least representative of the query.

\begin{figure}[t]
\centering
\setlength{\tabcolsep}{1.5pt}
\renewcommand{\arraystretch}{0.4}
\begin{tabular}{ccc@{\hspace{10pt}}ccc}
\footnotesize Reference & \footnotesize GT & \footnotesize Ours &
\footnotesize Reference & \footnotesize GT & \footnotesize Ours \\[2pt]
\includegraphics[width=0.155\linewidth]{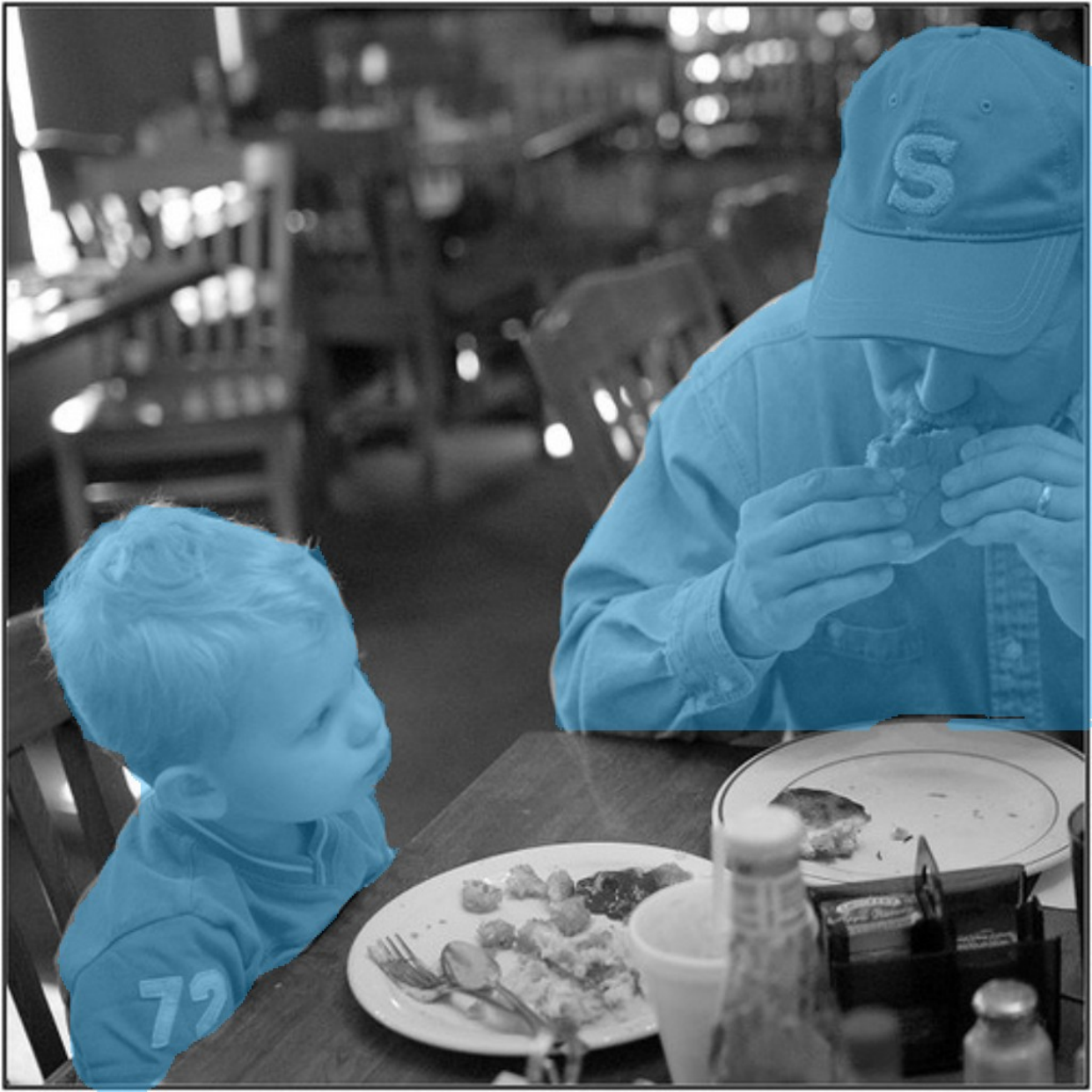} &
\includegraphics[width=0.155\linewidth]{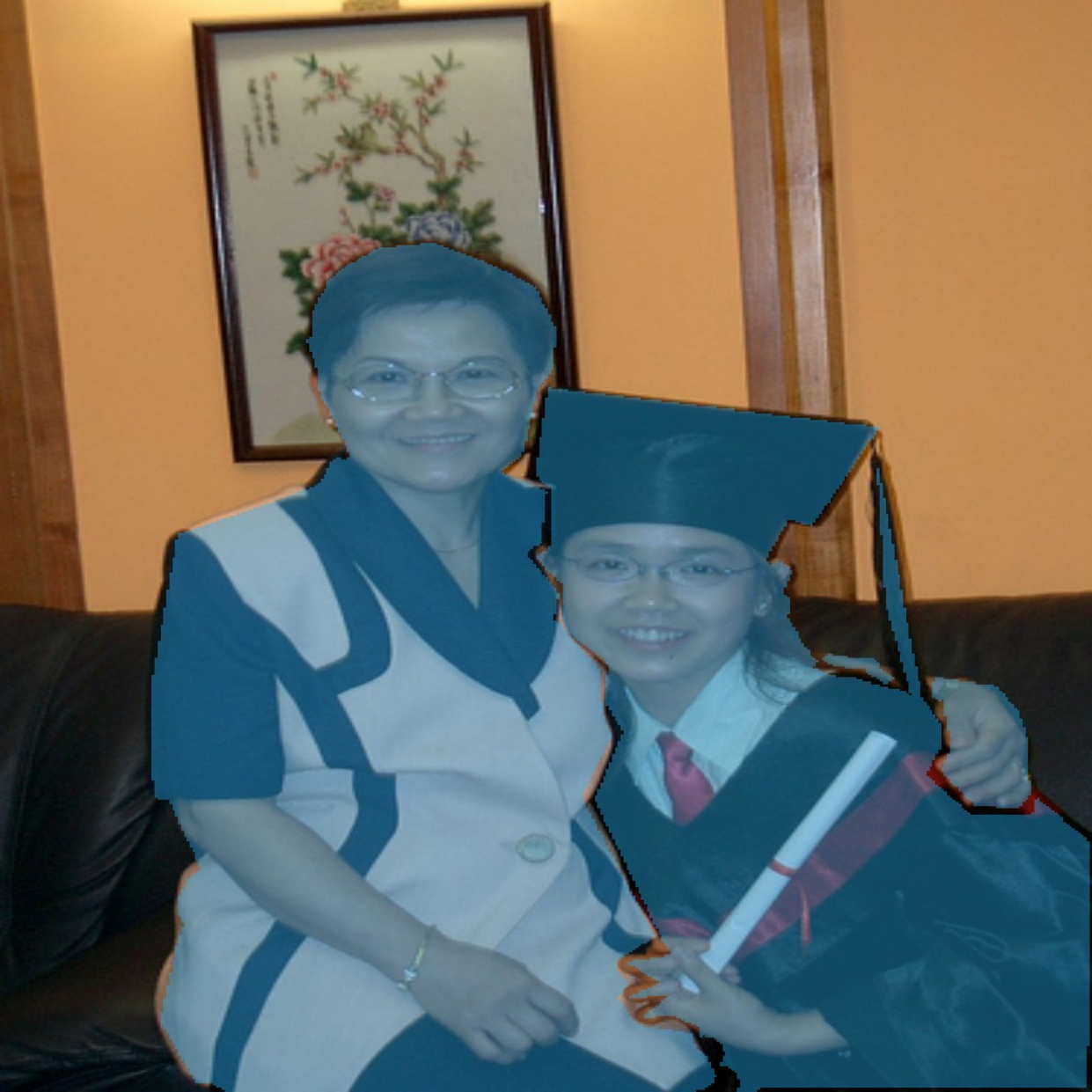} &
\includegraphics[width=0.155\linewidth]{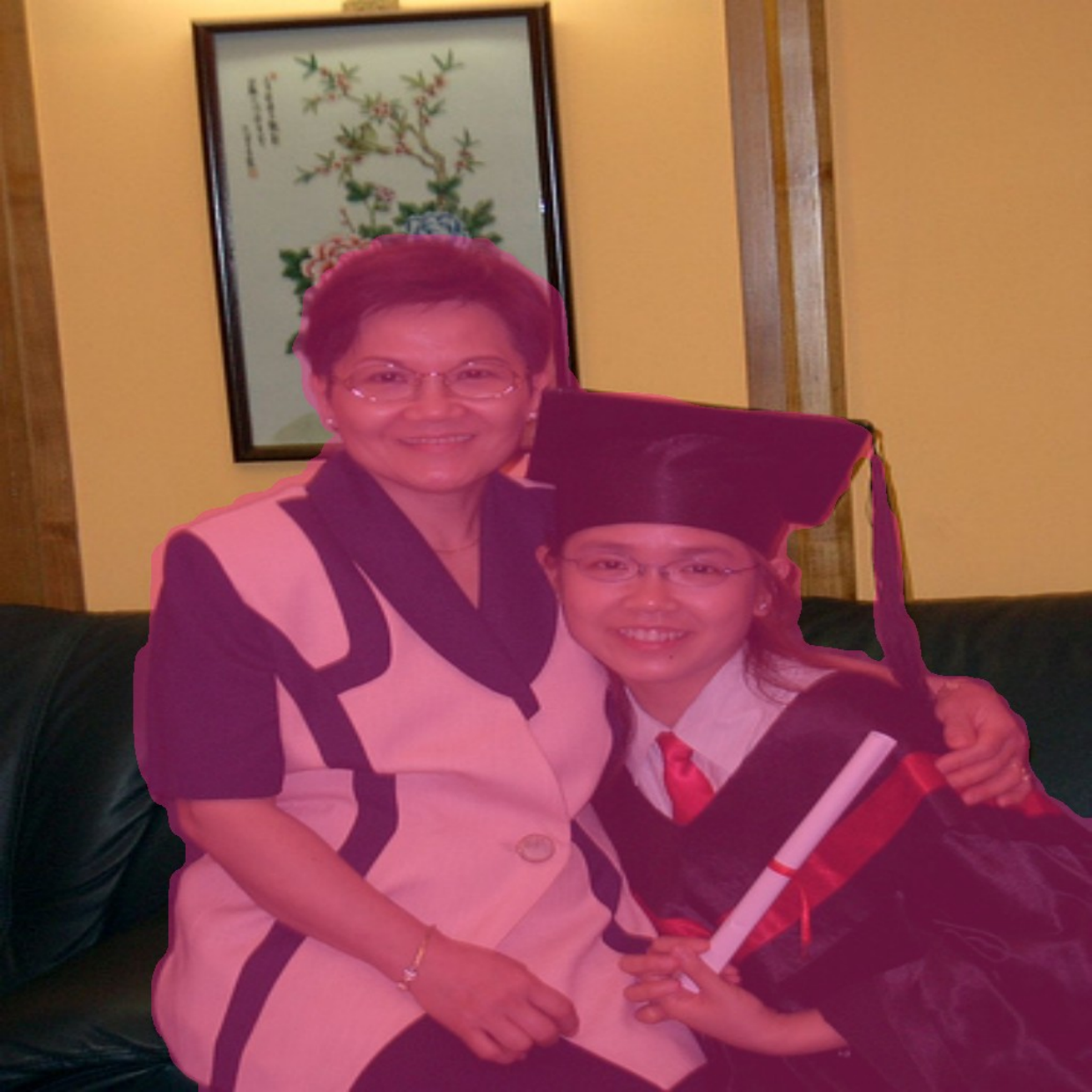} &
\includegraphics[width=0.155\linewidth]{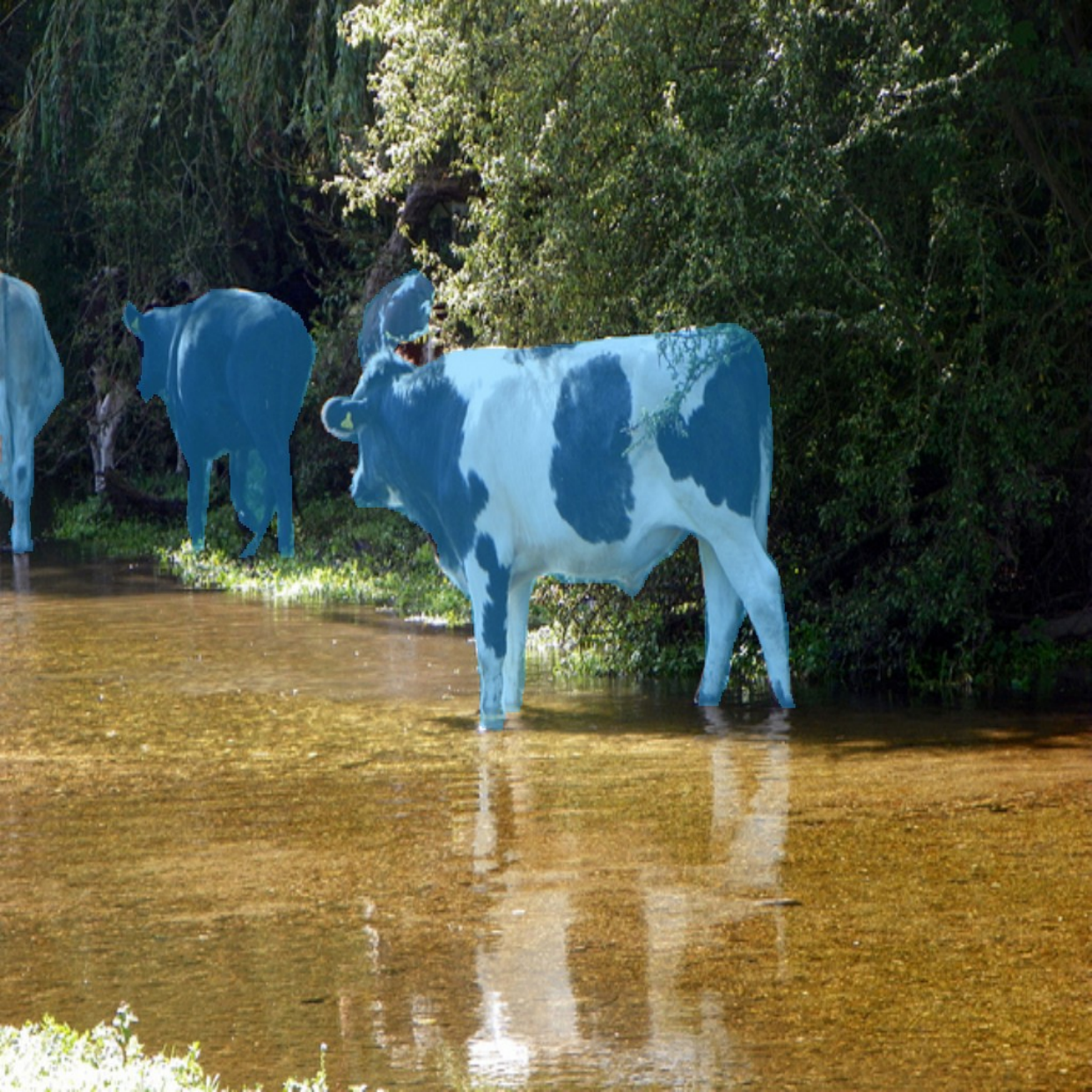} &
\includegraphics[width=0.155\linewidth]{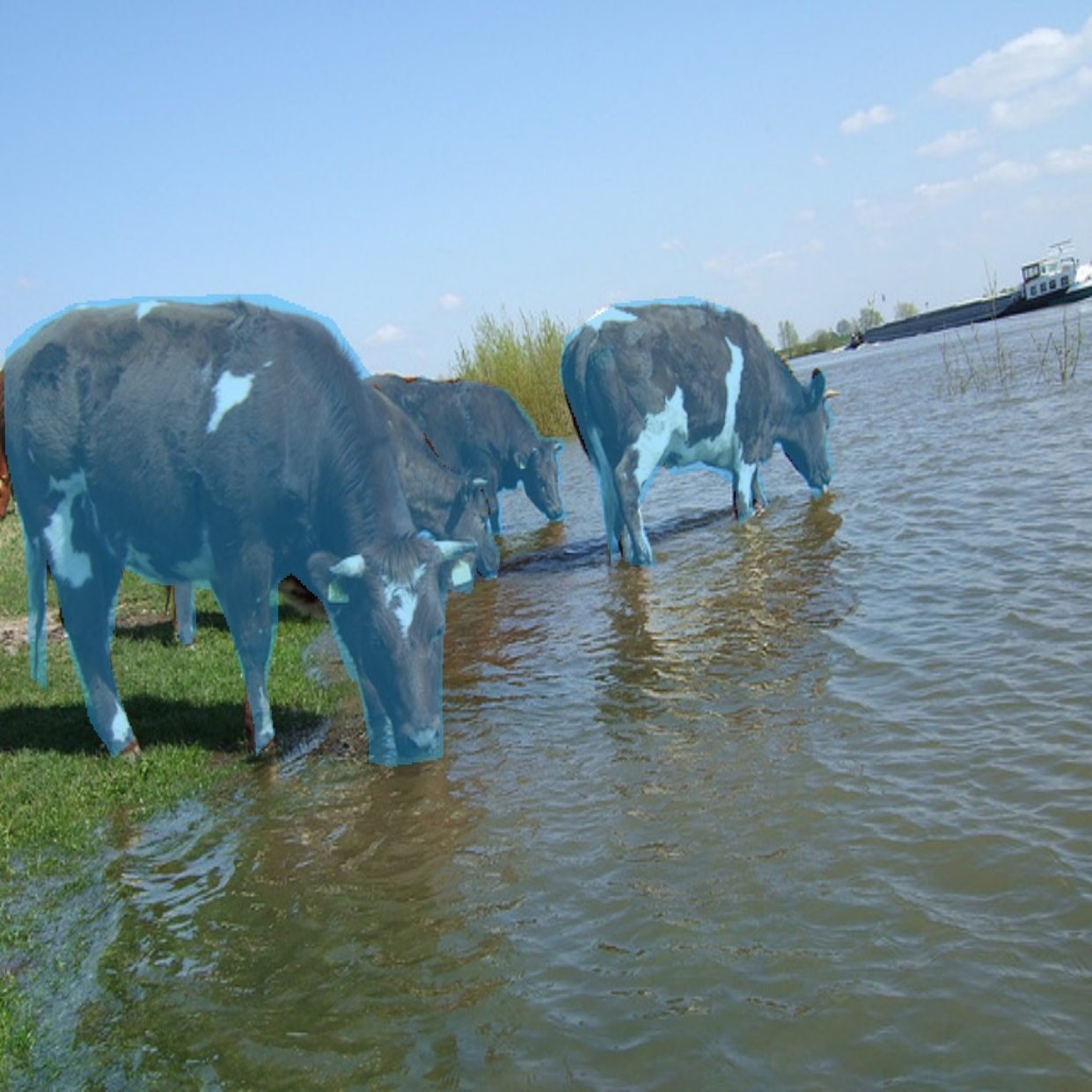} &
\includegraphics[width=0.155\linewidth]{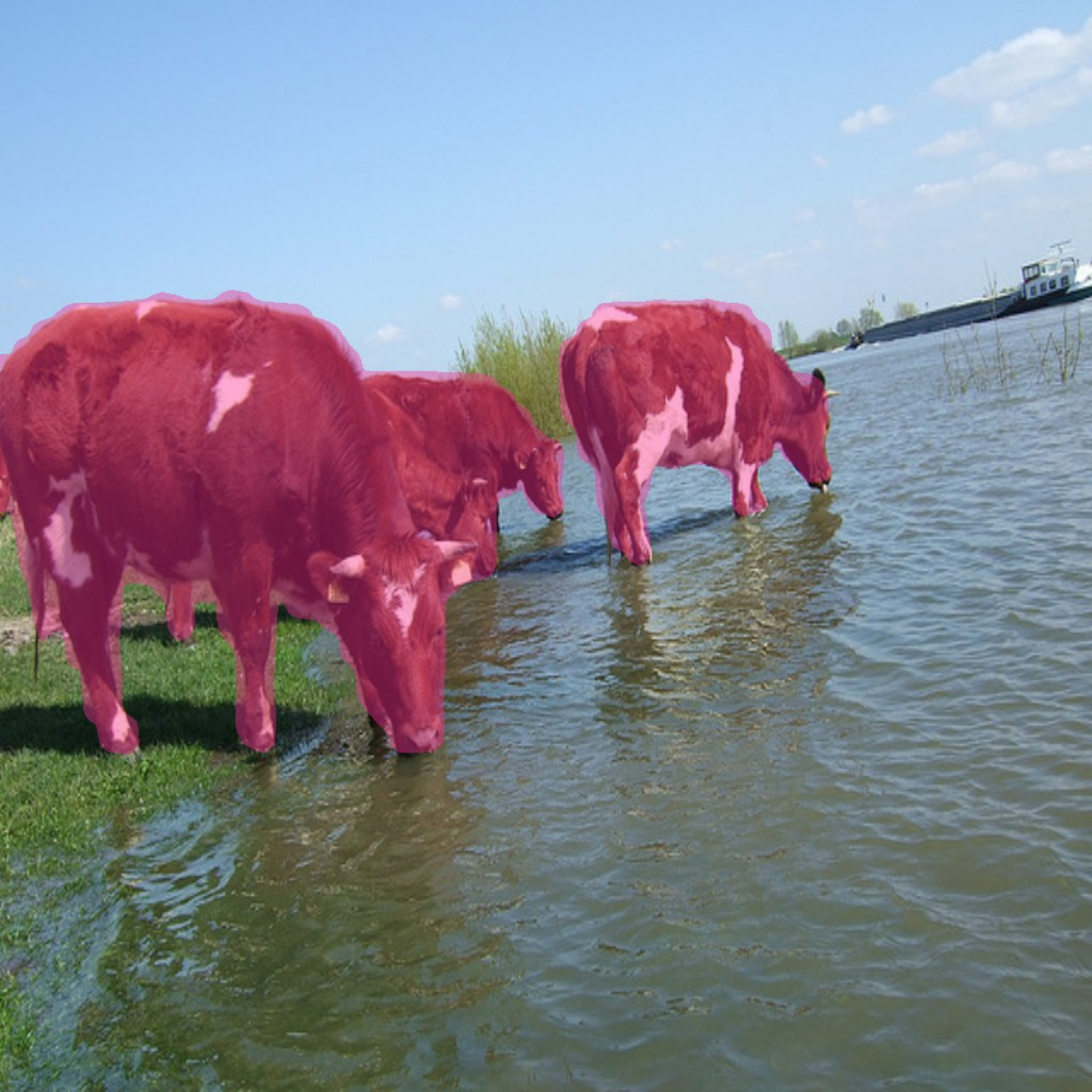} \\
\includegraphics[width=0.155\linewidth]{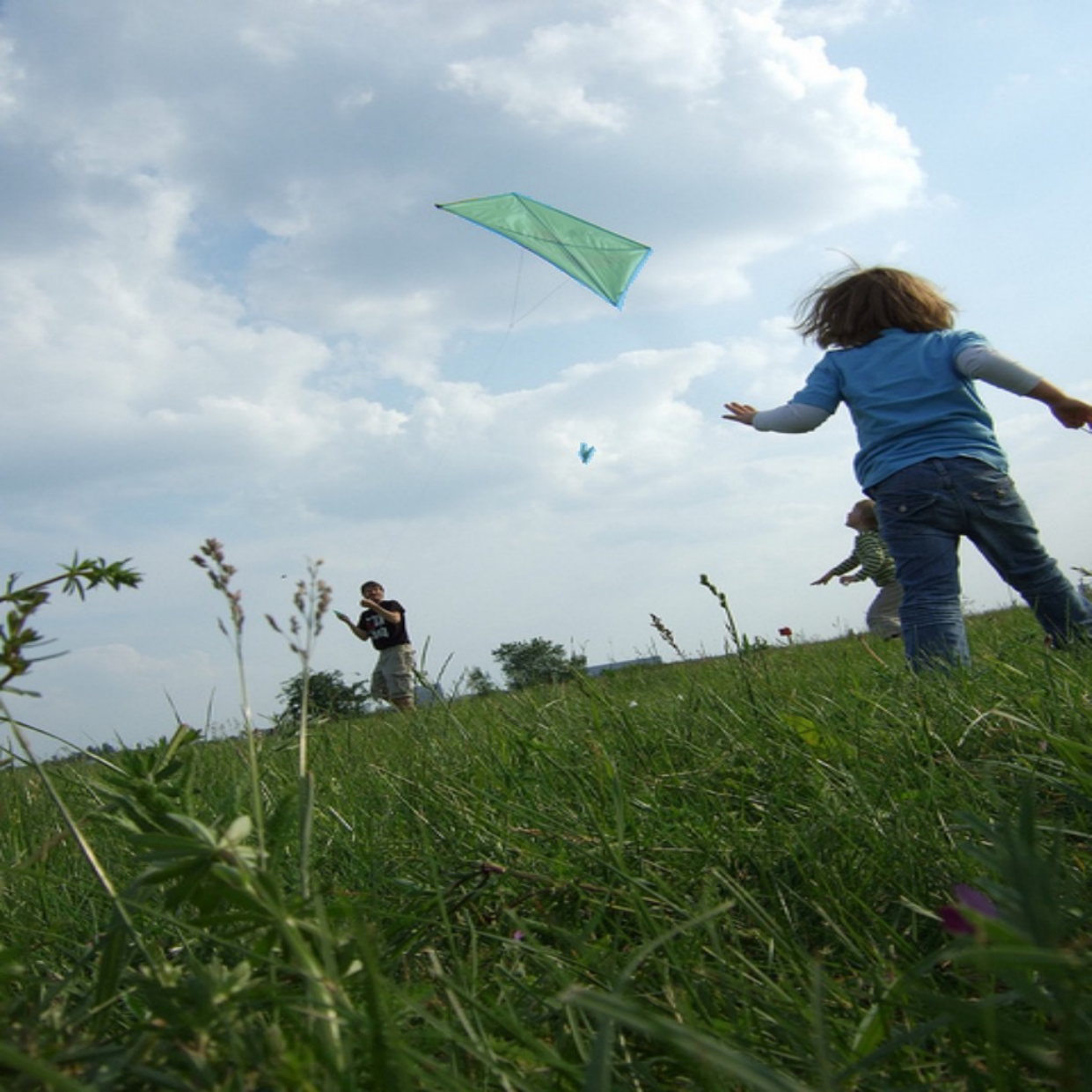} &
\includegraphics[width=0.155\linewidth]{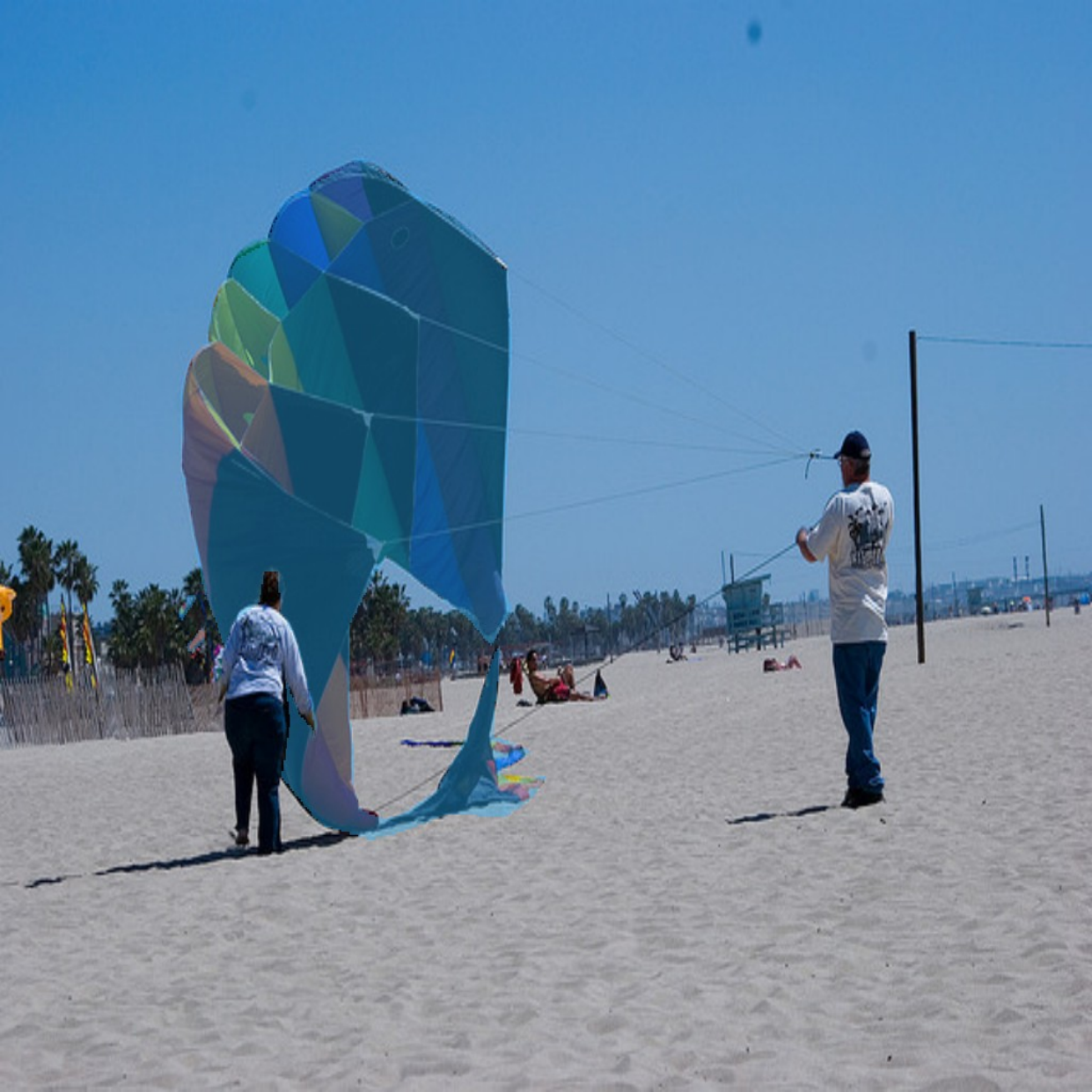} &
\includegraphics[width=0.155\linewidth]{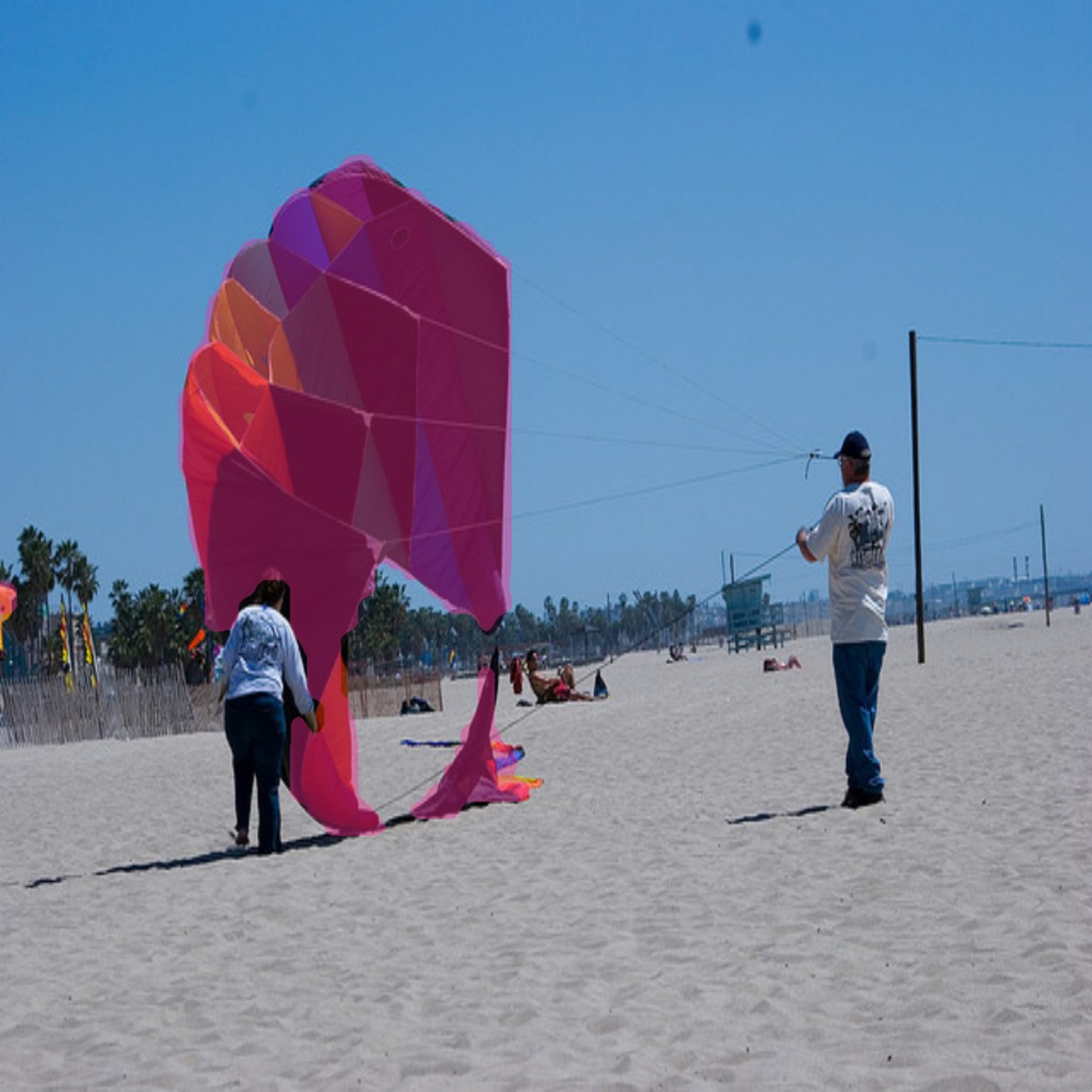} &
\includegraphics[width=0.155\linewidth]{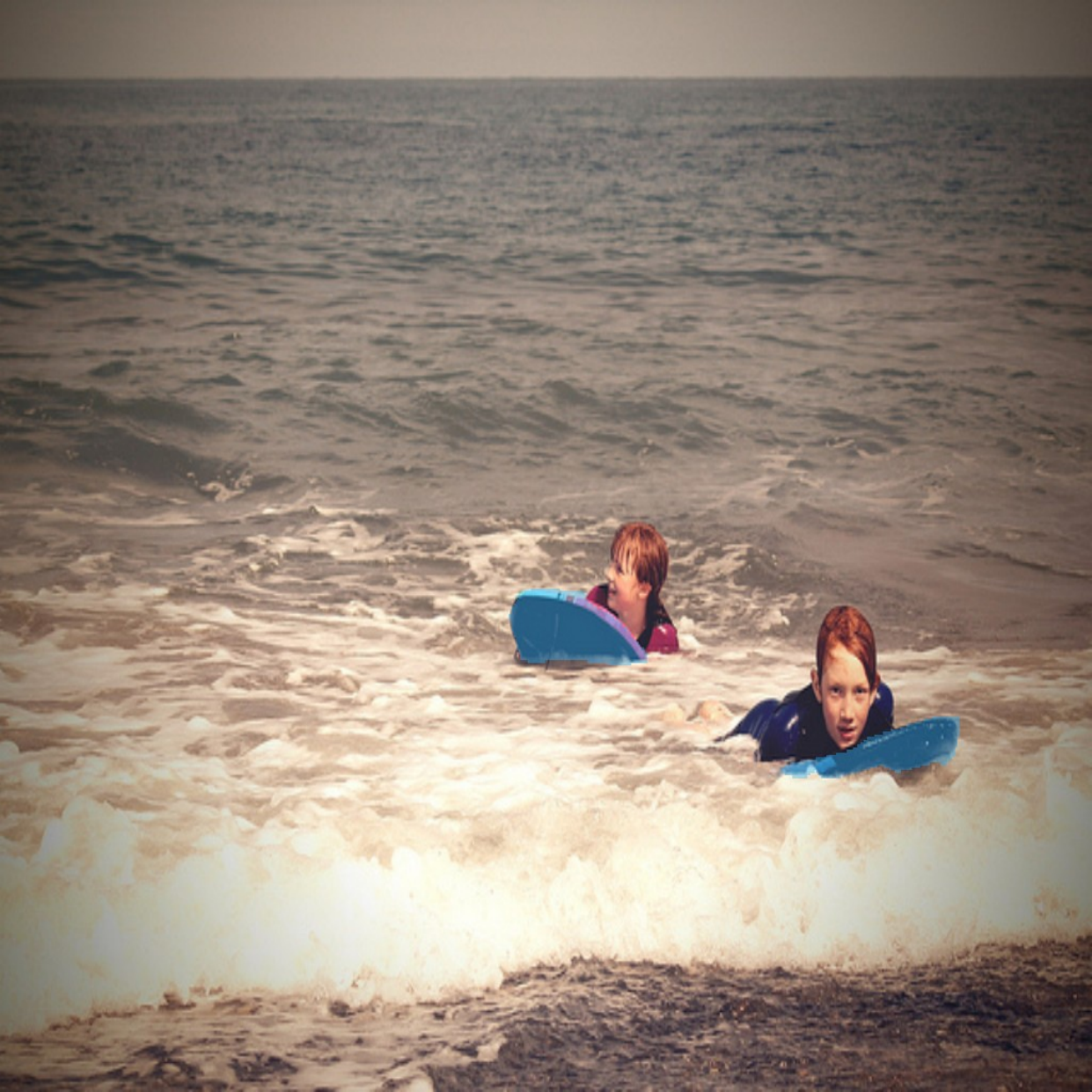} &
\includegraphics[width=0.155\linewidth]{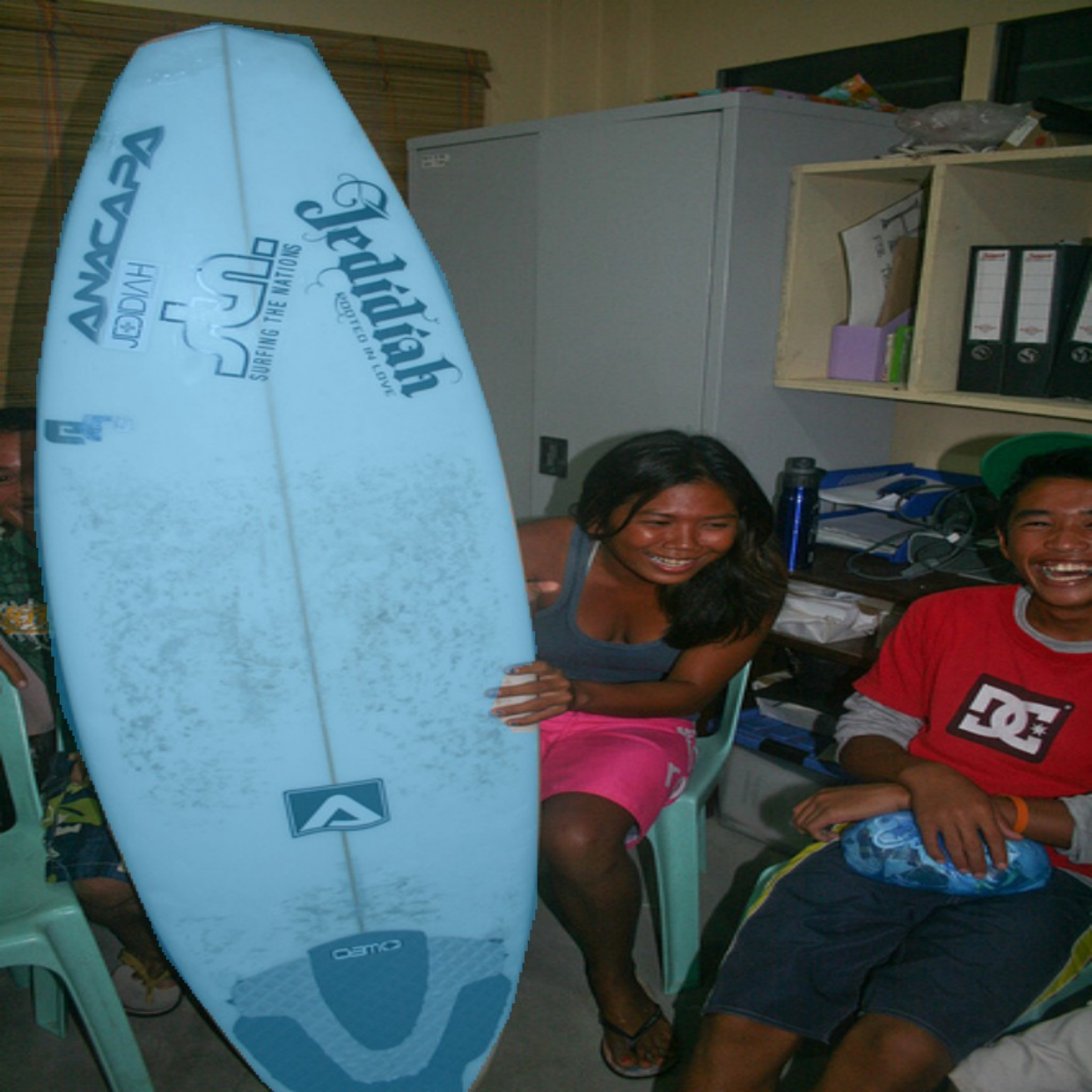} &
\includegraphics[width=0.155\linewidth]{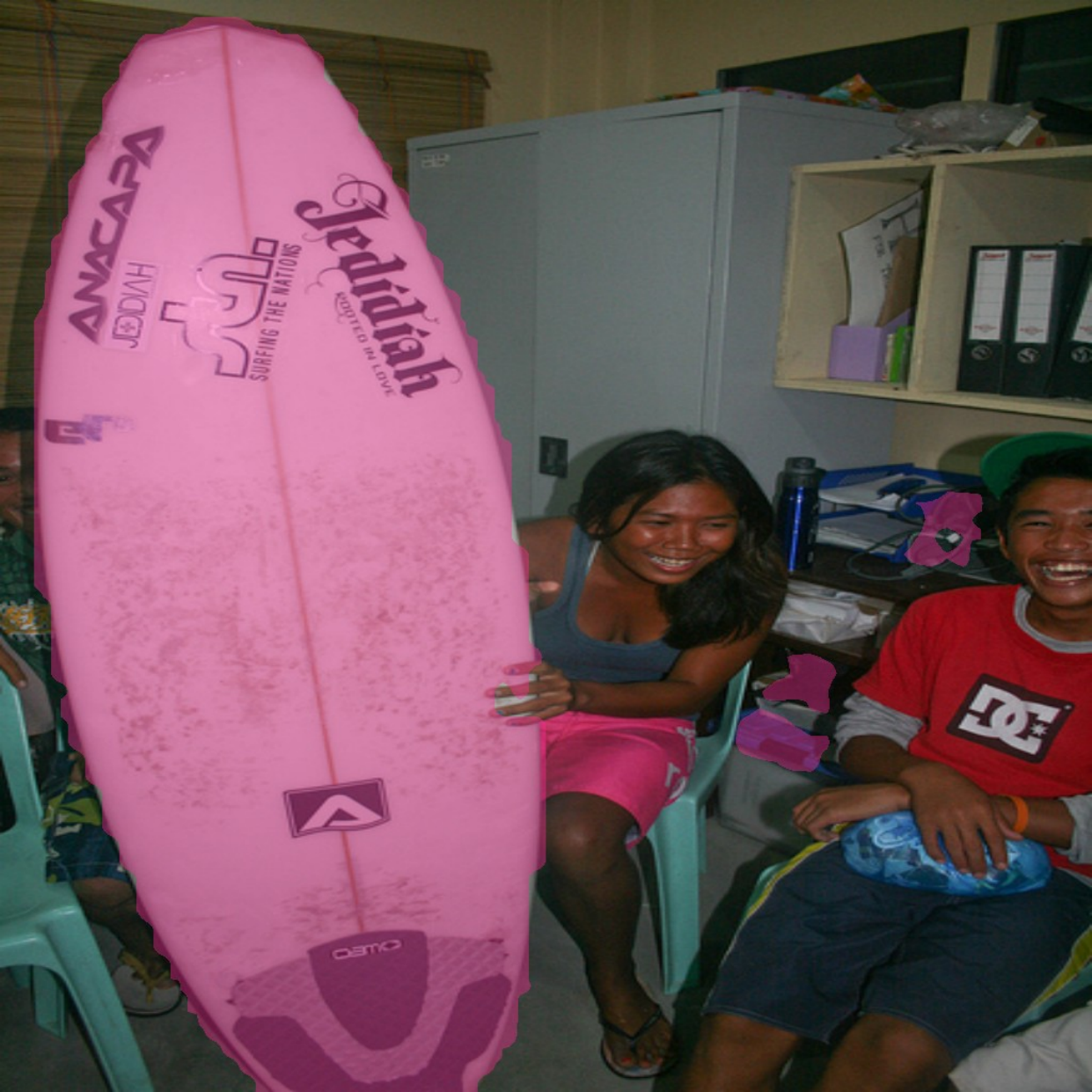} \\
\includegraphics[width=0.155\linewidth]{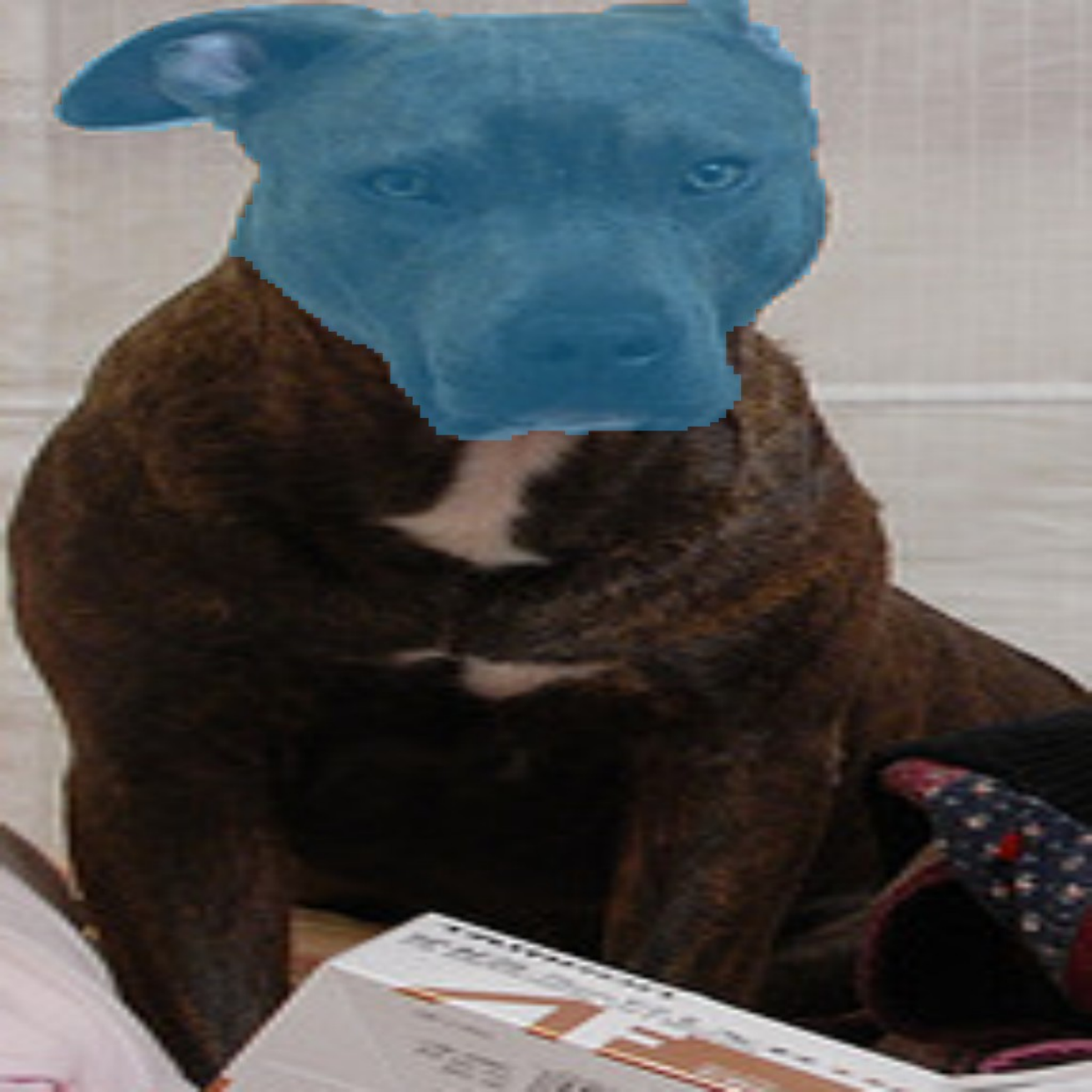} &
\includegraphics[width=0.155\linewidth]{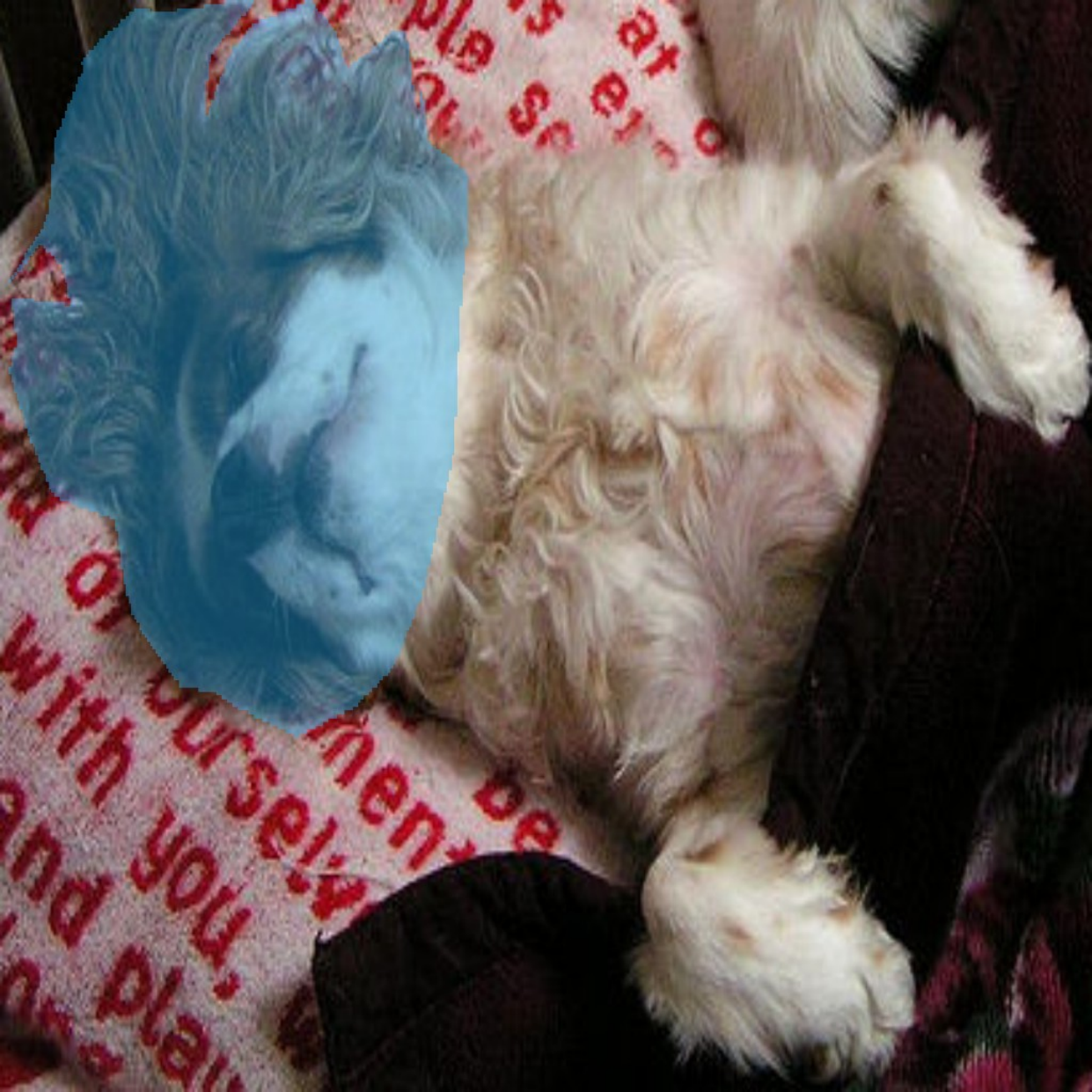} &
\includegraphics[width=0.155\linewidth]{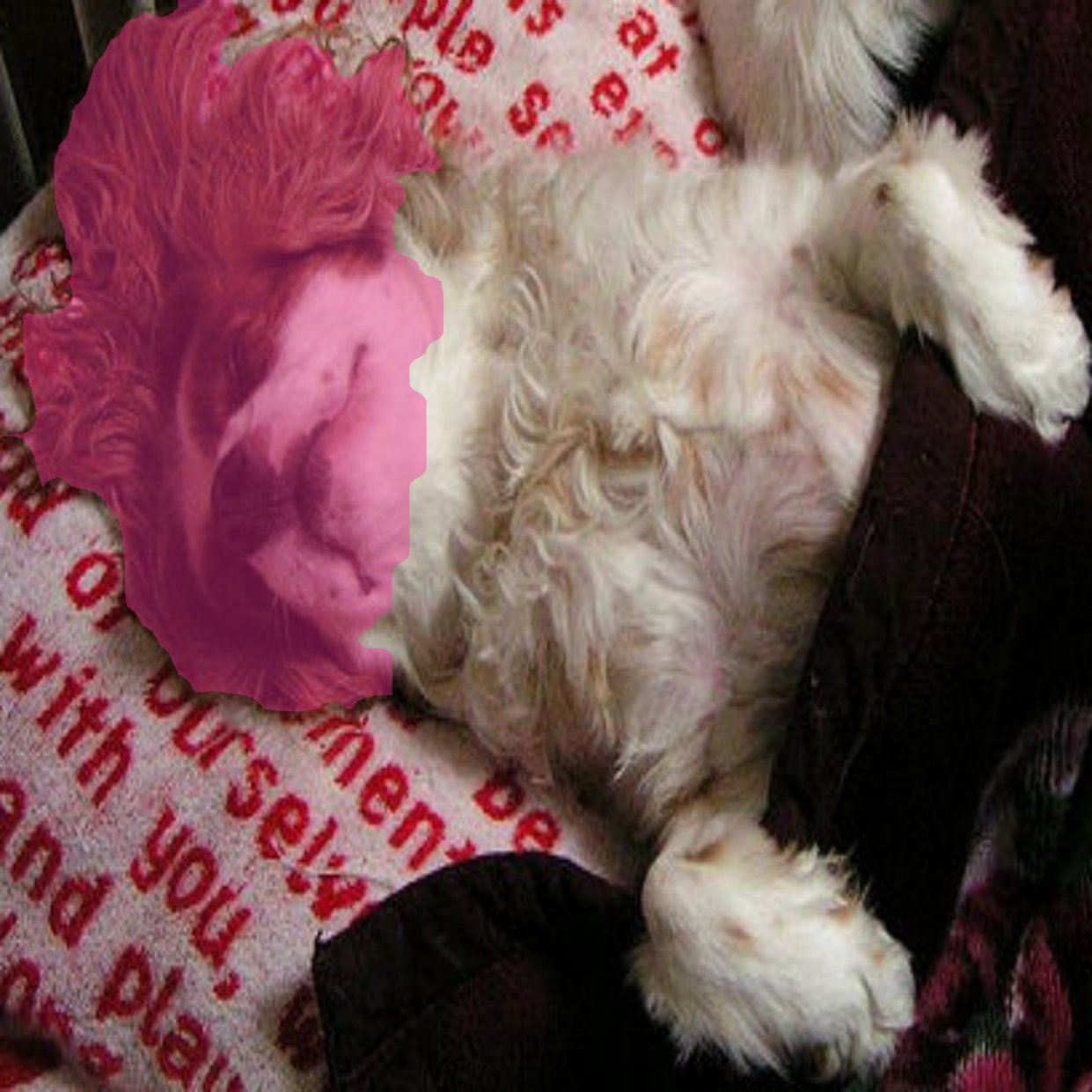} &
\includegraphics[width=0.155\linewidth]{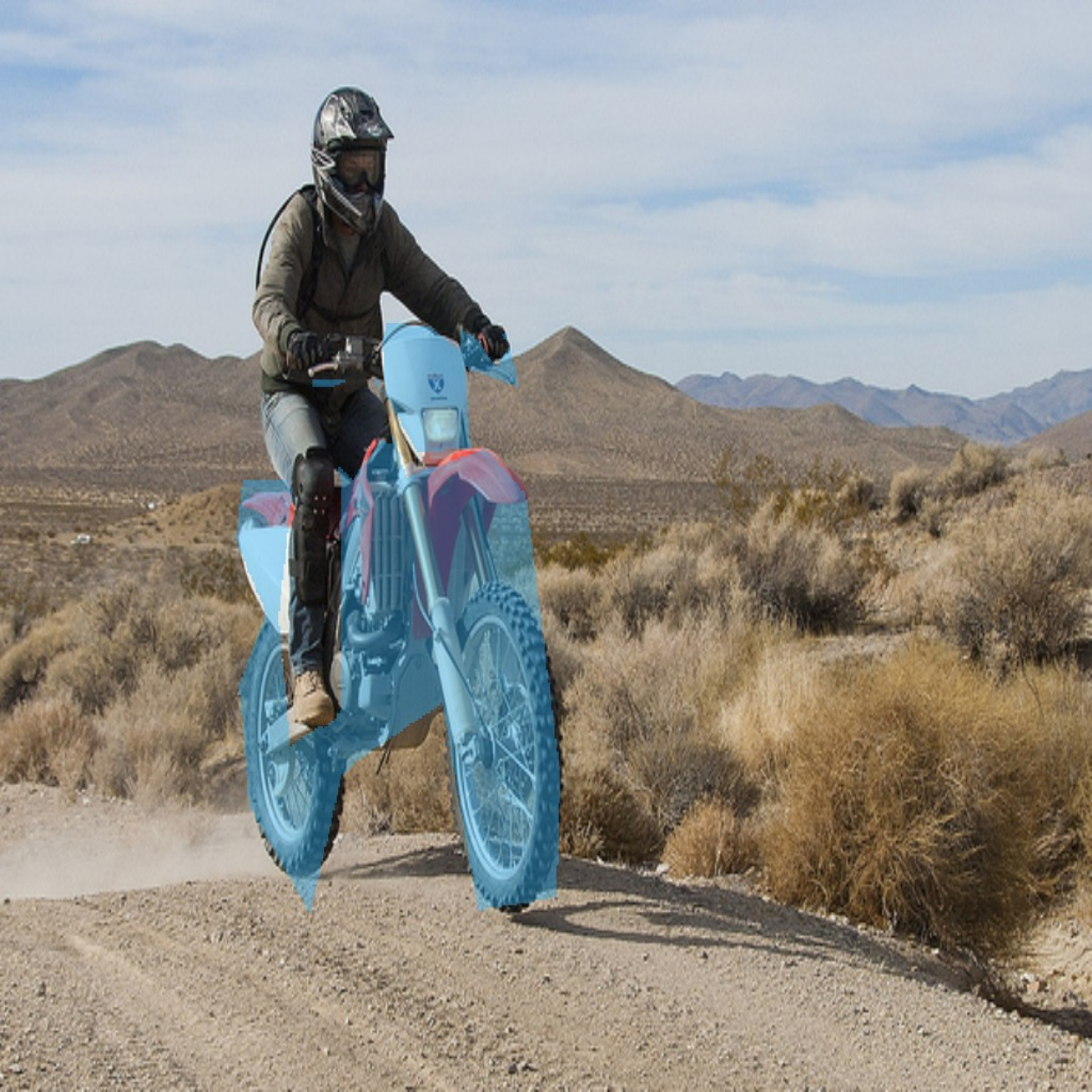} &
\includegraphics[width=0.155\linewidth]{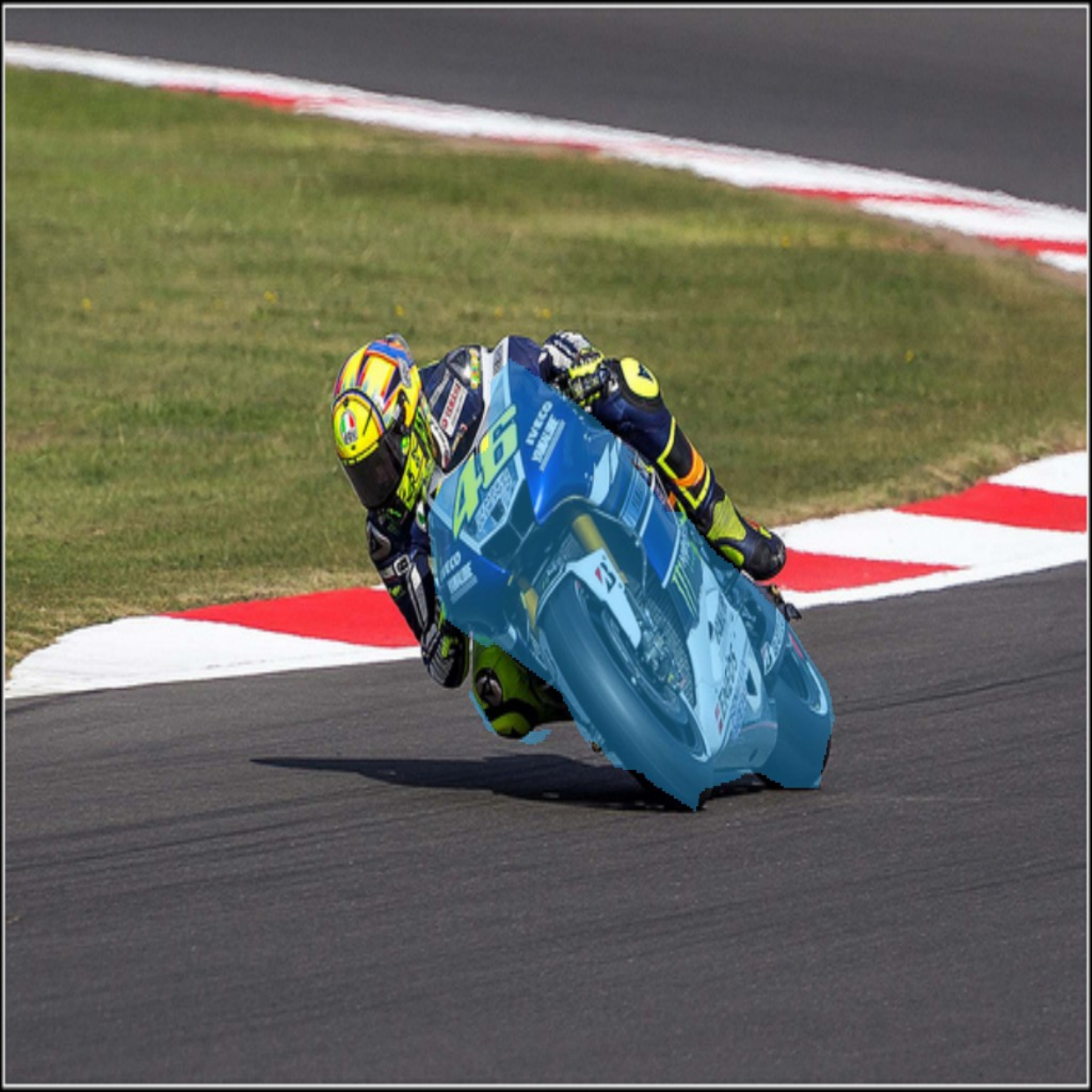} &
\includegraphics[width=0.155\linewidth]{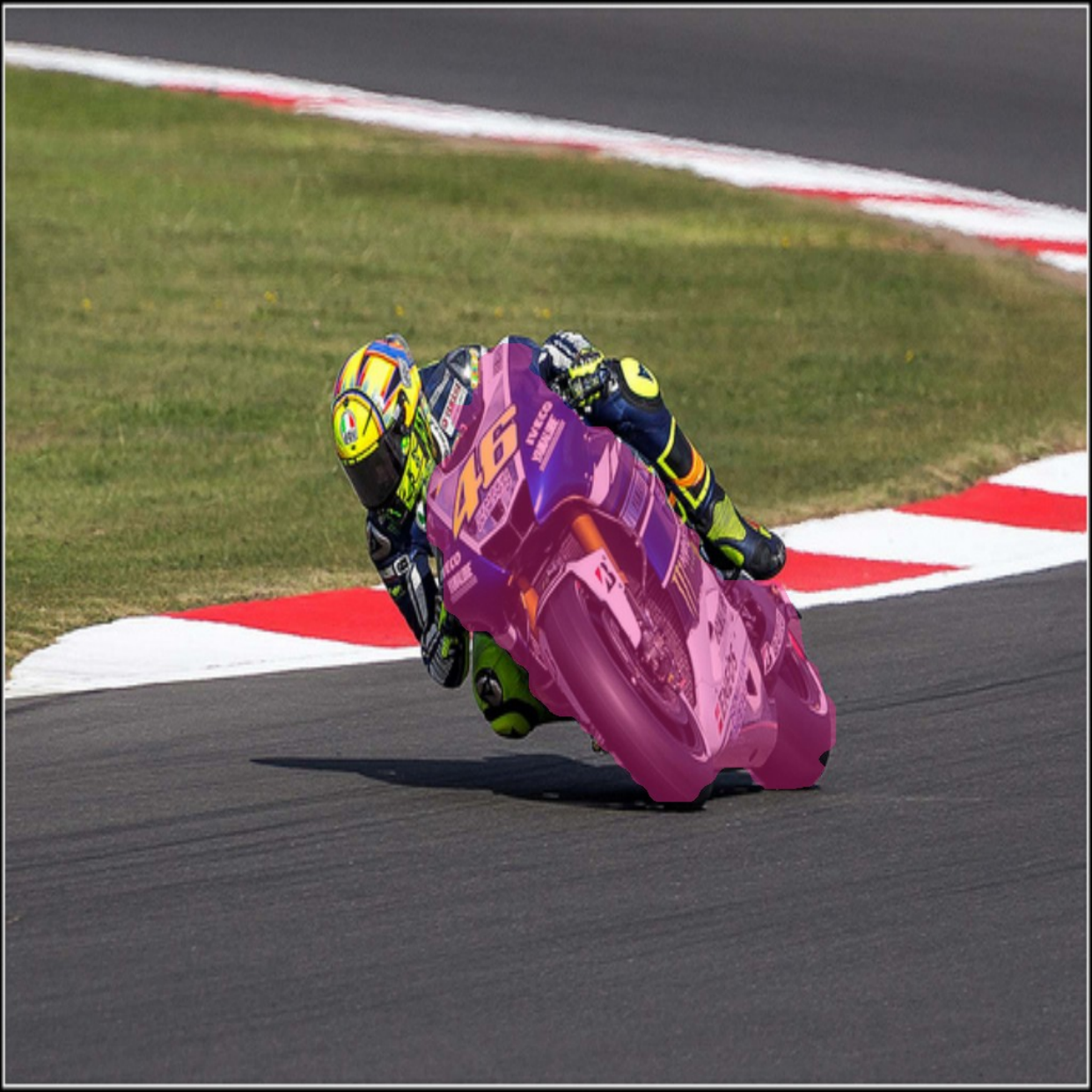} \\
\includegraphics[width=0.155\linewidth]{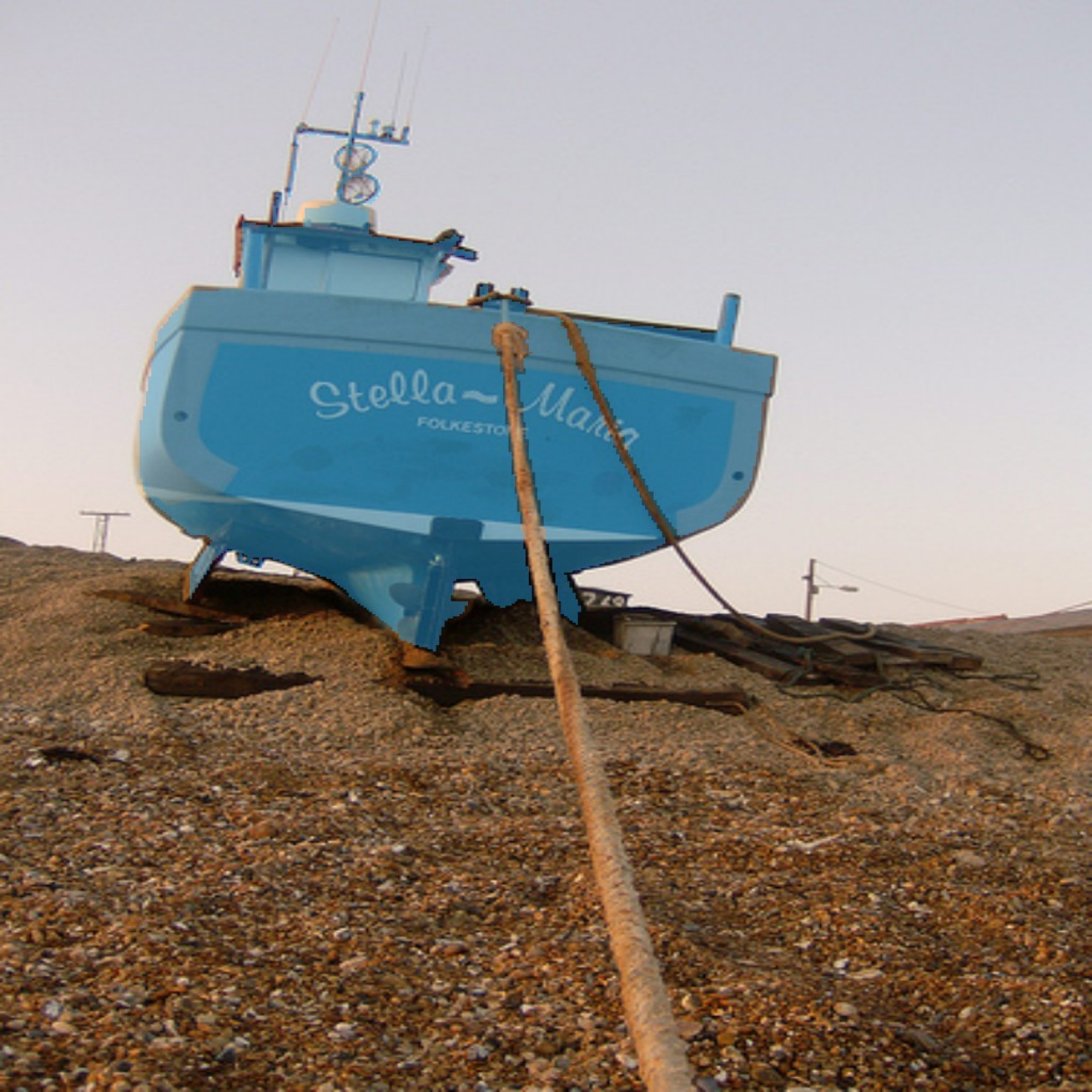} &
\includegraphics[width=0.155\linewidth]{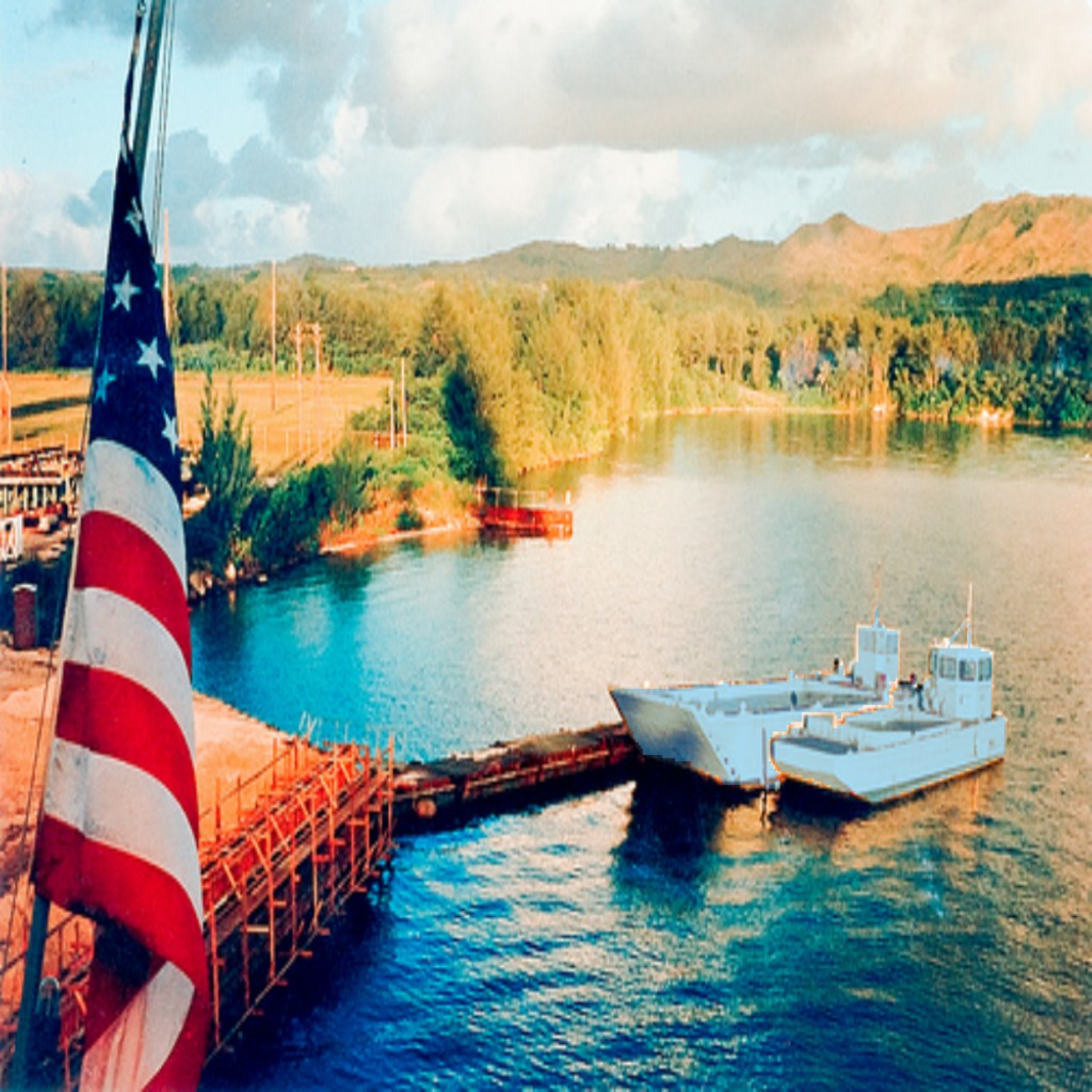} &
\includegraphics[width=0.155\linewidth]{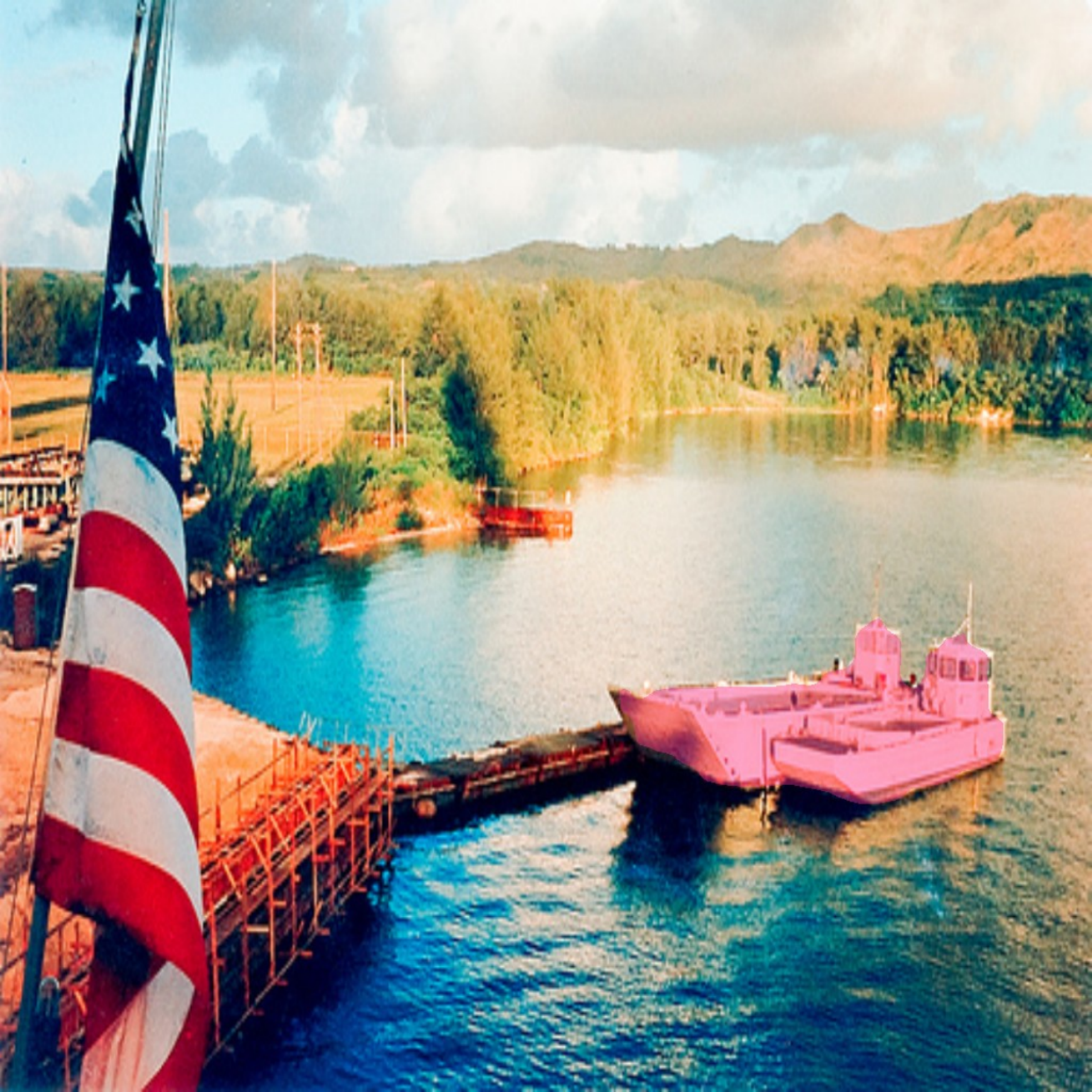} &
\includegraphics[width=0.155\linewidth]{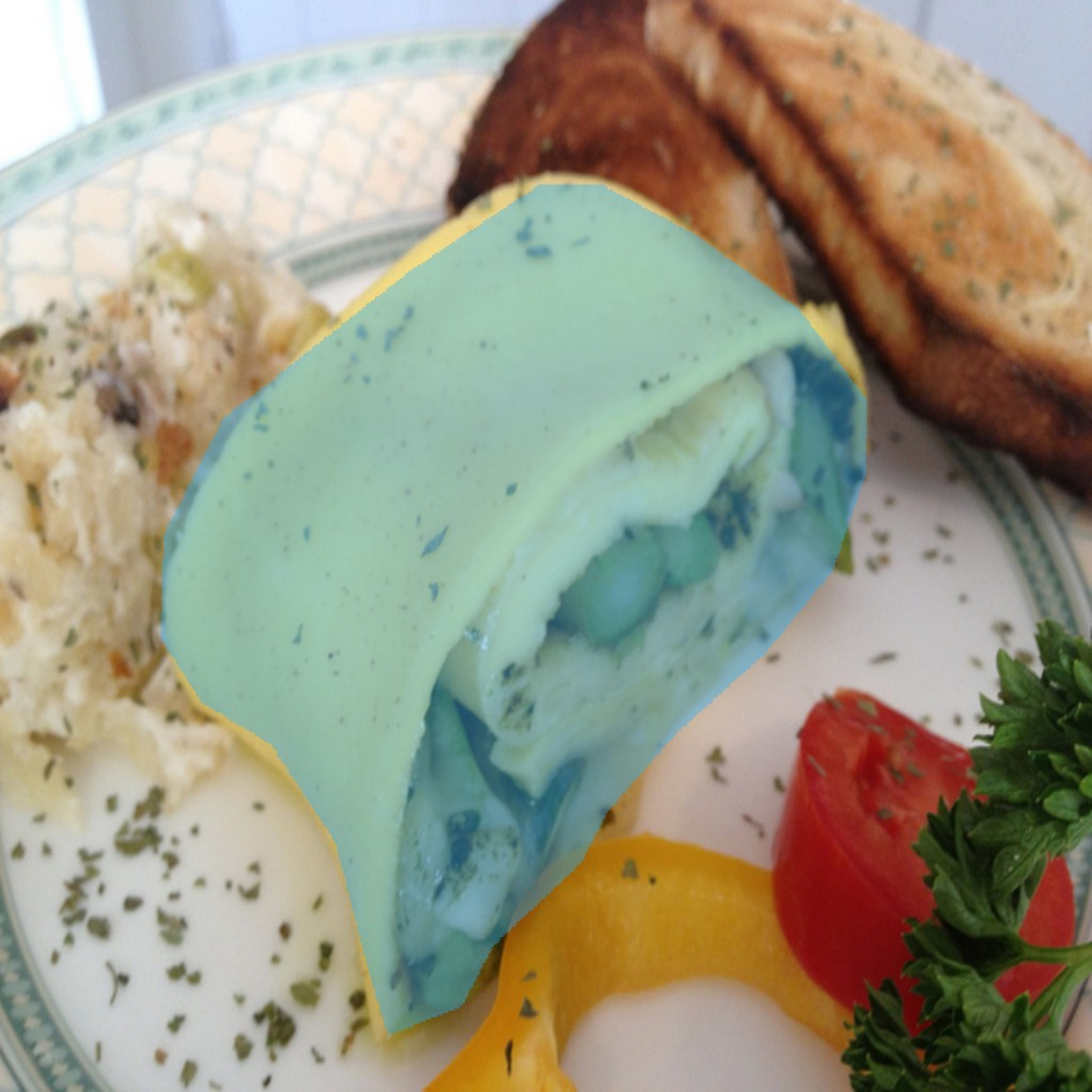} &
\includegraphics[width=0.155\linewidth]{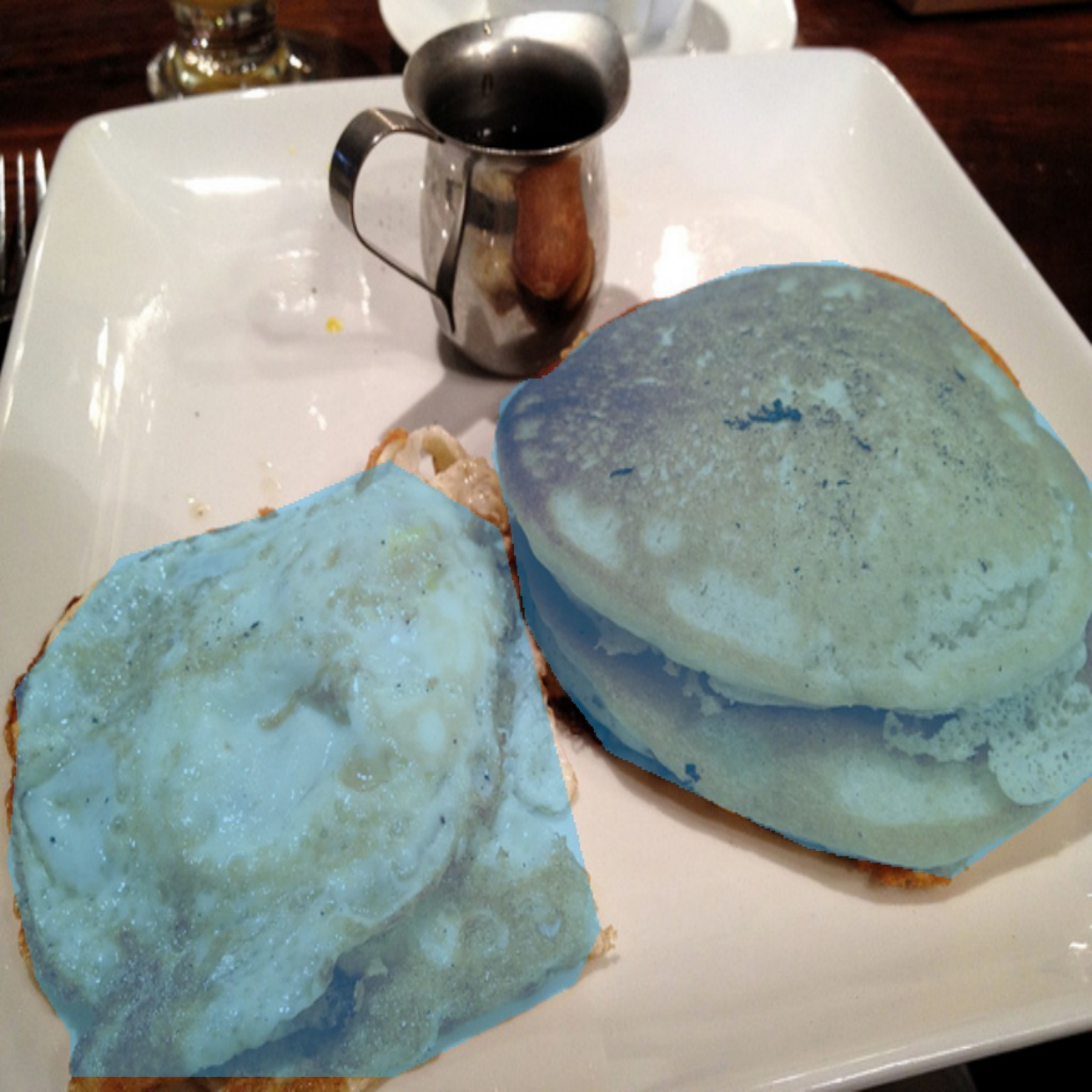} &
\includegraphics[width=0.155\linewidth]{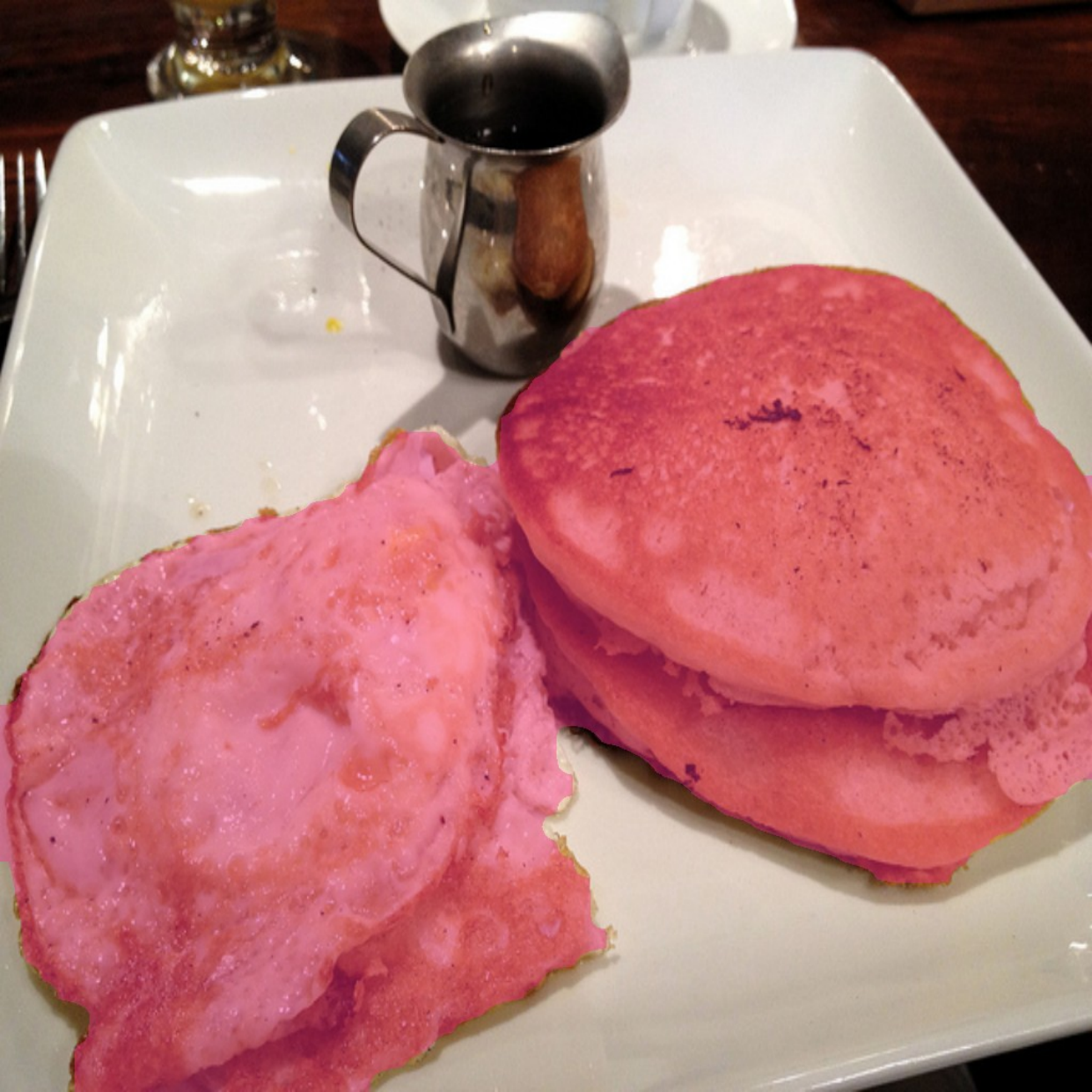} \\
\includegraphics[width=0.155\linewidth]{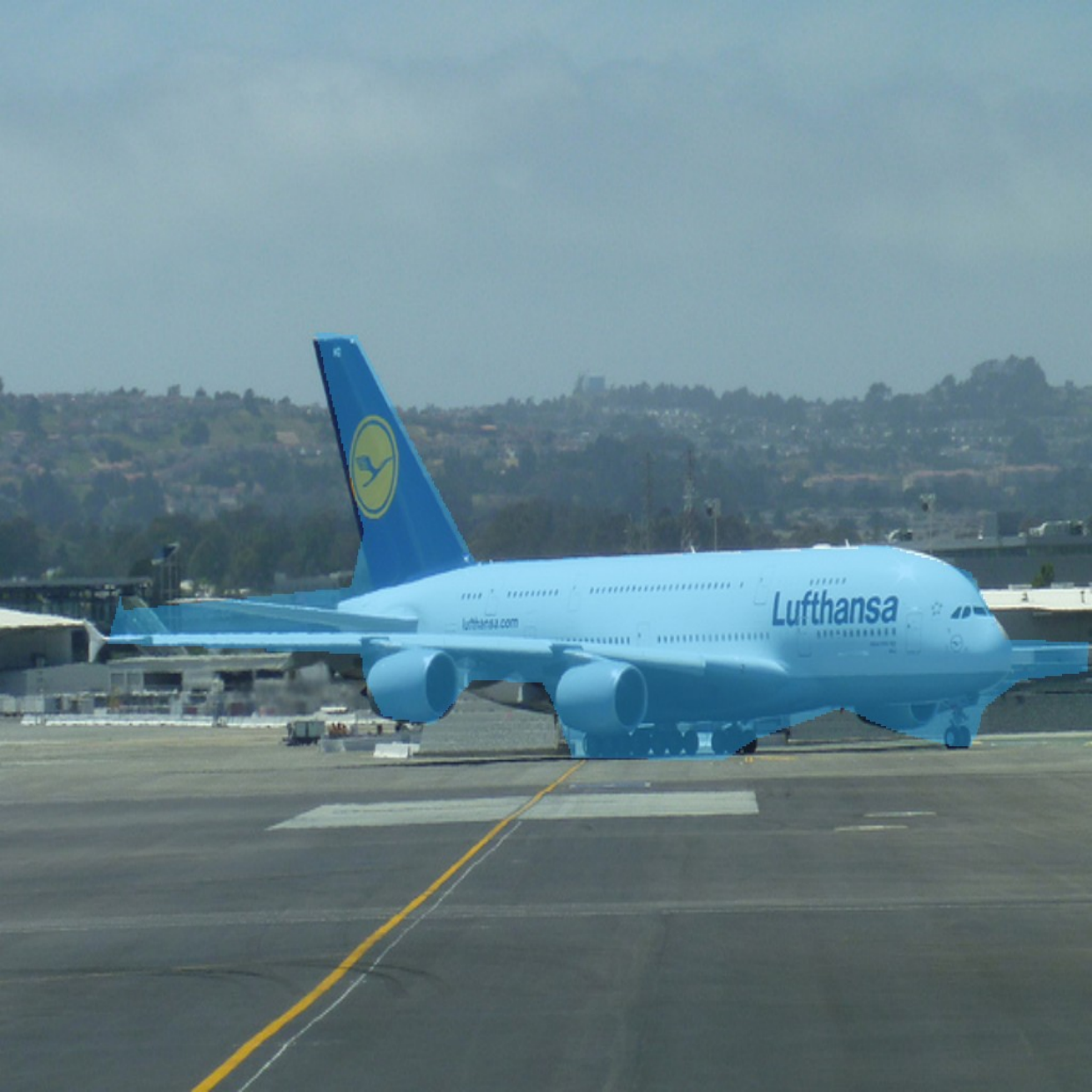} &
\includegraphics[width=0.155\linewidth]{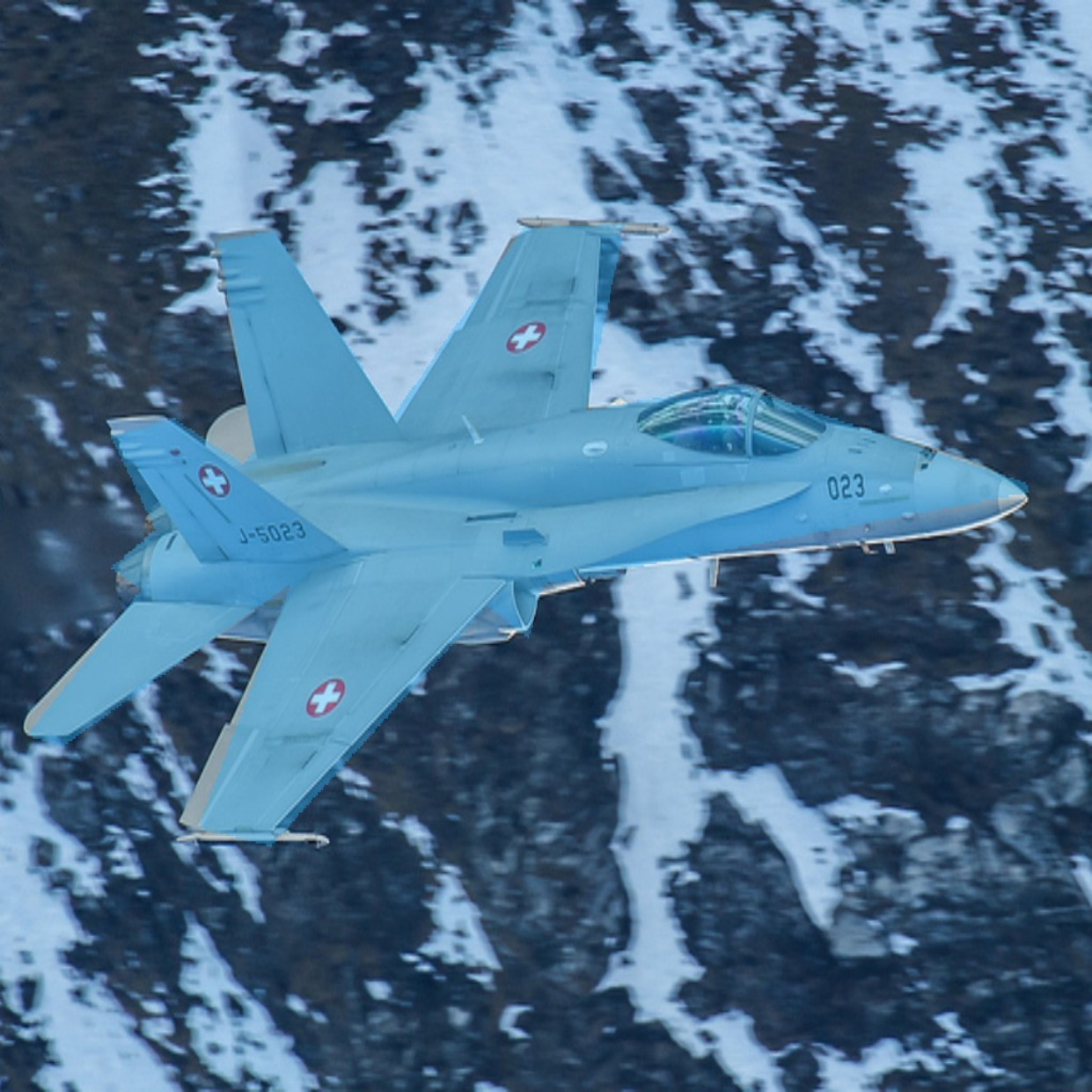} &
\includegraphics[width=0.155\linewidth]{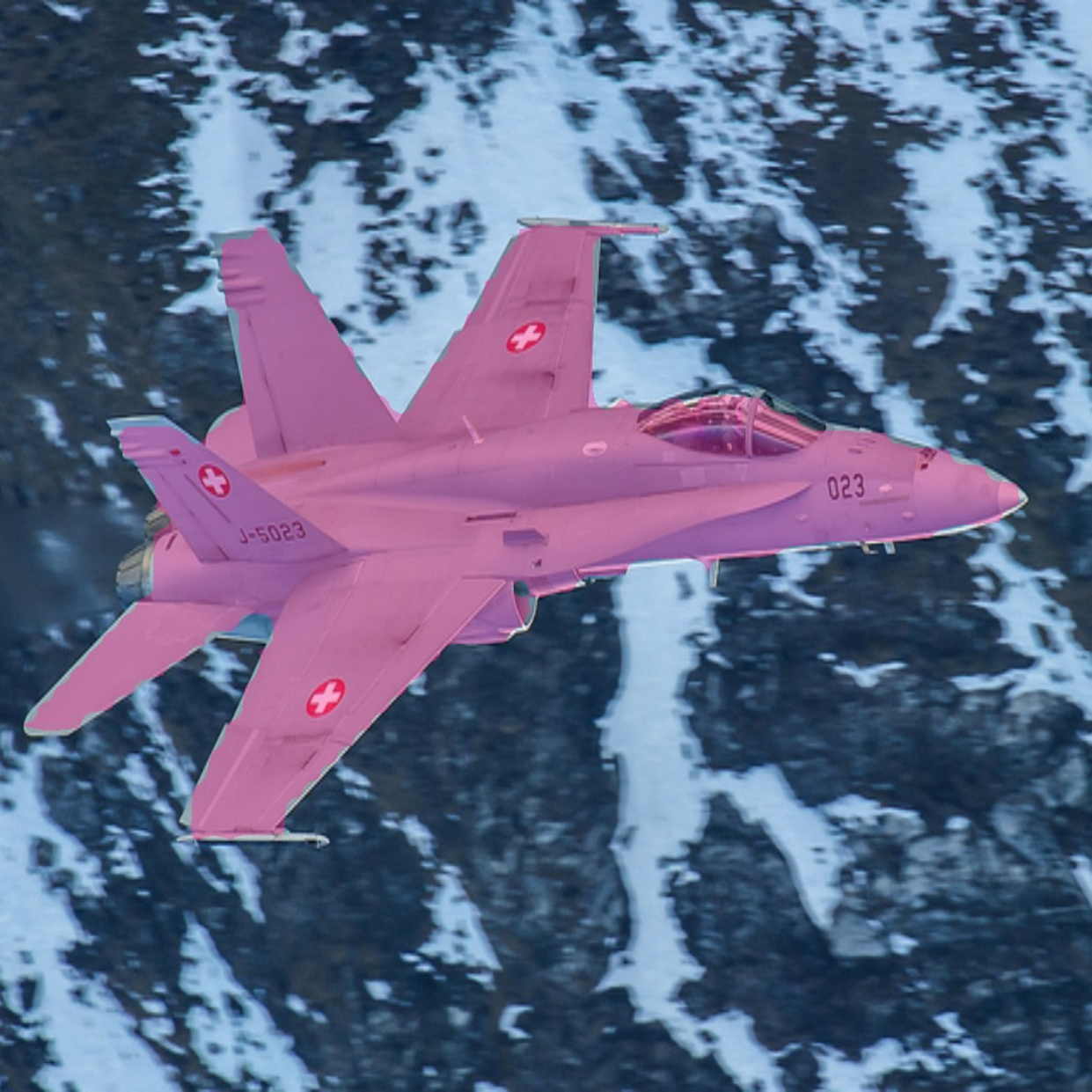} &
\includegraphics[width=0.155\linewidth]{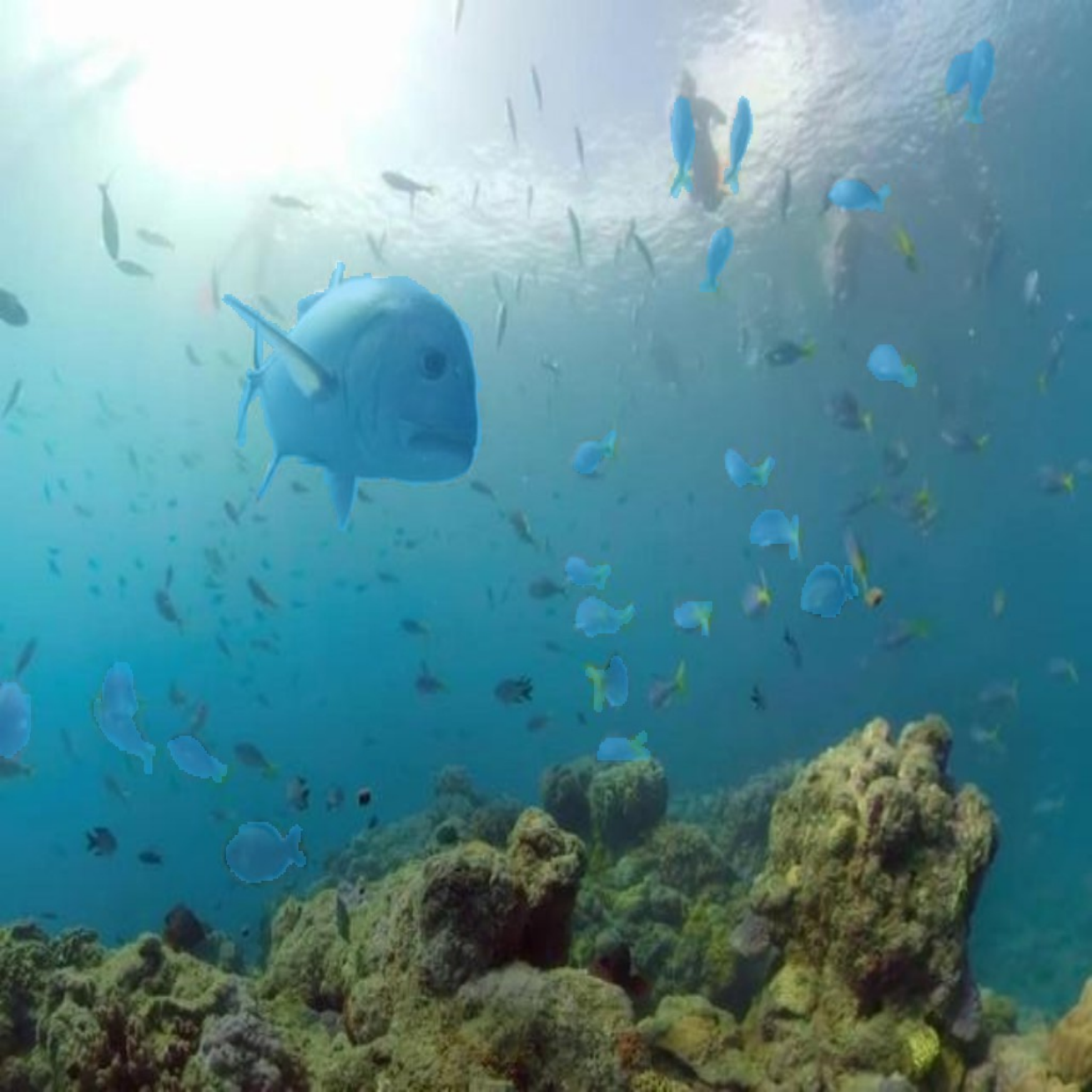} &
\includegraphics[width=0.155\linewidth]{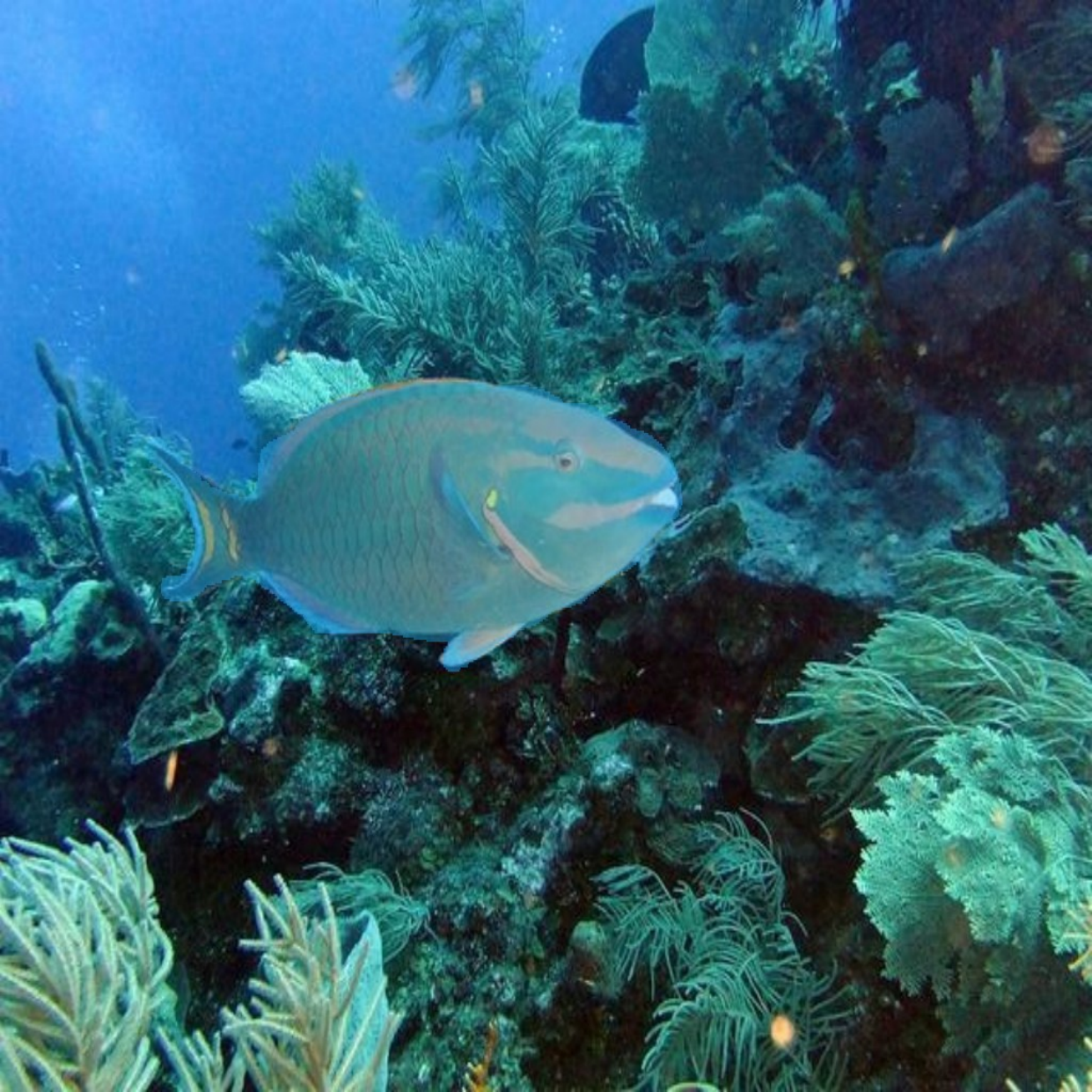} &
\includegraphics[width=0.155\linewidth]{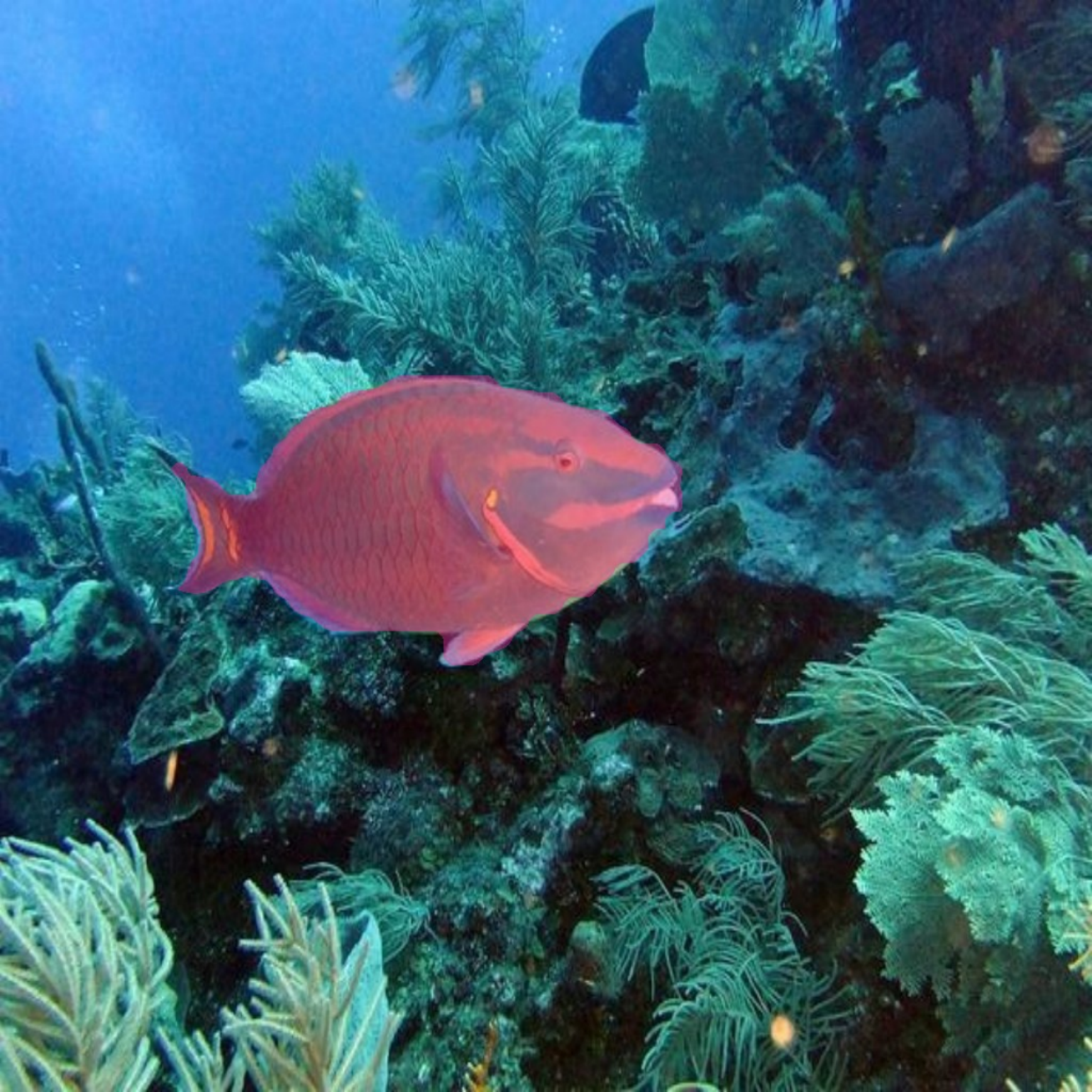} \\
\includegraphics[width=0.155\linewidth]{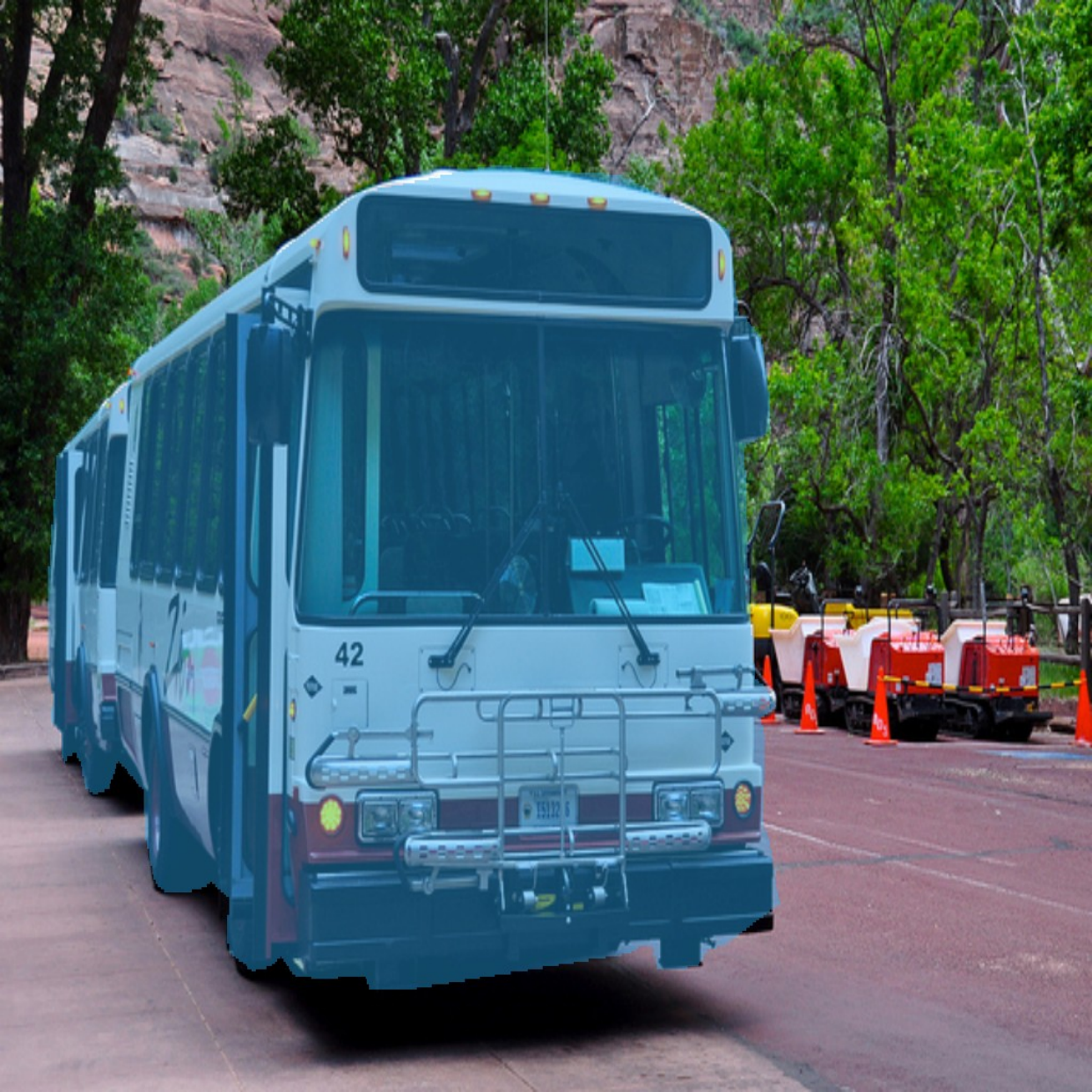} &
\includegraphics[width=0.155\linewidth]{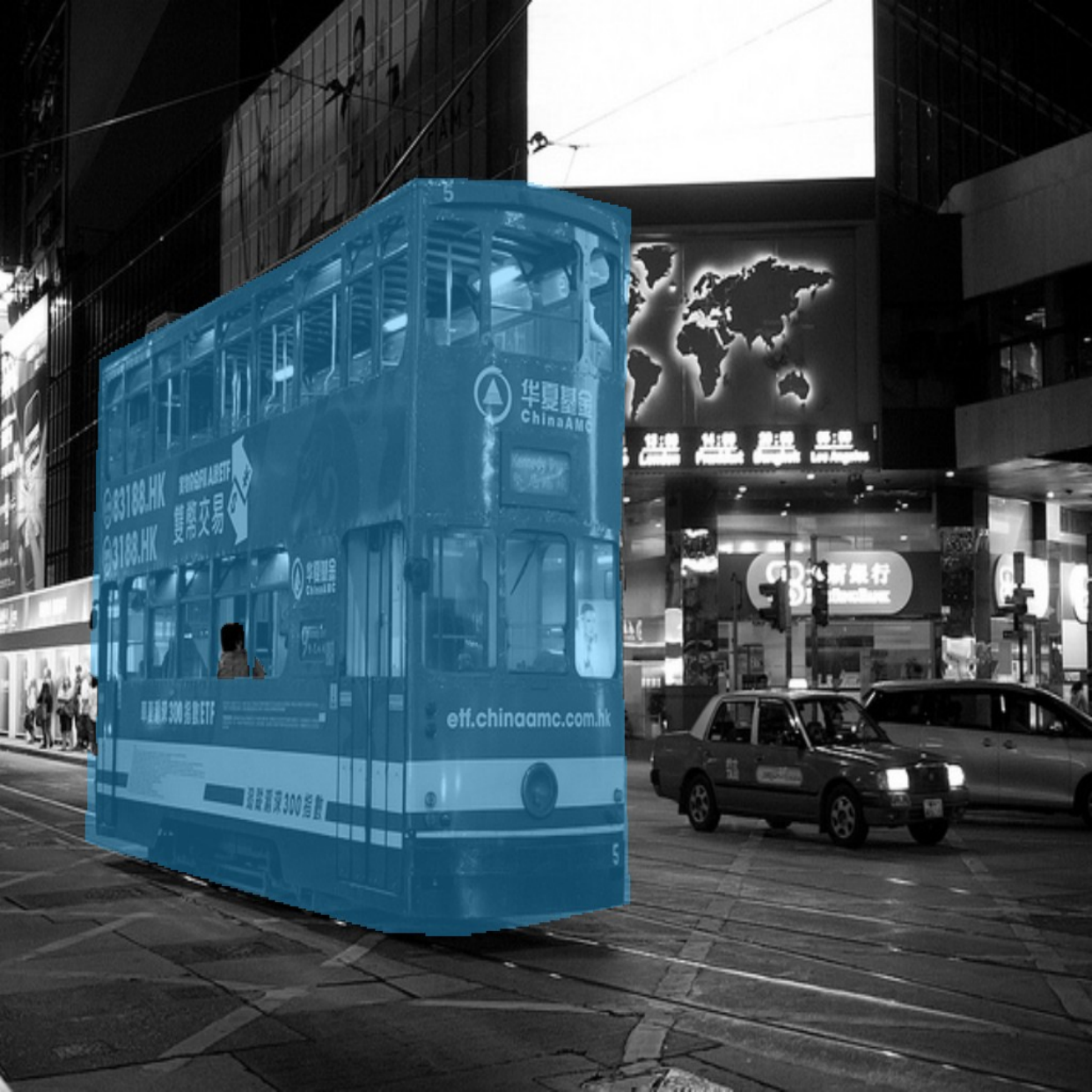} &
\includegraphics[width=0.155\linewidth]{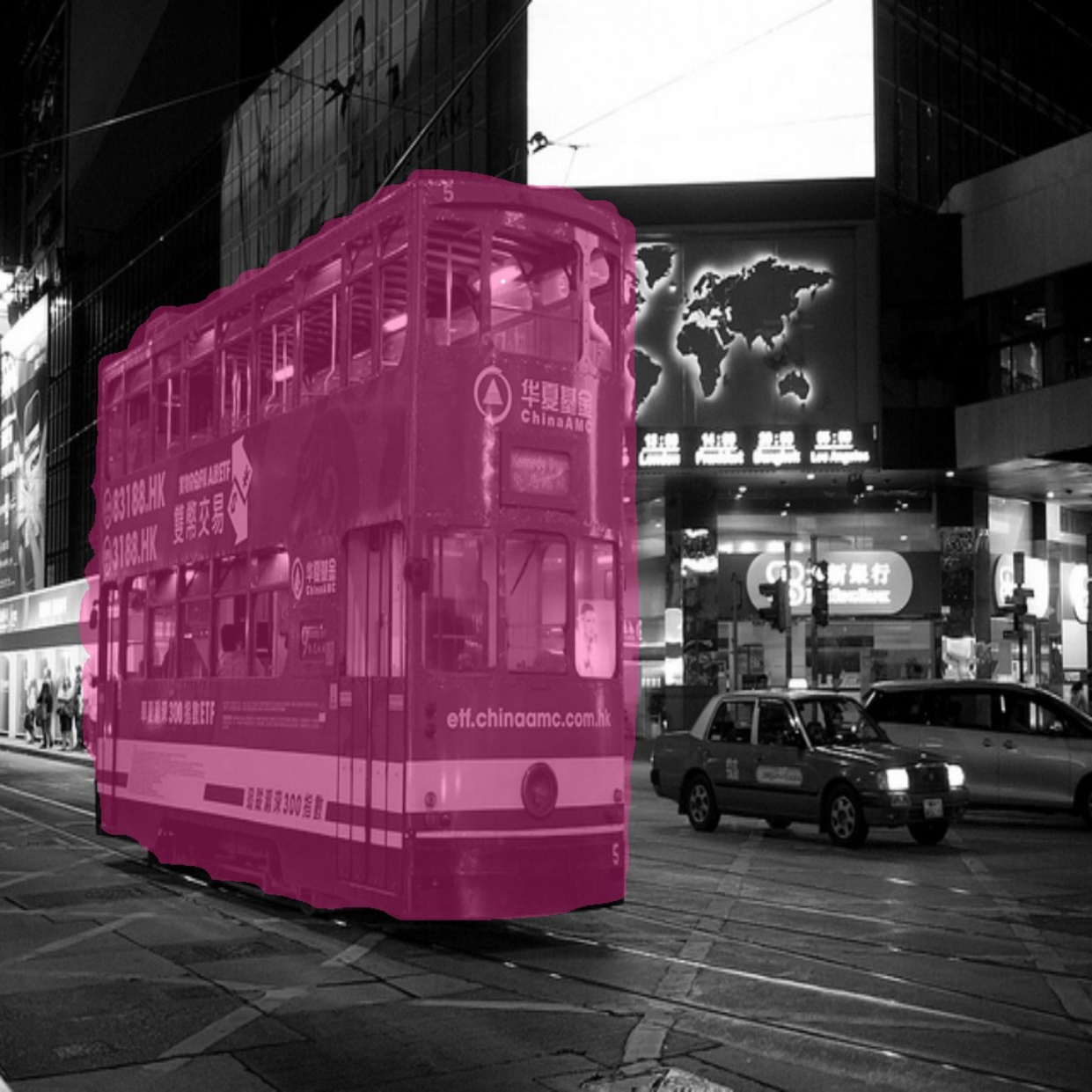} &
\includegraphics[width=0.155\linewidth]{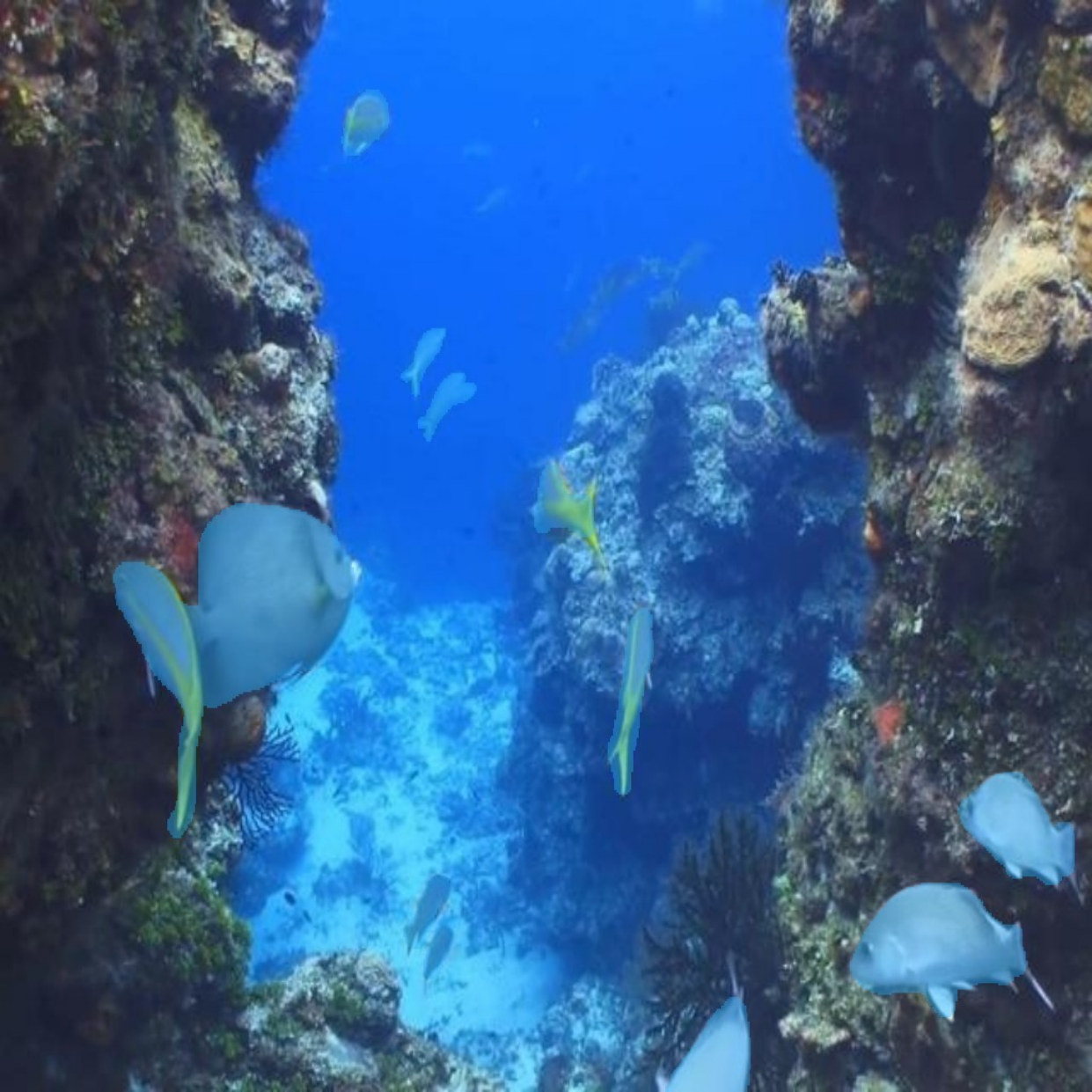} &
\includegraphics[width=0.155\linewidth]{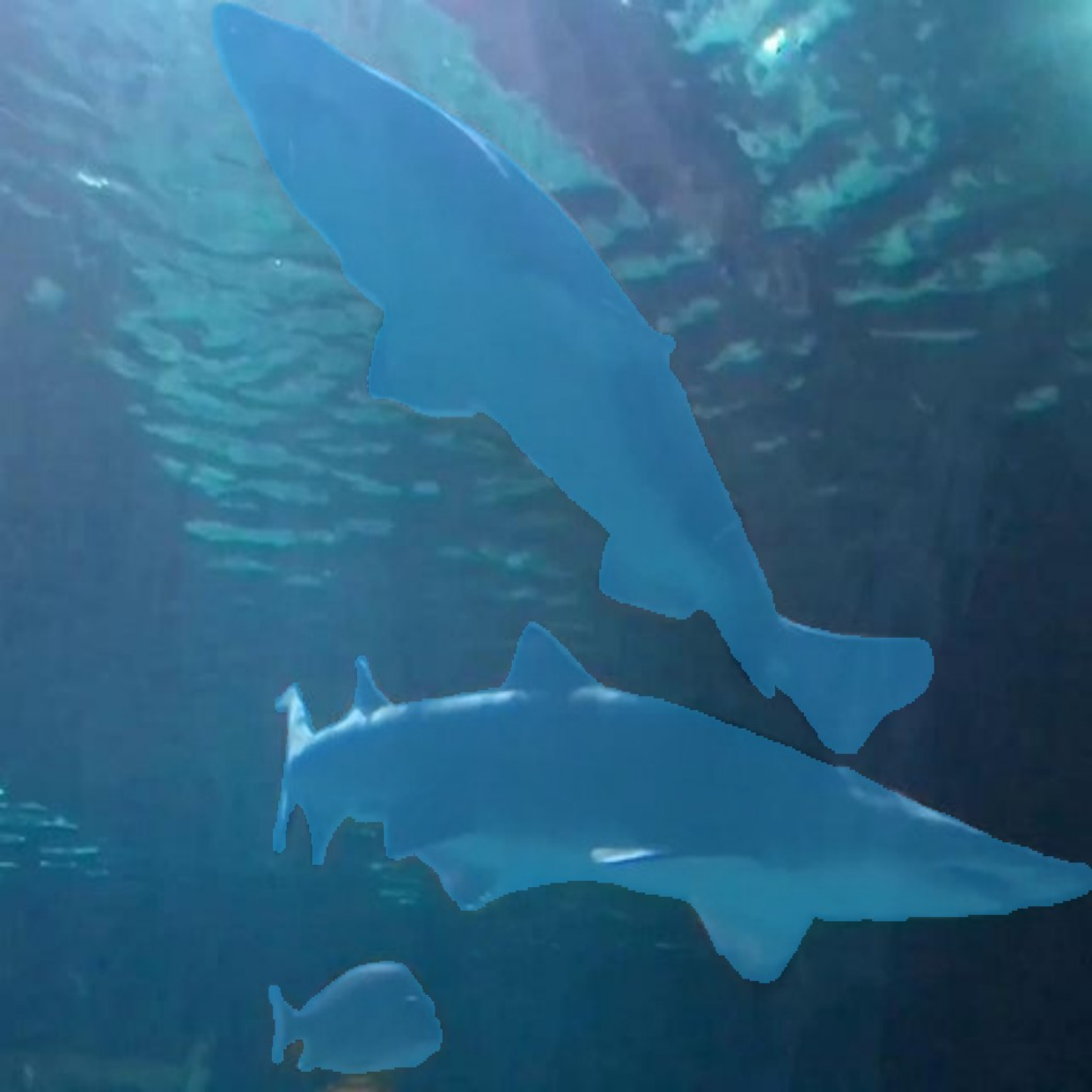} &
\includegraphics[width=0.155\linewidth]{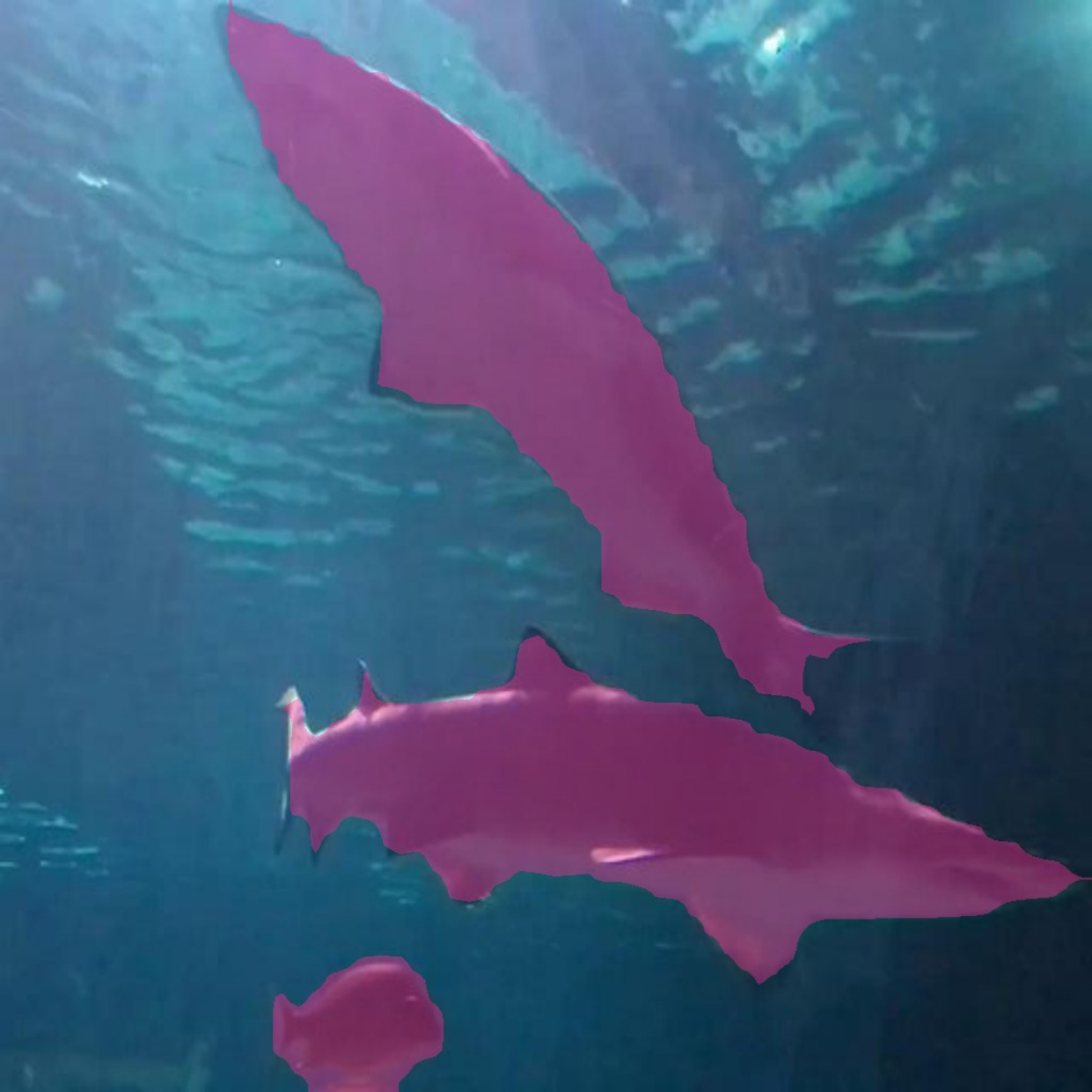} \\
\includegraphics[width=0.155\linewidth]{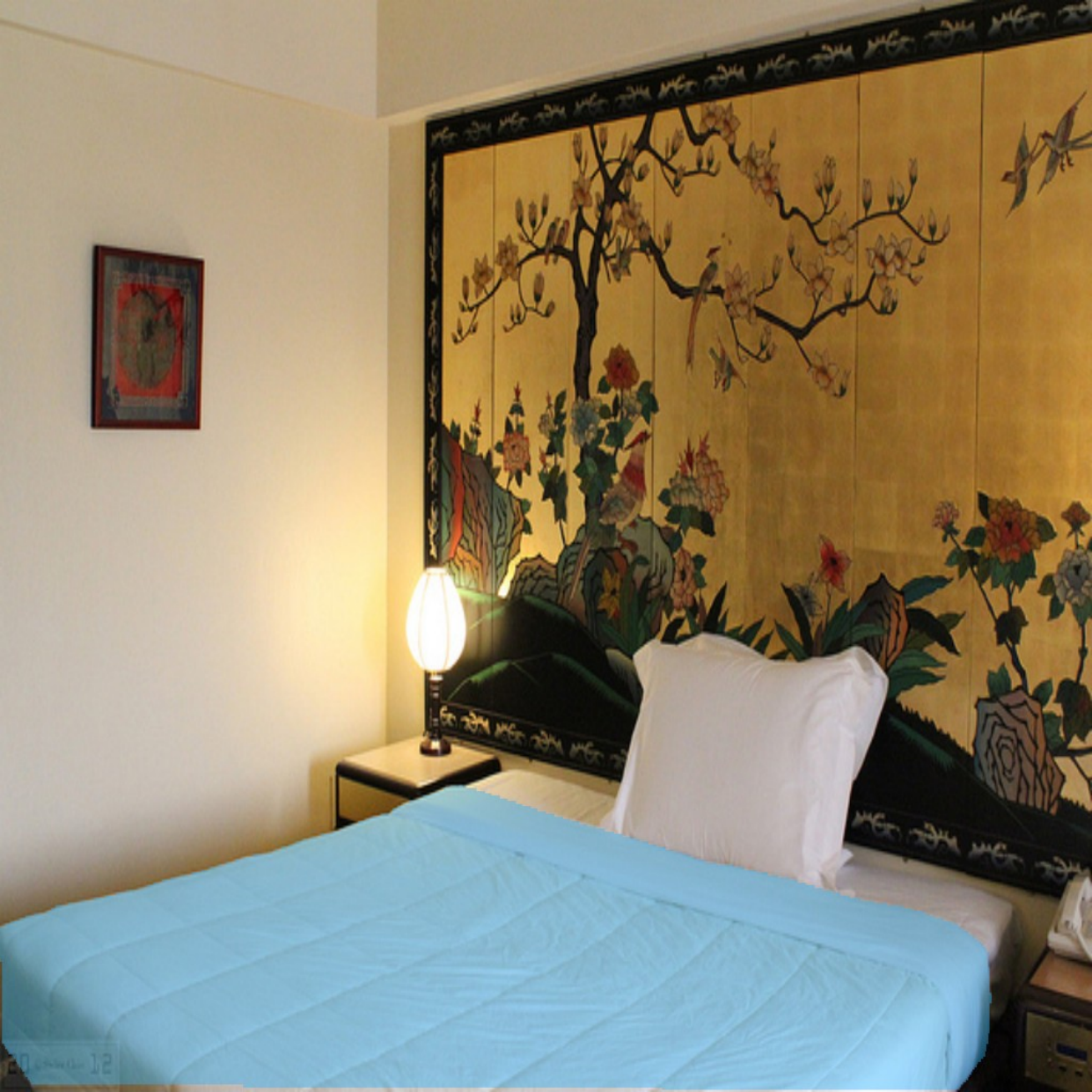} &
\includegraphics[width=0.155\linewidth]{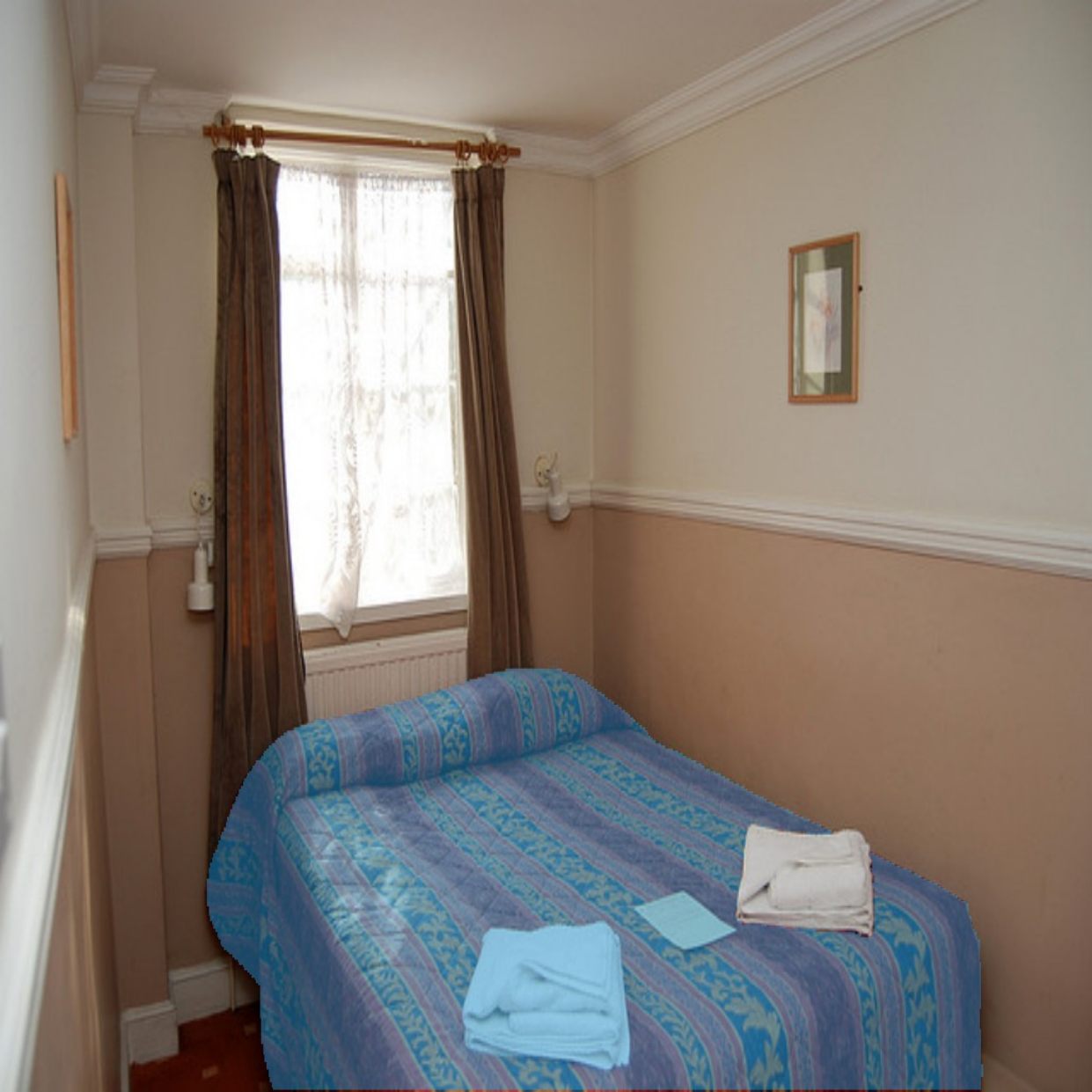} &
\includegraphics[width=0.155\linewidth]{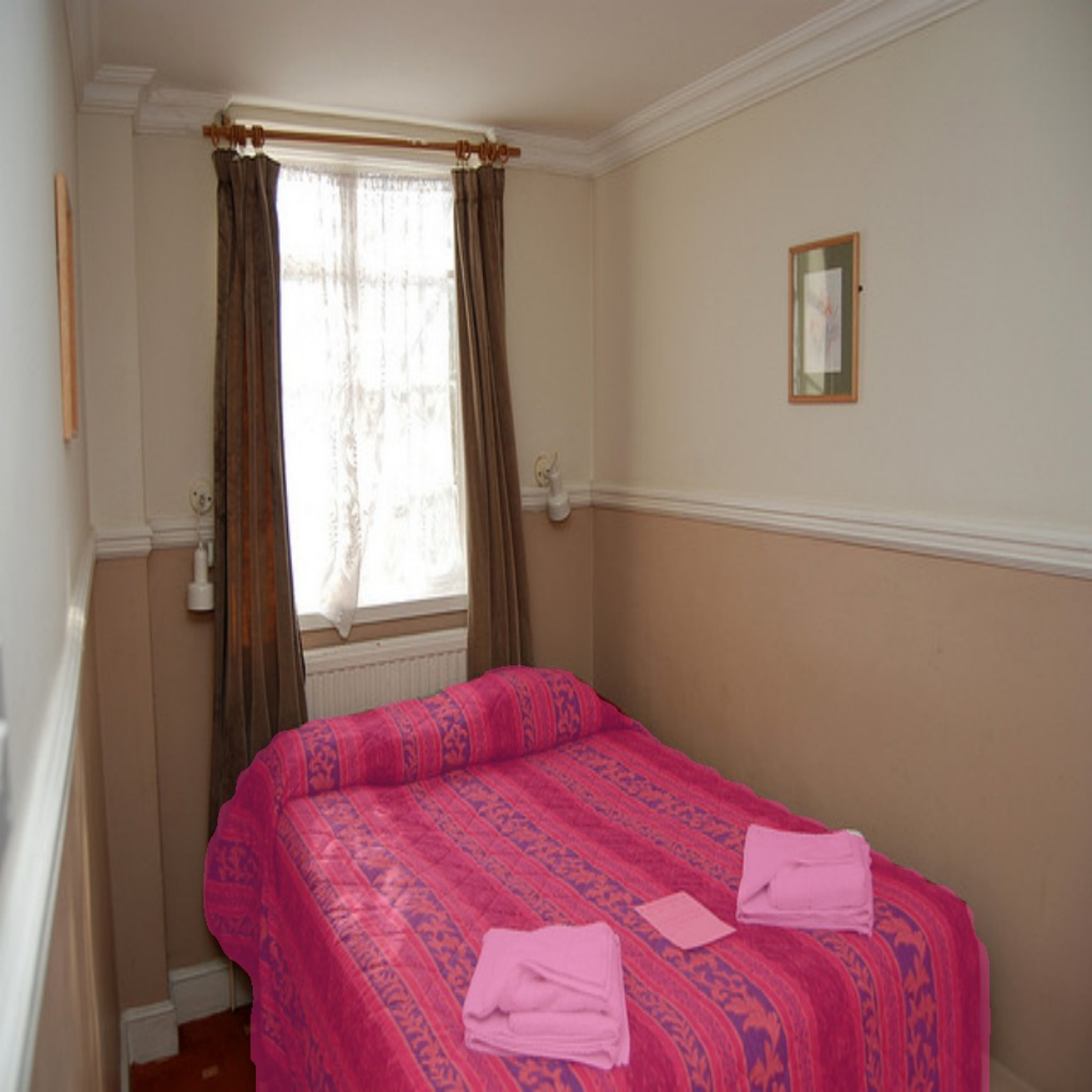} &
\includegraphics[width=0.155\linewidth]{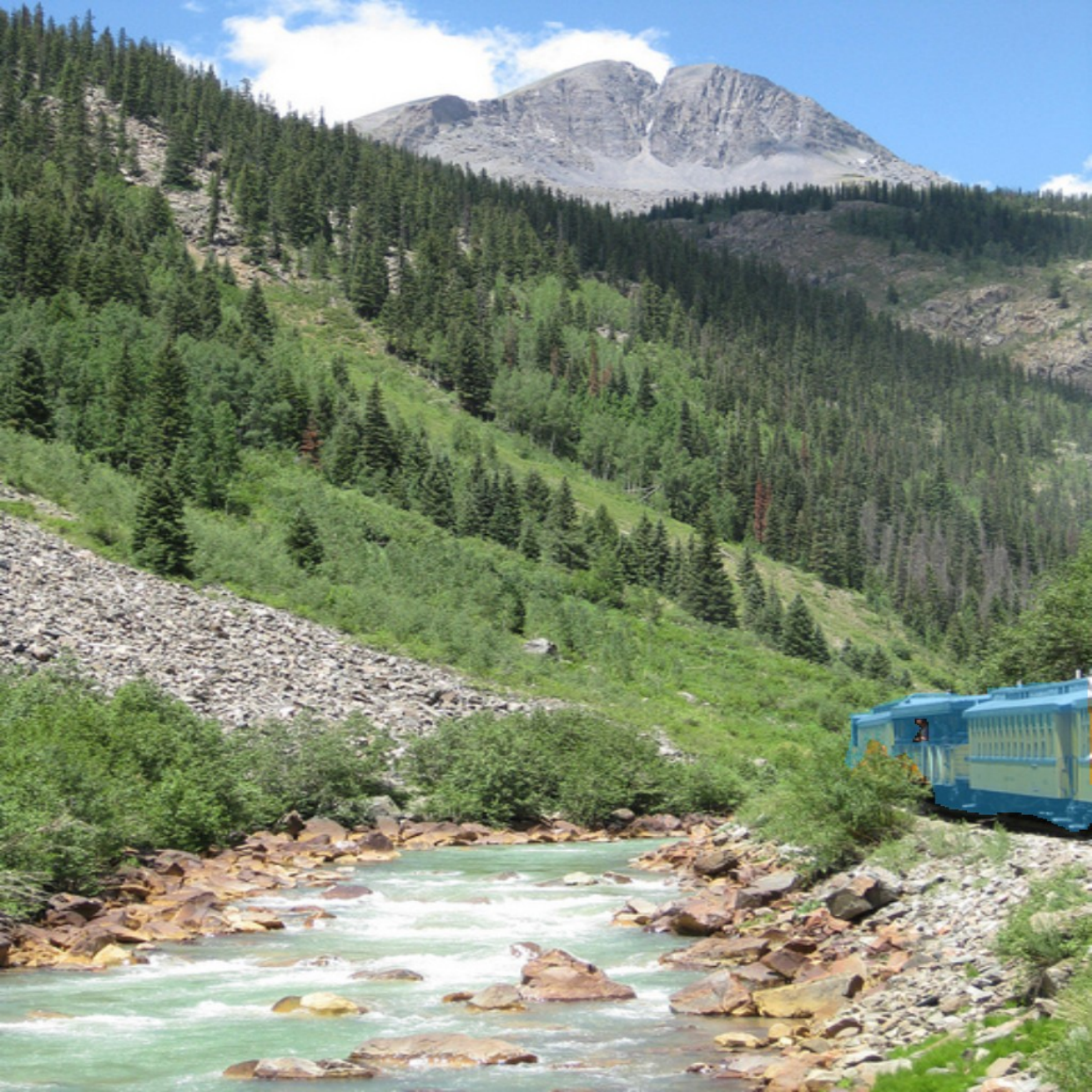} &
\includegraphics[width=0.155\linewidth]{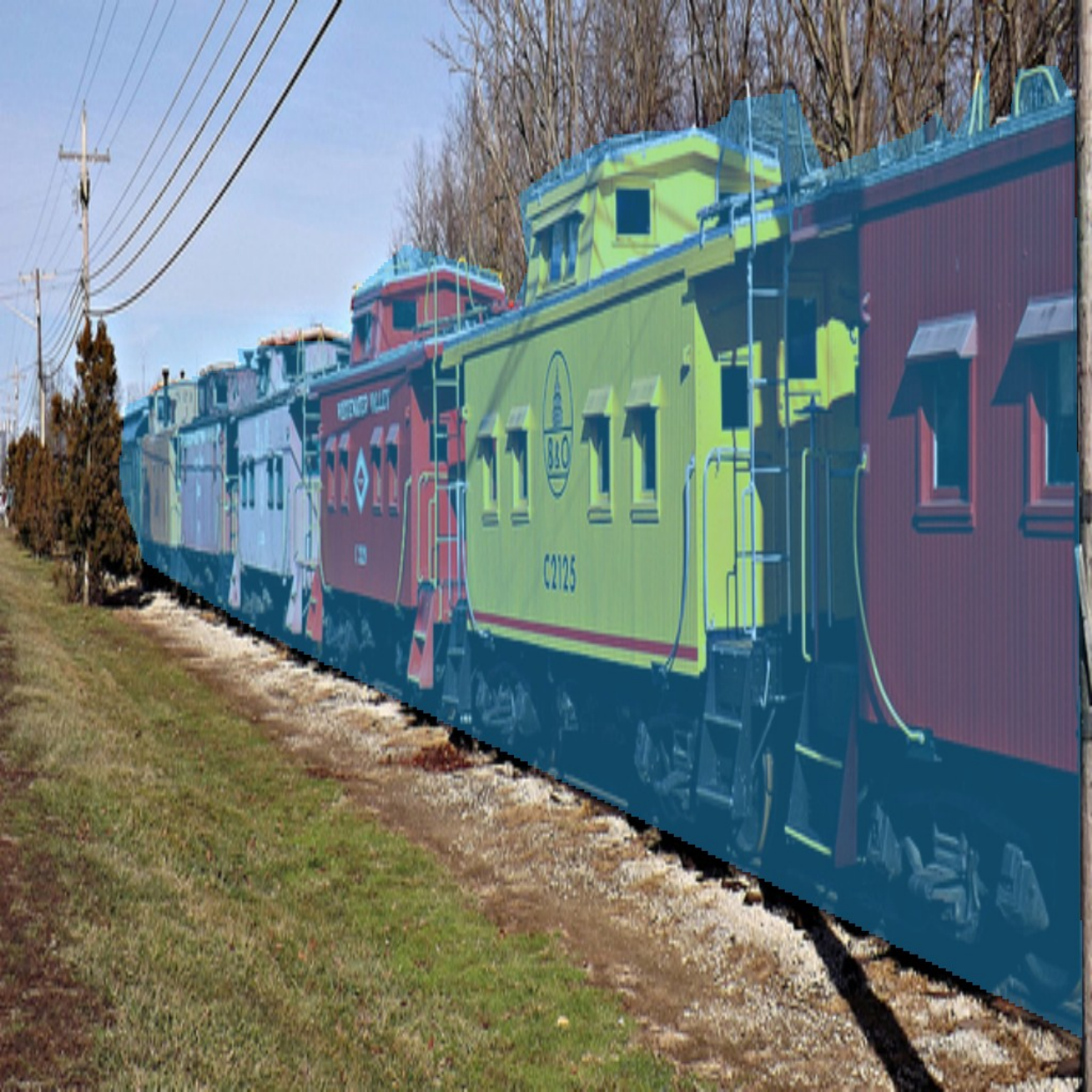} &
\includegraphics[width=0.155\linewidth]{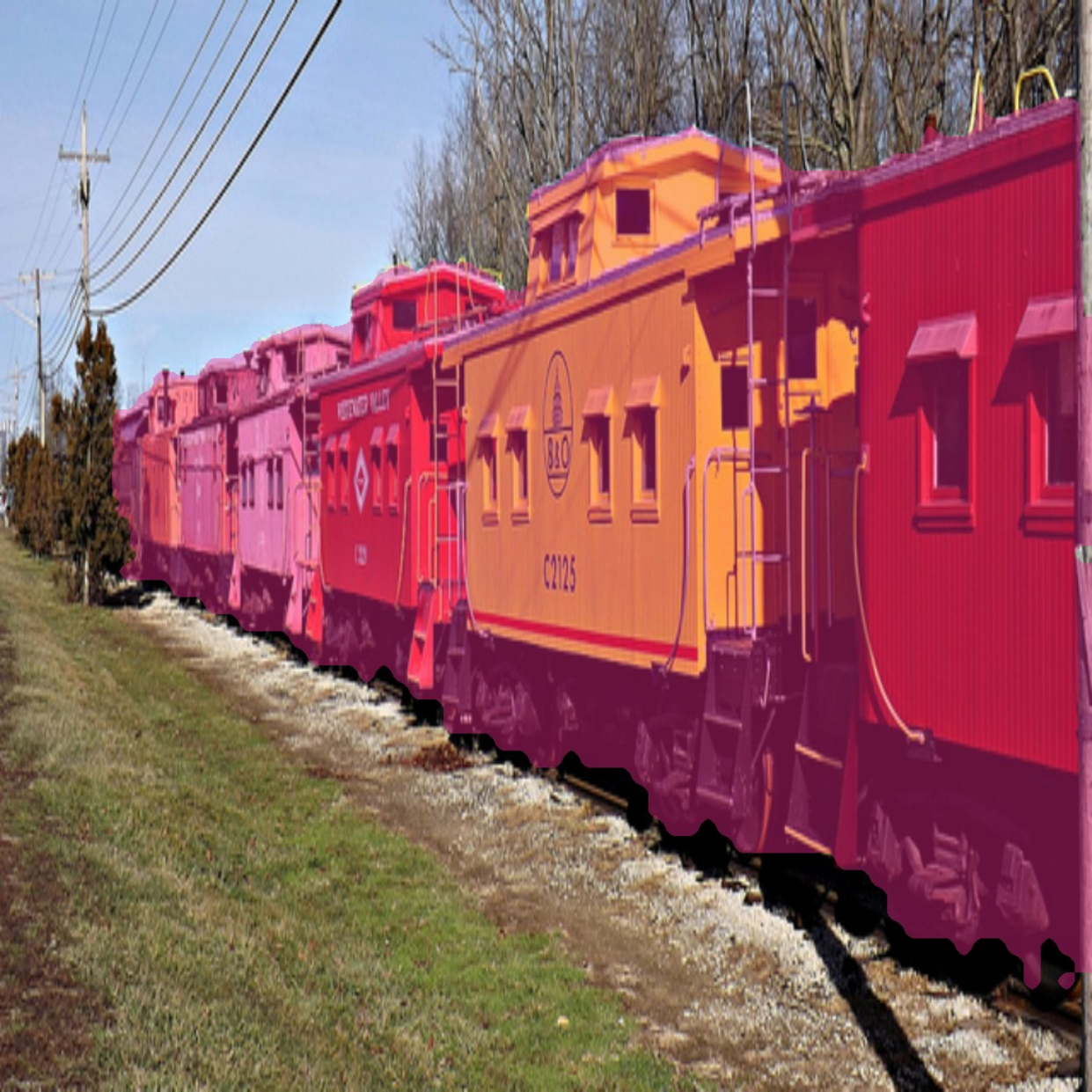} \\
\includegraphics[width=0.155\linewidth]{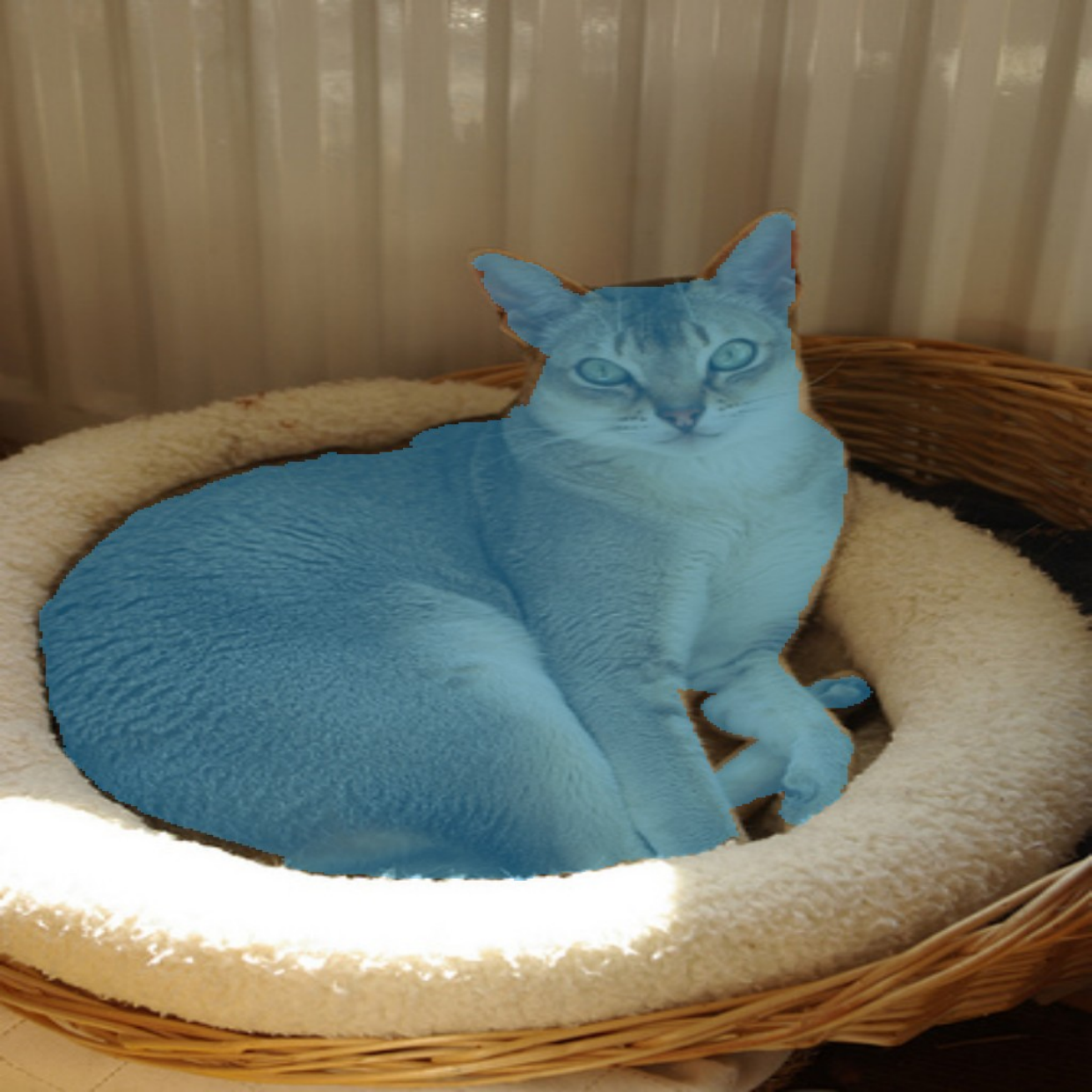} &
\includegraphics[width=0.155\linewidth]{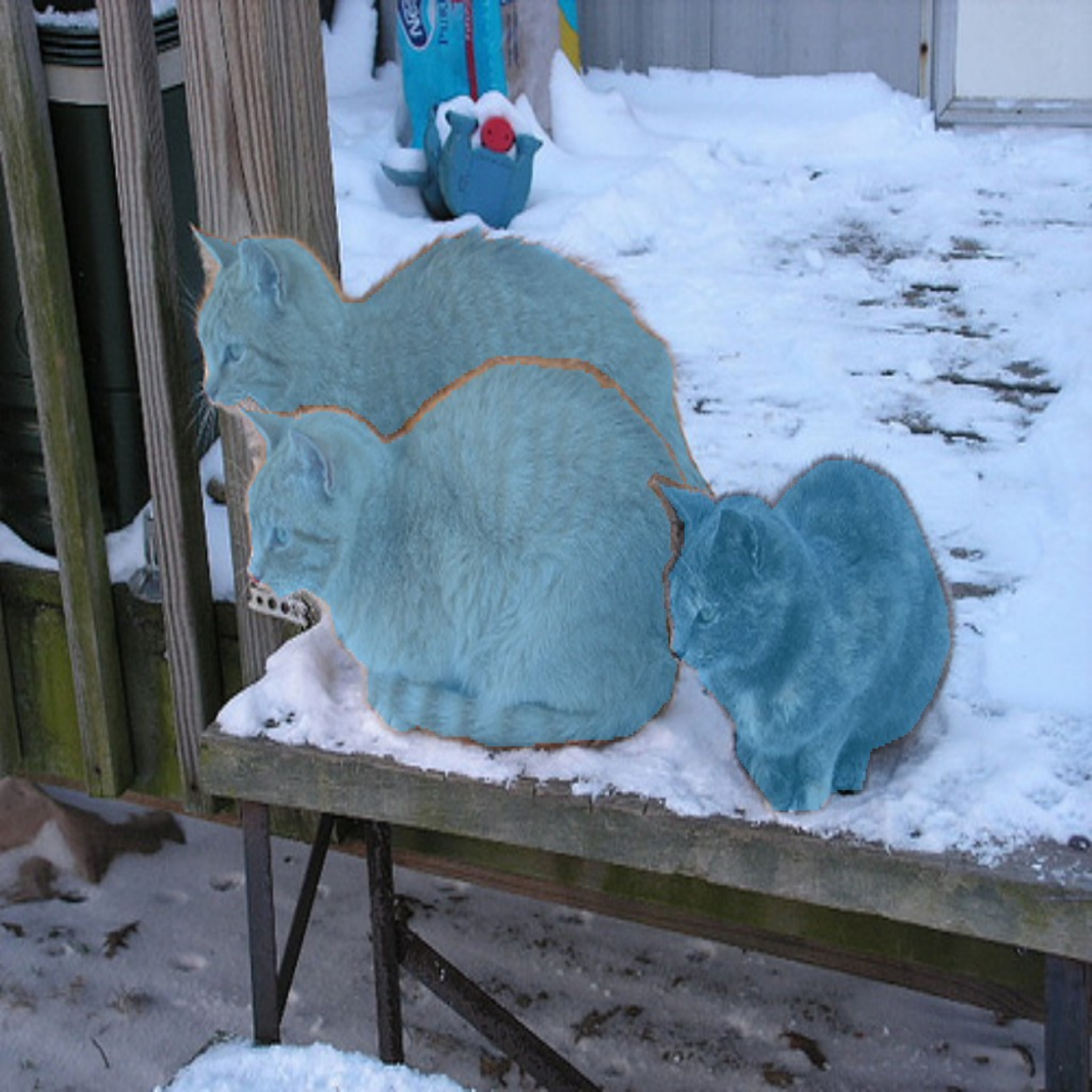} &
\includegraphics[width=0.155\linewidth]{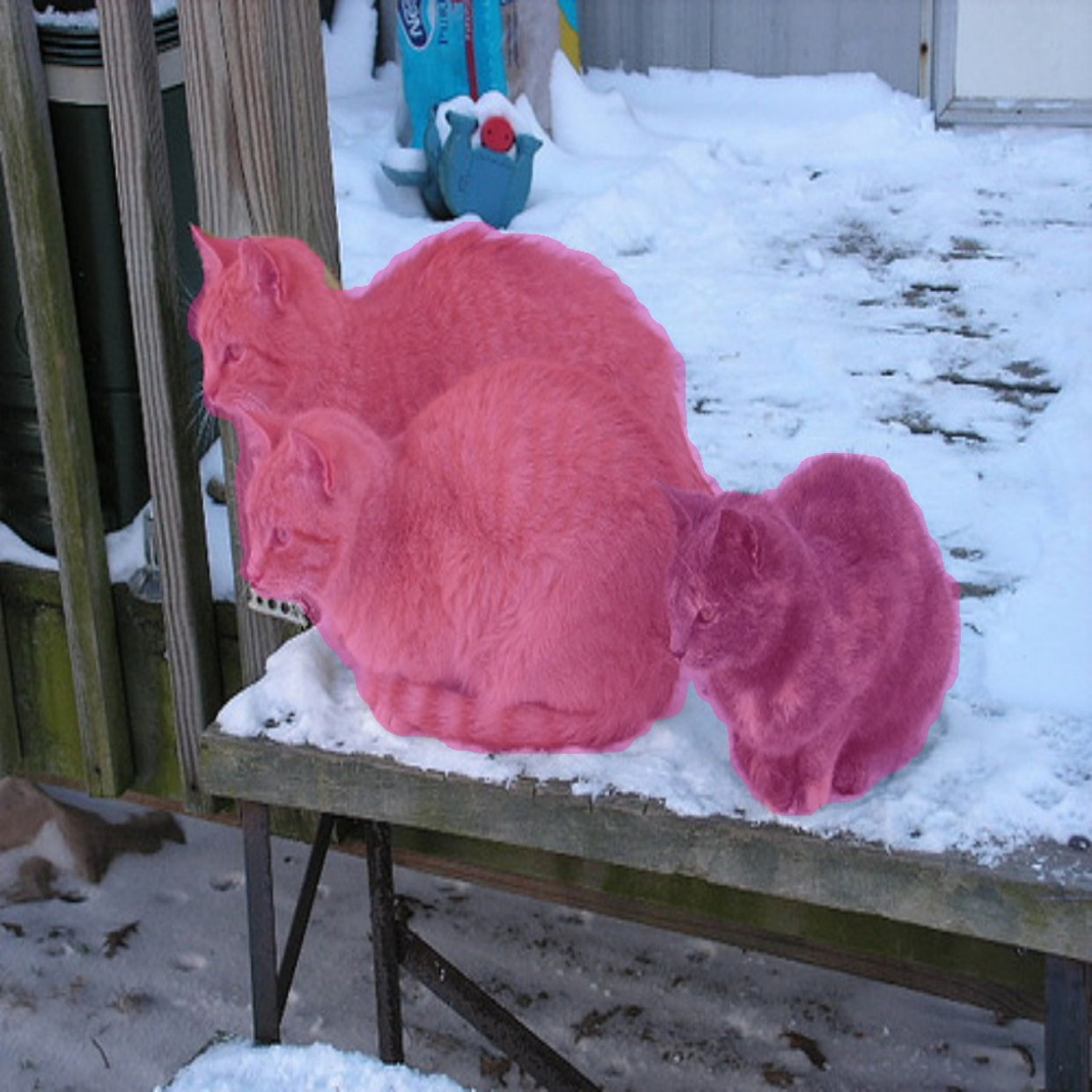} &
\includegraphics[width=0.155\linewidth]{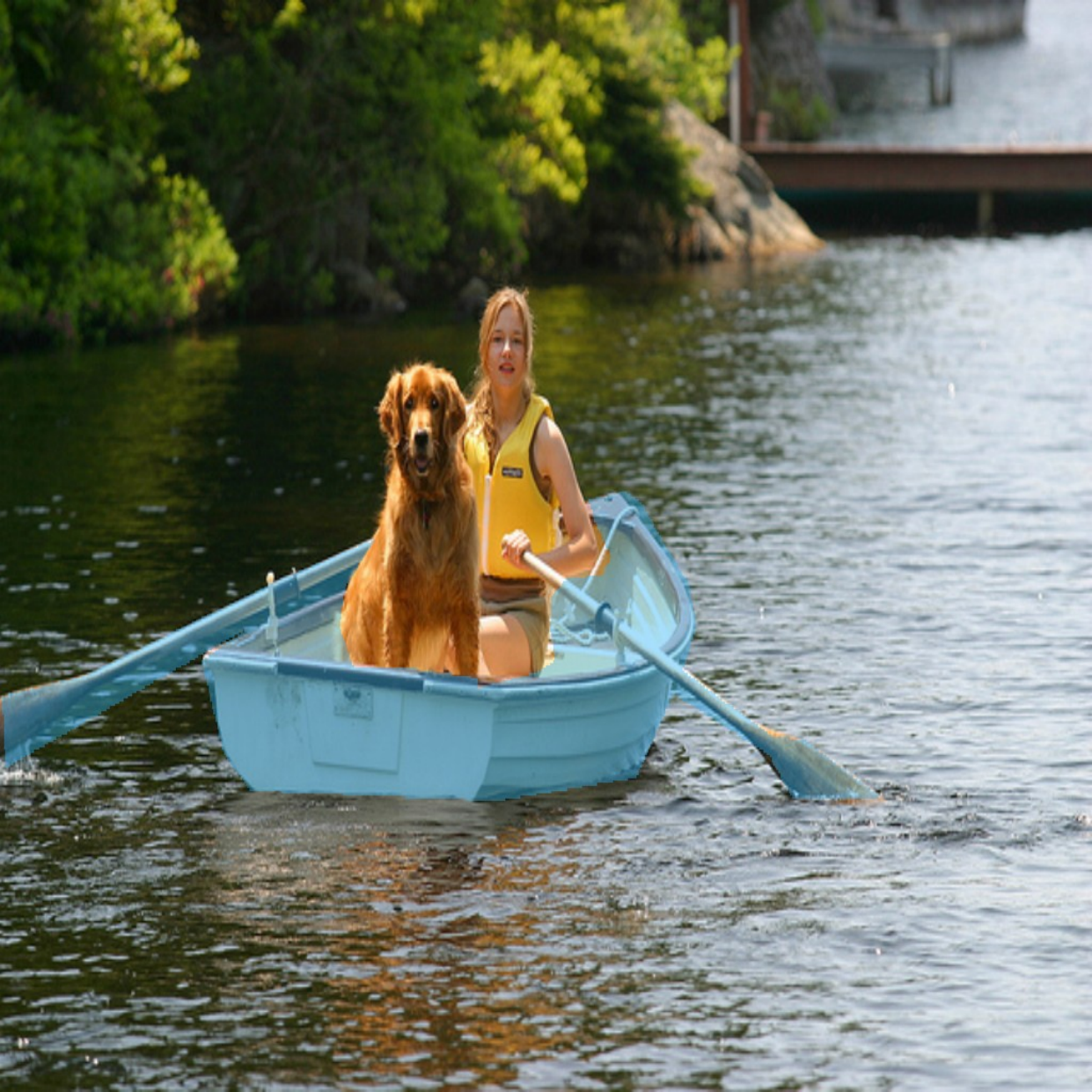} &
\includegraphics[width=0.155\linewidth]{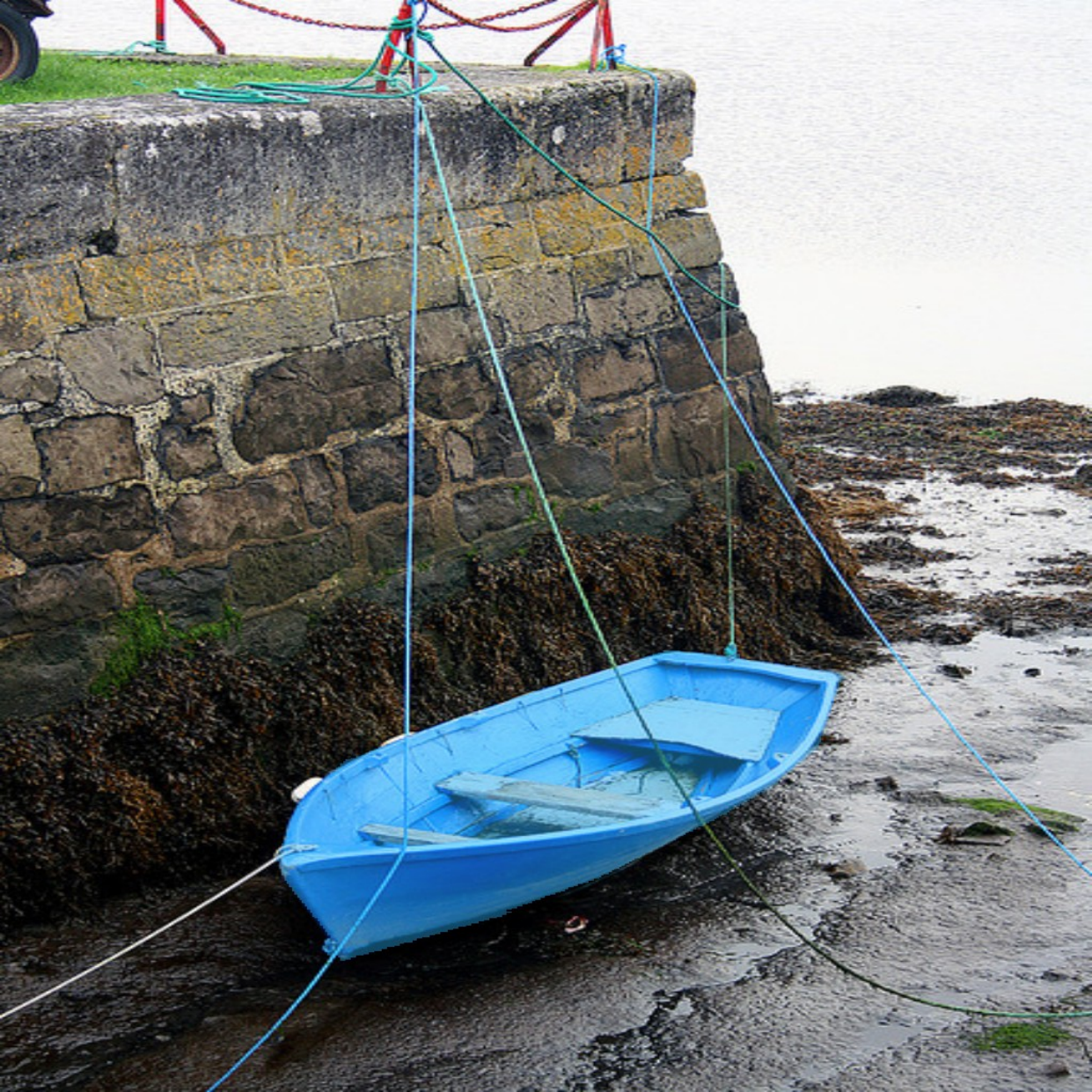} &
\includegraphics[width=0.155\linewidth]{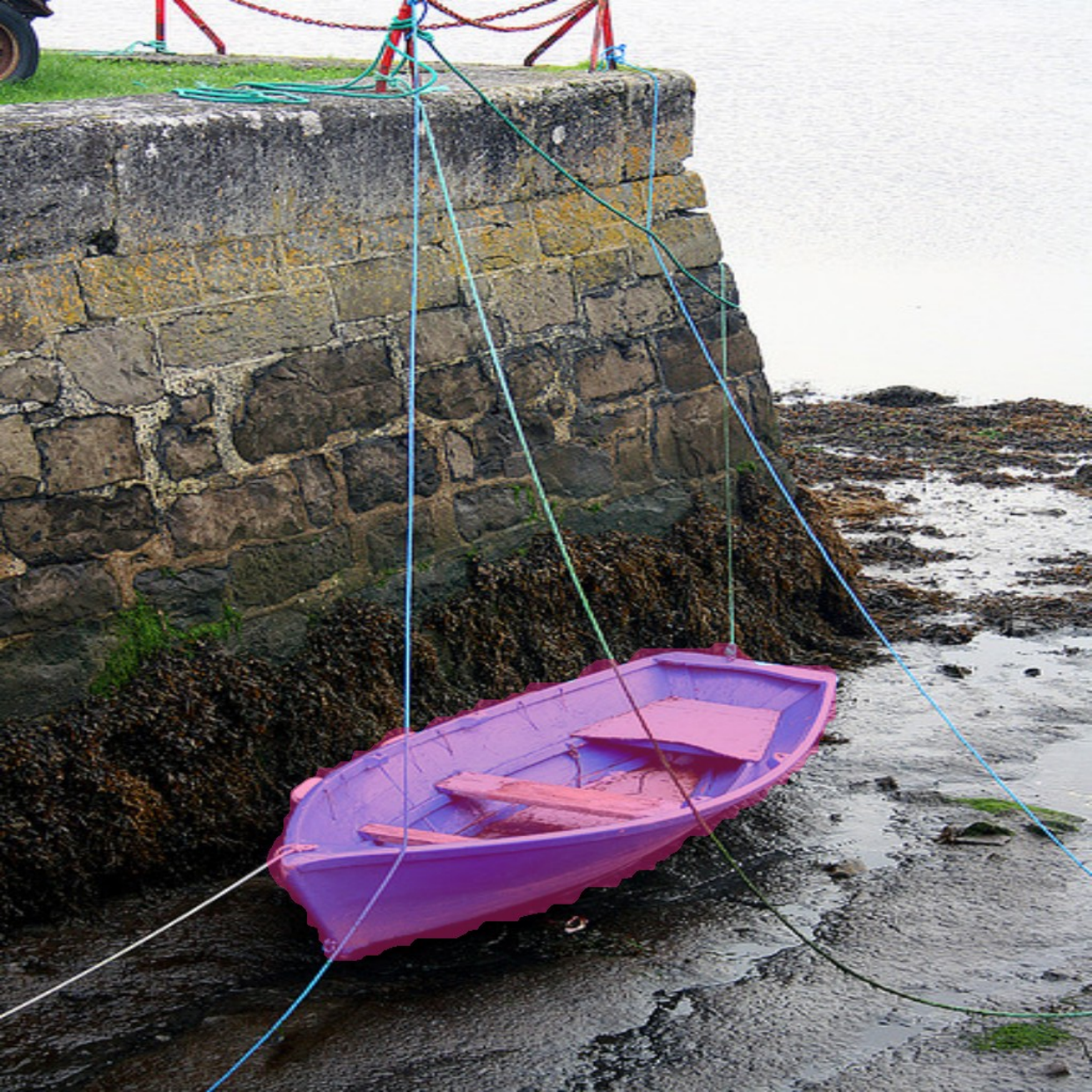} \\
\end{tabular}
\caption{Qualitative results of FROST on the general datasets. Each triplet shows, from left to right, the reference image with its mask, the query image with its mask, and the prediction of FROST, with two triplets per row separated by a gap. All examples are one-shot predictions obtained from a frozen DINOv3 backbone without any training.}
\label{fig:qual_general}
\end{figure}

Figure~\ref{fig:qual_general} repeats this inspection on the benchmarks beyond overhead imagery, with sixteen one-shot examples drawn from the natural-image and cross-domain benchmarks under the same protocol, the same frozen backbone, and the same overlay convention. The predictions confirm what the per-benchmark numbers report, since on the object-centric and part-centric targets FROST recovers the instance with boundaries that track the object, and on the cross-domain images its masks stay accurate even though the appearance is far from anything seen during the pretraining of the backbone. As in the overhead case, the residual gap on these benchmarks comes from harder scenes with several co-occurring instances or with a query whose appearance departs strongly from the single reference, rather than from gross failures.

\clearpage
\section{Backbone ablation}
\label{app:backbone}

The decision rule of FROST reads from whatever features its backbone provides, so a natural question is how much of its accuracy is owed to the specific frozen encoder we use and how much to the density-ratio formulation that surrounds it. To separate the two, we hold the entire FROST pipeline at its default configuration and replace only the backbone, comparing six frozen encoders that vary along three axes, the pretraining domain, the model scale, and the architecture family. We evaluate every encoder at one shot on five of the seventeen remote-sensing benchmarks, the land-cover benchmarks Potsdam, LoveDA, and Vaihingen together with the instance-dense iSAID and the drone benchmark UDD5, chosen to span the range of foreground scales the full suite covers, and Table~\ref{tab:backbone} reports the foreground mIoU of each. The domain axis contrasts the default DINOv3 encoder, pretrained on general-domain imagery, with a DINOv3 variant pretrained on satellite imagery that we write DINOv3-SAT, the scale axis contrasts a ViT-L of about $0.3$ billion parameters with a ViT-7B of about $6.7$ billion, and the architecture axis contrasts the vision transformer of DINOv3 with its ConvNeXt-L variant and with the earlier DINOv2 ViT-L. The first row is the default DINOv3 ViT-L backbone of FROST, and the column $\Delta$ reports each encoder's change in mean foreground mIoU relative to it.

Three findings follow from the table. The first concerns the pretraining domain, where the general-domain DINOv3 encoders surpass their satellite-pretrained DINOv3-SAT counterparts by a wide and scale-independent margin, the DINOv3 ViT-L leading the DINOv3-SAT ViT-L by $14.3$ mean mIoU and the DINOv3 ViT-7B leading the DINOv3-SAT ViT-7B by $14.4$. Pretraining on satellite imagery therefore harms in-context segmentation on remote-sensing imagery at both scales rather than helping it, a counterintuitive result we read as evidence that the broad visual coverage of general-domain pretraining yields features that transfer to overhead scenes better than the narrower satellite distribution, which the density ratio then exploits without any adaptation of its own. The second finding concerns model scale, where growing the DINOv3 encoder from ViT-L to ViT-7B raises the mean by only $0.6$ and wins on three of the five benchmarks, an improvement small against the roughly twentyfold increase in parameters and the larger inference cost it carries, so the ViT-L sits at a favourable operating point at which accuracy has largely saturated. The third finding concerns the architecture and the pretraining recipe at a fixed scale, where the DINOv3 ViT-L exceeds the DINOv2 ViT-L by $6.1$ and the DINOv3 ConvNeXt-L by $7.0$, so the gain is not explained by the transformer architecture alone, since the ConvNeXt variant shares the DINOv3 recipe yet emits features on a native $32 \times 32$ grid at half the spatial resolution of the ViT and trails it on every benchmark. Together these results indicate that the accuracy of FROST rests on a general-domain transformer backbone of moderate scale rather than on a domain-specialized or larger one, and that its default DINOv3 ViT-L encoder is the configuration that the density ratio is best served by.

\begin{table}[h]
\centering
\caption{One-shot foreground mIoU of the FROST pipeline with the backbone replaced, on five remote-sensing benchmarks, where DINOv3 is pretrained on general-domain imagery and DINOv3-SAT on satellite imagery. The first row is the default backbone of FROST, the column params reports the approximate parameter count in billions, the column MEAN reports the foreground mIoU averaged over the five benchmarks, and the column $\Delta$ reports the change in that mean relative to the default backbone.}
\label{tab:backbone}
\setlength{\tabcolsep}{4pt}
\resizebox{\linewidth}{!}{%
\begin{tabular}{l c ccccc cc}
\toprule
Backbone & params & Potsdam & iSAID & LoveDA & UDD5 & Vaihingen & MEAN & $\Delta$ \\
\midrule
DINOv3 ViT-L (default) & $0.3$B & 64.4 & 27.7 & 44.8 & 49.6 & 61.9 & 49.7 & --- \\
DINOv3 ViT-7B & $6.7$B & 68.1 & 29.2 & 42.9 & 48.8 & 62.5 & 50.3 & $+0.6$ \\
DINOv2 ViT-L & $0.3$B & 58.1 & 22.4 & 37.7 & 42.8 & 56.9 & 43.6 & $-6.1$ \\
DINOv3 ConvNeXt-L & $0.2$B & 57.0 & 19.8 & 38.4 & 44.5 & 53.6 & 42.7 & $-7.0$ \\
DINOv3-SAT ViT-7B & $6.7$B & 43.0 & 13.9 & 38.1 & 34.0 & 50.6 & 35.9 & $-13.8$ \\
DINOv3-SAT ViT-L & $0.3$B & 43.6 & 12.5 & 37.9 & 33.5 & 49.7 & 35.4 & $-14.3$ \\
\bottomrule
\end{tabular}
}
\end{table}

\end{document}

%% file: math_commands.tex
\usepackage{amsmath,amsfonts,bm}

\def\eqref#1{equation~\ref{#1}}
\def\1{\bm{1}}

\DeclareMathAlphabet{\mathsfit}{\encodingdefault}{\sfdefault}{m}{sl}
\SetMathAlphabet{\mathsfit}{bold}{\encodingdefault}{\sfdefault}{bx}{n}